\documentclass[nonacm, format=acmsmall, review=false, screen=true]{acmart}

\usepackage{graphicx}
\usepackage{textcomp}
\usepackage{xcolor}
\usepackage{amsmath}
\usepackage{url}
\usepackage{bm}
\usepackage{multirow}
\usepackage{algorithm}
\usepackage{makecell}
\usepackage{subcaption}
\usepackage{algpseudocode}
\usepackage{verbatim}
\usepackage{tabularx}
\usepackage{booktabs}
\usepackage{listings}

\usepackage{wrapfig}        % To wrap text around a figure

\newcommand{\RNC}[1]{\MakeUppercase{\romannumeral #1}}
\usepackage{pifont}

\newcommand{\cmark}{\ding{51}}%
\newcommand{\xmark}{\ding{55}}%

\definecolor{codegreen}{rgb}{0,0.6,0}
\definecolor{codegray}{rgb}{0.5,0.5,0.5}
\definecolor{codepurple}{rgb}{0.58,0,0.82}
\definecolor{backcolour}{rgb}{0.95,0.95,0.92}
\usepackage{amsmath,amsfonts,bm}

\def\gS{{\mathcal{S}}}

\def\sY{{\mathbb{Y}}}

\def\Figref#1{Figure~\ref{#1}}
\def\Secref#1{Section~\ref{#1}}

\def\Tabref#1{Table~\ref{#1}}

\everypar{\looseness=-1}

\acmJournal{CSUR}
\setcopyright{acmlicensed}
\begin{document}

% Title portion. Note the short title for running heads
\title{Automated Deep Learning: Neural Architecture Search Is Not the End}
% \subtitle{Neural Architecture Search Is Not the End}

% \author{Xuanyi Dong$^{*}$}
% \affiliation{%
%   \institution{University of Technology Sydney}
% %   \city{Sydney}
% %   \state{New South Wales}
% %   \postcode{2007}
%   \country{Australia}
% }
% % \email{xuanyi.dong@uts.edu.au}
\author{Xuanyi Dong}
\affiliation{%
  \institution{Complex Adaptive Systems Lab, University of Technology Sydney}
%   \city{Sydney}
%   \state{New South Wales}
%   \postcode{2007}
  \country{Australia}
}
\affiliation{
  \institution{Brain Team, Google Research}
%   \city{Latexville 2}
  \country{USA}
}
\email{xuanyi.dong@uts.edu.au}

\author{David Jacob Kedziora}
\affiliation{%
  \institution{Complex Adaptive Systems Lab, University of Technology Sydney}
%   \city{Sydney}
%   \state{New South Wales}
%   \postcode{2007}
  \country{Australia}
}
% \authornote{Both authors contributed equally to this research work.}
\email{david.kedziora@uts.edu.au}

\author{Katarzyna Musial}
\affiliation{%
  \institution{Complex Adaptive Systems Lab, University of Technology Sydney}
%   \city{Sydney}
%   \state{New South Wales}
%   \postcode{2007}
  \country{Australia}
}
\email{katarzyna.musial-gabrys@uts.edu.au}

\author{Bogdan Gabrys}
\affiliation{%
  \institution{Complex Adaptive Systems Lab, University of Technology Sydney}
%   \city{Sydney}
%   \state{New South Wales}
%   \postcode{2007}
  \country{Australia}
}
\email{bogdan.gabrys@uts.edu.au}

\thanks{$\dagger$ Corresponding authors: Xuanyi Dong, David Jacob Kedziora, and Bogdan Gabrys.
\\Authors' emails: \{xuanyi.dong,~david.kedziora,~katarzyna.musial-gabrys,~bogdan.gabrys\}@uts.edu.au
}
%
% The code below should be generated by the tool at
% http://dl.acm.org/ccs.cfm
% Please copy and paste the code instead of the example below.
%

\makeatletter
\let\@authorsaddresses\@empty
\makeatother

% The default list of authors is too long for headers.
\renewcommand{\shortauthors}{Xuanyi Dong, David Jacob Kedziora, Katarzyna Musial, Bogdan Gabrys}

\begin{abstract}
Deep learning (DL) has proven to be a highly effective approach for developing models in diverse contexts, including visual perception, speech recognition, and machine translation.
However, the end-to-end process for applying DL is not trivial.
It requires grappling with problem formulation and context understanding, data engineering, model development, deployment, continuous monitoring and maintenance, and so on.
Moreover, each of these steps typically relies heavily on humans, in terms of both knowledge and interactions, which impedes the further advancement and democratization of DL.
Consequently, in response to these issues, a new field has emerged over the last few years: automated deep learning (AutoDL).
This endeavor seeks to minimize the need for human involvement and is best known for its achievements in neural architecture search (NAS), a topic that has been the focus of several surveys.
That stated, NAS is not the be-all and end-all of AutoDL.
Accordingly, this review adopts an overarching perspective, examining research efforts into automation across the entirety of an archetypal DL workflow.
In so doing, this work also proposes a comprehensive set of ten criteria by which to assess existing work in both individual publications and broader research areas.
These criteria are: novelty, solution quality, efficiency, stability, interpretability, reproducibility, engineering quality, scalability, generalizability, and eco-friendliness.
Thus, ultimately, this review provides an evaluative overview of AutoDL in the early 2020s, identifying where future opportunities for progress may exist.
\end{abstract}

%%
%% The code below is generated by the tool at http://dl.acm.org/ccs.cfm.
%% Please copy and paste the code instead of the example below.
%%
% \begin{CCSXML}
% <ccs2012>
%   <concept>
%       <concept_id>10010147.10010257</concept_id>
%       <concept_desc>Computing methodologies~Machine learning</concept_desc>
%       <concept_significance>500</concept_significance>
%       </concept>
%  </ccs2012>
% \end{CCSXML}

% \ccsdesc[500]{Computing methodologies~Machine learning}


\maketitle

\section{Introduction}\label{sec:introduction}

% Introduce AutoDL as a fusion of DL and AutoML.
In the quest for artificial intelligence (AI), history may judge the early 2010s as a mental reset, stimulating a new era of research and development with unrivaled intensity.
Within those reformative years, the field of machine learning (ML) witnessed a shifting of priorities and approaches.
Two threads of aspiration stand out:
\begin{itemize}
	\item Deep Learning (DL) -- The idea that multi-layered artificial-neuron networks are central to pushing the capabilities of ML.
	\item Automated Machine Learning (AutoML) -- The idea that no part of an ML workflow should necessarily depend on human involvement.
\end{itemize}

\noindent{It} was inevitable that these two ideologies would eventually converge, fusing into the novel subject of automated deep learning (AutoDL).

% Give a brief history of ML/DL.
Admittedly, while AutoDL is a ``hot topic'' in 2021, the foundations underlying this surge of activity stretch back for decades.
The notion of ML itself~\cite{michie1994machine} was established in the 1950s, aiming to tune mathematical models of desirable functions via automated data-driven algorithms.
In time, by the turn of the 21st century, numerous ML models and algorithms would be in practical use, with support vector machines and other kernel methods proving particularly popular~\cite{hofmann2008kernel}.
However, the concept of a neuron, inextricably linked to human intelligence, always seemed an obvious basis for ML.
Depicted computationally as early as in the 1940s~\cite{mcculloch1943logical}, their representational power in multi-layered arrangements was evident by the late 1960s, exemplified by the proto-DL ``group method of data handling'' (GMDH)~\cite{ivakhnenko1971polynomial}.
Since then, with stutters around AI winters, numerous types of neural layers and architectural variants have been proposed and adopted.
These include recurrent structures~\cite{little1974existence,hopfield1982neural}, convolutional and downsample layers~\cite{fukushima1982neocognitron}, auto-encoder hierarchies~\cite{ballard1987modular}, memory mechanisms~\cite{pollack1989implications}, and gating structures~\cite{hochreiter1997long}.
As a result, the historical successes of artificial neural networks (ANNs) are undeniably many, encompassing handwriting recognition~\cite{lecun1989backpropagation}, time series prediction~\cite{weigend1993results}, video retrieval~\cite{yang2009detecting,ji20123d}, mitosis detection~\cite{cirecsan2013mitosis}, and so on.
Yet the advantages of deep neural networks (DNNs), including their status as universal approximators~\cite{hornik1989multilayer}, are countered by the unwieldy nature of their complexity.
For instance, while backpropagation was established as reverse automatic differentiation in the 1970s~\cite{linnainmaa1970representation}, this DNN training technique did not become generally feasible until relatively recently.
Thus, the rising dominance of DL in the 2010s~\cite{lecun2015deep,schmidhuber2015deep} is as much an outcome of big data infrastructure and hardware acceleration, specifically graphical processing units (GPUs), as it is the result of any one theoretical advance.

\begin{figure}[t!]
  \begin{center}
    \includegraphics[width=\linewidth]{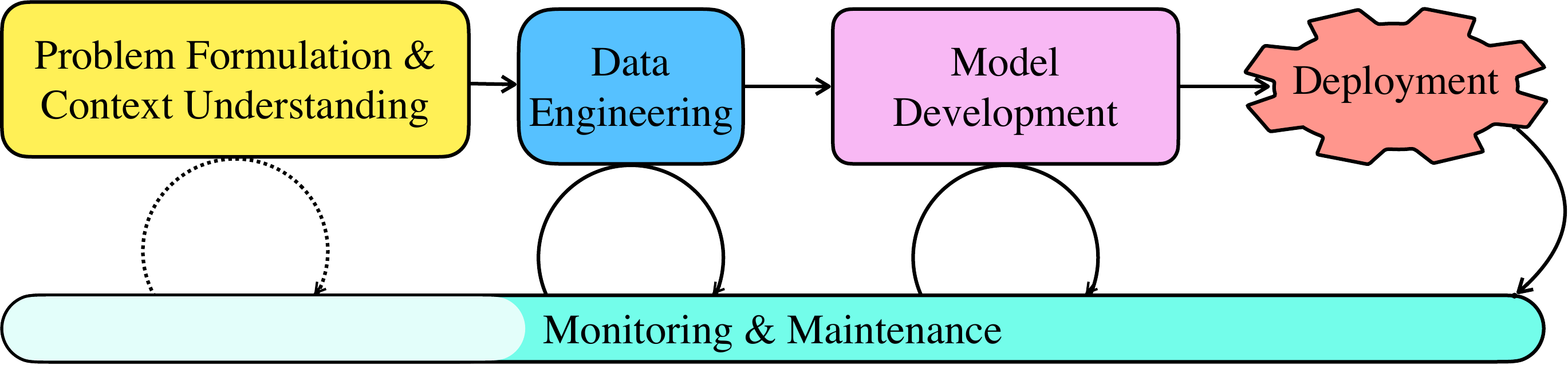}
  \end{center}
\caption{
Schematic of an end-to-end DL workflow, i.e., the processes involved in applying DL to a problem.
Traditionally, human decisions are required for every part of this workflow, such as analyzing a problem context, defining an ML task, designing a model, manually tuning hyperparameters, selecting training strategy, etc.\looseness=-1
}
\label{fig:dl-life-cycle}
\end{figure}

% Link a history of AutoML and mark NAS as the start of AutoDL.
In contrast, the evolution of AutoML is harder to pin down, primarily because the scope of automating higher-level ML mechanisms can be made extremely broad.
The extended history of this topic is grappled with elsewhere~\cite{kedziora2020autonoml}.
Nonetheless, the mainstream interpretation of AutoML -- and even the abbreviation itself -- has been forged on the back of advances in ML model/algorithm selection and the optimization of their user-defined hyperparameters.
Accordingly, if the success of a DNN in the 2012 ImageNet competition~\cite{krizhevsky2012imagenet} heralds the modern DL era, then the release of Auto-WEKA in 2013~\cite{thornton2013auto} marks the start of the modern AutoML era.
Within several years, by late 2016, these threads would start to entwine within the sub-field of neural architecture search (NAS)~\cite{baker2017designing,zoph2017NAS}.
This was not the first time that AutoML techniques had been applied to neural networks, but it was the moment that the broader data science community took notice.
It was also opportune; the website Papers-With-Code~\cite{paperswithcode} highlights that, while the number of DL publications has skyrocketed since 2012, year-by-year performance improvements on many benchmark datasets have diminished, i.e. those related to vision, text, audio, and speech.
There is a sense that, as state-of-the-art (SoTA) DL models have become highly sophisticated, a reliance on human design is locking out broader engagement behind steep learning curves, while also hindering further metric progress.
Automation through NAS is a vital step in enabling a broader community to push these technical limits.

% Claim that there is more to AutoDL than NAS.
Importantly, while NAS launches the modern AutoDL story, it does not encompass it.
Model selection, i.e. the design of a neural network, is but one stage of a DL workflow.
As illustrated in \Figref{fig:dl-life-cycle}, there are many other subtasks involved in ML/DL, such as defining a problem of interest, collecting and organizing data, generating features, deploying and adapting trained models, and so on.
This workflow may often be sequential in research and development, but real-world applications are much more agile and will typically reiterate through earlier operations and, in the case of large-scale systems, these processes may even be asynchronous.
In effect, AI based on DL cannot reach its full potential without considering the entire life cycle of a solution, from its design to the maintenance phase.

% Introduce AutonoML review and note AutoDL gap.
We now bring attention to a previous review~\cite{kedziora2020autonoml}, which surveyed efforts to automate all aspects of this workflow in the general context of ML, with additional focus on how the resulting mechanisms may be integrated into a single architecture.
The review touched on NAS and other elements of DL, but it could not cover the full extent of work in the AutoDL sphere.
It did not need to; on a high level, working with DNNs fits smoothly into the conceptual framework of both AutoML and its extension, autonomous machine learning (AutonoML).
However, on a practical level, the complexity of DNNs throws up many challenges that have arguably constrained the breadth of developments in AutoDL as compared to standard AutoML.
Instead, what is remarkable is the \textit{\textbf{depth}} of research in AutoDL, with numerous innovations brought about by attempts to surmount these obstacles, all with the aim of making the automation of DL feasible.
Certainly, it would be remiss to trivialize AutoDL as just a subset of AutoML.
Likewise, critically evaluating the limitations of present-day AutoDL is just as worthwhile as highlighting its accomplishments.
For instance, the field of DL is sometimes criticized for a tunnel-vision focus on model-performance metrics within a limited set of benchmarks, an attitude which, while valid, risks missing the broader perspective on all that AutoDL may become~\cite{gencoglu2019hark,dong2021nats}.
In essence, there is a need to consider several questions more thoroughly:
\begin{itemize}
    \item As we enter the 2020s, what is the current research landscape of DL?
    \item What makes a ``good'' DL model?
    \item How can automated systems best pursue and support this model ``goodness''?
    \item Is the field of AutoDL even advanced enough for such a meta-analysis?
\end{itemize}

\begin{figure}[t!]
  \begin{center}
    \includegraphics[width=\linewidth]{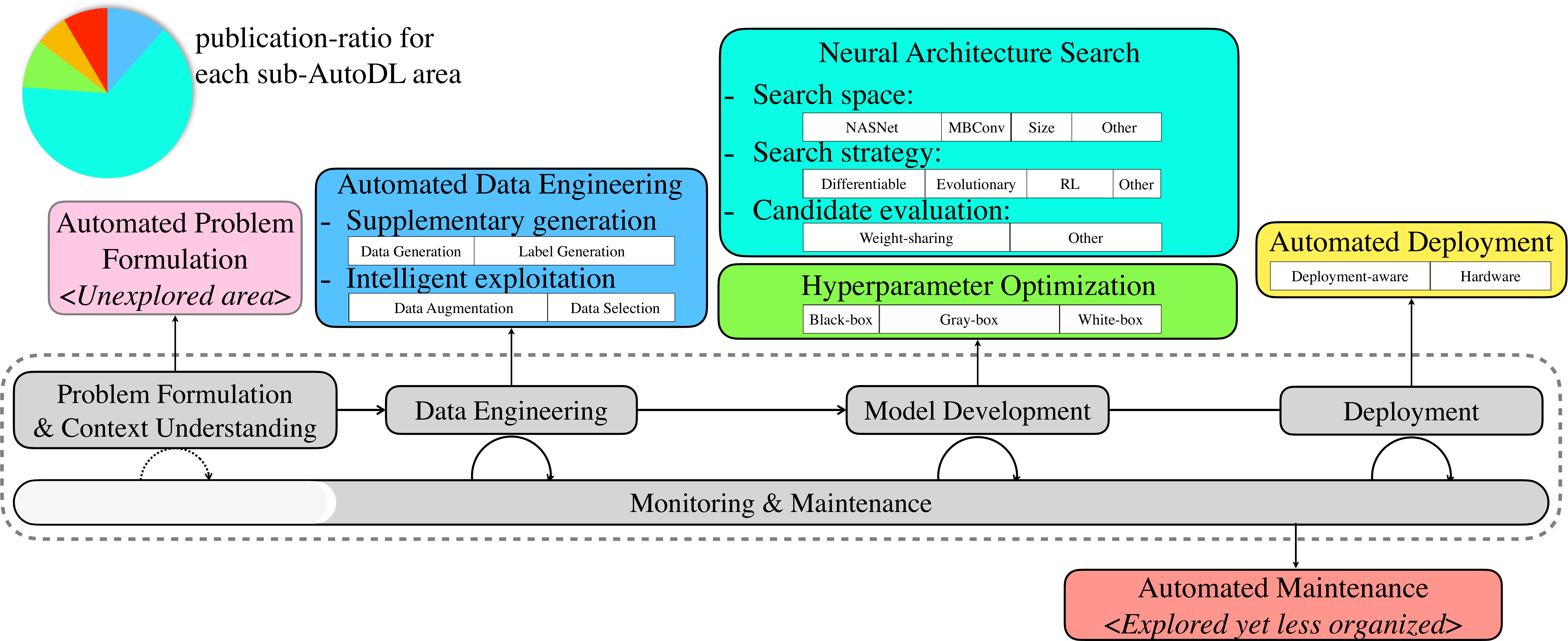}
  \end{center}
\caption{
Breakdown schematic for research activity in AutoDL, with surveyed publications attributed to different phases of a DL workflow and then further subcategorized.
The pie chart denotes the ratio of publications across all workflow phases, while the white stacked bars denote ratios within each subcategory.
% AutoDL aims to automate every human interaction traditionally required by the process of employing DL.
% For each sub-AutoDL topic, the length of each white block shows the relative ratio of papers published for each topic.
All statistics are derived from the Awesome-AutoDL project, at: \url{https://github.com/D-X-Y/Awesome-AutoDL}
}
\label{fig:what-is-autodl}
\end{figure}

% Motivate the AutoDL review and summarise sections.
This work is an extension of the broadly scoped AutonoML review~\cite{kedziora2020autonoml} with an in-depth focus on the newly popularized topic of AutoDL.
While there are many surveys in this sphere\footnote{One of the authors maintains a publicly accessible curated list of AutoDL resources at: \url{https://github.com/D-X-Y/Awesome-AutoDL}}~\cite{elsken2019neural,feurer2019hyperparameter,wistuba2019survey,yu2020hyper,ren2021comprehensive}, most focus on deep analysis within one or two sub-domains of AutoDL.
In contrast, we examine research along the entirety of DL workflow -- if it exists -- and try to assess, as of 2021, what the present role of AutoDL is and where its evolution is leading.
% To support this effort, this manuscript is structured hierarchically, as shown in \Figref{fig:what-is-autodl}.
We first provide an overview of AutoDL in \Secref{sec:autodl-overview}, introducing several fundamental concepts.
Then, partitioning major AutoDL research into sections inspired by a DL workflow, as per \Figref{fig:what-is-autodl}, we explore automation for: task management (\Secref{sec:auto-translate-context}), data preparation (\Secref{sec:auto-data-prepare}), neural architecture design (\Secref{sec:nas}), hyperparameter selection (\Secref{sec:hpo}), model deployment (\Secref{sec:deploy}), and online maintenance (\Secref{sec:maintenance}).

Crucially, a major component of this review is a reaction to the sheer quantity of publications in the space of AutoDL; we aim to provide summary assessments of surveyed AutoDL algorithms/research in terms of ten carefully designed criteria, not just accuracy alone.
These are introduced in \Secref{sec:criteria-intro} and form an evaluative framework for overviews within every aforementioned section, as well as, in \Secref{sec:criteria}, a final critical discussion around the entire field of AutoDL.

\section{AutoDL: An Overview}\label{sec:autodl-overview}

% Summarise the overview section.
The aim of AutoDL is to support, if not outright replace, the manual operations that data scientists undertake when applying DL to a problem.
\Secref{sec:workflow} elaborates what such a DL workflow may entail.
Of course, with AutoML having faced these same challenges for simpler ML models/algorithms, there is plenty of overlap between AutoDL and its more generic predecessor.
Thus, the basic concepts of AutoML are introduced in \Secref{sec:automl}, with a particular focus on an ``ML pipeline''.
Many of these notions are almost directly transferable.
Even so, the complexity of deep neural structures forces new challenges and different priorities upon AutoDL, which distinguish it as a research topic of its own.
These are summarized in \Secref{sec:beyond} and motivate the sections beyond.
Finally, \Secref{sec:criteria-intro} systematizes the ten criteria by which we propose AutoDL research should be judged; these will underpin brief evaluative overviews presented throughout this monograph.

\subsection{The DL Workflow}\label{sec:workflow}

% Talk DL before model construction.
A DL task typically starts with defining a problem of interest.
This primarily involves translating a desire conceived by humans into a computer-operable representation, e.g., the search for a predictive function from pixel maps to categorical classes, where sparse labeling may require semi-supervised learning techniques.
Once the problem is defined, the next step is usually to manage the input space for a prospective DNN model.
With a general assumption that input data should be independent and identically distributed (i.i.d.), strategies for data collection and organization need to be carefully considered.
Neural networks also train better when raw data is intelligently preprocessed, and there are many ways this can be done.
For example, principal component analysis (PCA) can be used to change the basis of high-dimensional data, i.e. instances with many features, such that data variances are maximized along a minimal number of dimensions; a subsequent projection eliminates the axes beyond these so-called principal components, achieving a dimensionality reduction with minimal information loss~\cite{hotelling1933analysis,pearson1901on}.
Preprocessing can also include the encoding of categorical data as integers or one-hot vectors~\cite{agresti2018introduction}, as well as feature scaling via normalization, standardization, or power transformation~\cite{box1964analysis}.

% Talk DL at model construction.
Eventually, the time comes to construct a DL model.
In standard formulation, this model is a DNN with multiple layers of neurons and various ways of connecting these layers together, e.g., using full connectivity, local convolutional connectivity, or even the outright layer skips of a residual neural network (ResNet)~\cite{he2016deep}.
Accordingly, a selection of a neural architecture and training algorithm is required; these are associated with what are commonly called ``model/architecture hyperparameters'' and ``algorithm/training hyperparameters'', respectively.
The DL model is then trained on input data, whereby the weights of this DL model are tuned\footnote{We avoid using the word ``optimize'' for network weights in this manuscript, so as to avoid confusion with the optimization of architectures, hardware, etc.} to best represent a desirable function.
Historically, there have been many proposals for how to do this, ranging from GMDH~\cite{ivakhnenko1971polynomial} to unsupervised winner-takes-all methods~\cite{lazzaro1988winner,carpenter1987massively,fukushima2013training}.
However, backpropagation\footnote{Technically, the term `backpropagation' refers to the computation of an error gradient, but it is often used loosely to also include a gradient-based optimization method that acts on this value.} has been the dominant training strategy of modern times for fully connected multi-layer neural networks.
With initial derivations and implementations~\cite{linnainmaa1970representation} tracing back to the 1960s and 1970s, respectively, it was popularized in the 1980s~\cite{rumelhart1985learning,lecun1985learning,williams1992simple} and effectively used for training multi-layer perceptrons (MLPs), even though it would be decades before advances in hardware would enable ubiquitous usage on large-scale problems.
Notably, the vanilla form of gradient descent through backpropagation has some drawbacks regarding speed, convergence, generalization, and so on.
Numerous upgrades have been proposed over the decades, such as stochastic gradient descent (SGD), momentum SGD~\cite{rumelhart1985learning}, resilient propagation~\cite{riedmiller1993direct}, and adaptive estimation~\cite{kingma2015adam}.
% Additionally, training multi-layered neural networks solely through the use of supervised learning algorithms like backpropagation has been beset by the so-called ``vanishing gradient'' problem~\cite{hochreiter2001gradient}, preventing effective training of ANNs with multiple layers. This has stimulated a new era of DL that mixes unsupervised learning algorithms -- e.g., to extract useful features in the early layers of a DNN -- with the more typical backpropagation-like forms of supervised learning for the final layer(s) of a DNN.
Neural Architecture Search (NAS) has also become one of the key research topics in DL model construction; see Section~\ref{sec:nas}.

% Talk DL after model construction.
Once the DL model is selected and trained, there is yet more to do within a typical DL workflow.
In practical applications, a DL model needs to be deployed, sometimes on custom devices and hardware.
For example, MobileNet-V2~\cite{sandler2018mobilenetv2} is deployed on an Edge Tensor Processing Unit (Edge TPU)~\cite{edge_tpu} to enable a 400 frames-per-second (FPS) inference speed, CycleGAN~\cite{zhu2017unpaired} targets Nvidia Graphics Processing Units (GPUs) for efficient execution, and face-recognition algorithms based on DL have been successfully deployed on smartphones~\cite{parkhi2015deep}.
In most cases, the deployed DL model is identical to the trained one in both structure and network weights.
However, in some resource-constrained scenarios, the models must be compressed via pruning, quantization, or sparsity regularization~\cite{han2016deep,han2016eie,dong2019tas,liu2015sparse}, before feeding them into real production.
Furthermore, there is a problem of learning from and adapting to changing environments with continuously streaming, non-stationary data that must be addressed by AutoDL approaches.
While this problem has been researched in the broader field of ML for a number of years~\cite{zliobaite2012next,kadlec2009architecture,gama2014survey}, even starting to be considered and addressed in the context of fully automated and autonomous ML systems~\cite{kedziora2020autonoml}, it continues to be a major challenge for DL. As the problem arises in many real-world scenarios such as stock markets~\cite{almgren2001optimal,yang2020qlib} and consumer recommendation systems~\cite{ricci2011introduction,he2017neural}, where DNNs are routinely deployed, robust adaptive capabilities are required to keep these models up-to-date.

\begin{table}[tp!]
\begin{center}\scriptsize
\setlength{\tabcolsep}{2pt}
\caption{
AutoDL algorithms dissected as optimization strategies.
}
\vspace{-4mm}
\begin{tabular}{ c | l | l | l | l | l }
\toprule
            & Algorithm & Search Space & Search Strategy & Boosts for Candidate Evaluation & Application \\
\midrule
\multirow{10}{*}{\rotatebox[origin=c]{90}{Auto. Data Engineering}}
      & \cite{maclaurin2015gradient} 2015.02 & training data                     & Differential  & optimal storage of discarded entropy & CLS: MNIST \\
    %   & \cite{konyushkova2017learning} 2017.03 & data selection &  \\
      & \cite{fan2018learning} 2018.05       & training data selection           & REINFORCE     & weight sharing; mini-batch sample    & CLS: MNIST/CIFAR/IMDB \\
      & \cite{ren2018learning} 2018.03       & training data re-weight           & Differential  & weight sharing; online approximation & CLS: MNIST/CIFAR \\
    %   & \cite{shu2019meta} 2019.02           & data weight      & Differential \\
      & \cite{cubuk2019autoaugment} 2018.05  & transformation in PIL~\cite{pil}  & PPO           & smaller model; reduced dataset & CLS: five datasets \\
      & \cite{lim2019fast} 2019.05           & transformation in PIL             & BayesOpt      & weight sharing                 & CLS: four datasets \\
      & \cite{zoph2020learning} 2019.06      & transformation in PIL             & PPO           & reduced dataset                & Object DET: VOC/COCO     \\
      & \cite{niu2019automatically} 2019.09  & classical NLP augmentations       & REINFORCE     & fewer epochs                   & Dialogue tasks       \\
      & \cite{cubuk2020randaugment} 2019.09  & transformation in PIL             & Random        & smaller model; reduced dataset & CLS + Object DET \\
      & \cite{such2020generative} 2019.12    & training data                     & Differential  & weight normalization           & CLS + Game: CartPole \\
      & \cite{li2020differentiable} 2020.03  & transformation in PIL             & Differential  & reduced dataset                & CLS + Object DET \\
\midrule
\multirow{27}{*}{\rotatebox[origin=c]{90}{Neural Architecture Search}}
      & \cite{bayer2009evolving} 2009.09         & cell topology in LSTM             & Evolution    &     N/A                                 & Grammar benchmarks      \\
      & \cite{jozefowicz2015empirical} 2015.07   & topology and operation in LSTM    & Evolution    & easy-to-hard tasks to filter            & Music + Language        \\
      &  \cite{zoph2017NAS} 2016.11              &  filter size + connectivity       & PPO          & fewer epochs                            & CLS + Language tasks    \\
      & \cite{baker2017designing} 2016.11        & MetaQNN space                     & Q-learning   & fewer epochs; early stop                & CLS: four datasets  \\
      & \cite{real2017large} 2017.03             & unrestricted CNN space            & Evolution    & weight inheritance                      & CLS: CIFAR-10/100   \\
      & \cite{zoph2018learning} 2017.07          & NASNet space                      & PPO          & fewer epochs                            & CLS + Object DET    \\
      & \cite{brock2018smash} 2017.08            & SMASH space                       & Random       & weight generation via HyperNet~\cite{ha2017hypernetworks} & CLS: five datasets \\
      & \cite{zhong2018practical} 2017.08        & BlockQNN space                    & Q-learning   & early stop                              & CLS: CIFAR/ImageNet    \\
      & \cite{ramachandran2017searching} 2017.10 & activation functions              & PPO          & smaller model                           & CLS + Translation      \\
      & \cite{real2019regularized} 2018.02       & NASNet space                      & Evolution    & smaller model; fewer epochs             & CLS: CIFAR/ImageNet    \\
      & \cite{pham2018efficient} 2018.02         & RNS        & REINFORCE    & weight sharing~\cite{pham2018efficient} & CLS + Language tasks \\
      & \cite{liu2019darts} 2018.06              & RNS              & Differential & weight sharing; smaller model; etc      & CLS + Language tasks  \\
      & \cite{cai2018path} 2018.06               & tree-structure                    & REINFORCE    & Net2Net~\cite{chen2016net2net}          & CLS: CIFAR/ImageNet \\
      & \cite{tan2019mnasnet} 2018.07            & MBS          & PPO          & fewer epochs                            & CLS + object DET       \\
      & \cite{bender2018understanding} 2018.07   & modified NASNet space             & Random       & weight sharing                          & CLS: CIFAR/ImageNet    \\
      & \cite{luo2018neural} 2018.08             & RNS              & Differential & weight sharing; neural predictor        & CLS + language  \\
    %   & \cite{zhang2019graph} 2018.10            & reduced NASNet space              & Random       &  weight generation via HyperNet & image classification \\
      & \cite{cai2019proxylessnas} 2018.12       & MBConv-based space                & Differential & weight sharing                          & CLS: CIFAR/ImageNet  \\
      & \cite{liu2019auto} 2019.01               & RNS + connectivity                & Differential & weight sharing; smaller model           & SEG: three datasets  \\
    %   & \cite{li2020random} 2019.02              & reduced NASNet space              & Random       & weight sharing; smaller model & image classification \\
      & \cite{chen2019detnas} 2019.03            & ShuffleNetv2-based backbone       & Evolution    & weight sharing                          & CLS + object DET     \\
    %   & \cite{quan2019auto} 2019.03 & reduced NASNet space + part-aware op & differentiable & weight sharing; smaller model & person re-ID \\
      & \cite{xie2019exploring} 2019.04          & architecture generator space      & manual       & fewer epochs                            & CLS + object DET     \\
      & \cite{ghiasi2019fpn} 2019.04             & FPN space                         & PPO          & smaller model; fewer epochs             & object DET: COCO     \\
      & \cite{dong2019tas} 2019.05               & depth + width                     & Differential & weight sharing                          & CLS: CIFAR/ImageNet  \\
    %   & \cite{dong2019search} 2019.06 & reduced NASNet space & Differential & weight sharing; Gumbel & image classification \\
      & \cite{fang2020densely} 2019.06           & densely connected search space    & Differential & weight sharing                          & CLS + object DET \\
    %   & \cite{xu2020pc} 2019.07 & reduced NASNet space & Differential & weight sharing; partial connection & image classification \\
    %   & \cite{du2020spinenet} 2019.12            & scale-permuted space              &  PPO         & smaller model; fewer epochs & object detection \\
      & \cite{ru2020neural} 2020.04              & architecture generator space      & BayesOpt     & fewer epochs                            & CLS: six datasets \\
      & \cite{liu2020evolving} 2020.04           & normalization+activation          & Evolution    & smaller dataset; fewer epochs           & CLS + SEG + GAN \\
      & \cite{wan2020fbnetv2} 2020.04            & width + resolution                & Differential & weight sharing                          & CLS: ImageNet \\
    %   & \cite{dai2021fbnetv3} 2020.06            & MBS + HPs                         & Evolution    & predictor; fewer epochs & image classification \\
    %   & \cite{tang2021auto} 2021.01              & reduced NASNet space              & Differential & weight sharing & navigation \\
\midrule
\multirow{10}{*}{\rotatebox[origin=c]{90}{Hyperparameter Opt.}}
      & \cite{bengio2000gradient} 1999.09     & a few differentiable HPs       & Differential    & N/A                                              & Synthetic data \\
      & \cite{snoek2012practical} 2012.06     & a few HPs                      & BayesOpt        & modeling costs; parallel                         & Diverse tasks \\
      & \cite{maclaurin2015gradient} 2015.02  & hundreds of differentiable HPs & Differential    & optimal storage of discarded entropy             & CLS: MNIST + Omniglot \\
      & \cite{li2018hyperband} 2016.03        & hundreds of HPs                & Random (Bandit) & adaptive resource allocation                     & Diverse tasks \\
      & \cite{loshchilov2016cma} 2016.04      & tens of of HPs                 & CMA-ES~\cite{hansen2001completely} & limited time budget; parallel & CLS: MNIST   \\
      & \cite{falkner2018bohb} 2018.07        & tens of of HPs                 & BayesOpt        & adaptive resource control; parallel              & Diverse tasks \\
      & \cite{houthooft2018evolved} 2018.02   & RL loss                        & Evolution       & truncated trajectory; parallel                   & Physics: MuJoCo \\
      & \cite{lorraine2020optimizing} 2019.11 & millions of differentiable HPs & Differential    & Neumann series                                   & Diverse problems \\
      & \cite{dong2021autohas} 2020.06        & MBS + tens of HPs              & REINFORCE       & weight sharing                                   & CLS: six datasets  \\
      & \cite{baik2020meta} 2020.10           & LR + WD      & Differential    & truncated trajectory                             & Few-shot CLS \\
\midrule
\multirow{8}{*}{\rotatebox[origin=c]{90}{Auto. Deployment}}
    %   & \cite{mao2016resource} 2016.11 & Resource management & REINFORCE &  & \\
      & \cite{mirhoseini2017device} 2017.06 & device placement                            & REINFORCE    & distributed training                      & CLS + translation \\
      & \cite{reagen2017case} 2017.07       & CMOS-based space + Arch. + HP                & BayesOpt     & N/A                                       & CLS: MNIST \\
      & \cite{nardi2019practical} 2018.10   & FPGA space + Arch. + HP                      & BayesOpt     & N/A                                       & N/A \\
    %   & \cite{kwon2018co} 2018.04 & SOC space + Arc.       &     & & image classification \\
      & \cite{parsa2019pabo} 2019.06        & PUMA space~\cite{ankit2019puma} + Arch. + HP & BayesOpt     & N/A                                       & CLS: Flower17/CIFAR \\
      & \cite{jiang2020hardware} 2019.07    & FPGA space + MBS                            & PPO          & fewer epochs; multi-level exploration     & CLS: CIFAR/ImageNet \\
      & \cite{choi2021dance} 2019.06        & Eyeriss space~\cite{chen2016eyeriss} + MBS  & Differential & weight sharing                            & CLS: CIFAR/ImageNet \\
      & \cite{yang2020co} 2020.02           & ASIC space + Arch.                           & REINFORCE    &  N/A                                      & CLS: three datasets \\
      & \cite{zhou2022rethinking} 2021.02   & edge accelerator space + MBS                & PPO          & weight sharing; neural predictor          & CLS: ImageNet + SEG \\
\midrule
\multirow{11}{*}{\rotatebox[origin=c]{90}{Auto. Maintenance}}
      & \cite{sutton1992adapting} 1992.07     & learning rate                    & Differential  & N/A                    & Synthetic data  \\
      & \cite{hochreiter2001learning} 2001.09 & learning algorithm               & Differential  & N/A                    & Synthetic data \\
      & \cite{white2016greedy} 2016.07        & trace hyperparameter             & Differential  & greedy strategy; approximation & Synthetic: Ringworld \\
      & \cite{wang2017learning} 2016.11       & RL algorithm                     & RL            & limit maximum trials   & Markov decision tasks \\
      & \cite{finn2017model} 2017.03          & initialization                   & Differential  & truncated trajectory   & Few-shot CLS \\
      & \cite{jaderberg2017population} 2017.11 & HPs                             & Evolution     & parallelization        & Game: Atari + StarCraft-{\uppercase\expandafter{\romannumeral2}}  \\
      & \cite{xu2018meta} 2018.05             & HPs in return function           & Differential  & gradient approximation & Game: Atari  \\
      & \cite{veeriah2019discovery} 2019.09   & auxiliary task as questions      & Differential  & truncated trajectory   & Synthetic + Game: Atari  \\
      & \cite{du2019sequential} 2019.06       & initialization + LR/WD + stop steps & REINFORCE     & limit maximum steps    & Recommendation \\
      & \cite{kirsch2020improving} 2019.10    & RL loss scalar                   & Differential  & truncated trajectory; parallelization   & Game + Physics  \\
      & \cite{xu2020meta} 2020.07             & RL target scalar                 & Differential  & truncated trajectory; single agent      & Game: Atari \\
    %   & \cite{liu2021return} 2021.02          &    &    &  & RL \\
\bottomrule
\end{tabular}\label{table:autodl-dissect}
\end{center}
{\scriptsize\textit{Algorithms are sorted chronologically per phase of a DL workflow.
Abbreviations are: ``Auto.'' for ``Automated'', ``Opt.'' for ``Optimization'', ``RNS'' for ``reduced NASNet space'', ``MBS'' for ``MBConv-based space'', ``Arch.'' for ``architectural search space'', ``HP'' for ``hyperparameters'', ``WD'' for ``weight decay'', ``CLS'' for ``classification'', ``DET'' for ``detection'', and ``SEG'' for ``segmentation''.
% To save space, we use ``RNS'', ``MBS'', ``Arch.'' and ``HP'' to indicate ``reduced NASNet space'', ``MBConv-based space'', ``architecture search space'', and ``hyperparameters'', respectively.
% ``CLS'', ``DET'', and ``SEG'' indicate ``classification'', ``detection'', and ``segmentation'', respectively.
% Most Auto-Adaptation methods could be categorized into HPO.
}}
\end{table}

Evidently, the success of a DL solution hinges on much more than model selection and the design of good neural architecture.
That stated, many phases of the DL workflow can still be expressed in similar ways, i.e., they can often be re-framed as an optimization problem.
\Tabref{table:autodl-dissect} dissects numerous seminal works in AutoDL according to such an interpretation, and these commonalities will be expanded upon over the course of this review.
Certainly, this representation is convenient, as a unified perspective across the entire DL workflow makes it arguably easier to build prescribed automated frameworks that manage a DL problem from end to end.
Moreover, this means that there is always a baseline way to assess mechanized approaches at almost any phase of the workflow.
Specifically, one can ask: what is the efficacy/speed of the search process?
Does it maximize DL model performance?
Of course, whether this is a sufficient form of evaluation is another matter, and we begin such discussion in \Secref{sec:criteria-intro}, but the optimization representation nonetheless serves to contextualize the historical evolution of AutoML/AutoDL.

\subsection{Connections to AutoML}\label{sec:automl}

% Motivate AutoML.
The automation of ML aims to minimize the need for human input at all stages of an ML workflow.
There are numerous motivations for this endeavor, both social and technical.
On the human side, expert technicians can turn their attention and energy to other productive areas, while democratization allows ML techniques to be employed by lay users.
However, on the technological side, automation can also improve the efficiency and speed of finding ML solutions, the quality and consistency of those ML solutions, the reusability of ML methodologies, and so on.
In pragmatic terms, this involves engineering a high-level AutoML system capable of managing the lower-level processes within an ML workflow, as schematized within \Figref{fig:what-is-autodl} for the DL case.

% Distinguish user workflow from transformation pipeline.
Now, terminology is key here.
We define an ML/DL workflow as the steps that a user must traditionally be involved in to produce and maintain a performant ML/DL solution; see \Secref{sec:workflow}.
This is distinct from what we introduce as an ML/DL pipeline, which describes the series of operations that inflow data undergoes in order to be transformed into useful outputs for descriptive, predictive, and prescriptive analytics.
In the ML sphere, such a pipeline can consist of data imputers, feature engineers, predictors, and ensemblers, among other options~\cite{sabu16,sabu19}.
It follows that the overarching purpose of an ML workflow is to find the best ML pipeline, as judged by some performance metrics.
Consequently, a portion of a modern AutoML system is always dedicated to what is effectively a direct optimization problem, i.e., selecting and tuning the components of such a pipeline.
However, AutoML can also encompass functionality that supports this goal indirectly, e.g., a natural-language user interface (UI) or a meta-model designed to analyze dataset similarity.
Thus, it is extremely challenging to distill both AutoML and its dynamical extension, AutonoML, into a simple engineering principle or a generic ML model development architecture~\cite{kadlec2009architecture,kedziora2020autonoml}.

% Introduce AutoML pipeline search.
Here, due to the constrained focus of present-day AutoDL, only the basics of AutoML need to be introduced.
Crucially, the core of every AutoML system is a module for hyperparameter optimization (HPO).
Its job is to explore a search space of configurations that define, for example, a predictor.
Each iteration is usually trained and then evaluated for model quality, e.g.~classification accuracy or detection precision, so as to find an optimal variant of the predictor.
If a user wants to then try other types of predictors and associated training algorithms, this becomes the combined algorithm selection and hyperparameter optimization (CASH) problem~\cite{thornton2013auto}; it is usually handled by treating the ML model/algorithm as just another high-level hyperparameter in configuration space.
Beyond this, the next level up in model complexity is a full multi-component pipeline search, which the CASH solvers of many SoTA AutoML packages strive to manage.

% Relate AutoML pipeline search to AutoDL.
Analogizing this search space within AutoDL is not trivial.
The early AutoDL community -- prior to the abbreviation being established -- have not always been aware of AutoML research.
When the term ``NAS'' was introduced~\cite{zoph2017NAS,baker2017designing}, the focus on neural architecture resembled a multi-component pipeline search with half the HPO, i.e., an HPO without algorithm hyperparameters.
In fact, inconsistently with AutoML, AutoDL researchers can often refer to HPO as the strict complement of NAS, i.e., solely an optimization of a training algorithm~\cite{feurer2019hyperparameter,dong2021autohas}.
We will adopt this terminology and emphasize these distinctions when needed for clarity.
Nonetheless, the analogy holds.
Much like a multi-component CASH-solver, a NAS strategy will often have a pool of possible components, i.e., layers, to build a DNN from.
The final layer of such an architecture is equivalent to a simple ML predictor, while all other layers can be seen as feature-space transformations.
Of course, that is not to say that DL does all its feature engineering within a DNN; the term ``auto-augmentation'' usually refers to optimizing data preprocessors outside of a neural network~\cite{cubuk2019autoaugment,cubuk2020randaugment}.

\begin{figure}[t!]
  \begin{center}
    \includegraphics[width=\linewidth]{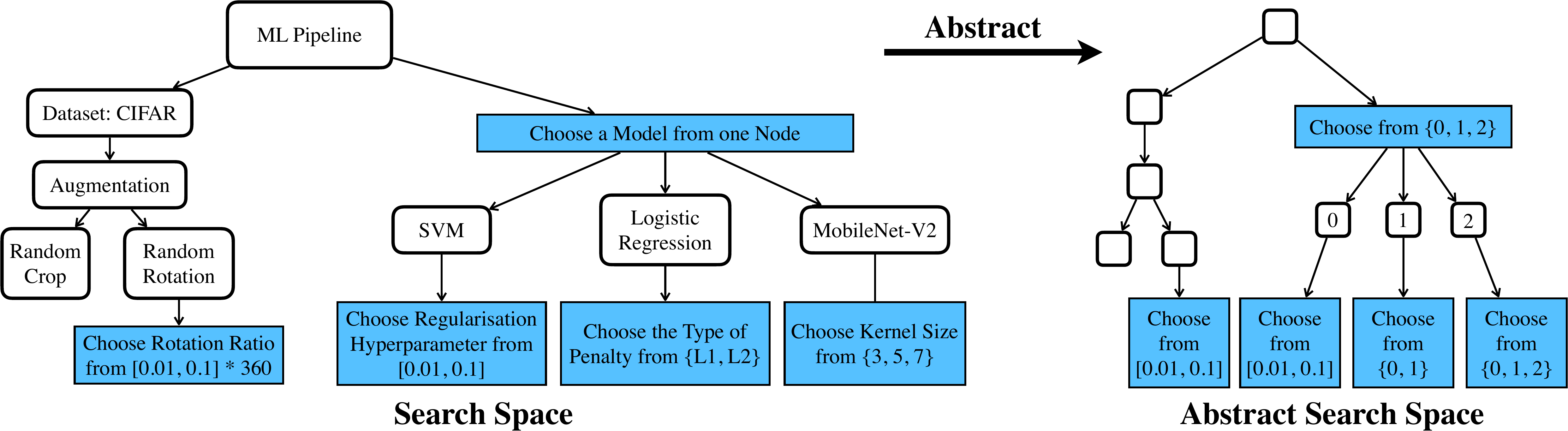}
  \end{center}
\caption{
An illustrative representation of an ML/DL pipeline search space, both concrete and abstract.
White rounded-rectangle nodes denote fixed components within the search space, while blue rectangle nodes denote optimizable options.
After an AutoML/AutoDL search, the resulting pipeline will have three components: a fixed crop, an optimized rotation, and one of three predictors with optimized hyperparameters.
% An illustrative example and an abstract view of an ML workflow search space.
% AutoML/AutoDL typically represents the search space as a tree structure.
% Each child node is a component of its parent, thus it depends on its parent.
% A white node (rounded rectangle) and a blue node represent a determined component and a to-be-determined component, respectively.
% To decouple the search space from AutoML algorithms, we can map the concrete tree into an abstract one.
}
\label{fig:search-space-abstract}
\end{figure}

% Exemplify search space.
Given the importance of a search space to ML/DL pipeline optimization, we provide a simple example of how configurations can be expressed.
First, a few diverse examples of hyperparameters are provided:
\begin{itemize}
    \item \textbf{Case$_1$}: Learning rate, as sampled from the range of $[0.01, 0.1]$ for each 90-epoch training run of ResNet.
    \item \textbf{Case$_2$}: The regularization variable for a Support-Vector Machine (SVM), possibly sampled from the range of $[0.01, 1.0]$.
    \item \textbf{Case$_3$}: The penalty for a logistic regression model, selected from \{L1, L2\}, as considered when maximizing classification accuracy for the Canadian Institute for Advanced Research (CIFAR) dataset~\cite{krizhevsky2009learning}.
    \item \textbf{Case$_4$}: Maximum rotation degree for a random rotation augmentation policy, as sampled from $[0.01, 0.1]\times360^{o}$ for each 90-epoch training run of MobileNet-V2 on CIFAR.
    \item \textbf{Case$_5$}: Kernel size, selected from \{3, 5, 7\}, when training MobileNet-V2 variants.
\end{itemize}
From this, it is clear that hyperparameters can be continuous (Case$_1$, Case$_2$ and Case$_4$), categorical (Case$_3$), discrete (Case$_5$), and so on.
Some will even be conditional, with, for instance, the choice of a predictor determining which variables are available to optimize.
Combining these all together can thus produce a very complex and high-dimensional configuration space for ML/DL pipelines.
This is exemplified in \Figref{fig:search-space-abstract}, which depicts the search space covered by four of the above cases in both explicit and abstract form~\cite{peng2020pyglove}.
Notably, these configurations still only represent a pipeline of no more than three components, i.e.~an image crop, an image rotation, and a monolithic model.

% Discuss basic search strategies.
Once a configuration space is defined, there are many ways to search through it.
Most solvers of the HPO/CASH/pipeline problem are black-box optimizers.
The simplest ones are based on grid search or random search, the former being quite standard for manual HPO among ML practitioners.
These strategies make no assumptions about the mapping from configuration space to solution quality, e.g., its derivatives, and are easy to scale up.
However, the evaluation of each candidate solution can be computationally expensive on its own; traversing a large configuration space can be extremely costly.
Thus, in practice -- although random search has proved remarkably efficient~\cite{bergstra2012random} -- a principled search strategy is desirable.

% Discuss principled search strategies.
Broadly stated there are three types of optimization routines typically employed in AutoML/AutoDL:
\begin{itemize}
    \item Population-based~\cite{goldberg1991comparative,loshchilov2016cma,real2017large,salimans2017evolution} -- 
	These approaches operate on sets of configurations named populations.
	Many seek gradual improvements through genetic-based processes between individuals within the population, e.g. crossover, mutation and selection.
    \item Bayesian Optimization (BayesOpt)~\cite{movckus1975bayesian,li2018hyperband,falkner2018bohb} -- 
	These approaches use a probabilistic approximation, also known as a surrogate, for the mapping between configuration space and ML pipeline quality.
	They alternate between two steps, the first being the use of an acquisition function on the surrogate to select the next most promising configuration to evaluate.
	The second step is evaluating an ML pipeline with that configuration and then using the new knowledge to update the fit of the surrogate.
    \item Distribution-based~\cite{williams1992simple,maclaurin2015gradient,schulman2017proximal,sutton2018reinforcement,liu2019darts,lorraine2020optimizing} -- 
	These approaches learn a parametric model of the probability distribution for whether an ML pipeline candidate will have a high metric score.
	The parametric models are usually tuned by reinforcement learning (RL)~\cite{williams1992simple,schulman2017proximal,sutton2018reinforcement} or gradient methods~\cite{maclaurin2015gradient,liu2019darts,lorraine2020optimizing}.
\end{itemize}
These three kinds of approaches can be further mixed into hybrid search strategies.
Nonetheless, even with SoTA optimization routines, the cost of ML/DL pipeline optimization can remain extreme.
There is plenty of ongoing research focused on efficiency gains within both ML and its narrower DL subset.
One way to boost search speed is to rely on low-fidelity approximations for pipeline evaluations, such as via dataset subsampling for training/testing or early-stopping for training algorithms.
However, there are many other proposed options as well.

% 
% Hint at meta-learning and MOO.
Crucially, these are just the basics of AutoML; model selection can be tweaked in various ways.
For instance, there are many investigations into the idea of meta-learning~\cite{schmidhuber1987evolutionary,schmidhuber2002bias,schaul2010metalearning,bengio1990learning}, where the historical application of ML workflows to ML problems may boost the efficiency and quality of a current solution search.
Then there is multi-objective optimization~\cite{marler2004survey,hwang2012multiple}, which appreciates the fact that model validity metrics, e.g., classification accuracy or detection precision, are not solely responsible for a good model.
Some alternative requirements, like short runtimes, can be aggregated with model accuracy easily~\cite{tan2019mnasnet,cai2019proxylessnas}, but others may be more challenging to evaluate, let alone Pareto-optimize, such as model interpretability or the convenience of user interactivity.

% Hint at AutonoML. 
The field of AutoML has also just started exploring the idea of dynamic environments in earnest, i.e., inflow data that changes over time to represent different information.
Unsurprisingly, the desire for an AutoML system to respond autonomously has produced developments in managing multi-pipeline solutions and adapting models.
We refer the reader to the AutonoML review for further information~\cite{kedziora2020autonoml}; it is an expansive subject.
Here, we will only elaborate topics if existing AutoDL research warrants it.

\subsection{AutoDL Beyond AutoML}\label{sec:beyond}

% Link DNN complexity to the challenges of AutoDL.
As this review will make clear, while boosting search efficiency is important in the field of AutoML, it is critical to AutoDL.
A DL model in the form of a DNN, combined with auto-augmentation, is far more flexible and complex than a typical ML pipeline.
This means that DL pushes the limits of computational resource usage, hardware provisioning, model search space, and so on.

% Mention NAS/HPO focus.
Given that model construction is so challenging, research in AutoDL has generally focused on a much narrower scope than AutoML.
The bulk of existing surveys assess NAS~\cite{elsken2019neural,wistuba2019survey,ren2021comprehensive} and HPO~\cite{feurer2019hyperparameter,yu2020hyper}.
Similarly, we have examined a spread of AutoDL approaches for model search, as dissected in \Tabref{table:autodl-dissect}; these will be discussed from various angles in the following sections.

% Exemplify necessity-driven innovations.
However, AutoDL research has its own notable fringes, occasionally expanding beyond even the current domain of AutoML.
For example, standard AutoML cares moderately about computational resources, whereas AutoDL \textit{needs} optimal infrastructures to run SoTA DL models effectively.
Thus, out of necessity, hardware search has become an attractive facet of AutoDL, with certain experiments varying these infrastructures~\cite{mirhoseini2017device,nardi2019practical} while keeping other elements constant, e.g., data preparation, the DL model, and the training strategy.
Then there is DL pipeline ensembling; the analog is not unheard of in AutoML/AutonoML~\cite{kedziora2020autonoml}, but its appeal has arguably driven its development in AutoDL much more strongly.
After all, if a single DL model is so computationally expensive to construct, would it not be beneficial to keep the completed product around in a pool or ensemble, so as to leverage whatever lessons it has learned?
With benefits in robustness, reusability, and generalizability, this approach has been adopted several times~\cite{caruana2004ensemble,xie2019exploring,zaidi2020neural,radosavovic2020designing,cai2020once}.

The take-away here is that AutoDL is worthy of independent consideration alongside AutoML; the field has had to embrace various innovations to face the challenges of DL model complexity.

\subsection{Assessment Criteria for AutoDL Research}\label{sec:criteria-intro}

Here lies a central problem: to varying degrees along the workflow, the field of AutoDL has been flooded with research.
It is extremely challenging for a would-be developer of an integrated AutoDL system to decide which techniques and mechanisms to favor.
In fact, we argue that the predominant focus on end-point accuracy/efficiency is insufficient to assess a piece of AutoDL research, even after accounting for the shortcomings of current benchmarking practices.
Thus, this review uniquely proposes that AutoDL researchers/developers should pay attention to a more encompassing set of ten criteria.

Altogether, the ten are listed below.

\noindent\textbf{\RNC{1}. Novelty:} How does the AutoDL algorithm distinguish itself from all existing works in AutoDL?

\noindent\textbf{\RNC{2}. Solution Quality}: How well does the AutoDL algorithm minimize the error of a target DL model?

\noindent\textbf{\RNC{3}. Efficiency}: Does the AutoDL algorithm achieve its aims with minimal resource expenditure, especially in terms of time costs? How does it impact the resource costs of a target DL model?

\noindent\textbf{\RNC{4}. Stability:} How consistent is the performance of the AutoDL algorithm with respect to statistical variability? How dependent is its performance on the choice of settings?

\noindent\textbf{\RNC{5}. Interpretability:} Is the AutoDL algorithm theoretically sound and human-understandable? How does it impact the explainability of a target DL model?

\noindent\textbf{\RNC{6}. Reproducibility:} Are reported results associated with the AutoDL algorithm easily reproduced? Have they been reproduced?

\noindent\textbf{\RNC{7}. Engineering Quality:} Does the AutoDL algorithm have an implementation? Is this codebase well managed, documented, accessible, and of a high standard?

\noindent\textbf{\RNC{8}. Scalability:} Is it feasible for the AutoDL algorithm to scale to a larger model or more data?

\noindent\textbf{\RNC{9}. Generalizability:} Can the AutoDL algorithm be applied to different tasks, datasets, search spaces, etc.?

\noindent\textbf{\RNC{10}. Eco-friendliness:} What is the environmental impact of both the AutoDL algorithm and its target DL model?

% Some of the above criteria are challenging to quantify, and this is discussed in \Secref{sec:criteria}.
% Nonetheless, they provide a broader assessment framework beyond the useful but limited representation in \Tabref{table:autodl-dissect}.

These criteria will form the basis for evaluative assessments in the sections to come.
However, although \Secref{sec:final-discussion-questionnaire} discusses relevant questions to ask of individual publications, motivating a detailed breakdown, most of the overviews in this monograph concern entire threads of AutoDL research related to individual phases of a DL workflow.
Thus, there is more to explain about how we formulate such assessments, which are necessarily aggregated and mostly qualitative.

First of all, there is an important caveat to acknowledge.
An AutoDL algorithm, i.e., a process, is distinct from an impacted target DL model, i.e., an outcome.
Some AutoDL algorithms will directly construct this model, e.g., NAS mechanisms, and others will simply influence it, e.g., automated mechanisms for data engineering or maintenance.
Now, an issue arises in that, while the notion of accuracy for an AutoDL algorithm is derived and will always refer to the performance of a target DL model, both an AutoDL algorithm and a target DL model will have their own distinct qualities in terms of efficiency, interpretability, etc.
The two need not be aligned either; an efficient NAS approach may produce an inefficient DNN, and, likewise, an inefficient mechanism can produce an efficient model.
However, in practice, this survey has not found such discrepancies to be particularly common.
If an ethos of efficiency or interpretability drives research in this field, as an example, then both AutoDL algorithm and target DL model tend to benefit from those improvements.
Thus, from an aggregate perspective, it is safe to make research-trend assessments based on evaluating the AutoDL process itself.

With that issue addressed, we now elaborate on the upcoming overviews.
To begin with, there is one major divergence for a criterion-based assessment between publications and entire research trends.
Specifically, the novelty (\RNC{1}) of a methodological category is represented by the years in which seminal works for the approach were published and, accordingly, how long the computer science community has been aware of its existence.
Distinct from novelty for an individual publication, which should always be significant, novelty for an entire topic is meant to be more informative than judgemental; older approaches are likely to be more robust and well-explored, while newer approaches are more likely to leverage SoTA breakthroughs.
Also, as a side note, it is very case-dependent on whether it is more instructional to present these histories in the context of DL or, more broadly, ML.
Associated table captions will clarify which historical context is used.

In all other criteria, assessments for a methodological paradigm are but averages of all surveyed publications that theoretically/experimentally employ that paradigm.
Most (\RNC{2}--\RNC{5} \& \RNC{8}--\RNC{10}) are given a rating of low, medium, and high.
Designations of mixed and unknown, i.e., ``?'', are also possible.

Solution quality (\RNC{2}) represents the contribution of an AutoDL approach towards the validity of a resulting DL model, according to self-reported but peer-reviewed claims.
For model-development techniques, this is baseline accuracy (suppose it is a classification task).
For maintenance techniques, this is accuracy integrated over time.
For everything else, this is improvement beyond the baseline accuracy.
Next up is efficiency (\RNC{3}), which, with adjustments for memory usage, primarily considers how quickly an AutoDL algorithm runs, again as self-reported.
For model-development techniques, this includes the training time for the DL model as well.
Stability (\RNC{4}) then follows, acknowledging how tight self-reported statistical bounds are on the accuracy of an AutoDL approach, as appropriately defined with respect to a phase of the DL workflow.

Interpretability (\RNC{5}) notes the degree to which either, one, the operation/impact of an AutoDL algorithm is immediately clear or, two, publications associated with the research trend make the effort to theoretically elucidate how and why the algorithm works; ablation studies are an example of a gold standard for this criterion.
Scalability (\RNC{8}) then assesses how resilient the efficiency of an AutoDL approach is when handling an increasingly difficult DL problem, as evidenced by self-reported complexity analyses or similar extrapolations.
Naturally, dataset size is a baseline metric for problem difficulty at all phases of a DL workflow, but model-development and deployment methodologies are also considered with respect to the size of relevant search spaces, and deployment and maintenance approaches additionally deal with model size.
As for generalizability (\RNC{9}), this criterion captures the diversity of DL problems that an AutoDL approach can be applied to as is.
This is often calculated as the inverse of how many context-specific assumptions are present in the self-reported theoretical foundations of the methodology.

Finally, we turn to the remaining criteria.
Notably, the ones already listed can be compiled directly from researchers reporting their own findings, albeit in an aggregate sense.
In contrast, reproducibility (\RNC{6}) and engineering quality (\RNC{7}) rely on secondary benchmarks and implementations, respectively.
It is beyond the scope of this review to grade these with a desirable rigor, so the associated criteria are simply marked ``\cmark'' if there is enough literature to assess relevant research trends appropriately, and ``?'' if not.
As for eco-friendliness (\RNC{10}), this criterion is stimulated by growing environmental concerns and can be graded within the low-to-high scheme but, in practice, will often be marked as unknown in this review.
One can assert that a ranking for an AutoDL algorithm is likely to correlate with efficiency and scalability, but, except in the most obvious cases, this review requires a direct analysis of power consumption to support a qualitative assessment.

With these ten criteria defined and their evaluation explained, we emphasise that such a set provides a broader assessment framework beyond the useful but limited representation of AutoDL algorithms in \Tabref{table:autodl-dissect}.
Indeed, this kind of consistency and thoroughness is needed to compensate for the idiosyncrasies at every phase of the DL workflow, and we posit that such an evaluative framework is a prerequisite to truly identifying promising directions in AutoDL.

\section{Automated Problem Formulation}\label{sec:auto-translate-context}

Employing the DL approach for real-world applications spans a wide range of processes, shown in \Figref{fig:dl-life-cycle}.
If an ideal AI system is to one day automate this entire procedure, then, for completeness, it is worth discussing the formulation of a learning task from a problem context.
Put simply, a problem context conceptualizes broad human-defined goals, such as creating ``an undefeatable computer opponent for the game of Go'' or ``a car that automatically drives people to their desired destination''.
It also covers the environment in which these goals apply, such as ``the rules of Go'' or ``the geography and physics of road transportation''.
Traditionally, data scientists have had to manually translate these conceptual contexts into computer-actionable tasks.
For instance, one may decide to frame the design of a Go agent as an RL-based optimization task for a DNN, where the probability of winning is the objective function and an appropriately constrained input space represents board positions~\cite{silver2016mastering}.

This ability to interpret a general problem context and forge a pragmatic pathway to a DL solution is a challenge; it may ultimately be the final obstacle for pure AutoDL, given how difficult it is to artificially mimic human creativity.
Unsurprisingly, there is no major literature on this topic currently, with the majority of existing work in both AutoML and AutoDL focusing more on model-construction aspects.
For that same reason, it is difficult to speculate whether AutoDL would treat task auto-formulation differently from AutoML.
Certainly, AutoDL opens up new types of learning tasks to map problems into, e.g., the development of convolutional long short-term memory (LSTM) networks for dynamic image recognition problems, but this is an issue of categorization rather than a fundamental contrast.

Nonetheless, speculation aside, this space is not untouched.
One example is the Libra system\footnote{See: \url{https://github.com/Palashio/libra}}, which aims to assist -- if not automate -- the act of declaring ML/DL tasks via natural language processing (NLP).
It enables this by constructing a semantic context around datasets and other objects, making it possible to interpret requests such as ``please model the median number of households'' or ``predict the proximity to the ocean''.
Likewise, the notion of problem-to-task translation links closely to scattered but growing research in the area of automated human-AI interfacing; interested readers are pointed to the section on user interactivity in the AutonoML review~\cite{kedziora2020autonoml}.
For now, however, progress in this area remains too sparse to evaluate in terms of the criteria introduced in \Secref{sec:criteria-intro}.

\section{Automated Data Engineering}\label{sec:auto-data-prepare}

In real-world applications, the path from a raw data source to a model input can be a long one.
At the earliest extreme, raw data can be encoded in numerous ways, e.g., vector versus pixel graphics, and can possess context-based idiosyncrasies, e.g., asynchronous timestamps for datastreams.
Given all the expert knowledge ingrained in these arbitrarily unique formats, data wrangling joins task formulation as a process that is intensely challenging to automate.
Of course, eventually, truly autonomous learning agents will need to be capable of seeking out problem-relevant data, at least as well as a human trawling the internet.
Perhaps these efforts will be aided by modern innovations around the ``extract, transform, load'' (ETL) paradigm, renowned in data engineering.
However, to date, it is rare to find even AutoML-based research/technology that does not assume some level of convenient formatting for collated data.

Nonetheless, there have been reasonable investigative attempts at automating most other phases of data engineering.
Traditional AutoML often works with relatively simple and quickly trained ML models, so it is arguably more advanced in this space; see the Automated Feature Engineering section of the AutonoML review~\cite{kedziora2020autonoml}.
Even so, AutoDL research has explored data preparation too, and its focus is driven by the unique demands of DL models.
Simply put, a DNN needs a lot of labeled data\footnote{Labels encompass manual annotations for supervised learning, pseudo-labels for unsupervised learning, and generated proxy-task labels for self-supervised learning.}.
Its complex nature as a universal approximator affords many representational benefits, but the numerous (effective) degrees-of-freedom requires a large number of training data instances to achieve good generalisation and avoid overfitting.
This can make it difficult to properly train, for example, GPT-3 with its 175 billion weights~\cite{brown2020language}.
Thus, most AutoDL research in this space prioritises one of two approaches: (i) generate more data or (ii) use data in a better way.

% \textbf{Make More Data} -- 
\subsection{Supplementary Generation}
When ground-truth observations are limited, one solution is to artificially generate new instances.
Many sophisticated generators can be employed in this capacity, e.g., deep Generative Adversarial Networks (GANs)~\cite{goodfellow2014generative}, deep Variational Autoencoders (VAEs)~\cite{vincent2008extracting}, or modern physics engines~\cite{unrealengine}.
Abstractly put, the idea is to automatically ``paint out'' the broader space represented by limited real data via interpolative or -- more riskily -- extrapolative procedures.
In practice, the assumptions underlying such estimations can be complex enough to seem arcane, e.g., the function learned by a GAN discriminator or the chaotic dynamics of complex physical equations.
Nonetheless, the procedure has proven merit, provided that a data generator, e.g., in the form of a neural network~\cite{such2020generative}, can, via optimization or similar, properly capture the salient characteristics of the real data.

This is key; automated data generation cannot supply a model with any useful information beyond its limited observations.
It can actually introduce false assumptions that degrades model performance, made clear once the model is tested on the real data environment.
Accordingly, to pursue a deeper understanding of this issue, several research efforts have examined just how reductively observed knowledge can be distilled~\cite{maclaurin2015gradient,wang2018dataset}.
Indeed, this kind of compression is the premise behind autoencoding.
But it remains an inescapable fact that, if a model requires new discriminative information, not just data estimates, an AutoDL system will need to seek it out.
For instance, while the system can itself be designed to pick reliable pseudo-labels for unlabeled data, known as self-training, a common approach is to query an oracle -- often a human -- for such annotations.
If these instances of data are selected intelligently, this is called active learning, and even the high-level strategies for active learning have been explored as targets for automated selection~\cite{konyushkova2017learning}.

% \textbf{Use Data Better} -- 
\subsection{Intelligent Exploitation}
Even if data is limited, the practicalities of ingesting data and training a DNN can still very much affect its performance.
Batch size is a classic hyperparameter to consider, but even the order of data ingestion matters~\cite{bengio2009curriculum}.
In effect, research here revolves around information content within data and how to maximally squeeze out its beneficial impact on an ML model.
So, as an example of automating data provision, a neural energy network has been explored as a teacher to select suitable data for another student network~\cite{dong2019teacher}.
Elsewhere, a weighting function -- represented later by an MLP~\cite{shu2019meta} -- has been used to flexibly control the influence of incoming data on model training; gradient-based optimization methods have attempted to learn the best values of these weights, so as to minimize model validation loss~\cite{ren2018learning}.

Of course, not all processes of automatically ``getting useful data'' can be divided so cleanly between generating new instances and using existing ones better.
The complex data types used by DNNs, such as images and freeform texts, often contain extra information content that can be pulled out by relatively simple transformations, e.g., image rotations or by replacing text with synonyms.
This concept is known as ``augmentation'' in the DL community, straddling the line between generative and transformative, and its use can significantly boost DL performance~\cite{cubuk2019autoaugment}.
Many forms of augmentation have been explored, including normalization, standardization~\cite{ioffe2015batch}, factor design~\cite{yang2020qlib}, image augmentation~\cite{shorten2019survey}, text augmentation~\cite{ma2019nlpaug}, etc.
However, while most of these simple transformations are well defined, it can still be difficult to manually choose their hyperparameters.
For trivial instance, without domain knowledge, when do image rotations go from a better discernment of a handwritten `6' to a misidentification of the digit `9'?
Thus, a new thread of research in AutoDL named auto-augmentation aims to take such choices out of human hands.
The original work explores image augmentation policies, seeking to optimize a mix of transformation types and magnitudes~\cite{cubuk2019autoaugment}.
Subsequent works have tried to improve accuracy, efficiency, and generalization ability~\cite{cubuk2020randaugment,niu2019automatically}.

\begin{table}[tp!]
\begin{center}\scriptsize
\setlength{\tabcolsep}{3pt}
\caption{
Evaluative assessment for trends in automated data engineering.
}
\begin{tabular}{l | l | l | l | l | l | l | l | l | l | l | l}
    \toprule
 &                   & Novelty           & Solution & Effic. & Stability & Interp. & Reprod.  & Engi. & Scalability   & General. & Eco. \\
\midrule
\multirow{2}{*}{\makecell[l]{Supplementary\\Generation}}
 & data generation   & $\approx$ 2 (30) & Low   &   Low     & Low     & High          & ?  & ?     & ?    & Low    & ?  \\\cmidrule{2-12}
 & label generation  & $\approx$ 2 (40) & High  &   Low     & High    & Low           & ?  & ?     & High   & Low    & ?  \\
\midrule
\multirow{2}{*}{\makecell[l]{ Intelligent\\Exploitation}}
 & data augmentation & $\approx$ 5 (30) & High  &   Mixed   & High    & Low           & ?  & ?     & High   & Low    & ?  \\\cmidrule{2-12}
 & data selection    & $\approx$ 4 (30) & Medium  &  Mixed     & High    & High       & ?  & ?     & High    & High   & ?  \\
    \bottomrule
\end{tabular}\label{table:ade-methods}
\end{center}
{\scriptsize\textit{
Each row marks an emergent trend in AutoDL, specifically automated data engineering. Each column marks a criterion -- see \Secref{sec:criteria-intro} -- by which the trend is assessed. The evaluations are mostly qualitative, averaged across the most significant works researching the trend. Where a graded value is not provided, ``?'' indicates that a rigorous assessment is not achievable without more research works to analyze.  Novelty denotes years since seminal works in DL (ML) were published. Abbreviations are: ``Solution'' for Solution Quality, ``Effic.'' for Efficiency, ``Interp.'' for Interpretability, ``Reprod.'' for Reproducibility, ``Engi.'' for Engineering Quality, ``General.'' for Generalizability, and ``Eco.'' for Eco-friendliness.
}}
\end{table}

\subsection{Overview}

% Comment on novelty.
In summary, the current state of automated data engineering in AutoDL -- what little of it exists -- is primarily about getting as much data as possible, i.e., data generation and label generation, and leveraging what exists with maximal efficiency, i.e., data augmentation and data selection.
First to note though, from \Tabref{table:ade-methods}, is that many of the related approaches have been developed in the context of ML over many decades, even if they have only being associated with DL for a handful of years.

% Comment on efficiency.
Traditionally, data engineering efforts have been guided with or without model performance in mind, and these are sometimes called ``wrapper'' or ``filter'' methods, respectively; filter methods, e.g., PCA, often lean on statistical properties of the datasets alone.
Although filter methods are much faster than wrapper methods, especially given that training a DNN for validation is so expensive, virtually all AutoDL endeavors in this area have thus far focused on model performance, i.e., they deal with wrapper methods.
Unsurprisingly, this preoccupation challenges the efficiency of automated data engineering in general, although other factors can of course impact the rating either way.
For instance, data generation often requires the expensive training of generator models, while, conversely, the computational demands for both data augmentation and data selection can be ameliorated somewhat by weight sharing -- see \Secref{sec:nas-boost-eval} -- and online approaches, respectively.

% Comment on accuracy.
Nonetheless, computational cost aside, there is a reason these approaches are employed to begin with.
The information content provided by both label generation and data augmentation has been showed to markedly improve model accuracy.
Data generation and selection both have promise too, although the latter refines rather than enriches a dataset and thus strains against its own performance ceiling, i.e., it is limited by what data is available.
To be fair, both have also only had limited exploration in AutoDL thus far, applied only to small datasets, which means there has not been the same degree of model-improvement claims in the literature.

% Comment on stability and scalability.
In fact, data generation in AutoDL can be considered particularly experimental at the current time, suggesting that approaches do not have the same degree of performance stability as the other three trends of interest, even if, admittedly, auto-augmentation has only been primarily tested within the context of vision problems.
Additionally, data generation sticks out in terms of scalability.
Specifically, algorithms for the other three AutoDL trends typically scale reasonably well with respect to the size of a problem dataset, i.e., the yardstick of interest for these approaches, as they simply involve operations applied to instances of data.
In contrast, it is unclear as to how an interpolating/extrapolating data generation algorithm depends on dataset size, given that some data instances paint out a far more informative picture of a global features-to-label space than others.
%there is no clear scalable relation between the size of a problem space and the resources required to run an interpolating/extrapolating data generation algorithm.
In short, more study is required.

% Comment on interpretability, generalisability and 'question marks'.
Turning to the remaining criteria in \Tabref{table:ade-methods}, the interpretability of data-engineering algorithms is roughly in opposition to their associated accuracy improvements.
It is hard to speculate whether the anti-correlation is coincidental or not, but it is evident that the practices of data generation and data selection have been contemplated heavily in the literature, supported by visualization in the former case, while algorithms for label generation and data augmentation have had very little theoretical analysis to date.
When it comes to whether these approaches generalize though, data selection is the sole winner, with procedures usually agnostic to data structure and format, problem formulation, etc.
In contrast, the other three AutoDL trends often manifest in algorithmic implementations that are tied to certain tasks or applications.
Finally, it is not possible to comment on reproducibility or engineering quality, as benchmarks and codebases are in short supply, while eco-friendliness has not yet been studied.
This is to be expected, given that existing research into automated data engineering within DL still seems mostly proof-of-principle or focused on theoretical design.

% Preface NAS by noting feature engineering is often network-internal.
For completeness, it is worth returning to how data engineering differs between ML and DL.
While auto-augmentation in the DL context is often related to generating transformed copies of existing data, there is a fuzzy overlap with feature engineering, which typically refers to in-place data transformations.
We do not stress the semantics here; auto-augmentation is nascent enough to be somewhat fluid.
Nonetheless, we emphasize that AutoML has often rigidly separated the task of optimizing preprocessing pipelines, i.e., data engineering, from the CASH problem, i.e., model development.
In the contrasting case of DL, while augmentation covers network-external preprocessing, much of the feature engineering is still relegated to the early layers of a DNN as part of a fuzzy overlap.
It is thus understandable why the topic of NAS dominates AutoDL, even if the field will eventually need to grapple with data preparation more broadly, especially as AutoDL moves beyond academic research and into real-world application.

\section{Neural Architecture Search}\label{sec:nas}

Flexible neural architecture design is arguably the core advantage/challenge offered up by DNNs.
Thus, unsurprisingly, automating that design has attracted significant attention from the DL community.
Fundamentally, this endeavor revolves around connecting layers of neurons into an extended architecture, where the full network can be considered as a directed acyclic graph (DAG).
Examples in \Figref{fig:nas-dag} depict this formulation, with input tensors being transformed into output tensors via a number of intermediate operations.
More precisely, intermediate feature tensors are produced via the transformation and fusion (with some learnable weights) of existing feature tensors.
The final output, resulting from the last operation and in tensor format, represents useful information, e.g., classification predictions or discriminative features in supervised and unsupervised learning, respectively.
Given this context, the modern surge in popularity of DL has been driven by proposals of new and effective forms for such architectures~\cite{krizhevsky2012imagenet,simonyan2015very,szegedy2015going,he2016deep,he2017mask}, advances that have significantly boosted the performance of numerous applications.
However, recent years have shown clear diminishing returns; SoTA architectures are just too complex to invent by hand.

Now, the automation of neural architecture design is not new~\cite{bayer2009evolving}.
However, early works focused on shallow networks and did not show promising empirical results when compared with manual designs, especially on large-scale benchmarks.
Then, in November of 2016, two concurrent works \cite{zoph2017NAS,baker2017designing} were publicly released, showing for the first time that automatically designed DNNs could be competitive with -- if not better than -- manually designed SoTA architectures.
It was at that time that the term ``Neural Architecture Search'' was proposed~\cite{zoph2017NAS}, referring to AutoDL algorithms specifically engineered to search for neural architectures.
Since then, NAS has been cemented as a core element of AutoDL.

In NAS, researchers mainly focus on three parts:
(1) the search space, containing all the possible candidate neural architectures that can be chosen,
(2) the search strategy, defining how to find a good candidate architecture from the search space, and
(3) the evaluation of the candidate architecture, which generates a performance metric to guide the search strategy.
Accordingly, these three aspects form the basis of comparison for 25+ NAS algorithms in \Tabref{table:autodl-dissect}, and the rest of this section reviews these concepts in greater depth.

\subsection{Search Space}\label{sec:nas-search-space}

\begin{figure}[t!]
  \begin{center}
    \includegraphics[width=\linewidth]{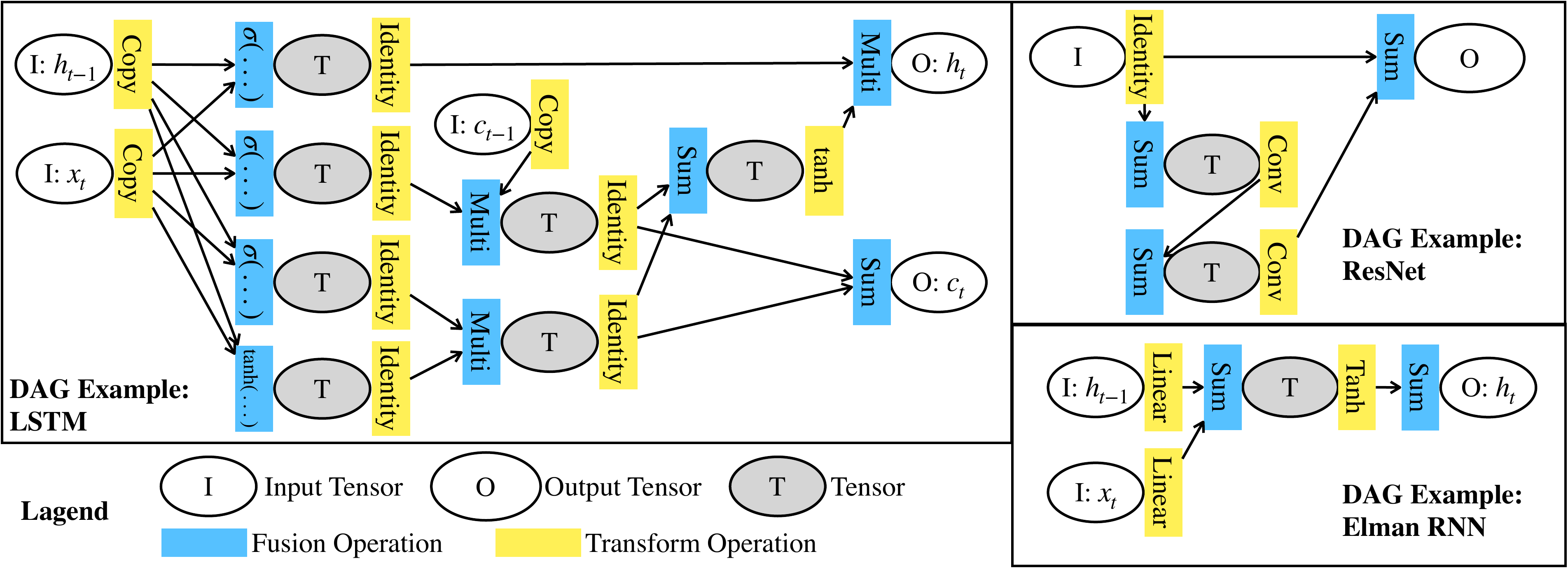}
  \end{center}
\caption{
Illustrative representations of three popular neural architectures in directed acyclic graph (DAG) format.
In each DAG, a node (circle) represents a feature tensor and an edge (arrow) indicates tensor flow. The blue block ahead of each tensor denotes a fusion operation to combine incoming flows, while the yellow block behind each tensor denotes a transform operation applied to the outgoing flow.
}
\label{fig:nas-dag}
\end{figure}

As shown in \Figref{fig:nas-dag}, NAS in the context of DNNs is essentially a search through possible DAG topologies, with variations in both connectivity and transformative operators.
The set of these possibilities is called a search space, and different NAS algorithms constrain this set in different ways according to both expert knowledge and domain characteristics.

\textbf{Size-related Search Spaces:}
As stressed before, while NAS has been popularized within the last handful of years, this is not to imply that no previous work has ever tried to optimize the structure of a neural network~\cite{snoek2012practical,thornton2013auto}.
Most attempts, however, have attacked the problem in a relatively rudimentary manner, focused primarily on network depth and layer width for feed-forward networks without substantially perturbing their topology.
In more recent phases of the DL era, efforts to automatically control the size of modern architectures were initially associated with the concept of structured network pruning, complete with learnable pruning ratios and dynamic networks~\cite{dong2017more,figurnov2017spatially}.
These approaches would decide on, for instance, the number of channels to use per layer, as well as other depth values.
Naturally, such work would eventually be recontextualized as part of NAS~\cite{frankle2018lottery,dong2019tas,cai2020once}, yet, despite this unification, the research thread still has its own priorities, e.g., how to preserve model performance after pruning or how to adjust DNN weights as part of a dynamic inference procedure~\cite{han2021dynamic}.

\textbf{Convolutional Search Spaces:}
The eponymous ``NAS'' work designed a search space for a convolution neural network (CNN) by allowing the kernel height, the kernel width, the number of kernels, and the connectivity between different layers to be searchable, while fixing the depth~\cite{zoph2017NAS}.
Concurrently, MetaQNN explored a different convolution search space~\cite{baker2017designing}.
The MetaQNN search space makes the depth searchable, allows the layer type to be selected from a \{convolution, pooling, fully connected\} set, and includes the hyperparameters associated with each layer type, such as the number of kernels for convolution.
The automatically discovered architectures within these two search spaces achieved competitive results on CIFAR compared to the popular ResNet and DenseNet~\cite{huang2017densely}.
In fact the high performance-to-manual-effort ratios of these works can be credited with the initial pull of researchers to the field of NAS.

However, in the wake of these seminal NAS works~\cite{zoph2017NAS,baker2017designing}, which proved more accurate than several popular deep CNNs on tested benchmarks, it was quickly realized that architectural search space could rapidly explode beyond feasible use.
A pragmatic philosophy arose, founded in the notion of transferability.
In essence, it asked: can good architectures be built by stacking together reusable cells/blocks, each larger than a single layer?
The NASNet algorithm was one of the first to use this approach, working with a cell-decomposable high-performance network tuned to CIFAR~\cite{zoph2018learning}.
Specifically, it hypothesized that stacking more cells on this network would make it adept at dealing with the larger and higher-resolution ImageNet, while still leveraging the previously optimized sub-structure of the CIFAR-based architecture.
Each type of reusable cell would be optimized by searching through a set of possible DAG topologies, similar to \Figref{fig:nas-dag}, with 13 possible options of basic pre-defined transformations for the operators.
Subsequent works have since improved the NASNet search space by removing useless operators~\cite{Liu_2018_ECCV,liu2019darts,pham2018efficient} or relaxing topological constraints~\cite{ying2019bench,dong2020bench}; we refer to these variants as NASNet-like search spaces.

The concept of reusability forces a dramatic reduction of a search space, although there remains some debate whether this constraint on DNN solutions is overly limiting~\cite{fang2020densely}.
For now, though, the consensus view is to continue minimizing computational costs.
In fact, even NASNet-like search spaces have been considered too bloated, with resulting networks inducing too many flows from input to output tensors, negatively impacting inference speed.
For instance, when comparing an architecture from a NASNet-like search space with ResNet, where the two have a similar count of floating-point operations (FLOPs) and are trained on ImageNet, ResNet has a far superior GPU-based inference speed~\cite{cai2019proxylessnas}.
In an effort to counter this, the MnasNet algorithm has since been proposed with the aim of discovering DNNs for edge devices~\cite{tan2019mnasnet}, such as mobiles, with a search space inspired by the mobile inverted bottleneck convolution (MBConv) used in MobileNet-V2~\cite{sandler2018mobilenetv2}.
The MnasNet search space still carries over the cell-stacking notion of NASNet, allowing both the number of cells and convolution kernels to be optimized, while also searching through different transformation operators.
However, the topology of each cell is fixed as an MBConv structure enhanced by squeeze-and-excitation (SE) principles.
Other algorithms like ProxylessNAS~\cite{cai2019proxylessnas} and FBNet~\cite{wu2019fbnet} have likewise gone down the MBConv route, showing significantly improved inference latency over models produced via NASNet-like search spaces.

Naturally, not all works in this topic align with NASNet-like and MBConv-based approaches.
Some have experimented with representation, e.g., exploring a tree-structured architecture space~\cite{cai2018path}, while others have designed search spaces with certain outputs in mind, e.g., densely connected networks~\cite{fang2020densely}.

% NOTE(xuanyidong): add the transformer-based search space.

\textbf{Other Search Spaces:}
The seminal work that introduced NAS in 2016 simultaneously presented results for both image-based CNNs and text-based recurrent neural networks (RNNs)~\cite{zoph2017NAS}.
Indeed, while RNNs have not been as heavily investigated as CNNs in the realm of NAS, possibly because training topologically complex RNNs to convergence is extremely challenging~\cite{liu2019darts,dong2019search}, they do have a history.
Efforts to optimize the topology of a memory cell in an LSTM network~\cite{hochreiter1997long} have been made as early as in 2009~\cite{bayer2009evolving}.
Arguably, the 2016 work keeps things relatively simple, effectively fixing cell topology beforehand and only searching through the type of operators involved in this cell.
However, variations in topology would later be included in the search space~\cite{pham2018efficient}.
Overall, NAS-discovered RNNs are shown to outperform the vanilla LSTM on some small benchmarks, but evidence is still lacking to compare against SoTA manual designs at the larger scale~\cite{liu2019darts}.
In addition, the complex topological structure of current NAS-generated RNNs makes it difficult to utilize the parallel-computation advantage of modern accelerators.
As a result, the realistic training/inference speed for these architectures is unexpectedly slow~\cite{liu2019darts,dong2019search}; more research is required to make NAS a feasible approach for generating high-performance RNNs.

Of course, not all search spaces are designed purely with CNNs or RNNs in mind.
For instance, researchers have developed a normalization-activation search space~\cite{liu2020evolving} where basic mathematical operations are employed in the architectural DAG, such as addition, multiplication, tanh, sigmoid, sqrt, etc.
As another example, attention-based sequence-to-sequence models have recently been explored~\cite{so2019evolved}, drawing inspiration from both a NASNet-like search space and Transformer architectures~\cite{vaswani2017attention}.
In fact, given the popularity of the Transformer, we are currently witnessing a gradual shift of focus from CNN/RNN-based search spaces to Transformer-based search spaces~\cite{so2021primer,chen2021searching}.

\textbf{Improving Search Space Design:}
It is becoming apparent that different types of problem/network may benefit from NAS searching on different instances of a search space, at least in terms of options available for model topology and transformation operators.
Accordingly, research attention has recently focused on seeking design principles for compact but encompassing search spaces, especially for particular classes of network~\cite{radosavovic2020designing}, e.g., ResNet-like models.
Other attempts to constrain search spaces continue, e.g., via the use of a network generator~\cite{xie2019exploring}.
These generators produce a large but controllable set of DNN candidates to sift through, which is a different approach from grappling with a complex and expansive multi-dimensional search space.
Taking this concept even further~\cite{xie2019exploring,ru2020neural,peng2020pyglove}, researchers could potentially factorize a huge search space by a small number of generators, each covering their own subspace.
Naturally, whether this is a good idea in practice remains to be seen; the NAS problem is then effectively elevated to one of ``generator search''.
If generators only need to be built once -- and once only -- to optimally capture the space for certain classes of problem/network, then this may be an appealing design principle.
However, if this is not the case, then assessing a subdivision of NAS into two levels depends again on how compact but encompassing the effective search space becomes.

Ultimately, it is evident that the automation provided by NAS simply shunts manual design into other processes.
For all its benefits, there is still a substantial reliance on human decisions to craft an effective search space, even if those choices are made at the developer level rather than by the user.
Certainly, several of the works reviewed in this section have worked on automating search space design~\cite{radosavovic2020designing,xie2019exploring,ru2020neural}, but, in practice, hyperparameters need to be selected at whatever level they are shifted to.
The only way to avoid making assumptions is if it is proven that certain search spaces are ideal for certain classes of network/problem, and this verges on the topic of meta-knowledge.
That stated, one can -- with respect to the architectural DAGs exemplified by \Figref{fig:nas-dag} -- consider a human choice to be (1) a topological constraint~\cite{zoph2018learning,liu2019darts,dong2019search} or (2) the inclusion of an advanced operator~\cite{liu2019darts,tan2019mnasnet}.
Thus, full generalization appears to be a prerequisite for maximally automating the NAS process, i.e., loosening all constraints and selecting a very basic set of mathematical operators~\cite{real2020automl,liu2020evolving}.
However, even with modern computational resources, NAS without some degree of human search-space design is currently infeasible.

\subsection{Search Strategy}\label{sec:nas-search-strategy}

Once a set of possible architectures is defined by a searchable and potentially complex space, it is up to a search strategy to explore this space efficiently and locate an optimal architecture.
Given that each candidate network can be evaluated for performance, typically accuracy, any black-box optimization method can be used as a search strategy in NAS, such as RL, an evolutionary algorithm, BayesOpt, etc.
Notably, because AutoDL as a field evolved organically to focus on neural architecture before optimizing hyperparameters more broadly, contrasting the developmental flow of general AutoML, \Secref{sec:hpo} is a more appropriate place to discuss the details of optimization methods.
Here, we mainly focus on how they are tailored for NAS.

\textbf{RL-based NAS} methods typically encode a neural architecture as a series of variables, which can be interpreted as building instructions for the model.
For instance, these variables may index existing node inputs within a DAG, exemplified in \Figref{fig:nas-dag}, as well as the types/configurations of operators to attach to the current network.
It is then up to a component called a ``controller'' to select candidate encodings within this space and have the performance of their represented models tested, with good outcomes guiding the controller in its continued exploration.

Originally, the eponymous NAS algorithm utilized an LSTM network as this controller~\cite{zoph2017NAS}, thus ascribing a sequential nature to the encodings; each sequence would represent the way to progressively grow out a network.
More recent efforts have instead simplified the controller to work with a collection of \textit{independent} multinomial variables that represent distributions over the transformative operators available to a candidate DNN~\cite{bender2020can,dong2021autohas}.
So, whereas an LSTM is one predictor that progressively predicts an optimal encoding, the simplified controller can be considered as a set of parallel predictors, each one responsible for one variable in the encoding.
Thus, the upgraded controller can immediately construct an architecture without building in sequence, but its effectiveness on broader search space still needs investigation.

Importantly, training the controller is orthogonal to both its design and that of the encodings it searches through.
For instance, Proximal Policy Optimization (PPO)~\cite{schulman2017proximal} is utilized for NASNet~\cite{zoph2017NAS,zoph2018learning}, REINFORCE~\cite{williams1992simple} is applied by both ENAS~\cite{pham2018efficient} and its subsequent works~\cite{bender2020can,dong2021autohas}, and Q-learning~\cite{watkins1989learning} is employed by MetaQNN~\cite{baker2017designing}.
To date, simple RL methods like REINFORCE appear sufficient for the purposes of NAS~\cite{bender2020can} in popular NASNet-like and MBConv-based settings, although the more challenging search spaces discussed in \Secref{sec:nas-search-space} may eventually require more advanced RL algorithms.

\textbf{Evolutionary NAS} methods deal with the same encoding issues -- the definition of architectural genotype -- that RL-based strategies do, but they otherwise employ standard procedures for evolving a population of networks into fit-for-purpose models.
Most of the time, researchers only vary two specific aspects: the mutation strategy for an individual architecture and the evolution process for the whole population~\cite{real2017large,real2019regularized,so2019evolved,jozefowicz2015empirical,so2021primer}.
Given the DAG representation in \Figref{fig:nas-dag}, mutation typically involves adding/removing edges and nodes, changing transformation operators, or merging two graphs.
Evolution processes can be more varied, although pairwise competition, a form of tournament selection~\cite{goldberg1991comparative}, has currently received a lot of attention~\cite{real2017large}.
A couple of works have employed this mechanism, although with an additional age-based mechanism to prioritize younger individuals, i.e., candidate networks more recently added to a population~\cite{real2019regularized,so2019evolved}.

A fairly common issue with applying evolution to NAS and other AutoDL sub-areas is the isomorphism of architectures.
Two candidate networks with different DAG representations may be mathematically identical to each other.
For example, the NATS-Bench topology search space~\cite{dong2021nats,dong2020bench} has 15K unique DAG encodings, yet there are only 6.5K unique architectures among them.
Unsurprisingly, the isomorphism of architectures severely impacts search effectiveness and consequently causes a significant waste of computational resources.
To combat this issue and identify/avoid isomorphisms, researchers have explored graph hashing~\cite{ying2019bench,ying2019enumerating}, customized DAG-to-string algorithms~\cite{dong2020bench}, and conjugate matrix ensembles~\cite{suzen2020equivalence}. Among them, hashing seems to be a particularly promising approach for helping AutoDL algorithms avoid duplicated computation~\cite{real2020automl}.

\textbf{BayesOpt-based NAS} methods are based on, as the name suggests, a Bayesian formalism~\cite{snoek2012practical,kandasamy2018neural,zhou2019bayesnas,white2021bananas}, inheriting the strengths of the strategy applied to HPO in the context of AutoML.
In fact, because BayesOpt was uniquely promoted into the NAS context from a DL-external community, it is more appropriate to elaborate on the approach when discussing HPO in \Secref{sec:hpo-black-box}.
However, with the understanding that BayesOpt relies on an iteratively updated prior probability distribution to estimate DL model performance, there have been a few adjustments to the approach when extended from HPO to NAS.
These have come in the form of succinctly encoding a candidate architecture as an input to the prior, as well as occasionally using a high-level neural network for the prior itself.
Additionally, given the cost of evaluating a candidate DNN at each iteration, custom proxy training recipes are often employed; methods for boosting candidate evaluation are discussed in \Secref{sec:nas-boost-eval}.

\textbf{Differentiable NAS} methods distinguish themselves from the aforementioned approaches by rejecting the limitation of a discrete and non-differentiable search space.
The DARTS algorithm is seminal in this research area~\cite{liu2019darts}, seeking a relaxation into a \textit{continuous} search space, so as to efficiently search architectures using optimization algorithms based on gradient descent.
In a way, the associated continuous encodings resemble fuzzy sets or quantum superpositions, and they are eventually defuzzified/collapsed back into a discrete representation.

Notably, while DARTS popularized differentiable optimization in NAS, many issues in the original work were left unaddressed, such as the accuracy of gradients with respect to architectural encoding,
the inconsistency between continuous and discrete search spaces, the implicit assumption behind linearly weighted sums for evaluating superposed networks, the bias of operator weights, etc.
These issues have gradually been addressed by subsequent investigations~\cite{zela2020understanding,xu2020pc,dong2019search,dong2019tas,xie2019snas}.
For example, sophisticated differentiable HPO methods have been used to more accurately calculate gradients with respect to architecture~\cite{lorraine2020optimizing}.
A relationship has also been established between the performance of a DARTS-produced architecture and the eigenvalues for the Hessian matrix of validation loss, again calculated with respect to architectural encoding~\cite{zela2020understanding}; this was used as a regularization factor in early-stopping a search.
Elsewhere, the Gumbel-softmax distribution~\cite{jang2017categorical} has been applied when discretizing architectural encodings~\cite{xie2019snas}, so as to alleviate the inconsistency between continuous and discrete search space.

Differentiable NAS remains popular, primarily due to three reasons: (1) it requires significantly decreased computational resources over alternative approaches; (2) the codebase for DARTS is open-source and easy to use; (3) DARTS is easily extendable.
However, it does have two main drawbacks.
First, the accuracy of architectures discovered by differentiable NAS is worse than those found via the RL-based procedures and evolutionary methods that currently claim the SoTA label for NAS~\cite{real2019regularized,tan2021efficientnetv2}.
Second, the most appropriate representation/evaluation of superposed candidate networks in a continuous search space is still an open question.

\subsection{Efficient Candidate Evaluation}\label{sec:nas-boost-eval}

Regardless of the black-box search strategy employed by a NAS algorithm, evaluating the accuracy of an architecture requires fully training it from scratch to convergence.
This may cost several GPU days for a modern DNN on a large-scale dataset~\cite{he2016deep,real2019regularized}.
Hence, it is computationally expensive to train/test even a single architecture, let alone the thousands of search-space samplings that may be needed to find an optimum, local or otherwise.

A straightforward and intuitive way to counter this cost is via a \textbf{low-fidelity approximation}, which is usually designed heuristically.
As shown in \Tabref{table:autodl-dissect}, many works scale down the model~\cite{zoph2017NAS,liu2019darts}, sub-sample the dataset~\cite{dong2019search}, reduce the number of training epochs~\cite{zoph2017NAS,zoph2018learning}, set a constrained time budget~\cite{loshchilov2016cma,falkner2018bohb}, early-stop the training~\cite{baker2017designing}, or explore different generalizable metrics~\cite{suzen2019periodic}.
These proxy strategies will decrease model accuracy as compared with a full training.
However, by assuming models maintain proportionality in their relative performance, a comparative ranking of architectures can be estimated.
Of course, the validity of low-fidelity approximations depends on whether this assumption holds true.

An alternative to low-fidelity approximation is the use of a \textbf{neural performance predictor}~\cite{wen2020neural,dai2021fbnetv3}.
This is a regression model that operates on architectural encodings~\cite{wen2020neural,dai2021fbnetv3}, a learning
curve~\cite{domhan2015speeding,chandrashekaran2017speeding}, or both~\cite{baker2018accelerating}, to predict performance, e.g.~final validation accuracy or latency~\cite{cai2019proxylessnas}.
The regression model can be non-DL-based~\cite{baker2018accelerating}, a simple MLP~\cite{dai2021fbnetv3}, or even something as advanced as a graph convolutional network~\cite{wen2020neural}.
After this predictor is optimized, it can be employed to boost the evaluation of candidate architectures in many ways, such as by replacing a low-fidelity approximation strategy~\cite{wen2020neural} or augmenting it via early-stopping~\cite{baker2018accelerating}.
Naturally, training a regression model can still take numerous network evaluations, which is a nontrivial computational cost.
Thus, these approaches are usually applied for NAS benchmarks~\cite{ying2019bench,dong2020bench,siems2020bench,dong2021nats} and are still considered expensive for large-scale datasets.

\textbf{Weight generation and weight inheritance} are more efficient solutions for NAS.
The SMASH algorithm~\cite{brock2018smash} exemplifies the former approach by training a hypernetwork~\cite{ha2017hypernetworks,schmidhuber1992learning} simultaneously with the search process, with the description of an architecture as input and its tuned weights as the target.
The idea here is that a model generator in a NAS process should eventually, with support from the hypernetwork, be able to immediately flesh out any candidate architecture with its ``correct'' weights, no training required.
Its effectiveness is only evaluated on a small-scale search space. When the search space becomes more complex and larger, training such hypernetwork would become evidently hard.
Weight inheritance provides an alternative shortcut, where, for NAS procedures that employ mutations, a new candidate network can adopt some of the weights from a former candidate~\cite{real2017large,yang2018netadapt,cai2018path}.
This strategy does not do away with retraining entirely, but there is an obvious speedup due to fewer model parameters that need to be tuned.
In fact, weight inheritance can be taken to a further extreme in the notion of weight sharing~\cite{pham2018efficient}, where all assessed candidates are sub-networks of -- and thus share their weights with -- a single giant ``super-network''.
In this way, the cost of training millions of candidate architectures is amortized.

\textbf{Weight sharing} in particular has attracted much attention due to its simplicity and efficiency~\cite{pham2018efficient}.
As shown in \Tabref{table:autodl-dissect}, it has become standard for efficient NAS.
However, unsurprisingly, the efficiency of weight sharing has a drawback: search effectiveness~\cite{dong2020bench,dong2021nats,zela2020bench}.
As the weights of the super-network are tuned for itself, there is no guarantee that any candidate model among its immense number of sub-networks shares the same optimality of performance.
It is certainly not clear how consistent/predictable performance degradation from weight sharing is, which, in turn, decreases the accuracy of relative rankings for all candidates.
Of course, many works have tried to improve upon this technique by, for example, reducing the correlation between super-network and sub-network via path dropout~\cite{bender2018understanding}, stabilizing model training via computing batch statistics on the fly~\cite{bender2018understanding}, accelerating super-network training in differentiable NAS via the use of a straight-through Gumbel-softmax estimator~\cite{dong2019search}, alleviating co-adaptation of shared weights via uniform sampling~\cite{guo2020single,dong2019one}, reducing inconsistent statistics between different candidates via switchable normalization~\cite{yu2019slimmable}, etc.
While these strategies have improved the empirical performance of weight sharing, the issues mentioned above are still unsolved and lack theoretical analysis.
Nonetheless, on the balance of efficiency versus effectiveness, weight sharing remains popular.

Indeed, for as long as DL wrestles with a paucity of computational resources, NAS research into shortcuts for both training and evaluating a candidate is expected to continue, and the most aggressive forms of evaluation are ones that \textbf{avoid training entirely}.
For instance, appearing on arXiv in June 2020~\cite{mellor2021neural}, a publication proposed that, for any architecture, a model quality score could be computed by analyzing the activation map between different batch samples of data.
This method showed competitive performance on both NAS-Bench-101~\cite{ying2019bench} and NAS-Bench-201~\cite{dong2020bench}, but the results on the large-scale dataset are missing.
Similarly, five metrics borrowed from the network pruning community have also been explored as potential performance estimators~\cite{abdelfattah2021zero}, with the analysis on more NAS benchmarks.
Elsewhere, two metrics, neural tangent kernel~\cite{jacot2018neural} and the number of linear regions for a CNN~\cite{xiong2020number}, have both been considered for ranking architectures, likewise without training~\cite{chen2021neural}.

Ultimately, it is currently difficult to assess the right balance between search efficiency and search effectiveness.
Some recent investigations report that applying NAS without wasting time on training weights has produced comparable -- sometimes even better -- DL models than previous SoTA NAS techniques~\cite{chen2021neural}.
But a general lack of comprehensive and diverse benchmarking in the field means that such a contentious debate is unlikely to be settled soon.

\begin{table}[tp!]
\begin{center}\scriptsize
\setlength{\tabcolsep}{2pt}
\caption{
Evaluative assessment for trends in NAS.
}
    \begin{tabular}{l | l | l | l | l | l | l | l | l | l | l | l}
    \toprule
      &                       &   Novelty     & Solution & Effic. & Stability & Interp. & Reprod.  & Engi. & Scalability   & General. & Eco. \\
\midrule
\multirow{4}{*}{\rotatebox[origin=c]{90}{\makecell{Search\\Space}}}
      & Size-related          &  $\approx$20  & Medium   & Medium & High      & ?     & \cmark   &  \cmark  & High      &  High & ?  \\
      & Conv: NASNet-like     &  $\approx$4   & High     & Low    & Medium    & ?     & \cmark   &  ?  & Medium    &  High & ?  \\
      & Conv: MBConv-based    &  $\approx$3   & High     & Mixed  & High      & ?     & ?        &  \cmark  & High      &  High & ?  \\
      & Others                &  $\approx$10  & Mixed    & Mixed  & Mixed     & ?     & ?        &  ?  & Mixed     &  Low  & ?  \\
\midrule
\multirow{4}{*}{\rotatebox[origin=c]{90}{\makecell{Search\\Strategy}}}
      & RL-based              &  $\approx$5   & High        & Mixed    &  ?      & Medium  & \cmark    &  \cmark & High   & High & ?      \\
      & Evolutionary          &  $\approx$10  & High        & Mixed    &  ?      & Medium  & \cmark    &  \cmark & High   & High & Mixed      \\
      & BayesOpt-based        &  $\approx$9   & High        & Mixed    &  ?      & Medium  & \cmark    &  \cmark & Medium & High & ?      \\
      & Differentiable        &  $\approx$3   & Medium        & High   &  ?      & Medium  & \cmark    &  ?   & High   & High & ?      \\
\midrule
\multirow{5}{*}{\rotatebox[origin=c]{90}{\makecell{Candidate\\Evaluation\\Boosts}}}
      &  Heuristic low-fidelity approx. & $\approx$10 &  High    & Low     & Mixed    & ?    & \cmark &  ?  &  Low    & High   & ?\\
      &  Neural predictor               & $\approx$6  &  Medium  & Low     & Mixed    & ?    & \cmark &  ?  &  Low    & High   & ?\\
      &  Weight generation/inheritance  & $\approx$6  &  Medium  & Medium  & Medium   & ?    & ? &  ?  &  Medium & Medium & ?\\
      &  Weight sharing                 & $\approx$3  &  Medium  & Medium  & Low      & ?    & ? &  ?  &  High   & Medium & ?\\
      &  NAS without training           & $\approx$1  &  Low     & High    & ?        & ?    & ? &  ?  &  Mixed  & ?      & ?\\
    \bottomrule
    \end{tabular}\label{table:nas-methods}
\end{center}
{\scriptsize\textit{
Each row marks an emergent trend in AutoDL, specifically NAS. Each column marks a criterion -- see \Secref{sec:criteria-intro} -- by which the trend is assessed. The evaluations are mostly qualitative, averaged across the most significant works researching the trend. Where a graded value is not provided, ``\cmark'' indicates a rigorous assessment is possible with analysis beyond the scope of this review, while ``?'' indicates that not even this is achievable without more research works to analyze. Novelty denotes years since seminal works in DL were published. Abbreviations are: ``Solution'' for Solution Quality, ``Effic.'' for Efficiency, ``Interp.'' for Interpretability, ``Reprod.'' for Reproducibility, ``Engi.'' for Engineering Quality, ``General.'' for Generalizability, and ``Eco.'' for Eco-friendliness. 
}}
\end{table}

\subsection{Overview}

% Start with search space; acknowledge 'others' and use size-related as baseline.
To assess the state of NAS research at the current time, consider first the search-space approaches evaluated in \Tabref{table:nas-methods}.
As is evident, there have been numerous exotic propositions in this area and, overarchingly, it is hard to conclude anything about them beyond the fact that they rarely generalize well, often being customized for specific applications and downstream tasks.
However, it is a little more straightforward to evaluate the mainstream approaches.
Size-related search spaces make for a good baseline in this comparison, as they are simple and stably optimized, thus being easily engineered, reproduced, scaled up, and applied wherever.
However, they do not lend themselves to particularly efficient searches, and their limitations preclude the discovery of high-accuracy neural networks.

% Compare NASNet-like.
For problems well handled by CNNs, NASNet-like search spaces were the first in the modern era of AutoDL to be employed.
Their main selling point is accuracy, enabling novel SoTA topologies for DNNs to be found.
On the other hand, as is to be expected for new experimental forays, this has come with a set of trade-offs that have not been entirely managed yet.
Traversing these complex spaces remains relatively slow and inefficient, and the strong coupling between sequential choices in the search process makes it difficult for the algorithm to stably converge and scale up.
There are also no popular libraries available that are suited to handling the graph representation required by NASNet-like search spaces, making it difficult to comment on engineering quality, although there has been interest in the trend for sufficiently long enough that research results have been benchmarked to a degree~\cite{ying2019bench,dong2020bench,siems2020bench,NAS-Bench-x11}.

% Compare MBConv-based.
Search spaces that are MBConv-based are a slightly more recent proposal, partially reacting to the inefficiencies present with NASNet-like spaces.
Arguably, their effectiveness still relies on secondary factors, such as whether MBConv-based searches leverage weight-sharing or rely on multiple trials.
Nonetheless, the representation of a candidate architecture in an MBConv-based space is not as complicated or tightly coupled, meaning that, with configurations expressed in list format, it is relatively easy to slot in well-engineered NAS/HPO libraries.
This ease of network representation also pays dividends in terms of scalability and the stability of optimization convergence.
However, research incorporating MBConv-based search spaces has still not been benchmarked heavily, and -- this is an issue across the board -- there have been no strong theoretical analyses of search spaces, meaning that it is not perfectly clear why certain arrangements produce better DL models than others.

% Move on to search strategies; talk accuracy and efficiency.
The onus of attaining good DL performance ultimately falls on search strategies though, and all four trends in \Tabref{table:nas-methods} are ranked as strong contributors to model accuracy.
Certainly, they are far more robust than manual searches, although the debate about random search and its own effectiveness remains~\cite{li2020random}.
However, reflecting the challenges of training complex DNNs, the efficiencies of most are rated as mixed, strongly dependent on candidate evaluation boosts.
In fact, applied in standard fashion or even with low-fidelity approximations, these search strategies are considered slow and resource-intensive in the literature, with only weight-sharing and related techniques managing to increase the efficiency score.
The one relatively speedy exception is differentiable NAS, which tries to relax search spaces so that gradient descent is a valid procedure; this experimental technique results in model-accuracy claims that are slightly less impressive, although these may improve as the youngest trend acquires more research focus.

% Discuss the other criteria for search strategies.
Overall, search strategies in NAS are often treated as generalizable black-box optimization methods, i.e., very high-level, that may as well have coincidentally been employed in the search for accurate DNNs.
They are thus subject to standard limitations that do not depend on the NAS context, such as BayesOpt struggling to scale well in search spaces with extremely high dimensions.
Moreover, there are also many good software libraries available for the four approaches, with the possible exception of differentiable NAS.
This general applicability also means that the strategies are regularly benchmarked, opening themselves up to reproducibility assessments, and, likewise, a moderate level of theoretical analysis exists.
The assertions made by these investigations, however, are limited by a reliance on conditions that may not hold in practice, especially once approximations and proxies are included in NAS.
To date, they also have little to say about the convergence stability of the studied search methods, at least with respect to NAS.

% Move on to efficiency boosts, anti-correlating accuracy and efficiency.
Moving down to a lower level, much more variation exists when assessing the diversity of options for boosting efficiencies in candidate evaluation.
As \Tabref{table:nas-methods} suggests, the passage of time has witnessed progressively more dramatic attempts to speed DL up, from low-fidelity approximations to the recent NAS-without-training proposal.
The assumptions underlying each new trend become seemingly looser and more radical, which lessens accuracy guarantees on the final DNN that NAS produces.
Conversely, as intended, the speed of NAS is substantially accelerated, especially after the introduction of weight generation, where candidate evaluations are no longer necessitated.

% Discuss the other criteria for efficiency boosts.
Given the number of years over which low-fidelity approximations and neural-predictor estimations have been researched in the context of DL, there already exists some commentary on the results and reproducibility of both the former~\cite{eggensperger2021hpobench} and the latter~\cite{NAS-Bench-x11}.
The stability of both methods is very dependent on multiple factors, e.g., the proxy task used, and they scale linearly with respect to the size of a search space.
They are also usually agnostic to architectural design and are thus very generally applicable.
In contrast, weight generation and sharing need customization for specific architectures and search spaces, with low guarantees on stable convergence.
However, the assumptions involved mean that, while the approaches do still need extra evaluations as a search space increases to provide optimal benefit, the scaling relation is attractively sublinear.

% Discuss eco-friendliness.
In terms of eco-friendliness, first impressions suggest that NAS has quite a problem~\cite{strubell2019energy}, and, certainly, minimizing power usage remains a significant challenge.
However, it remains contentious within the DL community as to how dire the impact truly is.
As expected, intelligent use of efficiency boosts can substantially reduce carbon emissions when searching through a large space for an optimal architecture~\cite{patterson2021carbon}.
The debate has been further confounded by an assertion in early research work~\cite{strubell2019energy} that NAS algorithms must be run from scratch every time a model is freshly trained.
In contrast, other publications~\cite{patterson2021carbon,li2021full} argue that, in practice, NAS is only ever performed once per combination of problem domain and architectural search space.
They also claim that other discounts in carbon emissions are often overlooked, especially when considering the billions of inferences that could have otherwise been made with an inefficient manually constructed DNN.
In truth, this debate is unlikely to be settled without many more extensive investigations.

% Highlight the question marks.
For now, throughout \Tabref{table:nas-methods}, a lot of unknowns remain.
The more novel a trend is, the less it has been rigorously studied.
Indeed, it is challenging to comment on reproducibility for weight generation and sharing due to a dearth of benchmarking.
In the case of NAS without training, which is effectively a brand new proposition, the stability and generalizability of the approach are entirely uncertain at the current time.
Nonetheless, even accounting for the passage of time, broader gaps in theory and practice remain.
For instance, how exactly these techniques relate to performance is not presently interpretable, beyond the common-sense understanding that making shortcut assumptions results in increased speed.
Likewise, the methods have not been implemented yet in any notable libraries.
So, for all the attention that NAS has received from the DL community, there is plenty more progress to make, even within this core aspect of AutoDL.

\section{Hyperparameter Optimization}\label{sec:hpo}

The definition of a hyperparameter is blurred in the literature.
In the broadest sense, they are traditionally human-chosen parameters -- model-based and algorithm-based -- that control a process of learning; they are not determined by that learning process.
However, for various historical reasons, the full implication of this definition has often gone unrecognized.
This is why the decision to treat ML model type as a hyperparameter was itself considered a surprising innovation of AutoML, allowing model selection to be repackaged into a broader CASH problem~\cite{thornton2013auto}.

Now, granted, some DL researchers have considered architectural structure to be a set of hyperparameters~\cite{dikov2019bayesian}, but early-NAS effectively developed without strong awareness of the AutoML community.
This siloed approach would eventually be rebuked, with a publication stating that, ``while the NAS literature casts the architecture search problem as very different from hyperparameter optimization, ... most NAS search spaces can be written as hyperparameter optimization search spaces''~\cite{zela2018towards}.
The paper would go on to challenge the then-predominant approach of running NAS first and then optimizing (the remaining) hyperparameters as an independent post-hoc step.
However, this distinction between two processes has stuck, and it is currently a convention in AutoDL that the word `hyperparameter' often relates solely to the training algorithm~\cite{dong2021autohas}, e.g., representing learning rate, weight decay, dropout rate, etc.
We maintain this convention in this review.

Within such a context, HPO in AutoDL has certain unique differences from HPO in AutoML, and they are not purely semantic.
For one thing, AutoML works with a diversity of model types and training algorithms that discourages an optimizer from making assumptions ahead of time.
In contrast, because all DNNs are based on the same universal approximator in the form of an artificial neuron, training algorithms are fewer in number; it is possible to have a favorite choice without neglecting outright better performers.
It follows that, while the type of training algorithm can still be made searchable~\cite{dong2021autohas,dai2021fbnetv3}, a parameter like learning rate may be more efficiently optimized~\cite{baydin2018online} if the selected training algorithm is known to be SGD~\cite{rumelhart1985learning} or Adam~\cite{kingma2015adam}.
We thus classify AutoDL HPO algorithms by how much they need to know about the training algorithm applied to a base model: black-box, gray-box, and white-box.

\subsection{Black-box HPO Approaches}\label{sec:hpo-black-box}

Black-box HPO approaches have a long-standing history \cite{snoek2012practical,conn2009introduction}.
In the context of HPO in AutoDL, they assume that a training procedure defined by a candidate set of hyperparameters can only be evaluated by the end result of the process, e.g., the accuracy of a DNN that is trained.
Thus, they are exceedingly general; the underlying search techniques have presence in AutoML, and they can be directly applied to other optimization problems in AutoDL, including Automated Data Engineering, NAS, and Automated Deployment.
Broadly speaking, there are three popular categories in AutoDL-based HPO: RL~\cite{schulman2017proximal}, evolutionary approaches~\cite{goldberg1991comparative}, and BayesOpt~\cite{snoek2012practical}.

\textbf{Reinforcement Learning:}
As with NAS in \Secref{sec:nas-search-strategy}, RL-based HPO can be enacted by designing/extending a controller to sample candidate hyperparameters.
Every time these hyperparameters are selected, a corresponding training procedure is instantiated to train a DL model, which may itself have been selected by a controller.
Evaluation metrics, such as accuracy on a validation set, are used to judge both the candidate DNN and training procedure in tandem.
These metrics represent a reward, and RL algorithms work to maximize this reward~\cite{ali2019meta,schulman2017proximal,williams1992simple,watkins1989learning}, thus teaching the controller to sample better candidates in the future, whether or not architectural choices are rolled into that search space.

An open challenge for RL-based HPO is how best to reformulate hyperparameter search into an RL problem.
This optimization has previously been treated as a sequential selection of hyperparameters~\cite{jomaa2019hyp} but, recently, it has also been simplified into a single-step decision-making problem~\cite{dong2021autohas}.
There are many design choices to be made with RL strategies, and broader discussions can be found elsewhere~\cite{sutton2018reinforcement}.

\textbf{Evolution:}
To apply evolutionary algorithms~\cite{goldberg1991comparative,de2016evolutionary,salimans2017evolution,schmidhuber1987evolutionary}, a population is created by randomly sampling candidates from a hyperparameter search space.
In this population, each candidate is represented as an encoding, treated analogously to a string of Deoxyribonucleic Acid (DNA).
Typical evolutionary algorithms will iteratively (1) alter candidate encodings within the population, (2) train and evaluate a DL model subject to each set of candidate hyperparameters, obtaining validation accuracy as a fitness metric, and (3) remove low-fitness candidates from the population, replacing them with higher-fitness encodings~\cite{goldberg1991comparative}.
In this way, the population is progressively improved and, at some stopping point, the highest-fitness candidate is selected as an optimal set of hyperparameters.
Of course, there are many ways to alter/replace encodings, e.g., via mutation or crossover, so there are many variants of evolutionary algorithms in existence~\cite{real2017large}.

\textbf{BayesOpt:}
Despite the sophistication of RL and evolution-based techniques, random search is also a common option, and it can be surprisingly effective in practice~\cite{bergstra2012random,li2020random,yu2020evaluating}.
In general, though, it is assumed that principled search methods can navigate to optima more efficiently.
With all the potential ``messiness'' of hyperparameter space, from discontinuities to conditional variables, BayesOpt methods have proven particularly popular and effective~\cite{bergstra2011algorithms,snoek2012practical,jenatton2017bayesian,zela2018towards}.
These consist of two components: a Bayesian-based surrogate model for estimating how a candidate set of hyperparameters maps to a performance metric, based on evaluations already made, as well as an acquisition function that decides where to sample next, so as to iteratively rein in the performance estimates of the surrogate.

While RL and evolutionary algorithms have been around for a while, the development of BayesOpt is what propelled AutoML into a broader spotlight within the early 2010s~\cite{hutter2009paramils,hutter2011sequential}.
However, all three types of techniques have representation in AutoDL.
For now, it remains an open question as to whether one approach is better than another, and in which problem settings.
Some preliminary works have benchmarked different HPO algorithms on small datasets and search spaces~\cite{eggensperger2014surrogate,klein2019tabular}, but more investigation is required to generalize these conclusions to large-scale scenarios, especially to bolster confidence in any comparative rankings.

\subsection{Gray-box HPO Approaches}\label{sec:hpo-gray-box}

Black-box optimization is flexible, but if one can be confident in assumptions/knowledge about what lies ``inside the box'', it is often possible to search through a space of solutions far more efficiently.
This is often unofficially referred to as \textit{gray-box optimization}.
By definition, its applicability is very dependent on the search problem of interest, and associated methods are often just upgraded forms of the generic black-box techniques described in \Secref{sec:hpo-black-box}.

Within HPO specifically, multi-fidelity optimization is among the most popular gray-box approaches~\cite{forrester2007multi,kandasamy2016gaussian,jamieson2016non,karnin2013almost,li2018hyperband,falkner2018bohb}, where variably cheap and accurate proxies/estimates of model performance are leveraged to aid the search.
For instance, if one is able to train models on small amounts of data, i.e., low-fidelity approximations, it is possible to quickly extrapolate these performances into a full learning curve~\cite{domhan2015speeding,chandrashekaran2017speeding,swersky2014freeze}.
This predictive curve can provide advice on many matters, e.g., whether to continue training, whether to add more computational resources, or whether to `early-stop' an unpromising set of hyperparameters.
The strategy of successive halving is another technique that similarly starts with low-fidelity approximations~\cite{karnin2013almost}.
It evaluates candidates trained on minimal data/time, throws away the worst half, evaluates the remnants on an increasing amount of data and computational budget, throws away the worst half, and so on; eventually one high-fidelity evaluation is left.
This process has since been refined into an algorithm named Hyperband by hedging its aggressiveness~\cite{li2018hyperband}, and Hyperband has subsequently been fused with BayesOpt techniques into BOHB~\cite{falkner2018bohb}, which has proven itself a highly efficient and effective HPO strategy for certain ``well-behaved'' datasets~\cite{yang2020hyperparameter,zela2020bench,dong2020bench,klein2019tabular}.
Why then are these approaches considered gray-box?
Their performance depends on the extrapolation of low-fidelity approximations to be predictable and well-behaved, which benefits from some understanding of -- or confidence in -- hyperparameter space.

There are other gray-box search approaches that likewise make assumptions on the behavior of DNN training algorithms when hyperparameters are varied.
For instance, certain shortcuts can be made if a DNN is assumed to be trained via gradient-based means~\cite{jaderberg2017population}.
Elsewhere, there has been a study of what happens when intelligently decomposing a black-box objective into composite functions, one of which is cheap to evaluate~\cite{astudillo2019bayesian}.
Of course, if a hyperparameter space is reasonably familiar or well-understood, HPO methods can also be warm-started with good hyperparameter candidates.
This has been done for both an RL controller~\cite{dong2021autohas} and evolutionary algorithms~\cite{so2019evolved,li2019generalized}.

Further works are listed in the HPO row of \Tabref{table:autodl-dissect}, where the column of ``Boosts for Candidate Evaluation'' indicates the shortcuts being employed; these imply which inside-the-box assumptions about search space/strategy are being made.
Importantly, while gray-box approaches have been very successful in trading off generality for efficiency, they still require some level of sampling, i.e., fully training/evaluating a candidate model, and this computational expenditure is not negligible.
Thus, in practice, black-box and gray-box methods can both be infeasible for large DL models.

\subsection{White-box HPO Approaches}\label{sec:hpo-white-box}

What if we fully open up the black box?
Unlike AutoML, which often juggles many disparate ML models/algorithms, a significant portion of DL involves feed-forward neural networks that all share the same fundamental principles.
Many of these principles relate to the layered nature of a DNN.
Chief among them is that, via the chain rule, one can calculate how a change in any weight parameter corresponds to a change in network performance, i.e., an error gradient.
Indeed, while this notion of backpropagation has been explored in AutoML~\cite{milutinovic2017end}, with parallels between ML pipelines and DNNs discussed in \Secref{sec:automl}, it remains particularly appropriate for DL due to the mathematics involved.

So then, can error gradients with respect to hyperparameters -- hypergradients -- also be computed/leveraged?
After all, if training a model gradually tunes model parameters, then why not hyperparameters too?
Sure enough, researchers have pursued this thread from before the 2000s.
For instance, the gradient of cross-validation error with respect to weight decay has previously been calculated within a simple single-layer network, and this hypergradient was used to adjust weight decay during network training~\cite{larsen2002adaptive}.
Other contemporary work made computing hypergradients somewhat simpler by developing a relation between hyperparameters and network weights, then leveraging this via the implicit function theorem~\cite{bengio2000gradient}.
Since then, hypergradient descent methods have continued to see strong attention, with, for instance, an algorithm being proposed to update hyperparameters by computing reverse-mode derivatives across truncated gradient descent steps~\cite{domke2012generic}.
A subsequent effort would upgrade this approach via the computation of exact hypergradients, additionally wrestling with the substantial memory-based storage costs of the procedure~\cite{maclaurin2015gradient}.

Notably, many early attempts focus on calculating ``exact'' hypergradients, which is computationally expensive for a DL model.
Thus, to improve scalability and generalizability, researchers have recently developed different approaches involving approximate hypergradients.
Some have considered gradually tightening the accuracy of such an approximation during the course of training, i.e., via an exponentially decreasing tolerance sequence~\cite{pedregosa2016hyperparameter}.
Others have analyzed truncated backpropagation for use in approximating the gradients of weight parameters with respect to hyperparameters~\cite{shaban2019truncated}.
Another lingering issue is that hypergradient calculations often rely on the expensive computation of an (inverse) Hessian, i.e., the second-derivatives of model error, which is infeasible for large-scale networks and/or a large number of hyperparameters.
Efforts to surmount this challenge include approximating the Hessian matrix by an identity matrix~\cite{luketina2016scalable} and approximating the inverse Hessian matrix by a Neumann series~\cite{lorraine2020optimizing}.

In summary, it is clear that white-box HPO can be far more efficient than black-box/gray-box HPO; hyperparameter updates can be applied per forward/backward pass during model training rather than after the model is evaluated.
In effect, white-box approaches roll HPO \textit{into} the process of model training.
However, hypergradient methods rely on mathematical equations that embody several assumptions, e.g., the continuity and differentiability of model loss with respect to hyperparameters, and, based on HPO setup, these do not always hold true.

\begin{table}[tp!]
\begin{center}\scriptsize
\setlength{\tabcolsep}{5pt}
\caption{
Evaluative assessment for trends in HPO.
}
\begin{tabular}{l | l | l | l | l | l | l | l | l | l | l}
    \toprule
                                  & Novelty & Solution & Effic. & Stability & Interp. & Reprod.  & Engi. & Scalability   & General. & Eco. \\
\midrule
    Black-box HPO & $\approx$ 30  & High    & Low    & High   & High         & \cmark    & \cmark  & Low  & High  & Low  \\
    Gray-box HPO  & $\approx$ 30  & Medium  & Medium & Medium & Medium       & \cmark    & \cmark  & Low  & High  & Medium  \\
    White-box HPO & $\approx$ 21  & Medium  & High   & Low    & Medium       & ?         & ?    & High & Low   & High  \\
    \bottomrule
\end{tabular}\label{table:hpo-methods}
\end{center}
{\scriptsize\textit{
Each row marks an emergent trend in AutoDL, specifically HPO. Each column marks a criterion -- see \Secref{sec:criteria-intro} -- by which the trend is assessed. The evaluations are mostly qualitative, averaged across the most significant works researching the trend. Where a graded value is not provided, ``\cmark'' indicates a rigorous assessment is possible with analysis beyond the scope of this review, while ``?'' indicates that not even this is achievable without more research works to analyze. Novelty denotes years since seminal works in ML were published. Abbreviations are: ``Solution'' for Solution Quality, ``Effic.'' for Efficiency, ``Interp.'' for Interpretability, ``Reprod.'' for Reproducibility, ``Engi.'' for Engineering Quality, ``General.'' for Generalizability, and ``Eco.'' for Eco-friendliness. 
}}
\end{table}

\subsection{Limitations in Applicability}\label{sec:hpo-gap-theory-practice}

Many aforementioned HPO methods and upgrades are grounded in strong theoretical bases.
None, to date, stand out exclusively among the rest.
This is no surprise, as the no-free-lunch theorems apply to optimizers at any level~\cite{wolpert1997no}.
That does not mean that \textit{certain} sets of hyperparameter spaces do not have an optimal HPO strategy; this has been explored by optimizing a hyperparameter optimizer in the form of an RNN~\cite{andrychowicz2016learning,li2017learning,chen2017learning}\footnote{We have no official stance on whether ``hyperhyperparameter'' should be introduced into the AutoDL lexicon.}.
But the point stands: the applicability of AutoDL-based HPO mechanisms must be carefully considered when choosing one for a real-world problem.

Crucially, this section has shown that HPO methods contend with a trade-off between generality and efficiency.
Any principled strategies beyond purely random search need to leverage some degree of knowledge/assumptions about a search space, and, in return for quicker/better searches, these requirements become more restrictive along the spectrum from black-box to white-box.
Granted, hyperparameter space can already be significantly complex and messy, even with the AutoDL limitation to model training procedures.
For instance, black-box BayesOpt has long grappled with surrogates for dimensions that can be continuous, categorical, or conditional~\cite{hutter2009paramils,hutter2011sequential}; research continues in this area almost a decade later~\cite{ru2020bayesian,daxberger20mixed}.
However, white-box hypergradient-based HPO methods rely on differentiability, and this efficiency extreme can thus only be applied to certain selections of continuous hyperparameters~\cite{pedregosa2016hyperparameter,luketina2016scalable,lorraine2020optimizing}, such as learning rate~\cite{baydin2018online} or continuous regularization~\cite{luketina2016scalable}.

Nonetheless, while the generality-efficiency trade-off will likely always remain, HPO research continues to push the boundary.
For example, population-based training (PBT) proposes to train a group of models together under different sets of hyperparameters, where those individual sets are tuned depending on how the rest of the population is faring~\cite{jaderberg2017population}.
This is a joint optimization of parameters and hyperparameters that does not involve hypergradients, discarding their differentiability restrictions.
It is thus fast, but the computational cost now depends on the scale of parallel training involved.
Elsewhere, a bilevel optimization procedure has introduced so-called ``best-response functions'' as trainable mappings between the values of hyperparameters and corresponding optimal network parameters~\cite{mackay2019self}.
This work likewise avoids hypergradients and their limitations, allowing the training-simultaneous tuning of discrete hyperparameters, data augmentation hyperparameters, and dropout probabilities.
Also of note is another recent effort that aims to maintain general applicability to hyperparameters, encapsulating the procedure of applying hyperparameters to model weights as a black box~\cite{dong2021autohas}.

The take-away from this discussion is that, as with NAS, HPO in AutoDL continues to see a flurry of research, with numerous novel techniques being frequently proposed.
However, \textit{also} as with NAS, HPO in AutoDL is still arguably in a nascent stage.
Systematic benchmarking is limited, making consensus comparisons difficult.
The technical reason behind this is clear, namely the computational expense of running NAS and HPO.
This is why, as summarized in \Tabref{table:autodl-dissect}, existing HPO methods have mainly experimented on small-scale models, e.g., linear models or shallow networks, as well as datasets that are either small or synthetic.
Nonetheless, there have been recent HPO investigations on larger-scale datasets, such as the CIFAR-style AlexNet~\cite{lorraine2020optimizing} and the vision dataset ImageNet~\cite{dai2021fbnetv3,dong2021autohas}.
It is simply a matter of time.
As computational resources increase and the demand for NAS/HPO in real-world applications grows, circumstances will eventually drive more rigorous assessments of applicability.

\subsection{Overview}

% Discuss the accuracy/efficiency trade-off for HPO, with stability/interpretability/eco implications.
Black-box optimization methods, gray-box shortcuts and even the fundamentals behind white-box approaches~\cite{bengio2000gradient} were all introduced to the ML community several decades ago, as noted within \Tabref{table:hpo-methods}.
In general, black-box methods rely on thorough optimizations with complete evaluations of candidate networks and are thus most accurate and stable.
They are slow, however, and the assumptions underlying gray-box and white-box approaches -- every shortcut used arguably weakens the interpretability of the method -- sacrifice accuracy for progressive improvements in search efficiency.
In this case, it is also reasonable to associate the efficiencies with a reduced reliance on computational resources and, accordingly, a better ranking for eco-friendliness.

% Remark on white-box HPO being scalable but not generalisable.
Crucially, gray-box HPO can be considered as black-box HPO with efficiency boosts, while white-box HPO relies on intrinsically different optimization methodologies, leveraging implicit/explicit assumptions that are particular to neural networks.
This means that the accuracy trade-off for white-box HPO, fine-tuned for DNNs, is not as severe as might be expected.
The efficiencies of white-box HPO, a result of dodging multiple candidate evaluations, also makes related approaches highly scalable.
However, this close tie-in to DNN formalism does mean that white-box HPO is heavily dependent on problem context and search space, while black-box and gray-box HPO are relatively generalizable.
Accordingly, black-box HPO, whether modified to be gray or not, has been benchmarked heavily and implemented within many software packages; future in-depth surveys may comment further on reproducibility or engineering quality.
White-box HPO needs more experimentation and analysis.

\section{Automated Deployment}\label{sec:deploy}

The topic of deploying an ML model into a production environment is an immense one, straddling theoretical principles and real-world practicalities.
Generalized commentary is further complicated by just how many ways an ML model may be used.
Will it serve as the predictive back-end of a queriable web app?
Will it be hooked into a robotic framework as a prescriptive system?
Will it interface with a high-fidelity ``digital twin'' of physical reality~\cite{gelernter1993mirror}?
The field of AutoML has barely begun grappling with the notion of automated deployment, and much of this discussion occurs beyond academia, with best practices for machine learning operations (MLOps) being hashed out by commercial entities~\cite{martel2021software}.

Nonetheless, when it comes to DL specifically, particular trends of research stand out, driven primarily by resource concerns; DNNs are heavyweight models in terms of both storage and inference.
Given that deployment settings can range from edge devices to the cloud, AutoDL strives to answer two mirrored questions:
\begin{enumerate}
\item Can a DL model be optimized for a specific production environment?
\item Can a production environment be optimized for a specific DL model?
\end{enumerate}

\subsection{Deployment-aware AutoDL}\label{sec:deploy-aware-autodl}

Many DL projects have rigid deployment constraints; the onus is on the model to accommodate these requirements.
Thus, while maximal predictive accuracy is still a primary objective, secondary objectives may involve inference latency, memory footprint, and energy cost.
In AutoDL-related literature, model-construction efforts that focus on these considerations are given differing names.
For example, there is ``platform-aware NAS'' for accommodating mobile devices~\cite{yang2018netadapt,tan2019mnasnet}, ``energy-aware pruning'' for constraining network connectivity~\cite{yang2017designing}, and other research published as latency-aware~\cite{cai2019proxylessnas} or resource-aware~\cite{xiong2019resource}.
Accordingly, we generalize such approaches under the banner of deployment-aware AutoDL.

There are several common approaches to dealing with multiple objectives~\cite{marler2004survey,hwang2012multiple}.
One of the simplest is constrained optimization, where a target metric such as inference latency is given an upper bound, and any architectures that do not operate within the tolerable range are discarded or, if possible, adjusted back into that range~\cite{yang2017designing,yang2018netadapt}.
However, if the constraints are poorly behaved, i.e., highly nonlinear, too many unsatisfactory candidates may be constructed during exploration, which is inefficient.
In addition, sampling fully constructed models to evaluate other objective functions negates the innovative shortcuts behind white-box HPO.

%Of course, if deployment objectives are easy to evaluate or estimate,
Other options for multi-objective optimization via NAS and HPO are also available.
A common alternative is to bundle all target metrics into a single one, using this combined objective function to guide AutoDL search algorithms.
Such efforts often focus on network latency, although these values must usually be estimated; efficient AutoDL algorithms cannot spend time evaluating every candidate architecture within an actual production environment.
A typical estimation process then is to (1) pre-compute the latency for hundreds of candidates, (2) train a small DNN on these values to predict latency in general, and (3) use this predictor to approximate the latency of candidate architectures during the AutoDL process~\cite{cai2019proxylessnas,zhou2022rethinking,bender2020can}.
Consequently, previous investigations have explored algebraic combinations of model accuracy and latency, including a re-scaled multiplication~\cite{tan2019mnasnet} and an addition~\cite{cai2019proxylessnas}.
The latter of these used the metric for differentiable NAS~\cite{cai2019proxylessnas}, but latency has also been factored into a reward function for RL-based NAS~\cite{bender2020can}.
Other examples of combined metrics also exist, e.g., applying a piece-wise function for the secondary objective~\cite{dong2019tas} or an absolute function for the re-scaled secondary objective~\cite{bender2020can}.

Sometimes the demands of a production environment can be a little more niche, such as when a device does not support full-precision computing; this can be desirable when aiming for cheap and fast DL applications.
The Infineon XC800 family of microcontrollers exemplifies this, operating in 8-bit.
So, to convert a model trained on a higher-precision processor, the typical solution is to use the so-called quantization technique~\cite{han2016deep}, which approximates the original network by another one with low-bit weights.
However, it is highly possible that, even for the same DL task, the optimal architectures on two platforms with different computing precision may be structured/connected completely differently~\cite{yang2020searching,wang2020apq}.
Ideally, NAS should be undertaken with quantization already in mind, and several works have explored this angle~\cite{wu2018mixed,wang2019haq,bulat2020bats,wang2020apq}.

Ultimately, it is an inescapable fact that numerous different deployment environments exist in the real world, each possibly having its own requirements in terms of latency, energy, etc.
Thus, while there are many approaches for building bespoke DL models, it becomes infeasible to constantly reconstruct models for any applications that are designed with broad release in mind.
Unsurprisingly, a cross-platform ethos has been embraced by certain investigative works.
For instance, one attempt proposes designing a once-for-all (OFA) network that supports diverse architectural settings, such that any supported environment works well with a unique sub-network of the OFA model, no further training required~\cite{cai2020once}.
There are other efforts that likewise seek to obtain multiple models for target environments with but one search~\cite{yu2020bignas}.
Naturally, while these shortcuts boost search efficiency, the accuracy of each sub-network cannot be guaranteed.
Thus, these methods often utilize techniques of knowledge distillation~\cite{hinton2014distilling} to transfer knowledge between super-networks and sub-networks.
Nonetheless, further research is required to compensate for the accuracy drawbacks of OFA approaches.
For now, it appears that planning a DL project cannot remain agnostic with respect to its eventual deployment environment.

\subsection{Hardware Search}\label{sec:deploy-hard}

What if the opportunity arose to mold a deployment environment around a DL model?
Such a circumstance would seem relatively rare at the current time, due to the typical rigidity of hardware constraints, but some commercial/industrial entities have both the capacity and will to be flexible in how they provision resources.

This becomes, as with the majority of research into automation along the DL workflow, an optimization problem.
Indeed, all discussions about black-box/gray-box optimizers in \Secref{sec:hpo} remain relevant here, but the search space now includes hardware.
For instance, multi-objective BayesOpt has been used to co-design both a neural network and an associated energy-efficient AI accelerator, with the latter using Complementary Metal–Oxide–Semiconductor (CMOS) technology~\cite{reagen2017case}.
This effort explored a 14-dimensional search space with one NAS variable, eight (training-based) HPO variables, and five hardware variables, e.g., bit-length and memory bandwidth.

In fact, this co-design approach is fairly typical in the field, and it is not the only investigation that has leveraged multi-objective BayesOpt.
The search technique has been applied to accelerators built from a Field-Programmable Gate Array (FPGA)~\cite{nardi2019practical}, while a variant of BayesOpt has also optimized Programmable Ultra-efficient Memristor-based Accelerators (PUMAs) alongside DNNs trained on AlexNet/VGG~\cite{parsa2019pabo}; the architecture configurations here were relatively simple, covering kernel size, width, and depth.
Of course, other search techniques have been applied too, with RL being used to co-optimize an FPGA-based accelerator and an MBConv-based architecture~\cite{jiang2020hardware}.
Likewise, RL has been employed in producing chip floorplans for the next generation of Google-designed AI accelerators~\cite{mirhoseini2021graph}.
Additionally, given that black-box methods can be computationally expensive, white-box methods have also seen their share of usage~\cite{zhou2022rethinking,choi2021dance}, e.g., one effort~\cite{choi2021dance} applies differentiable NAS/HPO to optimize accelerators based on Eyeriss~\cite{chen2016eyeriss} while searching for MBConv-based architectures~\cite{tan2019mnasnet}.

Importantly, while hardware search presently appears to be a promising research direction for AutoDL, especially with its comparative novelty versus NAS, long-term impact is arguably weakened by the current lack of consensus about AI accelerators.
Basically, AI acceleration is an emerging technology with no dominant design, and research associated with a particular format relies heavily on the success and uptake of that format.
% AI acceleration => act of using accelerators
Hence, there is ongoing debate around what type of systems to focus on.
For instance, one work disregards FPGAs in favor of an ``industry-standard'' edge accelerator~\cite{zhou2022rethinking}, while another explores co-optimization for Application-Specific
Integrated Circuits (ASICs)~\cite{yang2020co}, which are flexibly designed and thus powerful but often tricky to standardize.
Simply put, the evolution of this research sub-field will depend on how the standards of computational hardware themselves evolve.

Beyond training/inference acceleration, the management of hardware resources is also a potential target of automation, explored as early as in the 1990s~\cite{zhang1995reinforcement}.
This broad topic has many particular incarnations in AutoDL, e.g., how to delegate pipelined DL operations among various devices~\cite{mirhoseini2017device,addanki2019placeto,mirhoseini2018hierarchical}, how to effectively administer different DL jobs within a computing cluster~\cite{peng2018optimus}, how to schedule execution for a DL compiler~\cite{chen2018tvm}, etc.
In essence, optimization algorithms here are challenged by the extremely distributed nature of a production environment, potentially managed at different levels and operating in distinct ways, e.g., online versus offline.
These sorts of issues may become more and more important as time goes on and new principles of computing become mainstream.
% federated NAS~\cite{hean20, xuzh20}, for instance, is a reaction to a novel requirement that decentralized servers contribute to model development/deployment without sacrificing data privacy.

As a final assessment, the endeavor of automating hardware search is a promising extension of AutoDL beyond the theoretical focus of NAS to the practicalities of production environments.
It is not a dramatic change in perspective, with many of the works discussed above and summarized in \Tabref{table:autodl-dissect} reusing existing NAS/HPO algorithms; this bodes well for the future design of autonomous AutoDL systems that aim to integrate all DL workflow processes under one umbrella.
However, the real world is messy and definitely not standardized.
While it is reasonable to expect a data scientist to have familiarity with the fundamentals of DL, many of the reviewed works require specialized expert knowledge of hardware accelerators, distribution systems, and so on~\cite{mirhoseini2017device,mirhoseini2021graph,mirhoseini2018hierarchical,reagen2017case,zhou2022rethinking}.
For instance, domain knowledge is required when deciding how to avert any undue impact on search efficiency caused by the simulation cost of a hardware accelerator~\cite{mirhoseini2021graph,zhou2022rethinking}.
In effect, this means that there is a high barrier to entry for research and development around this form of automated deployment, simply in terms of hardware-related expert knowledge, computational resources, and software infrastructure.
Hopefully, this will be ameliorated in time, as existing automation work has already proved itself very valuable to ML-focused engineering and the next generation of AI accelerators~\cite{mirhoseini2021graph}.

\begin{table}[tp!]
\begin{center}\scriptsize
\setlength{\tabcolsep}{3pt}
\caption{
Evaluative assessment for trends in automated deployment.
}
\begin{tabular}{l | l | l | l | l | l | l | l | l | l | l}
    \toprule
                            & Novelty      & Solution & Effic. & Stability & Interp. & Reprod.  & Engi. & Scalability   & General. & Eco. \\
\midrule
    Deployment-aware AutoDL & $\approx$ 4  &   High   &  Mixed     & Medium    & High    & \cmark   &   ?    & Mixed         & Mixed            & Mixed  \\
    Hardware Search         & $\approx$ 5  &   High   &  Low       & ?         & Low     & ?        &   ?    & Low           & Low              & ?  \\
    \bottomrule
\end{tabular}\label{table:dep-methods}
\end{center}
{\scriptsize\textit{
Each row marks an emergent trend in AutoDL, specifically automated deployment. Each column marks a criterion -- see \Secref{sec:criteria-intro} -- by which the trend is assessed. The evaluations are mostly qualitative, averaged across the most significant works researching the trend. Where a graded value is not provided, ``\cmark'' indicates a rigorous assessment is possible with analysis beyond the scope of this review, while ``?'' indicates that not even this is achievable without more research works to analyze. Novelty denotes years since seminal works in DL were published. Abbreviations are:  ``Solution'' for Solution Quality, ``Effic.'' for Efficiency, ``Interp.'' for Interpretability, ``Reprod.'' for Reproducibility, ``Engi.'' for Engineering Quality, ``General.'' for Generalizability, and ``Eco.'' for Eco-friendliness. 
}}
\end{table}

\subsection{Overview}

% Highlight that hardware search is less advanced than deployment-aware AutoDL.
The two trends in automated deployment focus on optimizations in opposing directions.
Put simplistically, one adjusts software for the sake of hardware and one adjusts hardware for the sake of software.
Thus, despite these endeavors entering the domain of DL at roughly the same time, it is not surprising that hardware manipulation currently scores much worse over a broad range of assessment criteria, as listed in \Tabref{table:dep-methods}.
Research in both trends is of course equally valuable, each demonstrating that model performance and thus accuracy is maximized when production and deployment environments are considered together.
Nonetheless, progress in hardware search is stymied by the great variety of physical architectures that exist; the proponents of any novel procedure will also need to promote a particular system to the rest of the DL community.

% Elaborate on the present weaknesses of hardware search.
Accordingly, research in the sub-field of hardware search remains more scattered and exploratory.
A lack of benchmarking and standardized implementation makes it difficult to comment about the reproducibility of results and their stability, let alone how complicated proposed techniques are to engineer.
The presence of multiple layers between physical systems and abstracted DL algorithms also affects interpretability, obscuring how exactly hardware manipulations contribute to positive DL outcomes.
Moreover, because physical environments can typically only be optimized via gray-box procedures at best, efficiency boosts are limited and associated search procedures do not scale well with the complexity of hardware systems.

% Discuss deployment-aware AutoDL and its dependency on HPO.
In contrast, research in deployment-aware AutoDL is a little more unified, enough for some initial analysis of both stability and reproducibility~\cite{li2021hw}.
Admittedly, there are only a limited number of software libraries that currently implement related techniques, but the fundamentals underpinning the adjustment of DL models for deployment environments are fairly well understood, at least in contrast to hardware search.
Beyond that, the efficiency, scalability and generalizability of deployment-aware AutoDL cannot be simply summarized, as these criteria depend far too much on the specifics of the algorithm in question.
The assessments in \Tabref{table:hpo-methods} are informative in this case; white-box optimization methods will be efficient and scalable, but not generalizable, while black-box and gray-box methods will reverse those qualities.

\section{Automated Maintenance}\label{sec:maintenance}

The future of AutoML, at least in the short to medium term, is continuous learning; this is what a previous review identified and argued~\cite{kedziora2020autonoml}.
In fact, there is currently a disconnect between how the majority of ML models are designed/used -- ``one-and-done'' -- and what the growing demands of real-world applications are.
Recent global events have shown that models built on static assumptions can be very fragile\footnote{\url{https://www.dominodatalab.com/blog/how-covid-19-has-infected-ai-models}}, highlighting a dire need, depending on industry and problem context, for continuous monitoring and maintenance.
We label systems that are able to automatically provide this support as AutonoML frameworks, with the understanding that true autonomy is impossible without the capacity for persistent automated management of models.
For now, there is no archetypal AutonoML package that we can highlight, but many vendors have listed adaptation capability on their development roadmaps, researchers have published prototypes of adaptive ML systems even before the modern wave of AutoML~\cite{kadlec2009architecture,ruta2011generic}, and there is a growing number of academic research efforts in this direction~\cite{madrid2019towards,celik2021adaptation}.
Evidence shows that initial steps are being taken in the evolution from AutoML to AutonoML.

So, what about ``AutonoDL''?
Does such a research parallel exist?
The answer to this is very nuanced.
At zeroth order, it is reasonable to assert that continuous learning is not currently a major priority in the DL community.
This is understandable, as the major selling point of DL is in fact its representational power.
Manageably complex models are great for approximating complicated input-output mappings.
The corresponding downside though is that training a DNN well often relies on the repeated presentation of concepts in data.
These models are not agile, at least traditionally, and they fare poorly with conceptual instability~\cite{mccloskey1989catastrophic}.
Nonetheless, although continuous learning may not currently be practical, a lot of recent effort has gone into making repeat model development as flexible as possible, designed to accommodate changes in both data and task.

Therefore, after briefly considering the ideal of online learning in \Secref{sec:maintenance-challenges}, and why it is difficult, we discuss relying on continuous monitoring instead in \Secref{sec:maintenance-scenarios}, covering possible data-environment dynamics to watch out for, before reviewing popular paradigms for automatic model maintenance in \Secref{sec:maintenance-method}.

\subsection{The Challenges of Online Learning}\label{sec:maintenance-challenges}

Learning within dynamic data environments is not a new topic.
Over several decades, many bespoke ML models/algorithms have been proposed with the aim of operating on streams of data, e.g., incremental decision trees~\cite{quinlan1986induction}.
In the ideal case, these ML models should continue to develop their structures and tune their parameters and all associated pre-processing or data transformation steps~\cite{zliobaite2014adaptive} while they are deployed, either on delayed training data or on some feedback to query-based model responses, e.g., from a user, a digital-twin simulation, a robotic system, etc.
So, why not simply mimic this approach and leave a DNN in training mode during deployment?
Unfortunately, all the challenges of standard DL training immediately apply, but now in an environment that is difficult to control.
For instance, in the case of backpropagation, deciding on batch size is already a challenge~\cite{smith2018don}, as it determines how much a single instance of data affects a model.
Unsurprisingly, it is even more difficult to curate an informative batch of queries and feedback-derived labels on the fly, especially one that provides sufficient repeat presentation of any important but infrequently observed concept.

Even then, best practices of training aside, the fundamental obstacle that typical DL models face is catastrophic interference/forgetting~\cite{mccloskey1989catastrophic, thrun1995lifelong}.
Neither network depth nor connectivity, despite other benefits, helps isolate learned concepts; the latter actively competes with information localization.
Thus, when an input-output mapping is approximated by a monolithic model, new concepts can easily overwrite old ones.
This can be highly undesirable for a sequential learner, e.g., if an NLP text generator encounters low-quality writing after training on a high-quality corpus.
Simply put, monolithic models are particularly susceptible to experiencing ``garbage in, garbage out'' (GIGO).

Admittedly, there has been plenty of contemplation around how neural networks may operate in a persistent manner, engaging in lifelong learning while carefully managing the stability-plasticity dilemma~\cite{pake19}.
Some attempts to address this draw inspiration from neuroscience, even if the associated adaptive mechanisms are often better suited to models built from more biologically realistic neurons, e.g., spiking neural networks~\cite{dusu19}.
All the same, there are several strategies for standard artificial neurons that recur throughout the literature.
For instance, one approach relies on regularization~\cite{kirkpatrick2017overcoming,li2017learning_without_forget}, seeking to avoid catastrophic interference by controlling network weight updates with constraints.
This might be a quadratic penalty scaled to the difference between old and new parameters, which slows down how quickly previously learned information is overwritten~\cite{kirkpatrick2017overcoming}.
Alternatively, one can try to force information locality via some form of ensembling, often employed for data stream analysis~\cite{krawczyk2017ensemble}, or even fully dynamic architectures~\cite{rusu2016progressive}, which grow out new sub-networks as required.
There are yet other efforts that attempt to codify dual memories~\cite{hinton1987using}, first proposed early on~\cite{hinton1987using}, so as to have one set of weights maintain certain concepts in long-term storage and have another set of weights adapt to changing data dynamics.
Nonetheless, with few exceptions, such strategies for lifelong learning remain relatively unknown within the DL community, given that managing a standard DNN is already computationally expensive enough as it is.

\begin{wraptable}{r}{55mm}
\vspace{-7mm}
\begin{center}
\caption{
Scenarios in a dynamic data environment.
}
\begin{tabular}{ll}
\toprule
Scenario             & \makecell[l]{Changes\\ $(\gS_1\rightarrow\gS_2)$} \\ \midrule
The i.i.d. Assumption & None                                \\
Covariate Shift       & $P(x)$                              \\
Prior Shift           & $P(y)$                              \\
Concept Drift         & $P(y|x)$                            \\
Task Redefinition     & $\mathbb{X}$ and/or $\mathbb{Y}$    \\
\bottomrule                  
\end{tabular}\label{table:dynamics}
\end{center}
\textit{
Set $\gS_i$ is defined here as a collection of data that samples an input-output distribution of interest to a DL model, i.e., a domain indexed by $i$. Input $x$ and output label $y$ belong to input and output spaces $\mathbb{X}$ and $\mathbb{Y}$, respectively, while $P$ denotes a probability with respect to sampling representation.
}
\vspace{-6mm}
\end{wraptable}
In short, for many theoretical and practical reasons, continuous model updates are not presently feasible in the field of AutoDL, yet the need to respond to dynamic data environments remains.
Thus, rather than online development, the focus turns to a more relaxed form of continuous learning: updates as required.
The challenge of catastrophic interference never disappears, but it can now be faced with deliberate intention.
However, as a consequence of this approach, the process of adaptation now splits into two, namely continuous monitoring and maintenance, or ``when do I do it?'' and ``what do I do?'', respectively.

\subsection{Scenarios for Continuous Monitoring}\label{sec:maintenance-scenarios}

Many scenarios are possible in a dynamic data environment.
To ground this discussion, we define ``domain'' to be some distribution of input-output data that is desirable for learning, possibly a ground truth.
Under this definition, the fundamental notion underlying ML is that any ML model can only observe and learn from an $n_i$-sized sample of a domain indexed by $i$, i.e., ${(x_k,y_k)}_{k=1}^{n_i}$, although whether the outputs are accessible to the model depends on training or deployment.
Additionally, given a specific task, an input $x$ and an output $y$ drawn from this domain exist within prescribed spaces $\mathbb{X}$ and $\mathbb{Y}$, respectively.
Now, DL typically relies on the i.i.d. assumption, where the joint probability distribution of sampling $(x_k,y_k)$ is static, i.e., the samples arise from the same memoryless generative process.
In essence, this is what makes an inductive model trained on sample $\gS_1$ applicable to sample $\gS_2$; it is hoped that their corresponding domains, indexed by $1$ and $2$, are identical.
Of course, the i.i.d. assumption may be flawed to begin with~\cite{hendrycks2017baseline,zhou2009multi}, but, regardless, any significant deviation from these statistics can challenge the validity of a trained model.
We emphasize the adjective ``significant'' here, as anomaly detection is its own topic of research~\cite{chandola2009anomaly}.

Naturally, the stationary assumption can be broken in many ways, which have been surveyed extensively elsewhere~\cite{gama2014survey}.
The Bayes theorem suggests a few possibilities, especially when written as:\looseness=-1
\begin{align}
P(x \cap y)=P(x|y)P(y)=P(y|x)P(x),
\end{align}
and these are listed in \Tabref{table:dynamics}.
Prior and covariate shift -- variations in $P(y)$ and $P(x)$, respectively, between domain samples $\gS_1$ and $\gS_2$ -- are types of data drift relating to sampling representation.
This can be a problem, given that predictive DL models strive to determine the conditional probability $P(y|x)$ for all $x \in \mathbb{X}$ and $y \in \mathbb{Y}$; the accuracy of such an estimate suffers for subspaces in $\mathbb{X}$ and $\mathbb{Y}$ that have not previously been encountered with sufficient frequency.
Nonetheless, on its own, these kind of changes in a data environment do not negate the utility of an $\gS_1$-trained model deployed on $\gS_2$.
Sometimes, they simply suggest concept-space interpolations/extrapolations need to be tightened.
In essence, the two domains may still be identical.
On the other hand, concept drift is a scenario where $P(y|x)$ itself changes between domain samples $\gS_1$ and $\gS_2$, e.g., in the form of changing classification boundaries.
Sometimes concept drift even overlaps with the notion of task redefinition, where the label space $\mathbb{Y}$ may change.
In such a case, extensive model retraining appears almost inevitable.

For all such scenarios, it is the responsibility of a prospective AutoDL monitoring mechanism to identify when model performance may suffer, so as to trigger a maintenance procedure.
The monitoring may be directly reactive, assessing the degradation of a loss metric, or be indirectly preemptive, examining non-stationary statistics of data.
In fact, there are many drift detection mechanisms and strategies in existence, some of which have already been used in the context of adaptive AutoML~\cite{kedziora2020autonoml,celik2021adaptation}.
We do not discuss these in depth, as, for now, the field of AutoDL has not yet embraced the idea of agile monitoring.
However, as a compromise, there have been several past efforts to acknowledge diverse domains and fold their identification into a DL model.
For example, the simplest form of the long-established adaptive resonance theory (ART)~\cite{grossberg2020toward} dynamically sizes a layer of ``recognition neurons'' to cluster encountered data into distinct categories.
Far more recently, research efforts have merged a domain classifier into a DNN~\cite{ganin2015unsupervised} and have explored domain encodings via a so-called Memory-based Parameter Adaptation (MbPA) method~\cite{sprechmann2018memory}.
While the reactive potential is limited in both these modern examples to the memory space reserved for domains, at least data drifts between those domains can easily be adapted to.

\begin{table}[tp!]
\begin{center}\footnotesize
\setlength{\tabcolsep}{9pt}
\caption{
The most common approaches underlying adaptive strategies for AutoDL.
}
\begin{tabular}{llll}
\toprule
Maintenance Approach & \makecell[l]{Domain\\ Constraints\\$(\gS_1\rightarrow\gS_2)$} & \makecell[l]{Data Availability\\for Maintenance\\$(\gS_2)$} 
& Notes
\\
\midrule
Continuous Learning    & None                                                               & Any 
& \makecell[l]{Constantly processes a data stream.\\Can combine with other approaches.}
\\
\midrule
Domain Adaptation      & $\mathbb{Y}_1=\mathbb{Y}_2$                                        & $x$: \cmark, $y$: \xmark &  \multirow{4}{*}{\makecell[l]{Often generalizes more than one domain.\\Prioritizes minimal sampling of $\gS_2$.}}\\
Domain Generalization  & $\mathbb{Y}_1=\mathbb{Y}_2$                                        & $x$: \xmark, $y$: \xmark & \\
Few-shot Learning      & $\mathbb{Y}_1\neq\mathbb{Y}_2$                                     & $x$: \cmark, $y$: \cmark & \\
Zero-shot Learning     & $\mathbb{Y}_1\neq\mathbb{Y}_2$                                     & $x$: \xmark, $y$: \cmark & \\
\bottomrule
\end{tabular}\label{table:paradigms}
\end{center}
{\footnotesize\textit{
Sample $\gS_i$ represents a domain indexed by $i$, where $\mathbb{Y}_i$ is its corresponding output/label space. The inputs and output labels within sample data are represented by $x$ and $y$, respectively. Domain adaptation/generalization is well-suited for statistical shifts and concept drifts, given a fixed task, while low-shot learning quickly adapts to new classes, a consequence of task redefinition. Note that there are proposed variants to these approaches, e.g., differing degrees of label-based supervision from $\gS_2$.
}}
\end{table}

\subsection{The Current Paradigms of Maintenance}\label{sec:maintenance-method}

The vast majority of DL models are monolithic and do not contain specialized domain-memorization structures.
Thus, in most cases, the standard form of maintenance is simply retraining on a newly encountered domain.
Doing so from scratch, however, is not ideal, especially in the computationally expensive context of DL.
Consequently, there has been an increasing focus on developing efficient maintenance strategies for various scenarios.
The most common principles underlying such approaches are listed out in \Tabref{table:paradigms}.

First of all, if domains represented by $\gS_1$ and $\gS_2$ differ in statistics but not in $\sY$, the topics of domain adaptation and domain generalization apply~\cite{ben2007analysis,wang2021generalizing}.
An example of this is extending an image classifier of birds trained on sketches and cartoons to a more photorealistic domain.
In the case of domain adaptation, maintenance occurs reactively, with the model able to learn from newly encountered data that is possibly even labeled.
Domain generalization covers more preemptive strategies, improving model adaptability without encountering what it will have to adapt to.
Elsewhere in the DL field, low-shot learning attempts to deal with more dramatic task redefinitions, i.e., where label space $\sY$ changes.
In the extreme, these endeavors may form a prerequisite towards eventual general intelligence, but, in current practice, the tasks applied to differing domains sampled by $\gS_1$ and $\gS_2$ are closely related, e.g., both are image recognition problems using CNNs.
For few-shot learning~\cite{snell2017prototypical,ravi2017optimization,finn2017model,liu2019learning}, maintenance mechanisms do have access to limited data from the new domain, including for unseen classes, and adapt accordingly.
For zero-shot learning~\cite{romera2015embarrassingly,xian2018zero,liu2021isometric}, an AutoDL system does not expect to encounter any examples of unseen classes, although auxiliary information about these classes is leveraged instead, e.g., information based on attributes or embedding similarity.

In any case, for many reasons described earlier, these updating strategies are often triggered manually and applied offline.
However, should the priority focus of the DL community move to AutonoDL, these approaches are well-suited to being appropriated for online adaptation.
Thus, it is still worth highlighting the most common paradigm that underpins DL efforts in this space: the meta-model.

Meta-learning~\cite{schmidhuber1987evolutionary,bengio1990learning,schmidhuber2002bias,schaul2010metalearning} in the context of AutoML has been reviewed elsewhere~\cite{kedziora2020autonoml}, but, in AutoDL, it refers to identifying some similarity, often in terms of ``meta-features'', between a new domain/task and an old one; this allows previous knowledge to be leveraged while optimizing a DL model for a new environment, not too unlike transfer learning.
% A meta-model can then be seen as but a high-level context-aware recommendation system, and, in modern times, it is often a neural network of its own, thus being optimizable via gradient descent~\cite{andrychowicz2016learning,xu2018meta}, RL~\cite{li2017learning}, evolutionary approaches~\cite{houthooft2018evolved,jaderberg2017population}, etc.
A meta-model can then be seen as but a high-level context-aware recommendation system. This idea dates back to the year 1987~\cite{schmidhuber1987evolutionary}, although, at that time, the application was relatively simple and small-scale.
In modern times, it is often a DNN of its own, thus being optimizable via gradient descent~\cite{andrychowicz2016learning,xu2018meta}, RL~\cite{li2017learning}, evolutionary approaches~\cite{houthooft2018evolved,jaderberg2017population}, etc.
These meta-models are usually trained in offline mode, experiencing many domains/tasks -- debate endures around just how much is needed~\cite{gemp2017automated,ali2018cross} -- and they ideally learn which recommendations are optimal for training a base model in each setting.
They can also be developed in online mode, alternating in updates with a base DL model~\cite{xu2018meta,xu2020meta}, but this compounds the risks of learning-based instability.
  
Importantly, meta-models are highly varied in their usage.
For instance, there is a strand of DL research popularly known as ``learning to learn''~\cite{schmidhuber1998reinforcement,hochreiter2001learning}, which is motivated by certain questions: when is Adam better than standard SGD? What is the quickest way to train a model for sparse data? And so on.
In such a scenario, the meta-model -- typically an RNN such as an LSTM -- learns correlations between diverse operating contexts and best values for select variables relating strictly to optimization procedures~\cite{du2019sequential}, i.e., algorithmic hyperparameters, that can then boost model development in new domains.
Of course, meta-model input-output details vary widely across research efforts, whether the approach is applied to standard optimizers~\cite{andrychowicz2016learning} or deep RL~\cite{duan2016rl}.

Notably, if there is one area of DL that does focus heavily on online operations and adapting to new domains/tasks, it is in fact deep RL.
This approach is heavily favored in the AI sub-field of ``general game playing'', and many related research works are tested in diverse contexts, such as on collections of Atari 2600 video games~\cite{xu2018meta,veeriah2019discovery}.
As an example of meta-learning for deep RL, one publication~\cite{houthooft2018evolved} focuses on the fact that, in RL, there is often no intrinsic relation between a task objective and a loss/reward function.
It thus applies a context-aware temporal CNN to optimize this loss function so that an RL agent, the base DL model, is able to learn a task with maximal efficiency.
In this particular approach, the meta-model is closely integrated with the base model, influencing it via backpropagation.
Elsewhere, adaptation of the return function has similarly been explored with a gradient-based meta-learning algorithm, tuning hyperparameters, such as discount factor and a bootstrapping parameter, in online fashion~\cite{xu2018meta}.
This particular approach, using ``meta-gradients'', has partially inspired subsequent work, such as an effort to have RL agents seek out useful questions, in general value function (GVF) format, that, when answered, optimally support their learning process~\cite{veeriah2019discovery}.

It is now clear that, should an AutoDL monitoring system throw an alert that a DL model needs to be updated, prior experience can definitely accelerate model adaptation.
However, it is often difficult to assess the quality of proposed meta-models and meta-learning algorithms.
Many works are based on handcrafted update rules targeting select hyperparameters, such SGD learning rate~\cite{sutton1992adapting}, SGD weight decay~\cite{du2019sequential}, the decay factor in RL~\cite{xu2018meta}, etc.
Others gradually move further and further to full automation, leveraging neural networks to simulate the loss function~\cite{houthooft2018evolved,kirsch2020improving}, the target value to maximize in RL~\cite{xu2020meta}, the delta of weights~\cite{duan2016rl,andrychowicz2016learning}, etc.
So, how much of meta-model design should rely on humans, producing limited but decent strategies?
How much should be fully automated, producing generic but potentially unwieldy mechanisms?
These are open questions, as is whether and in which cases meta-learning is actually effective~\cite{gemp2017automated,ali2018cross,nguyen2021exploring}.

Ultimately, we emphasize that meta-learning is not the same thing as continuous learning, the latter of which is a fundamental requirement of AutonoDL.
Meta-learning is a principle that can be applied across the entire DL workflow in \Figref{fig:dl-life-cycle}, so as to leverage previous experience in speeding up the development/deployment process.
Similarly, there are ways to react intelligently to new domains that have nothing to do with typical notions of prior knowledge.
Nonetheless, this is where AutoDL presently sits on the automated-maintenance front; it is a nascent exploratory topic with little benchmarking beyond an assessment of dynamic hyperparameter control~\cite{eimer2021dacbench}.

\begin{table}[tp!]
\begin{center}\scriptsize
\setlength{\tabcolsep}{4pt}
\caption{
Evaluative assessment for trends in automated maintenance.
}
\begin{tabular}{l | l | l | l | l | l | l | l | l | l | l}
    \toprule
                        & Novelty         & Solution & Effic. & Stability & Interp. & Reprod.  & Engi. & Scalability & General. & Eco. \\
\midrule
    Domain Adaptation   & 3 ($\approx$30) & ?        &  High  &  ?        &    ?    &     ?    &   ?   &    High     &   Mixed  &   ?  \\
    Low-shot Learning   & 3 ($\approx$29) & ?        &  High  &  ?        &    ?    &     ?    &   ?   &    High     &   Mixed  &   ?  \\
    Continuous Learning & 5 ($\approx$30) & ?        &  High  &  ?        &    ?    &     ?    &   ?   &    High     &   Mixed  &   ?  \\
    % Auto-Concept Drift       & 2 ($\approx$33) & ? &  ?       &  ?  & ? &  ?  &  ? &  ?      &   ?     & ? \\
    % Deep RL             & 5               & ? & Mixed    &  ?  & ? &  ?  &  ? & Mixed   &  Medium   & ? \\
    \bottomrule
\end{tabular}\label{table:maint-methods}
\end{center}
{\scriptsize\textit{
Each row marks an emergent trend in AutoDL, specifically automated maintenance. Each column marks a criterion -- see \Secref{sec:criteria-intro} -- by which the trend is assessed. The evaluations are mostly qualitative, averaged across the most significant works researching the trend. Where a graded value is not provided, ``?'' indicates that a rigorous assessment is not achievable without more research works to analyze.  Novelty denotes years since seminal works in DL (ML) were published. Abbreviations are: ``Solution'' for Solution Quality, ``Effic.'' for Efficiency, ``Interp.'' for Interpretability, ``Reprod.'' for Reproducibility, ``Engi.'' for Engineering Quality, ``General.'' for Generalizability, and ``Eco.'' for Eco-friendliness. 
}}
\end{table}

\subsection{Overview}

% Discuss how AutonoDL trends are too novel/scattered to make many claims.
Given how adaptive ML, namely AutonoML, has only just become an emerging thread of research~\cite{kedziora2020autonoml}, it is of no surprise that there is little consolidated focus in the literature on automated mechanisms for the maintenance of DL models, let alone an AutonoDL field.
Indeed, \Secref{sec:maintenance} has instead discussed the most promising approaches, as listed in \Tabref{table:maint-methods}, that are likely to inform adaptation practices in the future.
This connection is immediate in the case of continuous learning, and the paucity of research in this area merely reflects the challenges of constantly updating a DNN, but domain adaptation and low-shot learning are often not considered by the DL community in terms of fully automated model maintenance, instead being associated with tackling new DL problems; related publications often miss or ignore the automation aspect in their commentaries.
Accordingly, the evaluative assessment in \Tabref{table:maint-methods} is necessarily replete with unknowns.
This is the case even though the theoretical foundations for the listed trends stretch back far beyond the DL era, e.g., low-shot learning being considered in the 1990s~\cite{schmidhuber1992learning}.
%However, they have only been associated with AutoDL for a handful of years, and research has been scattered over many bespoke experimental settings.
Ultimately, it is just not possible at this time to gauge how effectively an approach counters diminishing model accuracy, let alone whether such claims are reproducible over a representative benchmark of dynamic DL problems.

% Talk about what we can say.
So, if theoretical analyses are limited, making it difficult to comment on stability, and there are no widely adopted implementations of such mechanisms, restricting any eco-friendliness assessment, can anything be surmised about automated maintenance in AutoDL?
Well, the most common methodologies employed for domain adaptation, together with low-shot learning and continuous learning, are gradient-based, which means that such approaches can be considered relatively efficient ways to adapt a model.
Likewise, they are highly scalable.
As for generalizability, this depends on the specific methods used for each form of maintenance.
Some are specifically designed for updating a specific network component such as a model head, i.e., the classification layer, and thus may struggle when applied to novel search spaces.
Beyond this, one must reserve judgment as to how the automated maintenance endeavor will evolve.
There are certainly promising research directions, but serious consideration of AutonoDL as a field will likely be contingent on further advances in computational hardware.

% \section{Criteria of AutoDL}\label{sec:criteria}
\section{Critical Discussion and Future Directions}\label{sec:criteria}

% The overarching introduction and assessment for every sub-area of AutoDL has been delivered through \Secref{sec:auto-translate-context} to \Secref{sec:maintenance}.
% A nature idea is to assessment the entire AutoDL area as a whole by connecting every sub-area.
% To complete such an assessment, we first introduce a unified questionnaire for all AutoDL works instead of focusing on a single narrowed area.
% Later, an overarching evaluation of AutoDL is introdcued in \Secref{sec:final-discussion-evaluation}.
% Based on these evaluations, we rethink the current challenges and opportunities for AutoDL.

% Highlight just how oversaturated DL/AutoDL research is.
It is undeniable that, at present, the topic of DL continues to attract an unparalleled degree of attention within computer science communities, and its successes have spilled over into the sub-field of AutoDL.
This monograph has attempted to take a sample of that research, representing the most significant trends in the area, and categorically systematize it with respect to the simple but encompassing DL workflow depicted in \Figref{fig:dl-life-cycle}.
However, this has not been a trivial task; the sheer quantity of publications in AutoDL and related sub-fields can be considered daunting.
Certainly, popularity has positives, as a critical mass of attention is required to drive progress in a topic, but the negatives are just as evident, risking both a mob-mentality `clumping' around certain trends -- these endeavors may or may not have been exhausted of promise -- and a level of low-quality publication `noise' that obscures possible leads for future advancement.

% Stress the criteria and promote a questionnaire.
To combat the counterproductive dangers of research oversaturation, this work has emphasized the need to examine AutoDL research holistically, suggesting a broad set of criteria in \Secref{sec:criteria-intro} that may be used for evaluation.
Granted, this review can only provide a limited assessment under such a framework, or similar, without broader adoption and contemplation by the DL community.
Thus, \Secref{sec:final-discussion-questionnaire} promotes a practical questionnaire for AutoDL researchers to use, based on the same criteria, when evaluating how their proposed methodologies/experiments fit within the broader context.
It is not intended as a rigid document, serving more as a kick-starter for conversation around the topic.

% Bring it all back to an overview of AutoDL.
Finally, still guided by the assessment criteria, we consolidate all the assessments within the previous sections, each related to a single stage of a DL workflow, into an overarching overview of the AutoDL field, presented in \Secref{sec:final-discussion-evaluation}.
This high-level perspective enables subsequent commentary in \Secref{sec:future} on what this review finds the most significant challenges and greatest opportunities to be in terms of future directions.

\subsection{A Proposed Questionnaire for Self-assessment}\label{sec:final-discussion-questionnaire}

% Recapping the strong points and limitations of the introduced auto data-preparation in \Secref{sec:auto-data-prepare}, neural architecture search in \Secref{sec:nas}, hyperparameter optimization in \Secref{sec:hpo}, auto-deployment in \Secref{sec:deploy}, and auto-maintenance in \Secref{sec:maintenance}.
While the full scope of AutoDL research is expansive, especially beyond the bulk of work that exists within the topic of NAS, there are clear commonalities across the field.
Each investigation, whether theoretical or experimental, typically seeks to develop a high-level mechanized process that efficiently contributes to the existence of a ``good'', ideally persistent, DL model.
Accordingly, as \Tabref{table:autodl-dissect} indicated, these attempts can often be characterized as optimizations, regardless of the phase of a DL workflow in which they operate.
This also means that it is possible to construct a shared baseline framework by which all AutoDL research can be assessed, which is the purpose of the ten criteria introduced in \Secref{sec:criteria-intro}.

Naturally, a unified assessment framework is appealing for several reasons:
\begin{itemize}
    \item It helps researchers/developers to be comprehensive in understanding the strengths and limitations of their approach, which can then lead to subsequent improvements.
	\item It simplifies peer review, supporting the accelerated provision of accurate feedback.
	\item It enables the contextualization of an individual algorithm within the entirety of AutoDL, promoting informative high-level perspectives of the field.
\end{itemize}
Granted, critics may claim that a one-size-fits-all framework does not appreciate nuances, and this is a reasonable concern, but, given the state of mass publications in the field, it is arguably a higher priority of consideration to encourage improved rigor through standardization, at least at the current time.

So, how would a researcher go about engaging with the criteria listed in this monograph?
To answer this, we provide an example form in \Tabref{table:autodl-questionnaire} and \Tabref{table:autodl-questionnaire-sup}, filling out a series of questions as if they were asked of two seminal publications, NASNet~\cite{zoph2018learning} and DARTS~\cite{liu2019darts}, which we stress have been selected for their importance to AutoDL, not for any perceived deficiency.

Now, crucially, because this form represents a deeper dive than the quantitative research-trend overviews described in \Secref{sec:criteria-intro}, there are certain nuances to consider.
For instance, one significant difference alluded to in the earlier section is that a detailed assessment should capture the quality of both an AutoDL process and the target DL model that it impacts.
Accordingly, several criteria in the example form split their questions in such a manner.
As to the rationale behind listed questions, this matter is elaborated on in the following discussion.

\newcommand{\answerYes}[1][]{\textcolor{blue}{[Yes] #1}}
\newcommand{\answerNo}[1][]{\textcolor{orange}{[No] #1}}
\newcommand{\answerNA}[1][]{\textcolor{gray}{[N/A] #1}}
\newcommand{\LACK}[1]{{\textcolor{red}{#1}}}
\newcommand{\EX}[1]{{\textcolor{purple}{#1}}}
\newcommand{\answerTODO}[1][]{\textcolor{red}{\bf [TODO]}}
\newcommand{\didmark}{\ding{51}}
\newcommand{\notmark}{\ding{55}}

\begin{table}[tp!]
\begin{center}\scriptsize
\setlength{\tabcolsep}{1pt}
\caption{
[Part-1/2] An example self-assessment questionnaire for AutoDL algorithms.
}
\begin{tabularx}{\linewidth}{ p{0.04\linewidth} | p{0.35\linewidth} | X | X}
\toprule
       & Question & Example response for NASNet~\cite{zoph2018learning}
                  & Example response for DARTS~\cite{liu2019darts} \\
\midrule
\multirow{12}{*}{{\rotatebox[origin=c]{90}{\RNC{1}: Novelty}}}
       & Is there significant innovation in this work?        
       & Yes; the first cell-based search space in NAS.
       & Yes; the first differentiable search algorithm in NAS.
       \\
       & Is this work completely distinct from existing research in Automated Data Engineering?        
       & Yes; within reasonable scope.
       & Yes; within reasonable scope.
       \\
       & Is this work completely distinct from existing research in NAS?        
       & No, but without issue; reuses the search space in \cite{zoph2018learning}.
       & No, but without issue; reuses the search method in \cite{zoph2017NAS}.
       \\
       & Is this work completely distinct from existing research in HPO?
       & No, but without issue; the search method has been studied in HPO.
       & No, but without issue; the search method has been studied in HPO.
       \\
       & Is this work completely distinct from existing research in Automated Deployment?                  
       & Yes; within reasonable scope.
       & Yes; within reasonable scope.
       \\
       & Is this work completely distinct from existing research in Automated Maintenance?                  
       & Yes; within reasonable scope.
       & Yes; within reasonable scope.
       \\
\midrule
\multirow{7}{*}{{\rotatebox[origin=c]{90}{\RNC{2}: \makecell[c]{Solution\\Quality}}}}
      & Does the contribution of this AutoDL method to model accuracy (or similar metric) significantly improve upon equivalent SoTA approaches?
      & No, but without issue; the largest NASNet model (of the time) achieves 82.7\% top-1 accuracy on ImageNet, which is similar to previous SoTA model but with 40\% reduction in FLOPs.
      & No; the best DARTS model achieves  2.76\% error on CIFAR-10 with 3.3 million parameters, which is slightly worse than previous SoTA models.\\
      & Is the reported contribution to accuracy (or similar metric) fairly representative? 
      & Yes, decently; reported for CIFAR-10, ImageNet, COCO.
      & Yes, decently; reported for CIFAR-10, ImageNet, PTB, WikiText-2.
      \\
\midrule
\multirow{8}{*}{{\rotatebox[origin=c]{90}{\RNC{3}: Efficiency}}}
      & Does this AutoDL method converge with significant speed?                 
      & No; 2000 NVidia P100-hours to complete NAS on CIFAR-10.
      & Yes; 4 NVIDIA GTX 1080Ti GPU-days to complete NAS on CIFAR-10.
      \\
      & Does the contribution of this AutoDL method to model efficiency (i.e., minimization of computational-resource usage) significantly improve upon equivalent SoTA approaches?
      & Yes; this NAS method can produce a model with 3.3 million parameters that achieves 2.65\% error on CIFAR-10, \LACK{but latency is not evaluated.}
      & Yes; this NAS method can produce a model with 3.3 million parameters that achieves  2.76\% error on CIFAR-10, \LACK{but latency is not evaluated.}
      \\
      & Are the theoretical/practical efficiencies fairly analyzed?
      & Yes, mostly; the number of model parameters and FLOPs are analyzed, \LACK{although there is no comment on practical speed.}
      & Yes, mostly; the number of model parameters and FLOPs are analyzed, \LACK{although there is no comment on practical speed.}\\
\midrule
\multirow{9}{*}{{\rotatebox[origin=c]{90}{\RNC{4}: Stability}}}
       & Do repeated runs of this AutoDL method produce results (e.g., in terms of accuracy metric) with minimal variance?
       & Unknown; \LACK{analysis does not exist.}
       & Yes, moderately; the results of four search runs are reported.
       \\
       & Do repeated runs of this AutoDL method impact model structure (if applicable) in the same way?
       & Unknown; \LACK{although NAS methods intrinsically influence model structure by virtue of construction, analysis does not exist.}
       & Yes; different search runs may produce different models structures.
       \\
       & Is the AutoDL method stable (i.e., not sensitive) for certain values of AutoDL-method hyper-parameters?
       & Unknown; \LACK{only brief mention that a learning rate of 3.5e-4 is relatively stable.}
       & Unknown; \LACK{analysis does not exist.}\\
\midrule
\multirow{4}{*}{{\rotatebox[origin=c]{90}{\RNC{5}: \makecell[c]{Interpre-\\tability}}}}
       & Is this AutoDL method explainable in how it impacts the target model?
       & Unknown; \LACK{analysis does not exist.}
       & Yes, decently; analyzed under the differentiable bi-level optimization framework.
       \\
       & Are the results produced by the target model (after the AutoDL method takes effect) explainable?
       & Unknown; \LACK{analysis does not exist.}
       & Unknown; \LACK{analysis does not exist.}
       \\
\bottomrule
\end{tabularx}
\label{table:autodl-questionnaire}
\end{center}
{\scriptsize\textit{
Ideally, AutoDL algorithms should be evaluated w.r.t. all ten criteria.
If a publication answers with a high quantity of thorough ``yes'' and justified ``no'' responses, it can be considered high-quality research with comprehensive consideration of all relevant issues in AutoDL.
Seminal NAS works, NASNet~\cite{zoph2018learning} and DARTS~\cite{liu2019darts}, are used to exemplify usage of this questionnaire.
All comparisons are made against the available literature of the time, i.e., July 2017 and June 2018.
We note that the questions are neither exhaustive nor final.
}}
\end{table}

\begin{table}[tp!]
\begin{center}\scriptsize
\setlength{\tabcolsep}{1pt}
\caption{
[Part-2/2] An example self-assessment questionnaire for AutoDL algorithms.
}
\begin{tabularx}{\linewidth}{ p{0.04\linewidth} | p{0.35\linewidth} | X | X}
\toprule
       & Question & Example response for NASNet~\cite{zoph2018learning}
                  & Example response for DARTS~\cite{liu2019darts} \\
\midrule
\multirow{8}{*}{{\rotatebox[origin=c]{90}{\RNC{6}: Reproducibility}}}
      & Is all relevant experimental code involving this AutoDL method publicly available?
      & No.
      & Yes; links can be found in the paper.
      \\
      & Is all relevant experimental code involving the models resulting from this AutoDL method publicly available?    
      & Partially; \LACK{the model definition and pre-trained weights can be found, but lack instructions on how to reproduce them}.
      & Yes; at github.com/quark0/darts.\\
      & Are all relevant experimental information for AutoDL-method and resulting models (e.g., hyperparameters, random seed, hardware, or platform) reported in the paper/code?  
      & Yes, decently; most hyperparameters are mentioned, and hardware and platform are mentioned.
      & Yes; most information can be found in either paper or the released code.
      \\
    %   & Are all relevant experimental values for resulting models (e.g., hyperparameters) reported in the paper/code?
    %   & Yes, mostly.
    %   & Yes. \\
    % %   & Is miscellaneous experimental information (e.g., random seed, hardware, or platform) reported?
    %   & Yes, decently; hardware and platform are mentioned.
    %   & Yes; they are either reported or can be found in the open-sourced repo.
    %   \\
\midrule
\multirow{8}{*}{{\rotatebox[origin=c]{90}{\RNC{7}: \makecell[c]{Engineering\\Quality}}}}
      & Is this AutoDL method implemented for public use (e.g., off-the-shelf scripts)? 
      & Minimally; \LACK{there are only two template scripts for evaluation usage.}
      & Yes; the search, training, evaluation scripts can be found.
      \\
      & Is the implementation code commented?
      & Partially.
      & Rarely.
      \\
      & Is the implementation code tested?
      & Yes, mostly; \LACK{however, there is a lack of detailed test reports.}
      & No; \LACK{lacks unit tests.}\\
      & Is the implementation code documented?
      & Partially; model evaluation is documented.
      & No.
      \\
      & Is the implementation code modular, reusable, and extendable?
      & Yes; different config files can instantiate different versions of NASNet.
      & Yes; the minimal search cell can be reused.
      \\
\midrule
\multirow{11}{*}{{\rotatebox[origin=c]{90}{\RNC{8}: Scalability}}}
       & Are the theoretical/practical computational costs of this AutoDL method analyzed?
       & Yes, partially; empirical costs are analyzed, \LACK{theoretical costs are not.}
       & Yes, partially; empirical costs are analyzed, \LACK{theoretical costs are not.}
       \\
       & Is it feasible to scale up to large-scale datasets?
       & No, but issue considered; sub-networks (i.e., cells) discovered for small-scale datasets can be transferred to large-scale datasets.
       & Unknown; the AutoDL method is efficient and is ideally able to scale up to ImageNet, \LACK{but analysis does not exist.}
       \\
       & Is it feasible to scale up to large-scale models?
       & Unknown; \LACK{analysis does not exist.}
       & Unknown; the AutoDL method is efficient and is ideally able to scale up to ImageNet-scale models, \LACK{but analysis does not exist.}
       \\
       & Is it feasible to scale up to large-scale search spaces?
       & Unknown; \LACK{analysis does not exist.}
       & Unknown; \LACK{analysis does not exist.}
       \\
\midrule
\multirow{7}{*}{{\rotatebox[origin=c]{90}{\RNC{9}: \makecell[c]{Generaliz-\\ability}}}}
       & Can this AutoDL method generalize well to unseen datasets?
       & Yes; from CIFAR to ImageNet and COCO.
       & Yes; from CIFAR to ImageNet, and from PTB to WikiText-2.
       \\
       & Can this AutoDL method generalize well to different target models? 
       & Yes; can generalize to different macro structures.
       & Yes; can generalize to different macro structures.
       \\
       & Can this AutoDL method generalize well to different kinds of search spaces?
       & No, but without issue; the proposed NASNet search space is intrinsic to the NAS algorithm.
       & Unknown; \LACK{analysis does not exist.}
       \\
\midrule
\multirow{7}{*}{{\rotatebox[origin=c]{90}{\RNC{10}: \makecell[c]{Eco-friend-\\liness}}}}
       & Is the energy consumption minimal for this AutoDL method?
       & Unknown; \LACK{analysis does not exist.}
       & Unknown; \LACK{analysis does not exist.}
       \\
       & Is the energy consumption minimal for the target model when actively employed?          
       & Unknown; \LACK{analysis does not exist.}
       & Unknown; \LACK{analysis does not exist.}
       \\
       & Is the potential environmental impact of this AutoDL method (e.g., converted to carbon emissions) minimized?
       & Unknown; \LACK{analysis does not exist.}
       & Unknown; \LACK{analysis does not exist.}
       \\
\bottomrule
\end{tabularx}
\label{table:autodl-questionnaire-sup}
\end{center}
\vspace{-3mm}
{\scriptsize\textit{
Ideally, AutoDL algorithms should be evaluated w.r.t. all ten criteria. Please follow the same notation as \Tabref{table:autodl-questionnaire} to read this table.
}}
\end{table}

\textbf{\RNC{1}. Novelty} is a crucial prerequisite for the publication of any research in academia.
For the most part then, this criterion does not need to be spelled out explicitly, as a strong peer-review system will control for its quality automatically.
% Every research work will be evaluated regarding its novelty, and most publications were accepted due to the novelty.
% However, a phenomenon of reviewing is: the novelty of an AutoDL algorithm is only evaluated within the narrow range of a sub-area of AutoDL.
% For example, a gradient-based NAS method is compared with other gradient-based NAS methods. In contrast, its gradient-related strategy may have already been studied in gradient-based HPO methods.
% If we take gradient-based HPO methods into account, the novelty of such NAS methods will reduce to extending a gradient-based HPO method to the architecture search space.
% To fairly evaluate the novelty of each AutoDL work and the cross-domain thinking, we appeal to the AutoDL researchers and reviewers to consider the entire AutoDL area when studying, comparing to, and discussing with related works.
% To the endless point, the comparison with the works in the broader Auto(no)ML or ML area should also be considered.
% However, at the current stage, the challenges in other Auto(no)ML/ML areas are different from that in AutoDL, most AutoDL algorithms are not motivated from them. Thus, we temporarily remove such comparison from our questionnaire.
% The four points listed in the ``Novelty'' criteria is a minimum requirement and should be extended accordingly.
% but it does not mean it is unnecessary to compare if a work is motivated 
That stated, a recurring grievance in data science and beyond is that authors will often promote the novelty of their work, as justified within a narrow scope, without awareness of what exists across related disciplines or further back in academic history~\cite{smith2009cross}.
This limited perspective can be even more extreme.
For example, a gradient-based method employed in NAS may solely be compared against other gradient-based NAS methodologies without appreciating the possible existence of associated studies in HPO.
So, whereas an author may argue that the methodology is entirely new, the novelty may technically be its application to a search space of network architecture rather than a search space of training-algorithm hyperparameters.
Of course, it would be ideal for AutoDL researchers to review parallel work in the space of AutoML/AutonoML and even further afield, but, given the eventual goal of developing an integrated end-to-end AutoDL system, one should at least be cognizant of similar approaches across all phases of the DL workflow in \Figref{fig:dl-life-cycle}.
The questions in \Tabref{table:autodl-questionnaire} reflect this bare minimum.

\textbf{\RNC{2}. Solution quality} refers primarily to the ``correctness'' of a DL model, which is traditionally the predominant focus of academic research; it is the obvious criterion to include and needs no in-depth justification.
However, it must be noted that there are many different metrics by which this quality may be gauged, often depending on the DL problem that the model attempts to solve.
For instance, accuracy often refers to classification tasks, mean average precision (mAP) is used for object detection, and so on.
Even for the same type of task, a change in context may prioritize a different objective function.
For example, maximizing top-1 accuracy may be required for high-stakes predictive analytics, e.g., skin disease classification, while top-n accuracy for $n \neq 1$ may be fine for other recommendation systems, e.g., preference prediction involving musical genres.

So, while the matter of supporting future comparisons, i.e., reproducibility, will be addressed shortly, researchers are still advised to be diligent when making their own comparisons with previous SoTA approaches.
Indeed, the notion of ``ceteris paribus'', i.e., all other things being equal, is the ideal when assessing model correctness resulting from two competing AutoDL algorithms.
Of course, some forms of equivalence, or lack thereof, are trivially clear; AutoDL algorithms intended for different phases of a DL workflow will have different aims.
For instance, NAS and HPO works construct DL models from scratch and thus focus on their baseline quality, while data-engineering and deployment methods are best judged on how they modify the correctness of existing models.
Similarly, although automated maintenance is complicated by a time dimension that makes it difficult to pick a representative solution-quality metric, related works still have the common aim of minimizing model deterioration.

Regardless, where obvious and not-so-obvious, the questions in \Tabref{table:autodl-questionnaire} urge AutoDL researchers to make model-correctness claims with an eye to both equivalence and fairness, detailing evaluation contexts in depth.
Far too often, incremental improvements are claimed as SoTA -- one may speculate as to the reasons -- without strong justification.
Alternatively, superior model accuracies may not be adequately accompanied by an emphasis on the trade-offs involved.
In actuality, it is not a great weakness for an AutoDL algorithm to lack accuracy improvements over SoTA models if the benefits of the technique are to be found elsewhere.
The NASNet publication exemplifies this in \Tabref{table:autodl-questionnaire}, proving far more impressive in reaching similar-to-SoTA accuracy values via the construction of much smaller DNNs than contemporary hand-crafted models.

\textbf{\RNC{3}. Efficiency} is arguably the next highest priority for the AutoDL community after model correctness, emphasizing how most SoTA DNNs can be considered resource heavyweights, even in the current era of hardware.
Generally, there are two primary considerations in this category: speed and memory.
In the case of an AutoDL algorithm, these two issues are represented in how long an AutoDL algorithm takes to converge upon its solution, how much run-time memory is used, how many accelerators are required, and so on.
Model training time and memory footprint are included in this assessment if part of an AutoDL process, e.g., for NAS, HPO, deployment-aware AutoDL, and certain forms of automated maintenance.
% Apart from the convergence, the efficiency usually involves the run-time memory usage of the AutoDL algorithm and the number of accelerators required for the AutoDL algorithm.
% Ideally, all of these aspects should be analyzed in the paper and summarized in the questionnaire.
% Apart from analyzing these aspects, the efficiency of an AutoDL algorithm should also be compared with the literature (such as does it beat the previous SoTA solutions).
However, some AutoDL algorithms are also responsible for inducing efficiency in a DL model, whether constructed from scratch or already in existence.
In the latter case, this may not even involve any modifications to the direct structure of a DNN; intelligent exploitation in the category of automated data engineering can boost inference speed by optimally managing datastreams.
Whatever the case, the questionnaire in \Tabref{table:autodl-questionnaire} motivates assessing the efficiency of both the AutoDL process and, where applicable, the DL outcome.
Researchers will note that the speed and compactness of a DL solution often relates to (1) its number of parameters, (2) the memory footprint related to its activations, (3) its theoretical FLOPs usage, and (4) its practical inference speed.

Finally, we reiterate what was stated when discussing the correctness of a DL model; equivalence is key.
Hardware and development/deployment environments should be diligently detailed to make any SoTA efficiency claims.
For the same consideration, theoretical analyses and practical reporting are both beneficial.
Only with a fair assessment does it become clear where, in relation to both correctness and efficiency, the Pareto front for AutoDL algorithms exists.

\textbf{\RNC{4}. Stability}, predominantly of model correctness but also of any efficiency metric, is the first criterion that, as of the early 2020s, is usually overlooked by the AutoDL community.
This is not simply due to negligence either, at least when granted the benefit of the doubt.
Given how computationally expensive it is for even a single AutoDL run, it can often seem infeasible to apply the bare minimum commonly required for a stability assessment, i.e., a statistical sampling of performance followed by a report on associated variances.
Nonetheless, the field of AutoDL suffers from this lack of statistical rigor, so much so that conclusions on SoTA performance have been challenged~\cite{li2020random}.
Metric-based analysis is thus recommended, where possible, to provide confidence in the reporting of any results.
In the same vein, there are other more advanced analyses that can further inspire trust in published research.
For instance, AutoDL algorithms that impact model structure, e.g., those related to NAS, may produce dramatically different networks.
Stability in this case becomes even more complex to assess if those models are associated with indistinguishable performances, sometimes called ``the Rashomon effect''~\cite{br01}.
In such a scenario, it is worth knowing whether the AutoDL algorithm in question is truly effective at optimization or whether the DL problem it is employed upon is similarly solvable for any network

Finally, the questionnaire in \Tabref{table:autodl-questionnaire} also includes a higher-level question related to the hyperparameters of an AutoDL approach, although we avoid that terminology so as to avoid confusion with HPO.
The point here is that the stability of an AutoDL algorithm may depend on both stochastic or deterministic elements, and it is worth assessing how sensitive the approach is to the latter, namely the settings selected for AutoML-method parameters.

% Other advanced evaluation for stability included but limited to the variance of the auto-discovered DL solution or meta-hyperparameters.
% Although the poor stability of AutoDL algorithms has became a common concern~\cite{li2020random}, as of 2021, many AutoDL works did not report the simplest stability metric.

\textbf{\RNC{5}. Interpretability} is one of the newer criteria that the AutoDL community has had to consider, reflecting the permeation of DL into broader society.
It is no longer seen as acceptable in a variety of high-stakes settings for a DL model to hide its decision-making process within the complexity of its network, even if that complexity has often been associated with the performative strengths of DNNs.
Admittedly, enforcing interpretability within a model is much more complex than acknowledging its value, and the topic is thus relatively nascent in DL~\cite{guidotti2019survey}.
Nonetheless, these requirements and associated efforts to address them have similarly diffused into AutoDL~\cite{biedenkapp-lion18a}.
% It asks for the explanations for the final auto-discovered DL solutions as well as the predictions an AutoDL system made.
% Currently, it may be too early to study this for AutoDL, because the interpretability for DL is still in an early stage~\cite{guidotti2019survey}.
% Having said that, more and more efforts have been put into improving the interpretability for AutoDL~\cite{biedenkapp-lion18a}.
% In our questionnaire, we put two simple questions and appeal the future researchers can try to answer them for their designed algorithm.
Thus, for this criterion, researchers should question whether the impact of an AutoDL algorithm and, if relevant, the DL model that it targets are both explainable.
Certain theoretical analyses can also bolster associated commentary, such as the provision of ablation studies.

\textbf{\RNC{6}. Reproducibility} is a core principle underlying the scientific method.
Its inconsistent treatment has been a common concern in general ML for a while~\cite{pineau2020improving}, and these concerns have likewise been mirrored in the field of AutoDL~\cite{li2020random}.
After all, claims that an AutoDL algorithm is accurate or efficient are pragmatically useless if they can never be replicated.
Granted, while the criterion of solution quality requires a researcher to be diligent in comparing their method to those that have come before, and the criterion of stability arguably relates to assessing repeatability within the confines of the same experiment, the onus of confirming reproducibility belongs to the works that come after.
Nonetheless, there are plenty of ways that a publication can support an ethos of reproducibility, and most of these come down to a sufficient provision of details.
% Having said these restrictions, from the perspective of author, releasing the full codes and reporting the all technical details can definitely help encouraging the reproducibility of the AutoDL community.
Accordingly, the questions in \Tabref{table:autodl-questionnaire-sup} are motivated by considerations discussed elsewhere~\cite{lindauer2020best}.
% The extended part includes more fine-grained assessment of the released codes and the reported technical details.
They essentially urge a researcher to consider whether they have provided comprehensive parameter values and ideally code for any presented experiments.
This goes for both the AutoDL process and the DL model that it targets.
So, while the NASNet publication assessed as an example does reasonably well, the lack of publicly available scripts around the operation of the NAS process is a weakness, even though it is understandable for a proof-of-principle work to focus on the end-product DL model as evidence of a well-performing AutoDL algorithm.
In any case, researchers should also consider detailing the development/deployment environments that the experiments are run within.

\textbf{\RNC{7}. Engineering quality} is a criterion that reflects the translation of AutoDL algorithms from theory to application.
While reproducibility concerns code that can be used to verify the validity of certain claims, this criterion considers whether code exists that is usable beyond the contexts that were explored.
Of course, if some form of implementation exists, experts with enough skill may eventually be able to hack it for their purposes after expending enough time and energy.
However, given that the field of AutoDL is a proponent of automation, the worthiness of an algorithm can be judged, along one axis, by how accessible its implementation is to non-experts.
How this human-computer interaction (HCI) should be managed and whether it should be allowed at all in certain contexts are topics for a separate discussion.
Here, published codebases score well if they are informative, both internally via comments and externally via documentation, e.g., user guides.
The codebase should also be robust and reliable, i.e., well-tested, and, ideally, there should be some degree of modularity that facilitates plug-and-play extension.
After all, if a long-term goal of the AutoDL field is to have a fully integrated end-to-end system, the standard approach for achieving this is by fashioning together distinct pre-existing mechanisms that focus on different phases of a DL workflow.
% is a missing evaluation criterion for the current community.
% The preliminary goal of programming is to instruct machines to do the same as the designed algorithm.
% Beyond that, its value is to let other people use, learn, modify, and improve.
% These benefits require the high quality of codes, including good documentation, reliability, maintainability, portability, reusability, and testability.
% It is worthy to evaluate the codes associated with a (or multiple) research algorithms.
% Notably, even if a code repository has good reproducibility, it does not mean it has good engineering.
% Some reproducible codes are badly structured -- that may be unmaintainable, mislead the users, or waste users' time to fully understand the structure in order to make any modification.
% To solve these issues, we take the first step towards \textit{engineering-friendly} AutoDL, which introduced a few checkpoints for the AutoDL participants to self-assess their released codes.
% To quantitatively evaluate the quality of an AutoDL code repository, \Tabref{table:autodl-questionnaire} listed the questions for the documentation, off-the-shelf scripts, unit tests, and some other aspects.
% To some extent, these checkpoints quantitatively evaluate a code repository's quality.
% In this way, we can compare two pieces of code in a more fair fashion. The response for these checkpoints can also be used as a quality indicator for the novice, when they vacillate between hundreds of public AutoDL repositories on GitHub.

\textbf{\RNC{8}. Scalability} is the first measure of extensibility promoted by this monograph, denoting how versatile an AutoDL approach is for standard variations of its default usage.
Certainly, for proof-of-principle research, it is usually fine to test the operation of an algorithm on a small dataset, a small target DL model, and a small search space.
However, real-world scenarios can face numerous datastreams to process, an extreme depth of network to capture all relevant features, a multitude of adaptive mechanisms to select from, and so on.
Thus, it is worth asking whether an AutoDL mechanism is realistically suitable for scaling up in terms of data, model, and search space.
Given that this is generally a matter of efficiency, \Tabref{table:autodl-questionnaire-sup} questions whether computational costs have been analyzed, e.g., formalized in ``Big O'' notation.
% As these questions are strongly associated with the computational costs of the AutoDL algorithm, we also include the evaluation for the cost in the questionnaire.
% AutoDL algorithms should be scalable to the size of the base DL model, the number of hyperparameters, and the training steps for optimizing a DL model.
% Early gradient-based approaches involve the expensive computation with the Hessian matrix or yield the quadratic time complexity to the number of hyperparameters or square to the model size~\cite{maclaurin2015gradient}.
% Although later approaches alleviated this problem, different levels of approximation are introduced.
% A common strategy is to truncate the thousands of gradient descent steps for a moderns DL optimization algorithm into one step~\cite{luketina2016scalable,dong2021autohas} or a few steps~\cite{shaban2019truncated}.
% However, Wu~et~al.~\cite{wu2018understanding} pointed out such truncation caused a serious bias problem, such as choosing a small learning rate.
% Similarly, different approximation strategy has theoretical and empirical issues.
% Therefore, it is necessary to experiment with scaling the AutoDL algorithm until it fails at a certain level of scale (at least, we should not just conduct experiments on small-scale datasets). Many AutoDL algorithms still lack experiments in the large-scale setting.

\textbf{\RNC{9}. Generalizability} is the second measure of extensibility promoted by this monograph, denoting how versatile an AutoDL approach is for non-standard variations of its default usage.
It mostly mirrors the criterion of scalability in that the focus is on dataset, target model, and the search space relevant to the algorithm.
However, this time the proposed questions are more open-ended, gauging across these three aspects how broadly applicable a proposed approach is.
There is of course no issue with an AutoDL method being specialized, e.g., the NASNet example in \Tabref{table:autodl-questionnaire-sup} is intrinsically wedded to a particular search space.
Nonetheless, this criterion acknowledges that there is always value in general-purpose mechanisms, assuming the trade-offs, usually efficiency, are reasonable.
% Can the AutoDL algorithm work well on different datasets, different base DL models, different kinds of search space?
% For the pespective of dataset and base DL model, the generalizability can be divided into two parts: can the search algorithm generalize well and can the auto-discovered DL solution generalize well?
% Evaluating the generalizability requires answering these questions.
% However, the literature works usually fail to analyze all of them.

\textbf{\RNC{10}. Eco-friendliness}, as the final criterion suggested in this review, is a very new topic within data science, let alone AutoDL~\cite{patterson2021carbon}, that reflects the current concerns of global society.
While traditionally ignored for the most part, the evolution of computational hardware has made this a much more pressing consideration; greater processing capabilities have not been able to dissociate from greater energy expenditures.
Certainly, this is a problem as of the early 2020s due to the world relying heavily on non-renewable resources for power generation.
Unnecessary energy usage exacerbates the current global warming crisis via carbon emissions.
Moreover, these effects are argued to become non-negligible once large-scale SoTA models are being trained and operated across extensive computer clusters by many developers and users, respectively.
However, to be fair, a genuine assessment of this criterion is complex and comes down to trade-offs.
Any negative evaluation of a target DL model should also not be doubled up for the AutoDL process, and vice versa.
While we elaborate on this discussion in \Secref{sec:final-discussion-evaluation}, it is sufficient here to recommend that researchers attempt separate power-usage records for the AutoDL algorithm and the target DL model.
Estimating environmental impact is also worthwhile, even if simply applying a quick power-to-emission conversion based on a cited source.
% a new research direction in the AutoDL community~\cite{patterson2021carbon}, while it increasingly attracts more and more attention.
% Since the computational costs of ML, DL, and AutoDL have grown rapidly, it comes to a huge energy cost that may have a significant and potentially negative influence on the earth's environment.
% The energy cost to train an AutoDL or DL algorithm is undoubtedly large, while training is not the end of its lifetime.
% The more important thing is to deploy the trained model and use it to make the world become better. To this end, our questionnaire also asks for energy consumption after deployment.
% The early work studied the carbon dioxide equivalent emissions (CO$_{2}$e) of large-scale NAS experiments~\cite{patterson2021carbon}.
% These studies are based on the industrial-level system and professional tools, which may not be accessible to every researcher.
% To be responsible for the economic, environmental, and social impact of research, green AutoDL is desired.

In summary, this section has discussed the rationale behind ten criteria and associated example questions, in \Tabref{table:autodl-questionnaire} and \Tabref{table:autodl-questionnaire-sup}, that are designed to foster a comprehensive evaluation of an AutoDL algorithm, especially when considering its place within the field.

As a final point, the criteria have been ordered to reflect the mental process a reader may have when newly encountering an AutoDL publication. Such a sequence of thoughts may be as follows:
``Have I seen this stuff in the abstract before? No? Ok, I will skim the paper. Does the proposed approach do a good job? Yes? Well, does it take forever to run? No? Ok, I will read further. Does it do its job consistently? Yes? Well, can I understand how the job is being done? Still yes? Seems too good to be true. I would not mind testing the results myself. Can I do this easily? Yes? I am impressed. Well, is there an implementation that I can use for my own applications? Yes? Can I scale it up? Yes? Can I use it for something else? Yes? My power bill (and the world) will survive this, right? Yes? Wow. This truly is an amazing AutoDL algorithm.''
% \dxy{Do we need a summarization for this sub-section?}
% These questions fit for most AutoDL algorithms
% As shown in the NASNet example in \Tabref{table:autodl-questionnaire}, it is 
% Additionally, these ten aspects could also serve as assessments of research work for the AutoDL participants, such as reviewers. To this end, a comprehensive AutoDL specific questionnaire is proposed in the appendix.

\subsection{An Overarching Evaluation of AutoDL}\label{sec:final-discussion-evaluation}

Having dived deep into the criteria that form the basis of the assessment framework within this review, this section now zooms out once more for a final high-level perspective of AutoDL as of the early 2020s.
Many of the following discussion points, again sorted by criterion, arise when considering all the overviews within the previous sections in an aggregate manner.

But first, it is worth a reminder that, with respect to the workflow-based reconceptualization of AutoDL in \Figref{fig:what-is-autodl}, around three-quarters of surveyed work relates to model development, and the majority of this concerns NAS specifically.
In essence, and as with AutoML~\cite{kedziora2020autonoml}, there certainly are intensifying conversations around just what the scope and long-term goals of AutoDL are, given its organic and unplanned evolution, but it must be acknowledged that the field is still dominated by a single sub-category.
The AutoDL story is primarily a NAS story.
At the same time, it is arguable that, because the other phases of the DL workflow are less well-explored, they may be more fertile for ongoing progress.
With that context addressed, we now assess AutoDL through the lenses of particular criteria.
% Our questionnaire \Tabref{table:autodl-questionnaire} identified tens of key aspects to evaluate an AutoDL algorithm.
% With the help of this questionnaire, we can make an overarching evaluation for the current status of AutoDL.

\textbf{\RNC{1}. Novelty.}
An obvious take-away from \Tabref{table:autodl-dissect}, which lists seminal works in AutoDL, is that, excluding outliers, the field only truly became consolidated within the mid-to-late 2010s.
Additionally, although the scattering of research makes conclusions difficult, it is reasonable to claim that NAS and DL-related HPO started up in about 2016, while the extension of the automation ethos into the data-engineering and deployment phases followed in around 2017/2018.
As for automated maintenance, there has definitely been adaptation experimentation performed in parallel, but this research thread is not necessarily as advanced as the others; we are not aware of any DL model-adaptation mechanism that has the same standard of applicability.

Unsurprisingly though, elements of all AutoDL research are often inspired by previous work, some of it even external to the field.
This is most obvious when the automated approaches in \Tabref{table:autodl-dissect} are treated as computationally processed optimization problems, dissected into search space, search strategy, and boosting method for candidate evaluation.
% From the perspective of search strategy, the fundamental methods in AutoDL mainly came from the research in the field of HPO.
% The search strategies in other sub-areas either directly use the existing strategies in HPO or made a little modification by injecting prior human knowledge about specific tasks.
It follows that the foundations of all search strategies are well established, having been utilized for decades with the broader ML sphere, as indicated by \Tabref{table:hpo-methods}.
This means that, details aside, there is a convenient familiarity when working with, for example, gradient-based methods in both NAS~\cite{liu2019darts,dong2019search} and DL-based HPO.
% Having said that, not all search strategies in HPO are originated from itself, in which RL and evolution-based solutions mainly came from the broader RL and evolution areas.

Search spaces, on the other hand, are far more diverse, differing even in the abstract between various phases of a DL workflow, if not sub-categories.
For instance, automated data augmentation may sample sets of transformations, hardware search might optimize against series of programmable gates, and setting up low-shot learning for quick automated adaptation might have a search space of parameter-update rules.
Given the importance of good search space design for optimal performance, with the possible exception of strategy-focused HPO, a lot of innovations often arise in this facet of AutoDL research.
Granted, in research threads that have sufficient publication density, certain seminal works set the standards for such design, e.g., NASNet-like spaces~\cite{zoph2018learning}, etc.
These are usually tweaked by subsequent proposals rather than entirely replaced.
Nonetheless, there is plenty of experimentation in the aspect of search spaces, even for NAS, and breakthroughs that improve the quality of AutoDL algorithms are possible.
% From the perspective of search space, every sub-AutoDL area designs its own search spaces due to its distinct nature.
% The search space in automated data engineering, NAS, and auto-deployment is quite important. A good search space design in these three areas usually equals better final performance.
% Several seminal works~\cite{zoph2018learning} \dxy{more refs} laid the foundations of search space design for these areas, which have become the most popular search spaces.
% Most existing works directly only made a little modification by removing useless parts or adding a few customized components.
% There are also a few works exploring to design some completely new search spaces, which may potentially bring some breakthrough in the future but is still at a very preliminary at the current stage.
% As for HPO, only a little attention has been paid to the study of search space, as the main focus of HPO is search strategy.
% As for automated maintenance, the search space design depends on the scenarios. There are too many different scenarios, and each of them may apply a different search space.

Tactics for managing candidate evaluations are also a prime target for innovation, almost entirely driven by a focus on efficiency.
Indeed, search spaces may differ across numerous scenarios, but almost every type of AutoDL algorithm needs to evaluate its target DL model after modifying the DL process in whatever way it does, otherwise there is no objective function to improve against.
The obvious exception to this rule is filter-type data engineering, where features are transformed to optimize inherent properties of a dataset without involving the model in any way.
However, in general, the point stands; training a DNN per iteration of an optimization is computationally heavy, and many novel techniques are regularly proposed to mitigate this.
% Boosting candidate evaluation is the main focus of efficient AutoDL.
% The expensive computational cost of executing a DL program is the main barrier to efficient AutoDL. Thus researchers are exploring how to assess the performance of a DL program quickly.
Nonetheless, as with search spaces, there are recurring types of approaches.
Low-fidelity approximations, which may involve trimming some amount of data or reducing the training-time budget, are perhaps the oldest.
Weight sharing is an alternative that is much more preferred for the modern DL model, especially given its scalability, which is ranked high in \Tabref{table:nas-methods}.
It is very popular in NAS~\cite{pham2018efficient} despite its recent entrance into the field.
Yet, once again, one could argue that weight sharing is not particularly novel, having been implicitly utilized within the gradient-based HPO method introduced at least as early as in the 1990s~\cite{larsen1996design}.

In truth, AutoDL is often both a reappropriation and customization of existing concepts and methodologies, frequently applied with ingenuity, and thus novelty should always be assessed in that light.
% Generally speaking, most techniques are motivated by other non-AutoDL areas and are customized in the context of AutoDL.

\textbf{\RNC{2}. Solution quality.}
Current benchmarks for DL-model accuracy and related metrics are dominated by NAS, often in combination with other techniques.
It is clear that automatically searching for DNNs can regularly outperform purely handcrafted networks, even if these differences become less compelling for DL tasks where the limits of model correctness are already approached.
In such cases, the utility of an AutoDL algorithm becomes more reliant on whether the high accuracy is consistent or cherry-picked; this is a matter of stability.
But the point remains that AutoDL has proven value in producing SoTA models.

There is, however, a broader discussion to be had on how solution quality is benchmarked.
A common notion is to partake in a comprehensive leaderboard for a popular dataset, such as ImageNet for vision~\cite{russakovsky2015imagenet}, SQuAD for NLP~\cite{rajpurkar2016squad}, and LibriSpeech for speech recognition~\cite{panayotov2015librispeech}.
This does allow for direct comparisons between the outcomes of competing AutoDL algorithms.
However, such datasets may be amenable to certain characteristics of an algorithm that are useless or detrimental for other datasets, particularly in real-world applications.
Unfortunately, there is no good solution here at the current time, given that benchmarking is an arduous process for large-scale AutoDL.
It is simply worth highlighting that portions of the DL community put excessive weight on certain evaluations that may not be fairly representative of general solution quality.

In the other extreme, and with supporting evidence, there does exist skepticism around the performance of many NAS algorithms, e.g., whether they are better than random search~\cite{yu2020evaluating}.
While this concern relates to issues of reproducibility, it does perhaps support another assertion that this survey makes: the most important contributor in NAS to model accuracy is search space design, not search strategy.
In saying this, it should also not be lost on the reader that the current search spaces yielding best accuracy are derived heavily from popular architectures that are manually designed.
It is unclear whether there are better options out there, but this does suggest that, while humans cannot typically beat a NAS algorithm in selecting an optimal candidate DNN, their expert knowledge on general structure is not entirely worthless.

At this current stage, it is not clear whether these emergent principles are the same for other phases of a DL workflow, e.g., whether constraining a subset of FPGA configurations is more worthwhile than applying a different optimization method to hardware search.
Admittedly, the reason expert knowledge is so useful in NAS may simply be because the appropriate search space has been arrived at after plenty of human exploration; the benefit of expert knowledge may not be available in other non-NAS contexts.
Regardless, while AutoDL research is generally more rudimentary along the rest of the DL workflow, it has been shown that distinct mechanisms can work in concert to push solution quality even further.
% Taking one step back, NAS is just a single component in the entire DL pipeline.
% The SoTA accuracy should be achieved by considering every component in the DL pipeline.
For example, a recent work achieved 87.3\% accuracy on the challenging ImageNet dataset by combining, with the aid of manually designed hyperparameters, automated data engineering with NAS~\cite{tan2021efficientnetv2}.
This reaffirms the value of pursuing an integrated end-to-end AutoDL system, even if no such empirically sound framework currently exists.

% A compromise solution is to sequentially optimize each component~\cite{tan2021efficientnetv2,dong2021autohas}. However, some 

\textbf{\RNC{3}. Efficiency.}
While accuracy is always an important goal for any form of ML, the ethos of efficiency is particularly compelling for AutoDL as opposed to AutoML, given the computational expense that arises from embracing the complexity of a DNN.
Thus the drive for speed and, where applicable, compactness has traditionally come second only to solution correctness within AutoDL literature.
In fact, the priorities may have even swapped over the last couple of years due to, one, diminishing communal interest in incremental accuracy improvements and, two, a shift in requirements as AutoML and AutoDL are progressively translated from academia to industry.
The broader implications of this translation are worth discussing elsewhere.

As aforementioned, the push for efficiency applies to both the target DL model and the AutoDL process.
Given that the latter is usually responsible for training and retraining a DNN, model-development efficiency is usually bundled in as part of the AutoDL processing time.
However, the impact of the AutoDL process often outlives its operation and still needs to be accounted for, e.g., a deployment-aware model tuner has good value if, one, it works fast and, two, if the resulting model generates its predictions on a deployment environment in a timely manner.
In effect, AutoDL research aims to account for theoretical/realistic inference speed and memory footprint as well.
% Despite the performance side, the efficiency of the final DL model is also an important aspect.
% It can refer to the theoretical or realistic speed of a DL model, the training convergence rate, the memory footprint of activations or model weights, the real-time factor, etc.
% In the context of AutoDL, the commonly used metric is the number of hours and computational resources required to complete an AutoDL algorithm.
% Although other aspects of efficiency are usually overlooked in the literature, we should take a deep analysis on them due to their importance in real-world scenarios.
Regardless, simply in terms of AutoDL algorithms, the field has steadily made progress, at least in terms of preferred strategies.
For instance, NAS has shifted from low-fidelity approximations to weight sharing, HPO has done the same but into hyper-gradient approximations, automated deployment prefers neural predictors over deployment simulation, and so on.
% Theoretically speaking, all of these techniques can be applied in each AutoDL sub-area, while the popularity of these techniques in each sub-area is different due to their unique challenges.
% Not surprisingly, these efficient techniques are in different degrees of approximation to the vanilla solution (fully training and evaluating the DL model).
These transitions do not guarantee their benefits though, as efficiency boosts of any form often come with trade-offs, usually in the form of assumptions weakening solution quality.
It is thus a current focus within the AutoDL field to work out not just how to shift along a Pareto front but to push it out further.

\textbf{\RNC{4}. Stability.}
The impressive achievements of AutoDL in terms of both solution quality and model-specific efficiency have been tempered by a lack of trust in the results.
This cynicism is partially driven by deficiencies in published scientific practice that go beyond the scope of this monograph, but certain limitations are also somewhat understandable in the context of DL due to the computational resources required.
The limitation specific to this criterion is that many works do not go to the extra effort of establishing a statistical analysis for presented results, which makes it difficult to grapple with the stochastic nature of an AutoDL algorithm.
That stated, many publications are presently appealing for the reporting culture to be improved~\cite{lindauer2020best,li2020random,dong2020bench,ying2019bench}, and an evaluation of stability is starting to be seen within an increasing amount of AutoDL works.

\textbf{\RNC{5}. Interpretability.}
There is not much to say on this matter in AutoDL, given how the broader DL community is itself debating how best to instill some form of transparency in the DL process.
However, the absence of substantial focus in this area will likely need some serious consideration at some point.
Unlike AutoML, which is able to operate with explainable ML models, AutoDL is wedded to inherently complex connectionist structures.
Given that the automation of any process is often associated with another layer of black-boxing, AutoDL may struggle for democratization if it does not somehow avert undue technical obscuration.

\textbf{\RNC{6}. Reproducibility.}
Ensuring that an AutoDL algorithm is repeatable faces the same computational-resource challenges as stability, only this time the onus is on peer review to confirm it.
Thus, if a proposed method is to make a splash within the community, the proposer would be advised to facilitate such tests.
Unfortunately, this is another widespread deficiency in AutoDL research.
For instance, a recent investigation evaluated 12 NAS algorithms and found that most did not report enough details to fully reproduce the original results~\cite{li2020random}.
This concern has been echoed by many other researchers~\cite{lindauer2020best,dong2020bench}.
In fact, some have proposed potential solutions within the scope of NAS, such as comprehensively filling out a questionnaire~\cite{lindauer2020best} or enabling the approach to be easily benchmarked~\cite{ying2019bench,dong2020bench}.
Early signs indicate that such appeals may be fruitful, with a steadily increasing proportion of new NAS papers providing both code and details.

Repeatability issues are unsurprisingly mimicked in other phases of a DL workflow, not just NAS.
In general, they are somewhat alleviated by benchmarking suites~\cite{eimer2021dacbench,eggensperger2014surrogate,klein2019tabular}.
However, while these works encourage good practice in researchers, many do focus on small-scale models and datasets.
It thus remains an open question as to how the reproducibility of large-scale experiments, especially those requiring advanced hardware systems~\cite{real2020automl,real2019regularized}, can also be guaranteed.
Indeed, given their associated requirements for computational resources, there may be only two or three groups in the world that have the capability to do such experiments, and this does not even consider other limitations such as business restrictions or the availability of human expertise.

\textbf{\RNC{7}. Engineering quality.}
The upper echelons of DL experimentation are only accessible to entities with substantial computational resources, such as universities or, often more so, large corporations.
Nonetheless, DL has permeated into broader society and, just as AutoML has been embraced by hobbyists and commercial vendors, AutoDL is likely to follow suit over time.
There is, however, many differences between code scripted in academic settings and software written for real-world usage.
Technical debt is not too much of an issue for one-off experiments, but it can cause substantial problems for real-world AI systems~\cite{sculley2015hidden}.
Moreover, the risks are an order of magnitude greater for AutoDL than DL, simply due to its complex higher-level nature; the former needs more advanced software/library supports.

Granted, as discussed earlier, certain features of AutoDL algorithms may be well-established, such as black-box/gray-box optimization libraries, which have a long history of developmental improvements~\cite{akiba2019optuna,golovin2017google,song2022automated}.
In contrast, if AutoDL algorithms rely on more novel concepts, such as recent popular white-box optimization methods, the support may not be there.
Fortunately, given the size of the AutoDL community, there is a good possibility of an implementation arising sooner or later.
For instance, recent works have started to look into this issue for white-box AutoDL algorithms, providing examples as well~\cite{zhang2020retiarii,erickson2020autogluon}.

There are a number of new questions to ask once computational theory crosses over into software.
Two of them are as follows: can AutoDL algorithms already implemented in these libraries be easily maintained, and can these libraries easily support the implementation of future AutoDL algorithms?
Certain researchers doubt it, concerned about the practical modularity of such algorithms~\cite{real2020automl,peng2020pyglove}.
Worse yet, some high-level AutoDL libraries are tightly coupled with their associated low-level DL library~\cite{zimmer-tpami21a,jin2019auto}, and this constrains their utility.
To address these issues, the development of newer libraries has started to explore the use of symbolic programming to keep dependencies minimal~\cite{peng2020pyglove,pytorch_v1.10}.

In short, there is currently an impressive amount of activity in the AutoDL community on the engineering side.
However, to date, no AutoDL packages have cemented themselves in popular use, e.g., in the same vein as PyTorch or Tensorflow for DL.
It is also not clear how best to quantitatively compare AutoDL libraries beyond the minimum requirements proposed in \Tabref{table:autodl-questionnaire-sup}.

\textbf{\RNC{8}. Scalability.}
Given that this matter relates to efficiency, the ideal case for usability is if the time and space complexity of an AutoDL algorithm is linear, logarithmic, or constant with respect to the size/quantity of data used, the size of the target DL model, and the relevant search space.
Now, at a baseline, the model-evaluation element of an AutoDL process is well controlled, as model training, if relevant, will often involve linear factors in time and/or memory with respect to data inputs and number of model layers.
However, it is the search element of the process, whatever that may be according to the phase and sub-category of a DL workflow, that can easily blow out.
For instance, for auto-augmentation, the decision to search for an $n$-length pipeline of data transformations can exponentiate processing time by $n$.
Similarly, NAS can likewise face this exponentiation by simply increasing the depth of a target DNN to construct.
Thus, either with conscious or subconscious focus on scalability, AutoDL research does often grapple with how to make search manageable, e.g., by using repeatable NAS cells, even it remains a point of debate whether such constraints block out better solutions~\cite{hula18}.
% However, oftentimes, the overall complexity is not easy to analyze, especially, some factors in the complexity need to adaptively increase regarding the increasing of search space.
% For example, when the base model became deeper, the size of search space can exponentially increased, which consequently requires evaluating much more candidates.
% Therefore, not every efficient AutoDL algorithm is scalable.
Whatever the case, this issue will remain a challenge for AutoDL, and rigorous analysis of scalability must be done on a case-by-case basis.

\textbf{\RNC{9}. Generalizability.}
Like interpretability, this is a concept about which there is currently little to say within AutoDL as it is rarely rigorously considered.
This is perhaps because the field of AutoDL is currently preoccupied with making algorithms work well in an intended context, rather than extending them to work in other contexts.
% is an important but largely overlooked aspect in AutoDL.
This means that most AutoDL algorithms rely heavily on context-dependent assertions, especially white-box methods, which are preferable for their superior efficiency.
The downside is that these are then difficult to transfer from one task to another, e.g., visual classification to speech recognition, let alone from one phase of a DL workflow to another, e.g., from NAS to searching for adaptive mechanisms.
Granted, certain elements are more reusable than others, such as modular black-box optimization methods, but the state of computational hardware is not at a level where, on the whole, generalizability trumps efficiency.
% are usually customized for a single component in the DL pipeline.
% Almost no white-box-based AutoDL algorithms are capable of handling every kind of search space, not to mention automating the design of every component simultaneously.

\textbf{\RNC{10}. Eco-friendliness.}
This notion has quickly attracted increasing attention within the DL community, even if the associated body of published literature currently remains small~\cite{patterson2021carbon,li2021full} and many debates exist.
When contemplating this matter, in the most general and ideal sense, it is advisable to consider what level of energy usage is most appropriate for a computational task.
For instance, the diminishing returns and increasing power required to farm bitcoins is occasionally debated in the media, often accompanied by a dim view of the related profit-driven intent.
On the other hand, DL tasks with broad social benefits, e.g., improved health outcomes, would generally be considered worthy of a higher resource expenditure.
So, there is a baseline level for the energy value of every computational process, even if this is considerably difficult to quantify and highly subjective.
Then perhaps the key to considering whether AutoDL is eco-friendly should be based on counterfactual considerations: what would happen if the same process was not automated?

Most AutoDL processes are done once, whereas their resulting impact on a target DL model can be felt for the duration of its deployment.
If the kilowatt-hour cost of the algorithm is $y$ and this reduces the cost of executing a model prediction by $x$, to be done $k$ times over, then the eco-benefit of AutoDL is felt for $k>y/x$.
Admittedly, the rules change for a persistent AutoDL process, such as continuous monitoring and maintenance.
In such a case, the costs of adaptation must be weighed up more carefully~\cite{zlbu15}.
Regardless, analysis in this area is made challenging by how many levels of indirection exist between a line of code and a carbon emission.
Different platforms are optimized in different ways, every hardware system appears to have its own unique setup, and checking power usage at a socket is complicated when AutoDL operations are done remotely or in a decentralized manner.
Politics may obscure matters further, especially for corporations that are most likely to have the resources to cause significant impact, and even national grids vary in their proportion of renewable energy.
So, while there is an increasing drive towards eco-friendly solutions in the community, it is unlikely an associated metric will be benchmarked widely any time soon.

% In the scope of AutoDL, only very few works analyzed the carbon emission and energy cost of some NAS algorithms~\cite{patterson2021carbon,li2021full}, and many debates exist.
% For the AutoDL algorithm itself, many efficient algorithms are on the same par as the normal DL algorithm.
% The key lies in the impact of the final auto-discovered DL solution.
% As the productionalized DL solution will be executed millions of times, small improvements on the reduction of carbon emission and energy cost may finally be accumulated to have a significant impact on the environment.
% An issue is that each DL solution at different companies and productions is executed in a diverse way, and it is difficult (such as due to business confidence) to accurately quantify such impact.
% In addition, although deployment-aware AutoDL algorithms can generalize to find some eco-friendly DL solutions, such extension is still in lack.

\subsection{Challenges and Opportunities}\label{sec:future}

While the previous section emphasized that the AutoDL story has, up to this point, primarily been a NAS story, it does not have to remain that way.
In fact, the abbreviation of AutoDL exists to make that distinction, acknowledging that automating high-level operations for DL does not end at constructing a neural network, even if NAS will always remain a core facet of the endeavor.
Thus, given how nascent the entirety of the field is, the AutoDL community, however it continues to self-assemble, faces an arguably exciting opportunity to decide what matters most to it within the coming years.

Here, we briefly summarize the five most likely driving forces for future research, as extrapolated from the outcomes of this review.
% The challenges of AutoDL come from the unsatisfied assessments of the ten criteria mentioned in \Secref{sec:final-discussion-evaluation}.
% Based on the assessments, we summarized the following four challenges: 
They are: (1) democratized AutoDL, (2) large-scale AutoDL, (3) persistent AutoDL, (4) integrated AutoDL, and (5) self-assembling AutoDL.
The listing is in order of least-to-most challenging, although any such forecast is naturally subjective.

\textbf{Democratized AutoDL.}
This is the path towards usability.
It relies on making AutoDL approaches as accessible to the general public as possible.
In principle, it is also the easiest path to progress along, as most of the challenges here are organizational, related to engineering quality, rather than fundamental.
Certainly, there are already several AutoDL-related libraries in existence that have high-quality codebases~\cite{peng2020pyglove,zimmer-tpami21a,NNI,erickson2020autogluon}.
However, implementations are usually scattered and ad hoc; there is currently no automatic way to evaluate AutoDL code, even though this review has motivated several pertinent questions in \Tabref{table:autodl-questionnaire-sup}.
There is also a challenge to define protocols that can appropriately support AutoDL experiments and maintain related codebases.
It does not serve the community well if every new AutoDL algorithm is released in an isolated library that is abandoned after a year.
In short, as is progressively being shown for AutoML, there is plenty of interest in industry to leverage the capabilities of AutoDL, and plenty of real-world impact that can be achieved, if only AutoDL was easier to use.

\textbf{Large-scale AutoDL.}
This is the path towards capability.
It relies on finding ways to make AutoDL algorithms more powerful, capable of grappling with larger datasets, target models, and search spaces.
The driving impetus here is as for democratization, specifically the potential for broad real-world usage beyond academia.
However, while this path does not directly concern itself with who is employing AutoDL, it is concerned with what AutoDL will be employed upon, namely DL problems that do not hold back in scale.
Granted, the automation of data preparation, architecture design, and deployment has already shown some good progress on large-scale classification datasets, and this is why such a direction is not profoundly challenging.
Nonetheless, it is an ongoing process to push the power of AutoDL further without relying on computational resources to evolve in the background.
Simply put, intelligent search space design and other cost-cutting tricks will need to be developed even further to extend the capability of AutoDL.

\textbf{Persistent AutoDL.}
This is the path towards autonomy.
It relies on developing robust AutonoDL mechanisms that can adapt DL models, in the absence of human control and oversight, to time-dependent performance deterioration.
We stress that this is a weak form of autonomy, in that the AutonoDL system is still performing the DL task that a human requires of it, but the system is nonetheless autonomously improving and maintaining its predictive/prescriptive ability.
Of course, as this review has shown, the AutoDL community has not, by and large, put their focus on dynamic data environments; learning well in static environments is challenging enough.
This is clear from the computational resources required for regular updates, as well as the fundamental challenges associated with connectionist structures, e.g., catastrophic interference.
However, it is a major shortcoming to ignore this endeavor in the long run, especially as current global events have shown how detrimental it is to rely on fragile static models.
On the plus side, research activity into adaptive learning, both recent and historical, has been both intensive and extensive, despite its scattered nature.
Present efforts in topics such as domain adaptation and low-shot learning facilitate speedier model retraining for new contexts, and many proposals exist on how to elevate DL into a state of lifelong learning~\cite{pake19}.
It will naturally take a concerted effort by sections of the AutoDL community to push forward an AutonoDL agenda, but the theoretical foundations are there to mitigate inherent challenges.
% Most AutoDL algorithms focused on non-continuous scenarios -- auto-discovering a good DL component and re-run the entire DL workflow to get the final DL solution.
% When it comes to the continuous scenario, we need to monitor the quality of the DL solution over time and continuously adapt it to maintain its health.
% AutoDL algorithms that work well for the non-continuous scenarios can be applied to the continuous scenarios without or with little modification.
% The question is did they work?
% \dxy{Some useful references:
% Early work at 1995 mentioned that models catastrophically forget existing knowledge when learning from novel observations\cite{thrun1995lifelong}.
% Here are some popular solutions to overcome the catastrophic forgetting for DL.
% Regularization -- imposing the constraints during optimization. Example as \cite{kirkpatrick2017overcoming}, that used a quadratic penalty on the difference between the parameters for the old and the new tasks that slows down the learning for task-relevant weights coding for previously learned knowledge.
% Memory replay -- using extra memory to store knowledge of history. \cite{hinton1987using} proposed in 1987 tried to use a plastic weight with slow change rate which stores long-term knowledge and a fast-changing weight for temporary knowledge.
% Dynamic architectures -- changing architectures in response to new information learned from new tasks.
% }.

\textbf{Integrated AutoDL.}
This is the path towards consolidation.
It relies on researching and developing effective frameworks that are able to automate the entire DL workflow in \Figref{fig:dl-life-cycle}.
It is a significant step up from existing works that usually only focus on one phase of the workflow, typically model development, even though it is becoming clear that, in terms of both usability and even performance, combining multiple mechanisms can be optimal~\cite{tan2021efficientnetv2}.
% Current AutoDL algorithms are either designed or experimented on the automation of a few DL components.
% The challenges for an end-to-end AutoDL lie in two parts: how to construct a practical and reasonable scenario for the research of AutoDL, and how to maximize the generalizability ability of AutoDL to handle every DL component simultaneously?
Of course, the challenges of designing integrated architectures are numerous and have been discussed at length in the context of AutonoML~\cite{kedziora2020autonoml}.
Admittedly, ensuring that the constituent mechanisms are sufficiently general is a novel one to AutoDL.
Perhaps, the design of an overarching framework should contain a switching capability that activates white-box algorithms for when DL problems are amenable, and black-box/gray-box algorithms for where white-box assumptions fail.
Whatever the case, installing and using a single monolithic AutoDL system, rather than haphazardly cobbling together poorly interoperable mechanisms, remains an appealing goal within the field.

\textbf{Self-assembling AutoDL.}
This is the path towards emergence.
It relies on reassessing the handcrafted nature of AutoDL approaches themselves, seeking alternative ways to evolve better methods from scratch.
As of such, it is the most speculative pathway on account of being a paradigm shift in the way that AutoDL is researched.
Nonetheless, it is not a whimsical notion; many self-assembly algorithms are very robust, as they typically do not internalize assumptions about a solution space, i.e., in this case, what an AutoDL method should look like.
Of these mechanisms, many researchers are most familiar with evolutionary procedures, typically involving population-based mechanics to evolve solutions, but there are also alternative forms of ``algorithmic growth'' that are based on iteratively unfolding complexity to meet the demands of a task.
These latter processes have been discussed in the context of biological brain development~\cite{hi21}.
Whatever the mechanism, the idea is that, constrained primarily by the primitives involved, e.g., basic functions, self-assembling AutoDL will gravitate towards whatever works well.

Now, there are obvious reasons why self-assembling AutoDL is considered the most difficult goal of the five listed in this section.
A major prohibiting factor is computational complexity, made strongly evident when a DL model-optimizer is itself optimized.
It is also largely unnecessary to consider holistic self-assembly when AutoDL is still focused on individual mechanisms, rather than how best to mesh them together~\cite{kedziora2020autonoml}.
Nonetheless, existing works have explored this direction in limited domains~\cite{real2020automl,liu2020evolving}, and the idea is attractive.
While human expertise can prove beneficial, in that prescribed algorithms and search spaces are often speedy warm-starts towards good solutions, there is no guarantee that human biases do not obscure the very best solutions.
Indeed, perhaps superior NAS algorithms combine unusual optimization methods with bizarre search spaces and unpredictably delegated model-evaluation boosts.
A human would likely never consider this, but a suitable self-assembling process may well stumble upon an emergent design pattern that is both novel and effective.
So, although there are many more pressing aims for the field of AutoDL, such as consistently decent functionality, it would be remiss to not acknowledge that the high-level algorithms may \textit{themselves} turn out even better if sought out by an automated process.
% aims to automatically discover the DL workflow just using basic mathematical operations. As the current solution quality is bounded by the search space design, AutoDL-zero can help enlarge the scope of the current search space and reduce the human bias.
% Consequently, it can potentially find something completely different from the existing DL component, which may bring a significant gain. However, a trade-off is the enlarged search space would dramatically increase the search difficulty.
% Existing works have explored this direction in limited domains~\cite{real2020automl,liu2020evolving}. The challenge is how to enable such idea to more practical scenarios and outperforms the SoTA performance.

% In the current stage, there is no ``perfect'' AutoDL algorithm that can take care of efficiency, accuracy, stability, robustness, and generalizability at the same time.
% Towards perfect AutoDL, there are plenty of opportunities.
% The primary step would tend to allow AutoDL algorithms to be good at two aspects from the five.
% Then, we could gradually improve AutoDL algorithms to contain more characteristics.
%
%
% Another challenge is how to enable green AutoDL.
% The first step would be a comprehensive analysis of the CO$_{2}$e cost for different AutoDL algorithms, which is still a missing part for the current community.
% The next step could design CO$_{2}$e-friendly AutoDL algorithms and scientifically evaluate them.
% Following this direction, it is a blue ocean for AutoDL participators.

\section{Conclusions}\label{sec:conclusion}

% In this manuscript, we systematically reviewed the current status of Automated Deep Learning (AutoDL).
% Our review covers the automation within all facets of a DL life cycle instead of solely focusing on a singular aspect like NAS and HPO.
% As a result, we bridge the links between many areas, from task automation to data automation, to NAS, to HPO, to auto-deployment, and to auto-maintenance.
% We dissect existing AutoDL works in \Tabref{table:autodl-dissect} following the historical timeline, and critically evaluate AutoDL approaches in \Secref{sec:criteria}.
% We hope this survey could not only inspire new AutoDL directions but also help improve the quality of evaluations in future AutoDL works. 

In this monograph, we have surveyed a vast body of literature, mostly recent, aiming to automate DL at a high level.
Rather than focus solely on NAS, which is often treated as the be-all and end-all of this endeavor, we have emulated a previous review on AutonoML~\cite{kedziora2020autonoml} by taking a broader perspective of AutoDL research and examining it with respect to an encompassing conceptual framework: a DL workflow.
In the process of doing so, we have also noted deficiencies in the field stemming from both publication oversaturation and, more often than not, a lack of theoretical/experimental treatment that is both rigorous and holistic.
We have thus also proposed an extensive set of criteria by which, within this work, assessments of algorithm-focused trends in AutoDL are made; we hope to stimulate further contemplation within the DL community as to how a clearer picture of progress can be better supported.

All the foundations for this review were established in \Secref{sec:autodl-overview}, which not only discussed how the fields of AutoML and AutoDL relate and differ -- key considerations included the structure of ML/DL solutions and the computational cost of training -- but also elaborated the following notions:
\begin{itemize}
    \item The DL workflow, which covers the phases of problem formulation and context understanding, data engineering, model development, deployment, and monitoring and maintenance.
	\item The assessment criteria, which cover novelty, accuracy, efficiency, stability, interpretability, reproducibility, engineering quality, scalability, generalizability, and eco-friendliness.
\end{itemize}
The subsequent content of the review and certain key conclusions are summarized as follows:
\begin{itemize}
    \item \Secref{sec:auto-translate-context} noted that any DL application starts with problem formulation, and there is certainly room for elements of this to be automated to some degree.
	However, it was acknowledged that the reliance on human involvement would be most challenging to reduce here and, as of such, this phase of a DL workflow has negligible presence in AutoDL research.
	\item \Secref{sec:auto-data-prepare} examined automated data engineering, categorizing research trends into the supplementary generation of data/labels and the intelligent exploitation of DL model inputs via data augmentation/selection.
	Although relatively untouched under the lens of automation in DL, data engineering appears to have greater long-term returns on model accuracy than model engineering, suggesting that attention to this area will grow over time.
	This may potentially improve, as a generalization, the relatively weak efficiencies of associated approaches.
	\item \Secref{sec:nas} examined NAS, dissecting algorithms for constructing neural networks into the spaces that they search, the optimization strategies that they use, and the tactics that they employ to mitigate expensive model training, if any.
	This is where the bulk of publications in AutoDL exists, so the topic is well advanced.
	Even so, the assessment within this review has identified gaps in knowledge here and there, mostly revolving around prerequisites to the maturation of an exploratory field.
	Specifically, while NAS research is responsible for many algorithms capable of producing high-performance DL models, it is generally not interpretable which elements of a NAS algorithm contribute to that performance, nor is it clear in which contexts that performance is reproducibly attainable.
	\item \Secref{sec:hpo} examined HPO -- in AutoDL this conventionally refers to optimizing training-algorithm hyperparameters -- and categorized associated approaches by how much they leverage assumptions/knowledge about a training algorithm and hyperparameter space, i.e., black-box, gray-box, and white-box.
	Historically serving as the first expansion of AutoDL beyond NAS, the topic is similarly well studied, especially black-box HPO and, augmented by tactics to minimize the sampling/training of candidate models, gray-box HPO.
	White-box HPO is newer to the scene, folding HPO into model training for dramatic efficiency boosts, and its strong promise has yet to be validated by rigorous reproducibility studies.
	\item \Secref{sec:deploy} examined automated deployment, with deployment-aware AutoDL and hardware search describing, with respect to a DL model and its deployment environment, which of the two is modified to suit the other one.
	Relatively untouched, in similar fashion to automated data engineering, the topic nonetheless possesses enough research literature to form a preliminary overview, and it seems likely that attention to this area will intensify in the future as AutoDL continues to be translated from predominantly academic settings to real-world applications.
	For now, the assessment found that progress in hardware search moves slower than in deployment-aware AutoDL, although this is understandable given that hardware is extremely non-standardized; this does not diminish the importance of the endeavor.
	\item \Secref{sec:maintenance} examined automated maintenance, although it acknowledged the challenges of DL, primarily relating to computational costs, that currently make model adaptation frequently infeasible, at least in online fashion.
	Nonetheless, while there is too little automation-specific research in this area for a substantial assessment, the importance of this phase of the DL workflow to many real-world applications warranted far more in-depth consideration than for automated problem formulation.
	Accordingly, scenarios for how dynamic data environments may change were discussed, and presently favored paradigms for efficiently training a model around such dynamics were reviewed.
	These approaches were categorized into domain adaptation, low-shot learning, and continuous learning.
\end{itemize}
Following the survey, \Secref{sec:criteria} focused back on the assessment criteria proposed in this monograph, discussing a possible way in which AutoDL researchers may self-assess and report their theoretical/experimental investigations, so as to better foster a research environment that is conducive to accelerated progress in the field.
Then, driven by the same criteria, the section concluded by presenting and discussing an overarching perspective of AutoDL as of the early 2020s, as well as a summary of how the field may evolve in pursuit of: usability, capability, autonomy, consolidation, and emergence.

Ultimately, this review concludes that AutoDL sits in an unusual position, having made exceptional progress in certain areas while also remaining far short of its potential.
Granted, it is a nascent endeavor, and there are physical realities that keep general progress restrained; it is unlikely that an end-to-end integrated system for processing DL applications will be seen any time soon, even as the related field of AutoML seems to be moving in that direction~\cite{kedziora2020autonoml}.
Nonetheless, the trajectory of AutoDL in the broadest sense is also unnecessarily hindered by human factors, such as an all-too-common disregard, intentional or otherwise, for problems other than NAS, goals other than model accuracy, time periods other than the last few years, and so on.
We hope that this survey challenges such narrow perspectives, supporting the DL community with an encompassing overview of ``what is'' and ``what can be'' in AutoDL.

\vspace{4mm}
\Large\textbf{Acknowledgments:} 
% XD and DJK were recruited and received funding for carrying out the research described in this manuscript. The funding was secured by BG for the ongoing ``Automated and Autonomous Machine Learning'' research project within a larger ``Automating Analytics for Impact'' research programme carried out at the Complex Adaptive Systems Lab, University of Technology Sydney, Australia.
XD and DJK acknowledge financial support secured by BG, which funded their participation in this study and the ongoing ``Automated and Autonomous Machine Learning'' project as part of the ``Automating Analytics for Impact'' research programme carried out at the Complex Adaptive Systems Lab, University of Technology Sydney, Australia.

\Large\textbf{Author Contributions:} 
% XD and DJK (to be considered as joint first authors) led on the preparation of the first draft of the manuscript and implemented subsequent substantial revisions. KM and BG engaged in ongoing critical review and editing. BG was the initiator and coordinator of the work. All authors commented on previous versions, read and approved the final manuscript.
BG conceived, scoped and coordinated this study.
XD conducted the survey.
DJK, KM and BG provided guidance.
XD and DJK co-wrote and contributed equally to the monograph.
All authors reviewed and
approved the final version.

% Bibliography
\bibliographystyle{ACM-Reference-Format}
\bibliography{abrv,autodl}

%%% -*-BibTeX-*-
%%% Do NOT edit. File created by BibTeX with style
%%% ACM-Reference-Format-Journals [18-Jan-2012].

\begin{thebibliography}{325}

%%% ====================================================================
%%% NOTE TO THE USER: you can override these defaults by providing
%%% customized versions of any of these macros before the \bibliography
%%% command.  Each of them MUST provide its own final punctuation,
%%% except for \shownote{}, \showDOI{}, and \showURL{}.  The latter two
%%% do not use final punctuation, in order to avoid confusing it with
%%% the Web address.
%%%
%%% To suppress output of a particular field, define its macro to expand
%%% to an empty string, or better, \unskip, like this:
%%%
%%% \newcommand{\showDOI}[1]{\unskip}   % LaTeX syntax
%%%
%%% \def \showDOI #1{\unskip}           % plain TeX syntax
%%%
%%% ====================================================================

\ifx \showCODEN    \undefined \def \showCODEN     #1{\unskip}     \fi
\ifx \showDOI      \undefined \def \showDOI       #1{#1}\fi
\ifx \showISBNx    \undefined \def \showISBNx     #1{\unskip}     \fi
\ifx \showISBNxiii \undefined \def \showISBNxiii  #1{\unskip}     \fi
\ifx \showISSN     \undefined \def \showISSN      #1{\unskip}     \fi
\ifx \showLCCN     \undefined \def \showLCCN      #1{\unskip}     \fi
\ifx \shownote     \undefined \def \shownote      #1{#1}          \fi
\ifx \showarticletitle \undefined \def \showarticletitle #1{#1}   \fi
\ifx \showURL      \undefined \def \showURL       {\relax}        \fi
% The following commands are used for tagged output and should be
% invisible to TeX
\providecommand\bibfield[2]{#2}
\providecommand\bibinfo[2]{#2}
\providecommand\natexlab[1]{#1}
\providecommand\showeprint[2][]{arXiv:#2}

\bibitem[\protect\citeauthoryear{AB, Lundh, Clark, and Contributors}{AB
  et~al\mbox{.}}{2011}]%
        {pil}
\bibfield{author}{\bibinfo{person}{Secret~Labs AB}, \bibinfo{person}{Fredrik
  Lundh}, \bibinfo{person}{Alex Clark}, {and} \bibinfo{person}{Other
  Contributors}.} \bibinfo{year}{2011}\natexlab{}.
\newblock \bibinfo{title}{Python Imaging Library (PIL)}.
\newblock
\newblock


\bibitem[\protect\citeauthoryear{Abdelfattah, Mehrotra, Dudziak, and
  Lane}{Abdelfattah et~al\mbox{.}}{2021}]%
        {abdelfattah2021zero}
\bibfield{author}{\bibinfo{person}{Mohamed~S Abdelfattah},
  \bibinfo{person}{Abhinav Mehrotra}, \bibinfo{person}{{\L}ukasz Dudziak},
  {and} \bibinfo{person}{Nicholas~D Lane}.} \bibinfo{year}{2021}\natexlab{}.
\newblock \showarticletitle{Zero-Cost Proxies for Lightweight NAS}. In
  \bibinfo{booktitle}{\emph{International Conference on Learning
  Representations (ICLR)}}.
\newblock


\bibitem[\protect\citeauthoryear{Addanki, Venkatakrishnan, Gupta, Mao, and
  Alizadeh}{Addanki et~al\mbox{.}}{2019}]%
        {addanki2019placeto}
\bibfield{author}{\bibinfo{person}{Ravichandra Addanki},
  \bibinfo{person}{Shaileshh~Bojja Venkatakrishnan}, \bibinfo{person}{Shreyan
  Gupta}, \bibinfo{person}{Hongzi Mao}, {and} \bibinfo{person}{Mohammad
  Alizadeh}.} \bibinfo{year}{2019}\natexlab{}.
\newblock \showarticletitle{Placeto: learning generalizable device placement
  algorithms for distributed machine learning}. In
  \bibinfo{booktitle}{\emph{Proceedings of the International Conference on
  Neural Information Processing Systems (NeurIPS)}}.
  \bibinfo{pages}{3981--3991}.
\newblock


\bibitem[\protect\citeauthoryear{Agresti}{Agresti}{2018}]%
        {agresti2018introduction}
\bibfield{author}{\bibinfo{person}{Alan Agresti}.}
  \bibinfo{year}{2018}\natexlab{}.
\newblock \bibinfo{booktitle}{\emph{An introduction to categorical data
  analysis}}.
\newblock \bibinfo{publisher}{John Wiley \& Sons}.
\newblock


\bibitem[\protect\citeauthoryear{Akiba, Sano, Yanase, Ohta, and Koyama}{Akiba
  et~al\mbox{.}}{2019}]%
        {akiba2019optuna}
\bibfield{author}{\bibinfo{person}{Takuya Akiba}, \bibinfo{person}{Shotaro
  Sano}, \bibinfo{person}{Toshihiko Yanase}, \bibinfo{person}{Takeru Ohta},
  {and} \bibinfo{person}{Masanori Koyama}.} \bibinfo{year}{2019}\natexlab{}.
\newblock \showarticletitle{Optuna: A next-generation hyperparameter
  optimization framework}. In \bibinfo{booktitle}{\emph{Proceedings of the ACM
  SIGKDD International Conference on Knowledge Discovery and Data Mining
  (KDD)}}. \bibinfo{pages}{2623--2631}.
\newblock


\bibitem[\protect\citeauthoryear{Ali, Budka, and Gabrys}{Ali
  et~al\mbox{.}}{2019}]%
        {ali2019meta}
\bibfield{author}{\bibinfo{person}{Abbas~Raza Ali}, \bibinfo{person}{Marcin
  Budka}, {and} \bibinfo{person}{Bogdan Gabrys}.}
  \bibinfo{year}{2019}\natexlab{}.
\newblock \showarticletitle{A meta-reinforcement learning approach to optimize
  parameters and hyper-parameters simultaneously}. In
  \bibinfo{booktitle}{\emph{pricai}}. \bibinfo{pages}{93--106}.
\newblock


\bibitem[\protect\citeauthoryear{Ali, Gabrys, and Budka}{Ali
  et~al\mbox{.}}{2018}]%
        {ali2018cross}
\bibfield{author}{\bibinfo{person}{Abbas~Raza Ali}, \bibinfo{person}{Bogdan
  Gabrys}, {and} \bibinfo{person}{Marcin Budka}.}
  \bibinfo{year}{2018}\natexlab{}.
\newblock \showarticletitle{Cross-domain meta-learning for time-series
  forecasting}.
\newblock \bibinfo{journal}{\emph{Procedia Computer Science}}
  \bibinfo{volume}{126} (\bibinfo{year}{2018}), \bibinfo{pages}{9--18}.
\newblock


\bibitem[\protect\citeauthoryear{Almgren and Chriss}{Almgren and
  Chriss}{2001}]%
        {almgren2001optimal}
\bibfield{author}{\bibinfo{person}{Robert Almgren} {and} \bibinfo{person}{Neil
  Chriss}.} \bibinfo{year}{2001}\natexlab{}.
\newblock \showarticletitle{Optimal execution of portfolio transactions}.
\newblock \bibinfo{journal}{\emph{Journal of Risk}}  \bibinfo{volume}{3}
  (\bibinfo{year}{2001}), \bibinfo{pages}{5--40}.
\newblock


\bibitem[\protect\citeauthoryear{Andrychowicz, Denil, Gomez, Hoffman, Pfau,
  Schaul, Shillingford, and De~Freitas}{Andrychowicz et~al\mbox{.}}{2016}]%
        {andrychowicz2016learning}
\bibfield{author}{\bibinfo{person}{Marcin Andrychowicz}, \bibinfo{person}{Misha
  Denil}, \bibinfo{person}{Sergio Gomez}, \bibinfo{person}{Matthew~W Hoffman},
  \bibinfo{person}{David Pfau}, \bibinfo{person}{Tom Schaul},
  \bibinfo{person}{Brendan Shillingford}, {and} \bibinfo{person}{Nando
  De~Freitas}.} \bibinfo{year}{2016}\natexlab{}.
\newblock \showarticletitle{Learning to learn by gradient descent by gradient
  descent}. In \bibinfo{booktitle}{\emph{Proceedings of the International
  Conference on Neural Information Processing Systems (NeurIPS)}},
  Vol.~\bibinfo{volume}{29}.
\newblock


\bibitem[\protect\citeauthoryear{Ankit, Hajj, Chalamalasetti, Ndu, Foltin,
  Williams, Faraboschi, Hwu, Strachan, Roy, et~al\mbox{.}}{Ankit
  et~al\mbox{.}}{2019}]%
        {ankit2019puma}
\bibfield{author}{\bibinfo{person}{Aayush Ankit}, \bibinfo{person}{Izzat~El
  Hajj}, \bibinfo{person}{Sai~Rahul Chalamalasetti}, \bibinfo{person}{Geoffrey
  Ndu}, \bibinfo{person}{Martin Foltin}, \bibinfo{person}{R~Stanley Williams},
  \bibinfo{person}{Paolo Faraboschi}, \bibinfo{person}{Wen-mei~W Hwu},
  \bibinfo{person}{John~Paul Strachan}, \bibinfo{person}{Kaushik Roy},
  {et~al\mbox{.}}} \bibinfo{year}{2019}\natexlab{}.
\newblock \showarticletitle{{PUMA}: A programmable ultra-efficient
  memristor-based accelerator for machine learning inference}. In
  \bibinfo{booktitle}{\emph{Proceedings of the International Conference on
  Architectural Support for Programming Languages and Operating Systems}}.
  \bibinfo{pages}{715--731}.
\newblock


\bibitem[\protect\citeauthoryear{Astudillo and Frazier}{Astudillo and
  Frazier}{2019}]%
        {astudillo2019bayesian}
\bibfield{author}{\bibinfo{person}{Raul Astudillo} {and} \bibinfo{person}{Peter
  Frazier}.} \bibinfo{year}{2019}\natexlab{}.
\newblock \showarticletitle{Bayesian optimization of composite functions}. In
  \bibinfo{booktitle}{\emph{The International Conference on Machine Learning
  (ICML)}}. \bibinfo{pages}{354--363}.
\newblock


\bibitem[\protect\citeauthoryear{Baik, Choi, Choi, Kim, and Lee}{Baik
  et~al\mbox{.}}{2020}]%
        {baik2020meta}
\bibfield{author}{\bibinfo{person}{Sungyong Baik}, \bibinfo{person}{Myungsub
  Choi}, \bibinfo{person}{Janghoon Choi}, \bibinfo{person}{Heewon Kim}, {and}
  \bibinfo{person}{Kyoung~Mu Lee}.} \bibinfo{year}{2020}\natexlab{}.
\newblock \showarticletitle{Meta-Learning with Adaptive Hyperparameters}. In
  \bibinfo{booktitle}{\emph{Proceedings of the International Conference on
  Neural Information Processing Systems (NeurIPS)}}.
\newblock


\bibitem[\protect\citeauthoryear{Baker, Gupta, Naik, and Raskar}{Baker
  et~al\mbox{.}}{2017}]%
        {baker2017designing}
\bibfield{author}{\bibinfo{person}{Bowen Baker}, \bibinfo{person}{Otkrist
  Gupta}, \bibinfo{person}{Nikhil Naik}, {and} \bibinfo{person}{Ramesh
  Raskar}.} \bibinfo{year}{2017}\natexlab{}.
\newblock \showarticletitle{Designing Neural Network Architectures using
  Reinforcement Learning}. In \bibinfo{booktitle}{\emph{International
  Conference on Learning Representations (ICLR)}}.
\newblock


\bibitem[\protect\citeauthoryear{Baker, Gupta, Raskar, and Naik}{Baker
  et~al\mbox{.}}{2018}]%
        {baker2018accelerating}
\bibfield{author}{\bibinfo{person}{Bowen Baker}, \bibinfo{person}{Otkrist
  Gupta}, \bibinfo{person}{Ramesh Raskar}, {and} \bibinfo{person}{Nikhil
  Naik}.} \bibinfo{year}{2018}\natexlab{}.
\newblock \showarticletitle{Accelerating neural architecture search using
  performance prediction}. In \bibinfo{booktitle}{\emph{International
  Conference on Learning Representations (ICLR) Workshop}}.
\newblock


\bibitem[\protect\citeauthoryear{Ballard}{Ballard}{1987}]%
        {ballard1987modular}
\bibfield{author}{\bibinfo{person}{Dana~H Ballard}.}
  \bibinfo{year}{1987}\natexlab{}.
\newblock \showarticletitle{Modular learning in neural networks}. In
  \bibinfo{booktitle}{\emph{AAAI Conference on Artificial Intelligence
  (AAAI)}}. \bibinfo{pages}{279--284}.
\newblock


\bibitem[\protect\citeauthoryear{Baydin, Cornish, Rubio, Schmidt, and
  Wood}{Baydin et~al\mbox{.}}{2018}]%
        {baydin2018online}
\bibfield{author}{\bibinfo{person}{Atilim~Gunes Baydin},
  \bibinfo{person}{Robert Cornish}, \bibinfo{person}{David~Martinez Rubio},
  \bibinfo{person}{Mark Schmidt}, {and} \bibinfo{person}{Frank Wood}.}
  \bibinfo{year}{2018}\natexlab{}.
\newblock \showarticletitle{Online learning rate adaptation with hypergradient
  descent}. In \bibinfo{booktitle}{\emph{International Conference on Learning
  Representations (ICLR)}}.
\newblock


\bibitem[\protect\citeauthoryear{Bayer, Wierstra, Togelius, and
  Schmidhuber}{Bayer et~al\mbox{.}}{2009}]%
        {bayer2009evolving}
\bibfield{author}{\bibinfo{person}{Justin Bayer}, \bibinfo{person}{Daan
  Wierstra}, \bibinfo{person}{Julian Togelius}, {and}
  \bibinfo{person}{J{\"u}rgen Schmidhuber}.} \bibinfo{year}{2009}\natexlab{}.
\newblock \showarticletitle{Evolving memory cell structures for sequence
  learning}. In \bibinfo{booktitle}{\emph{Proceedings of the International
  Conference on Artificial Neural Networks (ICANN)}}.
  \bibinfo{pages}{755--764}.
\newblock


\bibitem[\protect\citeauthoryear{Ben-David, Blitzer, Crammer, Pereira,
  et~al\mbox{.}}{Ben-David et~al\mbox{.}}{2007}]%
        {ben2007analysis}
\bibfield{author}{\bibinfo{person}{Shai Ben-David}, \bibinfo{person}{John
  Blitzer}, \bibinfo{person}{Koby Crammer}, \bibinfo{person}{Fernando Pereira},
  {et~al\mbox{.}}} \bibinfo{year}{2007}\natexlab{}.
\newblock \showarticletitle{Analysis of representations for domain adaptation}.
  In \bibinfo{booktitle}{\emph{Proceedings of the International Conference on
  Neural Information Processing Systems (NeurIPS)}}.
\newblock


\bibitem[\protect\citeauthoryear{Bender, Kindermans, Zoph, Vasudevan, and
  Le}{Bender et~al\mbox{.}}{2018}]%
        {bender2018understanding}
\bibfield{author}{\bibinfo{person}{Gabriel Bender}, \bibinfo{person}{Pieter-Jan
  Kindermans}, \bibinfo{person}{Barret Zoph}, \bibinfo{person}{Vijay
  Vasudevan}, {and} \bibinfo{person}{Quoc Le}.}
  \bibinfo{year}{2018}\natexlab{}.
\newblock \showarticletitle{Understanding and simplifying one-shot architecture
  search}. In \bibinfo{booktitle}{\emph{The International Conference on Machine
  Learning (ICML)}}. \bibinfo{pages}{550--559}.
\newblock


\bibitem[\protect\citeauthoryear{Bender, Liu, Chen, Chu, Cheng, Kindermans, and
  Le}{Bender et~al\mbox{.}}{2020}]%
        {bender2020can}
\bibfield{author}{\bibinfo{person}{Gabriel Bender}, \bibinfo{person}{Hanxiao
  Liu}, \bibinfo{person}{Bo Chen}, \bibinfo{person}{Grace Chu},
  \bibinfo{person}{Shuyang Cheng}, \bibinfo{person}{Pieter-Jan Kindermans},
  {and} \bibinfo{person}{Quoc~V Le}.} \bibinfo{year}{2020}\natexlab{}.
\newblock \showarticletitle{Can weight sharing outperform random architecture
  search? an investigation with tunas}. In
  \bibinfo{booktitle}{\emph{Proceedings of the IEEE Conference Computer Vision
  Pattern Recognition (CVPR)}}. \bibinfo{pages}{14323--14332}.
\newblock


\bibitem[\protect\citeauthoryear{Bengio}{Bengio}{2000}]%
        {bengio2000gradient}
\bibfield{author}{\bibinfo{person}{Yoshua Bengio}.}
  \bibinfo{year}{2000}\natexlab{}.
\newblock \showarticletitle{Gradient-based optimization of hyperparameters}.
\newblock \bibinfo{journal}{\emph{Neural Computation}} \bibinfo{volume}{12},
  \bibinfo{number}{8} (\bibinfo{year}{2000}), \bibinfo{pages}{1889--1900}.
\newblock


\bibitem[\protect\citeauthoryear{Bengio, Bengio, and Cloutier}{Bengio
  et~al\mbox{.}}{1990}]%
        {bengio1990learning}
\bibfield{author}{\bibinfo{person}{Yoshua Bengio}, \bibinfo{person}{Samy
  Bengio}, {and} \bibinfo{person}{Jocelyn Cloutier}.}
  \bibinfo{year}{1990}\natexlab{}.
\newblock \bibinfo{booktitle}{\emph{Learning a synaptic learning rule}}.
\newblock \bibinfo{publisher}{Citeseer}.
\newblock


\bibitem[\protect\citeauthoryear{Bengio, Louradour, Collobert, and
  Weston}{Bengio et~al\mbox{.}}{2009}]%
        {bengio2009curriculum}
\bibfield{author}{\bibinfo{person}{Yoshua Bengio},
  \bibinfo{person}{J{\'e}r{\^o}me Louradour}, \bibinfo{person}{Ronan
  Collobert}, {and} \bibinfo{person}{Jason Weston}.}
  \bibinfo{year}{2009}\natexlab{}.
\newblock \showarticletitle{Curriculum learning}. In
  \bibinfo{booktitle}{\emph{The International Conference on Machine Learning
  (ICML)}}. \bibinfo{pages}{41--48}.
\newblock


\bibitem[\protect\citeauthoryear{Bergstra, Bardenet, Bengio, and
  K{\'e}gl}{Bergstra et~al\mbox{.}}{2011}]%
        {bergstra2011algorithms}
\bibfield{author}{\bibinfo{person}{James Bergstra}, \bibinfo{person}{R{\'e}mi
  Bardenet}, \bibinfo{person}{Yoshua Bengio}, {and} \bibinfo{person}{Bal{\'a}zs
  K{\'e}gl}.} \bibinfo{year}{2011}\natexlab{}.
\newblock \showarticletitle{Algorithms for hyper-parameter optimization}. In
  \bibinfo{booktitle}{\emph{Proceedings of the International Conference on
  Neural Information Processing Systems (NeurIPS)}}, Vol.~\bibinfo{volume}{24}.
\newblock


\bibitem[\protect\citeauthoryear{Bergstra and Bengio}{Bergstra and
  Bengio}{2012}]%
        {bergstra2012random}
\bibfield{author}{\bibinfo{person}{James Bergstra} {and}
  \bibinfo{person}{Yoshua Bengio}.} \bibinfo{year}{2012}\natexlab{}.
\newblock \showarticletitle{Random search for hyper-parameter optimization}.
\newblock \bibinfo{journal}{\emph{Journal of Machine Learning Research (JMLR)}}
  \bibinfo{volume}{13}, \bibinfo{number}{2} (\bibinfo{year}{2012}).
\newblock


\bibitem[\protect\citeauthoryear{Biedenkapp, Marben, Lindauer, and
  Hutter}{Biedenkapp et~al\mbox{.}}{2018}]%
        {biedenkapp-lion18a}
\bibfield{author}{\bibinfo{person}{A. Biedenkapp}, \bibinfo{person}{J. Marben},
  \bibinfo{person}{M. Lindauer}, {and} \bibinfo{person}{F. Hutter}.}
  \bibinfo{year}{2018}\natexlab{}.
\newblock \showarticletitle{{CAVE}: Configuration Assessment, Visualization and
  Evaluation}. In \bibinfo{booktitle}{\emph{Proceedings of the International
  Conference on Learning and Intelligent Optimization (LION)}}.
\newblock


\bibitem[\protect\citeauthoryear{Box and Cox}{Box and Cox}{1964}]%
        {box1964analysis}
\bibfield{author}{\bibinfo{person}{George~EP Box} {and}
  \bibinfo{person}{David~R Cox}.} \bibinfo{year}{1964}\natexlab{}.
\newblock \showarticletitle{An analysis of transformations}.
\newblock \bibinfo{journal}{\emph{Journal of the Royal Statistical Society:
  Series B (Methodological)}} \bibinfo{volume}{26}, \bibinfo{number}{2}
  (\bibinfo{year}{1964}), \bibinfo{pages}{211--243}.
\newblock


\bibitem[\protect\citeauthoryear{Breiman}{Breiman}{2001}]%
        {br01}
\bibfield{author}{\bibinfo{person}{Leo Breiman}.}
  \bibinfo{year}{2001}\natexlab{}.
\newblock \showarticletitle{Statistical Modeling: The Two Cultures (with
  comments and a rejoinder by the author)}.
\newblock \bibinfo{journal}{\emph{Statist. Sci.}} \bibinfo{volume}{16},
  \bibinfo{number}{3} (\bibinfo{date}{aug} \bibinfo{year}{2001}),
  \bibinfo{pages}{199--231}.
\newblock
\urldef\tempurl%
\url{https://doi.org/10.1214/ss/1009213726}
\showDOI{\tempurl}


\bibitem[\protect\citeauthoryear{Brock, Lim, Ritchie, and Weston}{Brock
  et~al\mbox{.}}{2018}]%
        {brock2018smash}
\bibfield{author}{\bibinfo{person}{Andrew Brock}, \bibinfo{person}{Theodore
  Lim}, \bibinfo{person}{James~M Ritchie}, {and} \bibinfo{person}{Nick
  Weston}.} \bibinfo{year}{2018}\natexlab{}.
\newblock \showarticletitle{{SMASH}: one-shot model architecture search through
  hypernetworks}. In \bibinfo{booktitle}{\emph{International Conference on
  Learning Representations (ICLR)}}.
\newblock


\bibitem[\protect\citeauthoryear{Brown, Mann, Ryder, Subbiah, Kaplan, Dhariwal,
  Neelakantan, Shyam, Sastry, Askell, et~al\mbox{.}}{Brown
  et~al\mbox{.}}{2020}]%
        {brown2020language}
\bibfield{author}{\bibinfo{person}{Tom~B Brown}, \bibinfo{person}{Benjamin
  Mann}, \bibinfo{person}{Nick Ryder}, \bibinfo{person}{Melanie Subbiah},
  \bibinfo{person}{Jared Kaplan}, \bibinfo{person}{Prafulla Dhariwal},
  \bibinfo{person}{Arvind Neelakantan}, \bibinfo{person}{Pranav Shyam},
  \bibinfo{person}{Girish Sastry}, \bibinfo{person}{Amanda Askell},
  {et~al\mbox{.}}} \bibinfo{year}{2020}\natexlab{}.
\newblock \showarticletitle{Language models are few-shot learners}. In
  \bibinfo{booktitle}{\emph{Proceedings of the International Conference on
  Neural Information Processing Systems (NeurIPS)}}, Vol.~\bibinfo{volume}{33}.
  \bibinfo{pages}{1877--1901}.
\newblock


\bibitem[\protect\citeauthoryear{Bulat, Martinez, and Tzimiropoulos}{Bulat
  et~al\mbox{.}}{2020}]%
        {bulat2020bats}
\bibfield{author}{\bibinfo{person}{Adrian Bulat}, \bibinfo{person}{Brais
  Martinez}, {and} \bibinfo{person}{Georgios Tzimiropoulos}.}
  \bibinfo{year}{2020}\natexlab{}.
\newblock \showarticletitle{{BATS}: Binary architecture search}. In
  \bibinfo{booktitle}{\emph{Proceedings of the European Conference on Computer
  Vision (ECCV)}}.
\newblock


\bibitem[\protect\citeauthoryear{Cai, Gan, Wang, Zhang, and Han}{Cai
  et~al\mbox{.}}{2020}]%
        {cai2020once}
\bibfield{author}{\bibinfo{person}{Han Cai}, \bibinfo{person}{Chuang Gan},
  \bibinfo{person}{Tianzhe Wang}, \bibinfo{person}{Zhekai Zhang}, {and}
  \bibinfo{person}{Song Han}.} \bibinfo{year}{2020}\natexlab{}.
\newblock \showarticletitle{Once-for-all: Train one network and specialize it
  for efficient deployment}. In \bibinfo{booktitle}{\emph{International
  Conference on Learning Representations (ICLR)}}.
\newblock


\bibitem[\protect\citeauthoryear{Cai, Yang, Zhang, Han, and Yu}{Cai
  et~al\mbox{.}}{2018}]%
        {cai2018path}
\bibfield{author}{\bibinfo{person}{Han Cai}, \bibinfo{person}{Jiacheng Yang},
  \bibinfo{person}{Weinan Zhang}, \bibinfo{person}{Song Han}, {and}
  \bibinfo{person}{Yong Yu}.} \bibinfo{year}{2018}\natexlab{}.
\newblock \showarticletitle{Path-level network transformation for efficient
  architecture search}. In \bibinfo{booktitle}{\emph{The International
  Conference on Machine Learning (ICML)}}. \bibinfo{pages}{678--687}.
\newblock


\bibitem[\protect\citeauthoryear{Cai, Zhu, and Han}{Cai et~al\mbox{.}}{2019}]%
        {cai2019proxylessnas}
\bibfield{author}{\bibinfo{person}{Han Cai}, \bibinfo{person}{Ligeng Zhu},
  {and} \bibinfo{person}{Song Han}.} \bibinfo{year}{2019}\natexlab{}.
\newblock \showarticletitle{{ProxylessNAS}: Direct neural architecture search
  on target task and hardware}. In \bibinfo{booktitle}{\emph{International
  Conference on Learning Representations (ICLR)}}.
\newblock


\bibitem[\protect\citeauthoryear{Carpenter and Grossberg}{Carpenter and
  Grossberg}{1987}]%
        {carpenter1987massively}
\bibfield{author}{\bibinfo{person}{Gail~A Carpenter} {and}
  \bibinfo{person}{Stephen Grossberg}.} \bibinfo{year}{1987}\natexlab{}.
\newblock \showarticletitle{A massively parallel architecture for a
  self-organizing neural pattern recognition machine}.
\newblock \bibinfo{journal}{\emph{Computer Vision, Graphics, and Image
  Processing}} \bibinfo{volume}{37}, \bibinfo{number}{1}
  (\bibinfo{year}{1987}), \bibinfo{pages}{54--115}.
\newblock


\bibitem[\protect\citeauthoryear{Caruana, Niculescu-Mizil, Crew, and
  Ksikes}{Caruana et~al\mbox{.}}{2004}]%
        {caruana2004ensemble}
\bibfield{author}{\bibinfo{person}{Rich Caruana}, \bibinfo{person}{Alexandru
  Niculescu-Mizil}, \bibinfo{person}{Geoff Crew}, {and} \bibinfo{person}{Alex
  Ksikes}.} \bibinfo{year}{2004}\natexlab{}.
\newblock \showarticletitle{Ensemble selection from libraries of models}. In
  \bibinfo{booktitle}{\emph{The International Conference on Machine Learning
  (ICML)}}. \bibinfo{pages}{18}.
\newblock


\bibitem[\protect\citeauthoryear{Celik and Vanschoren}{Celik and
  Vanschoren}{2021}]%
        {celik2021adaptation}
\bibfield{author}{\bibinfo{person}{Bilge Celik} {and} \bibinfo{person}{Joaquin
  Vanschoren}.} \bibinfo{year}{2021}\natexlab{}.
\newblock \showarticletitle{Adaptation strategies for automated machine
  learning on evolving data}.
\newblock \bibinfo{journal}{\emph{IEEE Transactions on Pattern Analysis and
  Machine Intelligence (TPAMI)}} \bibinfo{volume}{43}, \bibinfo{number}{9}
  (\bibinfo{year}{2021}), \bibinfo{pages}{3067--3078}.
\newblock


\bibitem[\protect\citeauthoryear{Chandola, Banerjee, and Kumar}{Chandola
  et~al\mbox{.}}{2009}]%
        {chandola2009anomaly}
\bibfield{author}{\bibinfo{person}{Varun Chandola}, \bibinfo{person}{Arindam
  Banerjee}, {and} \bibinfo{person}{Vipin Kumar}.}
  \bibinfo{year}{2009}\natexlab{}.
\newblock \showarticletitle{Anomaly detection: A survey}.
\newblock \bibinfo{journal}{\emph{ACM Computing Surveys (CSUR)}}
  \bibinfo{volume}{41}, \bibinfo{number}{3} (\bibinfo{year}{2009}),
  \bibinfo{pages}{1--58}.
\newblock


\bibitem[\protect\citeauthoryear{Chandrashekaran and Lane}{Chandrashekaran and
  Lane}{2017}]%
        {chandrashekaran2017speeding}
\bibfield{author}{\bibinfo{person}{Akshay Chandrashekaran} {and}
  \bibinfo{person}{Ian~R Lane}.} \bibinfo{year}{2017}\natexlab{}.
\newblock \showarticletitle{Speeding up hyper-parameter optimization by
  extrapolation of learning curves using previous builds}. In
  \bibinfo{booktitle}{\emph{The Joint European Conference on Machine Learning
  and Knowledge Discovery in Databases (ECML PKDD)}}.
  \bibinfo{pages}{477--492}.
\newblock


\bibitem[\protect\citeauthoryear{Chen, Wu, Ni, Peng, Liu, Fu, Chao, and
  Ling}{Chen et~al\mbox{.}}{2021b}]%
        {chen2021searching}
\bibfield{author}{\bibinfo{person}{Minghao Chen}, \bibinfo{person}{Kan Wu},
  \bibinfo{person}{Bolin Ni}, \bibinfo{person}{Houwen Peng},
  \bibinfo{person}{Bei Liu}, \bibinfo{person}{Jianlong Fu},
  \bibinfo{person}{Hongyang Chao}, {and} \bibinfo{person}{Haibin Ling}.}
  \bibinfo{year}{2021}\natexlab{b}.
\newblock \showarticletitle{Searching the Search Space of Vision Transformer}.
  In \bibinfo{booktitle}{\emph{Proceedings of the International Conference on
  Neural Information Processing Systems (NeurIPS)}}.
\newblock


\bibitem[\protect\citeauthoryear{Chen, Goodfellow, and Shlens}{Chen
  et~al\mbox{.}}{2016a}]%
        {chen2016net2net}
\bibfield{author}{\bibinfo{person}{Tianqi Chen}, \bibinfo{person}{Ian
  Goodfellow}, {and} \bibinfo{person}{Jonathon Shlens}.}
  \bibinfo{year}{2016}\natexlab{a}.
\newblock \showarticletitle{{Net2Net}: Accelerating learning via knowledge
  transfer}. In \bibinfo{booktitle}{\emph{International Conference on Learning
  Representations (ICLR)}}.
\newblock


\bibitem[\protect\citeauthoryear{Chen, Moreau, Jiang, Zheng, Yan, Shen, Cowan,
  Wang, Hu, Ceze, et~al\mbox{.}}{Chen et~al\mbox{.}}{2018}]%
        {chen2018tvm}
\bibfield{author}{\bibinfo{person}{Tianqi Chen}, \bibinfo{person}{Thierry
  Moreau}, \bibinfo{person}{Ziheng Jiang}, \bibinfo{person}{Lianmin Zheng},
  \bibinfo{person}{Eddie Yan}, \bibinfo{person}{Haichen Shen},
  \bibinfo{person}{Meghan Cowan}, \bibinfo{person}{Leyuan Wang},
  \bibinfo{person}{Yuwei Hu}, \bibinfo{person}{Luis Ceze}, {et~al\mbox{.}}}
  \bibinfo{year}{2018}\natexlab{}.
\newblock \showarticletitle{{TVM}: An Automated {End-to-End} Optimizing
  Compiler for Deep Learning}. In \bibinfo{booktitle}{\emph{The USENIX
  Symposium on Operating Systems Design and Implementation (OSDI)}}.
  \bibinfo{pages}{578--594}.
\newblock


\bibitem[\protect\citeauthoryear{Chen, Gong, and Wang}{Chen
  et~al\mbox{.}}{2021a}]%
        {chen2021neural}
\bibfield{author}{\bibinfo{person}{Wuyang Chen}, \bibinfo{person}{Xinyu Gong},
  {and} \bibinfo{person}{Zhangyang Wang}.} \bibinfo{year}{2021}\natexlab{a}.
\newblock \showarticletitle{Neural architecture search on imagenet in four gpu
  hours: A theoretically inspired perspective}. In
  \bibinfo{booktitle}{\emph{International Conference on Learning
  Representations (ICLR)}}.
\newblock


\bibitem[\protect\citeauthoryear{Chen, Hoffman, Colmenarejo, Denil, Lillicrap,
  Botvinick, and Freitas}{Chen et~al\mbox{.}}{2017}]%
        {chen2017learning}
\bibfield{author}{\bibinfo{person}{Yutian Chen}, \bibinfo{person}{Matthew~W
  Hoffman}, \bibinfo{person}{Sergio~G{\'o}mez Colmenarejo},
  \bibinfo{person}{Misha Denil}, \bibinfo{person}{Timothy~P Lillicrap},
  \bibinfo{person}{Matt Botvinick}, {and} \bibinfo{person}{Nando Freitas}.}
  \bibinfo{year}{2017}\natexlab{}.
\newblock \showarticletitle{Learning to learn without gradient descent by
  gradient descent}. In \bibinfo{booktitle}{\emph{The International Conference
  on Machine Learning (ICML)}}. \bibinfo{pages}{748--756}.
\newblock


\bibitem[\protect\citeauthoryear{Chen, Yang, Zhang, Meng, Xiao, and Sun}{Chen
  et~al\mbox{.}}{2019}]%
        {chen2019detnas}
\bibfield{author}{\bibinfo{person}{Yukang Chen}, \bibinfo{person}{Tong Yang},
  \bibinfo{person}{Xiangyu Zhang}, \bibinfo{person}{Gaofeng Meng},
  \bibinfo{person}{Xinyu Xiao}, {and} \bibinfo{person}{Jian Sun}.}
  \bibinfo{year}{2019}\natexlab{}.
\newblock \showarticletitle{{DetNAS}: Backbone Search for Object Detection}. In
  \bibinfo{booktitle}{\emph{Proceedings of the International Conference on
  Neural Information Processing Systems (NeurIPS)}}.
  \bibinfo{pages}{6642--6652}.
\newblock


\bibitem[\protect\citeauthoryear{Chen, Krishna, Emer, and Sze}{Chen
  et~al\mbox{.}}{2016b}]%
        {chen2016eyeriss}
\bibfield{author}{\bibinfo{person}{Yu-Hsin Chen}, \bibinfo{person}{Tushar
  Krishna}, \bibinfo{person}{Joel~S Emer}, {and} \bibinfo{person}{Vivienne
  Sze}.} \bibinfo{year}{2016}\natexlab{b}.
\newblock \showarticletitle{Eyeriss: An energy-efficient reconfigurable
  accelerator for deep convolutional neural networks}.
\newblock \bibinfo{journal}{\emph{IEEE Journal of Solid-state Circuits}}
  \bibinfo{volume}{52}, \bibinfo{number}{1} (\bibinfo{year}{2016}),
  \bibinfo{pages}{127--138}.
\newblock


\bibitem[\protect\citeauthoryear{Choi, Hong, Yoon, Yu, Kim, and Lee}{Choi
  et~al\mbox{.}}{2021}]%
        {choi2021dance}
\bibfield{author}{\bibinfo{person}{Kanghyun Choi}, \bibinfo{person}{Deokki
  Hong}, \bibinfo{person}{Hojae Yoon}, \bibinfo{person}{Joonsang Yu},
  \bibinfo{person}{Youngsok Kim}, {and} \bibinfo{person}{Jinho Lee}.}
  \bibinfo{year}{2021}\natexlab{}.
\newblock \showarticletitle{{DANCE}: Differentiable Accelerator/Network
  Co-Exploration}. In \bibinfo{booktitle}{\emph{The ACM/ESDA/IEEE Design
  Automation Conference (DAC)}}. \bibinfo{pages}{337--342}.
\newblock
\urldef\tempurl%
\url{https://doi.org/10.1109/DAC18074.2021.9586121}
\showDOI{\tempurl}


\bibitem[\protect\citeauthoryear{Cire{\c{s}}an, Giusti, Gambardella, and
  Schmidhuber}{Cire{\c{s}}an et~al\mbox{.}}{2013}]%
        {cirecsan2013mitosis}
\bibfield{author}{\bibinfo{person}{Dan~C Cire{\c{s}}an},
  \bibinfo{person}{Alessandro Giusti}, \bibinfo{person}{Luca~M Gambardella},
  {and} \bibinfo{person}{J{\"u}rgen Schmidhuber}.}
  \bibinfo{year}{2013}\natexlab{}.
\newblock \showarticletitle{Mitosis detection in breast cancer histology images
  with deep neural networks}. In \bibinfo{booktitle}{\emph{Proceedings of the
  International Conference on Medical Image Computing and Computer Assisted
  Intervention (MICCAI)}}. \bibinfo{pages}{411--418}.
\newblock


\bibitem[\protect\citeauthoryear{Conn, Scheinberg, and Vicente}{Conn
  et~al\mbox{.}}{2009}]%
        {conn2009introduction}
\bibfield{author}{\bibinfo{person}{Andrew~R Conn}, \bibinfo{person}{Katya
  Scheinberg}, {and} \bibinfo{person}{Luis~N Vicente}.}
  \bibinfo{year}{2009}\natexlab{}.
\newblock \bibinfo{booktitle}{\emph{Introduction to derivative-free
  optimization}}.
\newblock \bibinfo{publisher}{SIAM}.
\newblock


\bibitem[\protect\citeauthoryear{Cubuk, Zoph, Mane, Vasudevan, and Le}{Cubuk
  et~al\mbox{.}}{2019}]%
        {cubuk2019autoaugment}
\bibfield{author}{\bibinfo{person}{Ekin~D Cubuk}, \bibinfo{person}{Barret
  Zoph}, \bibinfo{person}{Dandelion Mane}, \bibinfo{person}{Vijay Vasudevan},
  {and} \bibinfo{person}{Quoc~V Le}.} \bibinfo{year}{2019}\natexlab{}.
\newblock \showarticletitle{{AutoAugment}: Learning augmentation strategies
  from data}. In \bibinfo{booktitle}{\emph{Proceedings of the IEEE Conference
  Computer Vision Pattern Recognition (CVPR)}}. \bibinfo{pages}{113--123}.
\newblock


\bibitem[\protect\citeauthoryear{Cubuk, Zoph, Shlens, and Le}{Cubuk
  et~al\mbox{.}}{2020}]%
        {cubuk2020randaugment}
\bibfield{author}{\bibinfo{person}{Ekin~D Cubuk}, \bibinfo{person}{Barret
  Zoph}, \bibinfo{person}{Jonathon Shlens}, {and} \bibinfo{person}{Quoc~V Le}.}
  \bibinfo{year}{2020}\natexlab{}.
\newblock \showarticletitle{{RandAugment}: Practical automated data
  augmentation with a reduced search space}. In
  \bibinfo{booktitle}{\emph{Proceedings of the IEEE Conference Computer Vision
  Pattern Recognition (CVPR) Workshop}}. \bibinfo{pages}{702--703}.
\newblock


\bibitem[\protect\citeauthoryear{Dai, Wan, Zhang, Wu, He, Wei, Chen, Tian, Yu,
  Vajda, et~al\mbox{.}}{Dai et~al\mbox{.}}{2021}]%
        {dai2021fbnetv3}
\bibfield{author}{\bibinfo{person}{Xiaoliang Dai}, \bibinfo{person}{Alvin Wan},
  \bibinfo{person}{Peizhao Zhang}, \bibinfo{person}{Bichen Wu},
  \bibinfo{person}{Zijian He}, \bibinfo{person}{Zhen Wei}, \bibinfo{person}{Kan
  Chen}, \bibinfo{person}{Yuandong Tian}, \bibinfo{person}{Matthew Yu},
  \bibinfo{person}{Peter Vajda}, {et~al\mbox{.}}}
  \bibinfo{year}{2021}\natexlab{}.
\newblock \showarticletitle{{FBNetV3}: Joint architecture-recipe search using
  neural acquisition function}. In \bibinfo{booktitle}{\emph{Proceedings of the
  IEEE Conference Computer Vision Pattern Recognition (CVPR)}}.
  \bibinfo{pages}{16276--16285}.
\newblock


\bibitem[\protect\citeauthoryear{Daxberger, Makarova, Turchetta, and
  Krause}{Daxberger et~al\mbox{.}}{2020}]%
        {daxberger20mixed}
\bibfield{author}{\bibinfo{person}{Erik~A Daxberger},
  \bibinfo{person}{Anastasia Makarova}, \bibinfo{person}{Matteo Turchetta},
  {and} \bibinfo{person}{Andreas Krause}.} \bibinfo{year}{2020}\natexlab{}.
\newblock \showarticletitle{Mixed-Variable Bayesian Optimization}. In
  \bibinfo{booktitle}{\emph{International Joint Conferences on Artificial
  Intelligence (IJCAI)}}. \bibinfo{pages}{2633--2639}.
\newblock


\bibitem[\protect\citeauthoryear{De~Jong}{De~Jong}{2016}]%
        {de2016evolutionary}
\bibfield{author}{\bibinfo{person}{Kenneth De~Jong}.}
  \bibinfo{year}{2016}\natexlab{}.
\newblock \showarticletitle{Evolutionary computation: a unified approach}. In
  \bibinfo{booktitle}{\emph{Proceedings of the 2016 on Genetic and Evolutionary
  Computation Conference Companion}}. \bibinfo{pages}{185--199}.
\newblock


\bibitem[\protect\citeauthoryear{Dikov and Bayer}{Dikov and Bayer}{2019}]%
        {dikov2019bayesian}
\bibfield{author}{\bibinfo{person}{Georgi Dikov} {and} \bibinfo{person}{Justin
  Bayer}.} \bibinfo{year}{2019}\natexlab{}.
\newblock \showarticletitle{Bayesian learning of neural network architectures}.
  In \bibinfo{booktitle}{\emph{The International Conference on Artificial
  Intelligence and Statistics (AISTATS)}}. \bibinfo{pages}{730--738}.
\newblock


\bibitem[\protect\citeauthoryear{Domhan, Springenberg, and Hutter}{Domhan
  et~al\mbox{.}}{2015}]%
        {domhan2015speeding}
\bibfield{author}{\bibinfo{person}{Tobias Domhan}, \bibinfo{person}{Jost~Tobias
  Springenberg}, {and} \bibinfo{person}{Frank Hutter}.}
  \bibinfo{year}{2015}\natexlab{}.
\newblock \showarticletitle{Speeding up automatic hyperparameter optimization
  of deep neural networks by extrapolation of learning curves}. In
  \bibinfo{booktitle}{\emph{International Joint Conferences on Artificial
  Intelligence (IJCAI)}}.
\newblock


\bibitem[\protect\citeauthoryear{Domke}{Domke}{2012}]%
        {domke2012generic}
\bibfield{author}{\bibinfo{person}{Justin Domke}.}
  \bibinfo{year}{2012}\natexlab{}.
\newblock \showarticletitle{Generic methods for optimization-based modeling}.
  In \bibinfo{booktitle}{\emph{The International Conference on Artificial
  Intelligence and Statistics (AISTATS)}}. \bibinfo{pages}{318--326}.
\newblock


\bibitem[\protect\citeauthoryear{Dong, Huang, Yang, and Yan}{Dong
  et~al\mbox{.}}{2017}]%
        {dong2017more}
\bibfield{author}{\bibinfo{person}{Xuanyi Dong}, \bibinfo{person}{Junshi
  Huang}, \bibinfo{person}{Yi Yang}, {and} \bibinfo{person}{Shuicheng Yan}.}
  \bibinfo{year}{2017}\natexlab{}.
\newblock \showarticletitle{{More is Less}: A More Complicated Network with
  Less Inference Complexity}. In \bibinfo{booktitle}{\emph{Proceedings of the
  IEEE Conference Computer Vision Pattern Recognition (CVPR)}}.
  \bibinfo{pages}{5840--5848}.
\newblock


\bibitem[\protect\citeauthoryear{Dong, Liu, Musial, and Gabrys}{Dong
  et~al\mbox{.}}{2021a}]%
        {dong2021nats}
\bibfield{author}{\bibinfo{person}{Xuanyi Dong}, \bibinfo{person}{Lu Liu},
  \bibinfo{person}{Katarzyna Musial}, {and} \bibinfo{person}{Bogdan Gabrys}.}
  \bibinfo{year}{2021}\natexlab{a}.
\newblock \showarticletitle{{NATS-Bench}: Benchmarking NAS Algorithms for
  Architecture Topology and Size}.
\newblock \bibinfo{journal}{\emph{IEEE Transactions on Pattern Analysis and
  Machine Intelligence (TPAMI)}} (\bibinfo{year}{2021}).
\newblock
\urldef\tempurl%
\url{https://doi.org/10.1109/TPAMI.2021.3054824}
\showDOI{\tempurl}
\newblock
\shownote{\mbox{doi}:\url{10.1109/TPAMI.2021.3054824}}.


\bibitem[\protect\citeauthoryear{Dong, Tan, Yu, Peng, Gabrys, and Le}{Dong
  et~al\mbox{.}}{2021b}]%
        {dong2021autohas}
\bibfield{author}{\bibinfo{person}{Xuanyi Dong}, \bibinfo{person}{Mingxing
  Tan}, \bibinfo{person}{Adams~Wei Yu}, \bibinfo{person}{Daiyi Peng},
  \bibinfo{person}{Bogdan Gabrys}, {and} \bibinfo{person}{Quoc~V Le}.}
  \bibinfo{year}{2021}\natexlab{b}.
\newblock \showarticletitle{{AutoHAS}: Efficient Hyperparameter and
  Architecture Search}. In \bibinfo{booktitle}{\emph{International Conference
  on Learning Representations (ICLR) Workshop}}.
\newblock


\bibitem[\protect\citeauthoryear{Dong and Yang}{Dong and Yang}{2019a}]%
        {dong2019tas}
\bibfield{author}{\bibinfo{person}{Xuanyi Dong} {and} \bibinfo{person}{Yi
  Yang}.} \bibinfo{year}{2019}\natexlab{a}.
\newblock \showarticletitle{Network Pruning via Transformable Architecture
  Search}. In \bibinfo{booktitle}{\emph{Proceedings of the International
  Conference on Neural Information Processing Systems (NeurIPS)}}.
  \bibinfo{pages}{760--771}.
\newblock


\bibitem[\protect\citeauthoryear{Dong and Yang}{Dong and Yang}{2019b}]%
        {dong2019one}
\bibfield{author}{\bibinfo{person}{Xuanyi Dong} {and} \bibinfo{person}{Yi
  Yang}.} \bibinfo{year}{2019}\natexlab{b}.
\newblock \showarticletitle{One-Shot Neural Architecture Search via
  Self-Evaluated Template Network}. In \bibinfo{booktitle}{\emph{Proceedings of
  the IEEE International Conference Computer Vision (ICCV)}}.
  \bibinfo{pages}{3681--3690}.
\newblock
\urldef\tempurl%
\url{https://doi.org/10.1109/ICCV.2019.00378}
\showDOI{\tempurl}


\bibitem[\protect\citeauthoryear{Dong and Yang}{Dong and Yang}{2019c}]%
        {dong2019search}
\bibfield{author}{\bibinfo{person}{Xuanyi Dong} {and} \bibinfo{person}{Yi
  Yang}.} \bibinfo{year}{2019}\natexlab{c}.
\newblock \showarticletitle{Searching for A Robust Neural Architecture in Four
  GPU Hours}. In \bibinfo{booktitle}{\emph{Proceedings of the IEEE Conference
  Computer Vision Pattern Recognition (CVPR)}}. \bibinfo{pages}{1761--1770}.
\newblock


\bibitem[\protect\citeauthoryear{Dong and Yang}{Dong and Yang}{2019d}]%
        {dong2019teacher}
\bibfield{author}{\bibinfo{person}{Xuanyi Dong} {and} \bibinfo{person}{Yi
  Yang}.} \bibinfo{year}{2019}\natexlab{d}.
\newblock \showarticletitle{Teacher Supervises Students How to Learn from
  Partially Labeled Images for Facial Landmark Detection}. In
  \bibinfo{booktitle}{\emph{Proceedings of the IEEE International Conference
  Computer Vision (ICCV)}}. \bibinfo{pages}{783--792}.
\newblock


\bibitem[\protect\citeauthoryear{Dong and Yang}{Dong and Yang}{2020}]%
        {dong2020bench}
\bibfield{author}{\bibinfo{person}{Xuanyi Dong} {and} \bibinfo{person}{Yi
  Yang}.} \bibinfo{year}{2020}\natexlab{}.
\newblock \showarticletitle{{NAS-bench-201}: Extending the scope of
  reproducible neural architecture search}. In
  \bibinfo{booktitle}{\emph{International Conference on Learning
  Representations (ICLR)}}.
\newblock


\bibitem[\protect\citeauthoryear{Du, Wang, Yang, Zhou, and Tang}{Du
  et~al\mbox{.}}{2019}]%
        {du2019sequential}
\bibfield{author}{\bibinfo{person}{Zhengxiao Du}, \bibinfo{person}{Xiaowei
  Wang}, \bibinfo{person}{Hongxia Yang}, \bibinfo{person}{Jingren Zhou}, {and}
  \bibinfo{person}{Jie Tang}.} \bibinfo{year}{2019}\natexlab{}.
\newblock \showarticletitle{Sequential scenario-specific meta learner for
  online recommendation}. In \bibinfo{booktitle}{\emph{Proceedings of the ACM
  SIGKDD International Conference on Knowledge Discovery and Data Mining
  (KDD)}}. \bibinfo{pages}{2895--2904}.
\newblock


\bibitem[\protect\citeauthoryear{Duan, Schulman, Chen, Bartlett, Sutskever, and
  Abbeel}{Duan et~al\mbox{.}}{2016}]%
        {duan2016rl}
\bibfield{author}{\bibinfo{person}{Yan Duan}, \bibinfo{person}{John Schulman},
  \bibinfo{person}{Xi Chen}, \bibinfo{person}{Peter~L Bartlett},
  \bibinfo{person}{Ilya Sutskever}, {and} \bibinfo{person}{Pieter Abbeel}.}
  \bibinfo{year}{2016}\natexlab{}.
\newblock \showarticletitle{RL2: Fast reinforcement learning via slow
  reinforcement learning}.
\newblock \bibinfo{journal}{\emph{arXiv preprint arXiv:1611.02779}}
  (\bibinfo{year}{2016}).
\newblock


\bibitem[\protect\citeauthoryear{Dura-Bernal, Suter, Gleeson, Cantarelli,
  Quintana, Rodriguez, Kedziora, Chadderdon, Kerr, Neymotin, McDougal, Hines,
  Shepherd, and Lytton}{Dura-Bernal et~al\mbox{.}}{2019}]%
        {dusu19}
\bibfield{author}{\bibinfo{person}{Salvador Dura-Bernal},
  \bibinfo{person}{Benjamin~A Suter}, \bibinfo{person}{Padraig Gleeson},
  \bibinfo{person}{Matteo Cantarelli}, \bibinfo{person}{Adrian Quintana},
  \bibinfo{person}{Facundo Rodriguez}, \bibinfo{person}{David~J Kedziora},
  \bibinfo{person}{George~L Chadderdon}, \bibinfo{person}{Cliff~C Kerr},
  \bibinfo{person}{Samuel~A Neymotin}, \bibinfo{person}{Robert~A McDougal},
  \bibinfo{person}{Michael Hines}, \bibinfo{person}{Gordon~MG Shepherd}, {and}
  \bibinfo{person}{William~W Lytton}.} \bibinfo{year}{2019}\natexlab{}.
\newblock \showarticletitle{{NetPyNE}, a tool for data-driven multiscale
  modeling of brain circuits}.
\newblock \bibinfo{journal}{\emph{{eLife}}}  \bibinfo{volume}{8}
  (\bibinfo{date}{apr} \bibinfo{year}{2019}).
\newblock
\urldef\tempurl%
\url{https://doi.org/10.7554/elife.44494}
\showDOI{\tempurl}


\bibitem[\protect\citeauthoryear{Eggensperger, Hutter, Hoos, and
  Leyton-Brown}{Eggensperger et~al\mbox{.}}{2014}]%
        {eggensperger2014surrogate}
\bibfield{author}{\bibinfo{person}{Katharina Eggensperger},
  \bibinfo{person}{Frank Hutter}, \bibinfo{person}{Holger~H Hoos}, {and}
  \bibinfo{person}{Kevin Leyton-Brown}.} \bibinfo{year}{2014}\natexlab{}.
\newblock \showarticletitle{Surrogate Benchmarks for Hyperparameter
  Optimization}. In \bibinfo{booktitle}{\emph{The European Conference on
  Artificial Intelligence (ECAI) Workshop}}. \bibinfo{pages}{24--31}.
\newblock


\bibitem[\protect\citeauthoryear{Eggensperger, M{\"u}ller, Mallik, Feurer,
  Sass, Klein, Awad, Lindauer, and Hutter}{Eggensperger et~al\mbox{.}}{2021}]%
        {eggensperger2021hpobench}
\bibfield{author}{\bibinfo{person}{Katharina Eggensperger},
  \bibinfo{person}{Philipp M{\"u}ller}, \bibinfo{person}{Neeratyoy Mallik},
  \bibinfo{person}{Matthias Feurer}, \bibinfo{person}{Ren{\'e} Sass},
  \bibinfo{person}{Aaron Klein}, \bibinfo{person}{Noor Awad},
  \bibinfo{person}{Marius Lindauer}, {and} \bibinfo{person}{Frank Hutter}.}
  \bibinfo{year}{2021}\natexlab{}.
\newblock \showarticletitle{HPOBench: A Collection of Reproducible
  Multi-Fidelity Benchmark Problems for HPO}. In
  \bibinfo{booktitle}{\emph{Proceedings of the International Conference on
  Neural Information Processing Systems (NeurIPS)}}.
\newblock


\bibitem[\protect\citeauthoryear{Eimer, Biedenkapp, Reimer, Adriaensen, Hutter,
  and Lindauer}{Eimer et~al\mbox{.}}{2021}]%
        {eimer2021dacbench}
\bibfield{author}{\bibinfo{person}{Theresa Eimer}, \bibinfo{person}{Andr{\'e}
  Biedenkapp}, \bibinfo{person}{Maximilian Reimer}, \bibinfo{person}{Steven
  Adriaensen}, \bibinfo{person}{Frank Hutter}, {and} \bibinfo{person}{Marius
  Lindauer}.} \bibinfo{year}{2021}\natexlab{}.
\newblock \showarticletitle{{DACBench}: A Benchmark Library for Dynamic
  Algorithm Configuration}. In \bibinfo{booktitle}{\emph{International Joint
  Conferences on Artificial Intelligence (IJCAI)}}.
  \bibinfo{pages}{1668--1674}.
\newblock


\bibitem[\protect\citeauthoryear{Elsken, Metzen, and Hutter}{Elsken
  et~al\mbox{.}}{2019}]%
        {elsken2019neural}
\bibfield{author}{\bibinfo{person}{Thomas Elsken}, \bibinfo{person}{Jan~Hendrik
  Metzen}, {and} \bibinfo{person}{Frank Hutter}.}
  \bibinfo{year}{2019}\natexlab{}.
\newblock \showarticletitle{Neural architecture search: A survey}.
\newblock \bibinfo{journal}{\emph{Journal of Machine Learning Research (JMLR)}}
  \bibinfo{volume}{20}, \bibinfo{number}{55} (\bibinfo{year}{2019}),
  \bibinfo{pages}{1--21}.
\newblock


\bibitem[\protect\citeauthoryear{{Epic Games}}{{Epic Games}}{2019}]%
        {unrealengine}
\bibfield{author}{\bibinfo{person}{{Epic Games}}.}
  \bibinfo{year}{2019}\natexlab{}.
\newblock \bibinfo{title}{Unreal Engine}.
\newblock \bibinfo{howpublished}{https://www.unrealengine.com}.
\newblock


\bibitem[\protect\citeauthoryear{Erickson, Mueller, Shirkov, Zhang, Larroy, Li,
  and Smola}{Erickson et~al\mbox{.}}{2020}]%
        {erickson2020autogluon}
\bibfield{author}{\bibinfo{person}{Nick Erickson}, \bibinfo{person}{Jonas
  Mueller}, \bibinfo{person}{Alexander Shirkov}, \bibinfo{person}{Hang Zhang},
  \bibinfo{person}{Pedro Larroy}, \bibinfo{person}{Mu Li}, {and}
  \bibinfo{person}{Alexander Smola}.} \bibinfo{year}{2020}\natexlab{}.
\newblock \showarticletitle{Autogluon-tabular: Robust and accurate automl for
  structured data}. In \bibinfo{booktitle}{\emph{The International Conference
  on Machine Learning (ICML) Workshop}}.
\newblock


\bibitem[\protect\citeauthoryear{Facebook AI Research}{Facebook AI
  Research}{[n.\,d.]}]%
        {paperswithcode}
Facebook AI Research \bibinfo{year}{[n.\,d.]}\natexlab{}.
\newblock \bibinfo{title}{Papers With Code}.
\newblock \bibinfo{howpublished}{https://paperswithcode.com}.
\newblock


\bibitem[\protect\citeauthoryear{Facebook Inc.}{Facebook Inc.}{2021}]%
        {pytorch_v1.10}
Facebook Inc. \bibinfo{year}{2021}\natexlab{}.
\newblock \bibinfo{title}{PyTorch V1.10.0}.
\newblock
  \bibinfo{howpublished}{https://github.com/pytorch/pytorch/releases/tag/v1.10.0}.
\newblock


\bibitem[\protect\citeauthoryear{Falkner, Klein, and Hutter}{Falkner
  et~al\mbox{.}}{2018}]%
        {falkner2018bohb}
\bibfield{author}{\bibinfo{person}{Stefan Falkner}, \bibinfo{person}{Aaron
  Klein}, {and} \bibinfo{person}{Frank Hutter}.}
  \bibinfo{year}{2018}\natexlab{}.
\newblock \showarticletitle{{BOHB}: Robust and efficient hyperparameter
  optimization at scale}. In \bibinfo{booktitle}{\emph{The International
  Conference on Machine Learning (ICML)}}. \bibinfo{pages}{1437--1446}.
\newblock


\bibitem[\protect\citeauthoryear{Fan, Tian, Qin, Li, and Liu}{Fan
  et~al\mbox{.}}{2018}]%
        {fan2018learning}
\bibfield{author}{\bibinfo{person}{Yang Fan}, \bibinfo{person}{Fei Tian},
  \bibinfo{person}{Tao Qin}, \bibinfo{person}{Xiang-Yang Li}, {and}
  \bibinfo{person}{Tie-Yan Liu}.} \bibinfo{year}{2018}\natexlab{}.
\newblock \showarticletitle{Learning to teach}. In
  \bibinfo{booktitle}{\emph{International Conference on Learning
  Representations (ICLR)}}.
\newblock


\bibitem[\protect\citeauthoryear{Fang, Sun, Zhang, Li, Liu, and Wang}{Fang
  et~al\mbox{.}}{2020}]%
        {fang2020densely}
\bibfield{author}{\bibinfo{person}{Jiemin Fang}, \bibinfo{person}{Yuzhu Sun},
  \bibinfo{person}{Qian Zhang}, \bibinfo{person}{Yuan Li},
  \bibinfo{person}{Wenyu Liu}, {and} \bibinfo{person}{Xinggang Wang}.}
  \bibinfo{year}{2020}\natexlab{}.
\newblock \showarticletitle{Densely connected search space for more flexible
  neural architecture search}. In \bibinfo{booktitle}{\emph{Proceedings of the
  IEEE Conference Computer Vision Pattern Recognition (CVPR)}}.
  \bibinfo{pages}{10628--10637}.
\newblock


\bibitem[\protect\citeauthoryear{Feurer and Hutter}{Feurer and Hutter}{2019}]%
        {feurer2019hyperparameter}
\bibfield{author}{\bibinfo{person}{Matthias Feurer} {and}
  \bibinfo{person}{Frank Hutter}.} \bibinfo{year}{2019}\natexlab{}.
\newblock \showarticletitle{Hyperparameter optimization}.
\newblock In \bibinfo{booktitle}{\emph{Automated Machine Learning}}.
  \bibinfo{publisher}{Springer, Cham}, \bibinfo{pages}{3--33}.
\newblock


\bibitem[\protect\citeauthoryear{Figurnov, Collins, Zhu, Zhang, Huang, Vetrov,
  and Salakhutdinov}{Figurnov et~al\mbox{.}}{2017}]%
        {figurnov2017spatially}
\bibfield{author}{\bibinfo{person}{Michael Figurnov},
  \bibinfo{person}{Maxwell~D Collins}, \bibinfo{person}{Yukun Zhu},
  \bibinfo{person}{Li Zhang}, \bibinfo{person}{Jonathan Huang},
  \bibinfo{person}{Dmitry Vetrov}, {and} \bibinfo{person}{Ruslan
  Salakhutdinov}.} \bibinfo{year}{2017}\natexlab{}.
\newblock \showarticletitle{Spatially adaptive computation time for residual
  networks}. In \bibinfo{booktitle}{\emph{Proceedings of the IEEE Conference
  Computer Vision Pattern Recognition (CVPR)}}.
\newblock


\bibitem[\protect\citeauthoryear{Finn, Abbeel, and Levine}{Finn
  et~al\mbox{.}}{2017}]%
        {finn2017model}
\bibfield{author}{\bibinfo{person}{Chelsea Finn}, \bibinfo{person}{Pieter
  Abbeel}, {and} \bibinfo{person}{Sergey Levine}.}
  \bibinfo{year}{2017}\natexlab{}.
\newblock \showarticletitle{Model-agnostic meta-learning for fast adaptation of
  deep networks}. In \bibinfo{booktitle}{\emph{The International Conference on
  Machine Learning (ICML)}}. \bibinfo{pages}{1126--1135}.
\newblock


\bibitem[\protect\citeauthoryear{Forrester, S{\'o}bester, and Keane}{Forrester
  et~al\mbox{.}}{2007}]%
        {forrester2007multi}
\bibfield{author}{\bibinfo{person}{Alexander~IJ Forrester},
  \bibinfo{person}{Andr{\'a}s S{\'o}bester}, {and} \bibinfo{person}{Andy~J
  Keane}.} \bibinfo{year}{2007}\natexlab{}.
\newblock \showarticletitle{Multi-fidelity optimization via surrogate
  modelling}. In \bibinfo{booktitle}{\emph{Mathematical, Physical and
  Engineering Sciences}}, Vol.~\bibinfo{volume}{463}.
  \bibinfo{pages}{3251--3269}.
\newblock


\bibitem[\protect\citeauthoryear{Frankle and Carbin}{Frankle and
  Carbin}{2018}]%
        {frankle2018lottery}
\bibfield{author}{\bibinfo{person}{Jonathan Frankle} {and}
  \bibinfo{person}{Michael Carbin}.} \bibinfo{year}{2018}\natexlab{}.
\newblock \showarticletitle{The lottery ticket hypothesis: Finding sparse,
  trainable neural networks}. In \bibinfo{booktitle}{\emph{The International
  Conference on Machine Learning (ICML)}}.
\newblock


\bibitem[\protect\citeauthoryear{Fukushima}{Fukushima}{2013}]%
        {fukushima2013training}
\bibfield{author}{\bibinfo{person}{Kunihiko Fukushima}.}
  \bibinfo{year}{2013}\natexlab{}.
\newblock \showarticletitle{Training multi-layered neural network
  neocognitron}.
\newblock \bibinfo{journal}{\emph{Neural Networks}}  \bibinfo{volume}{40}
  (\bibinfo{year}{2013}), \bibinfo{pages}{18--31}.
\newblock


\bibitem[\protect\citeauthoryear{Fukushima and Miyake}{Fukushima and
  Miyake}{1982}]%
        {fukushima1982neocognitron}
\bibfield{author}{\bibinfo{person}{Kunihiko Fukushima} {and}
  \bibinfo{person}{Sei Miyake}.} \bibinfo{year}{1982}\natexlab{}.
\newblock \showarticletitle{Neocognitron: A self-organizing neural network
  model for a mechanism of visual pattern recognition}.
\newblock In \bibinfo{booktitle}{\emph{Competition and Cooperation in Neural
  Nets}}. \bibinfo{publisher}{Springer}, \bibinfo{pages}{267--285}.
\newblock


\bibitem[\protect\citeauthoryear{Gama, {\v{Z}}liobait{\.e}, Bifet, Pechenizkiy,
  and Bouchachia}{Gama et~al\mbox{.}}{2014}]%
        {gama2014survey}
\bibfield{author}{\bibinfo{person}{Jo{\~a}o Gama}, \bibinfo{person}{Indr{\.e}
  {\v{Z}}liobait{\.e}}, \bibinfo{person}{Albert Bifet}, \bibinfo{person}{Mykola
  Pechenizkiy}, {and} \bibinfo{person}{Abdelhamid Bouchachia}.}
  \bibinfo{year}{2014}\natexlab{}.
\newblock \showarticletitle{A survey on concept drift adaptation}.
\newblock \bibinfo{journal}{\emph{ACM Computing Surveys (CSUR)}}
  \bibinfo{volume}{46}, \bibinfo{number}{4} (\bibinfo{year}{2014}),
  \bibinfo{pages}{1--37}.
\newblock


\bibitem[\protect\citeauthoryear{Ganin and Lempitsky}{Ganin and
  Lempitsky}{2015}]%
        {ganin2015unsupervised}
\bibfield{author}{\bibinfo{person}{Yaroslav Ganin} {and}
  \bibinfo{person}{Victor Lempitsky}.} \bibinfo{year}{2015}\natexlab{}.
\newblock \showarticletitle{Unsupervised domain adaptation by backpropagation}.
  In \bibinfo{booktitle}{\emph{The International Conference on Machine Learning
  (ICML)}}. \bibinfo{pages}{1180--1189}.
\newblock


\bibitem[\protect\citeauthoryear{Gelernter}{Gelernter}{1993}]%
        {gelernter1993mirror}
\bibfield{author}{\bibinfo{person}{David Gelernter}.}
  \bibinfo{year}{1993}\natexlab{}.
\newblock \bibinfo{booktitle}{\emph{Mirror worlds: Or the day software puts the
  universe in a shoebox... How it will happen and what it will mean}}.
\newblock \bibinfo{publisher}{Oxford University Press}.
\newblock


\bibitem[\protect\citeauthoryear{Gemp, Theocharous, and Ghavamzadeh}{Gemp
  et~al\mbox{.}}{2017}]%
        {gemp2017automated}
\bibfield{author}{\bibinfo{person}{Ian Gemp}, \bibinfo{person}{Georgios
  Theocharous}, {and} \bibinfo{person}{Mohammad Ghavamzadeh}.}
  \bibinfo{year}{2017}\natexlab{}.
\newblock \showarticletitle{Automated data cleansing through meta-learning}. In
  \bibinfo{booktitle}{\emph{AAAI Conference on Artificial Intelligence
  (AAAI)}}. \bibinfo{pages}{4760--–4761}.
\newblock


\bibitem[\protect\citeauthoryear{Gencoglu, van Gils, Guldogan, Morikawa,
  S{\"u}zen, Gruber, Leinonen, and Huttunen}{Gencoglu et~al\mbox{.}}{2019}]%
        {gencoglu2019hark}
\bibfield{author}{\bibinfo{person}{Oguzhan Gencoglu}, \bibinfo{person}{Mark van
  Gils}, \bibinfo{person}{Esin Guldogan}, \bibinfo{person}{Chamin Morikawa},
  \bibinfo{person}{Mehmet S{\"u}zen}, \bibinfo{person}{Mathias Gruber},
  \bibinfo{person}{Jussi Leinonen}, {and} \bibinfo{person}{Heikki Huttunen}.}
  \bibinfo{year}{2019}\natexlab{}.
\newblock \showarticletitle{HARK Side of Deep Learning--From Grad Student
  Descent to Automated Machine Learning}.
\newblock \bibinfo{journal}{\emph{arXiv preprint arXiv:1904.07633}}
  (\bibinfo{year}{2019}).
\newblock


\bibitem[\protect\citeauthoryear{Ghiasi, Lin, and Le}{Ghiasi
  et~al\mbox{.}}{2019}]%
        {ghiasi2019fpn}
\bibfield{author}{\bibinfo{person}{Golnaz Ghiasi}, \bibinfo{person}{Tsung-Yi
  Lin}, {and} \bibinfo{person}{Quoc~V Le}.} \bibinfo{year}{2019}\natexlab{}.
\newblock \showarticletitle{{NAS-FPN}: Learning scalable feature pyramid
  architecture for object detection}. In \bibinfo{booktitle}{\emph{Proceedings
  of the IEEE Conference Computer Vision Pattern Recognition (CVPR)}}.
  \bibinfo{pages}{7036--7045}.
\newblock


\bibitem[\protect\citeauthoryear{Goldberg and Deb}{Goldberg and Deb}{1991}]%
        {goldberg1991comparative}
\bibfield{author}{\bibinfo{person}{David~E Goldberg} {and}
  \bibinfo{person}{Kalyanmoy Deb}.} \bibinfo{year}{1991}\natexlab{}.
\newblock \showarticletitle{A comparative analysis of selection schemes used in
  genetic algorithms}.
\newblock In \bibinfo{booktitle}{\emph{Foundations of Genetic Algorithms}}.
  Vol.~\bibinfo{volume}{1}. \bibinfo{pages}{69--93}.
\newblock


\bibitem[\protect\citeauthoryear{Golovin, Solnik, Moitra, Kochanski, Karro, and
  Sculley}{Golovin et~al\mbox{.}}{2017}]%
        {golovin2017google}
\bibfield{author}{\bibinfo{person}{Daniel Golovin}, \bibinfo{person}{Benjamin
  Solnik}, \bibinfo{person}{Subhodeep Moitra}, \bibinfo{person}{Greg
  Kochanski}, \bibinfo{person}{John Karro}, {and} \bibinfo{person}{David
  Sculley}.} \bibinfo{year}{2017}\natexlab{}.
\newblock \showarticletitle{Google vizier: A service for black-box
  optimization}. In \bibinfo{booktitle}{\emph{Proceedings of the ACM SIGKDD
  International Conference on Knowledge Discovery and Data Mining (KDD)}}.
  \bibinfo{pages}{1487--1495}.
\newblock


\bibitem[\protect\citeauthoryear{Goodfellow, Pouget-Abadie, Mirza, Xu,
  Warde-Farley, Ozair, Courville, and Bengio}{Goodfellow et~al\mbox{.}}{2014}]%
        {goodfellow2014generative}
\bibfield{author}{\bibinfo{person}{Ian~J Goodfellow}, \bibinfo{person}{Jean
  Pouget-Abadie}, \bibinfo{person}{Mehdi Mirza}, \bibinfo{person}{Bing Xu},
  \bibinfo{person}{David Warde-Farley}, \bibinfo{person}{Sherjil Ozair},
  \bibinfo{person}{Aaron Courville}, {and} \bibinfo{person}{Yoshua Bengio}.}
  \bibinfo{year}{2014}\natexlab{}.
\newblock \showarticletitle{Generative adversarial networks}. In
  \bibinfo{booktitle}{\emph{Proceedings of the International Conference on
  Neural Information Processing Systems (NeurIPS)}}.
\newblock


\bibitem[\protect\citeauthoryear{Google}{Google}{2017}]%
        {edge_tpu}
\bibfield{author}{\bibinfo{person}{Google}.} \bibinfo{year}{2017}\natexlab{}.
\newblock \bibinfo{title}{{Edge TPU}}.
\newblock \bibinfo{howpublished}{\url{https://cloud.google.com/edge-tpu}}.
\newblock


\bibitem[\protect\citeauthoryear{Grossberg}{Grossberg}{2020}]%
        {grossberg2020toward}
\bibfield{author}{\bibinfo{person}{Stephen Grossberg}.}
  \bibinfo{year}{2020}\natexlab{}.
\newblock \showarticletitle{Toward Autonomous Adaptive Intelligence: Building
  Upon Neural Models of How Brains Make Minds}.
\newblock \bibinfo{journal}{\emph{IEEE Transactions on Systems, Man, and
  Cybernetics: Systems (TSMC)}} \bibinfo{volume}{51}, \bibinfo{number}{1}
  (\bibinfo{year}{2020}), \bibinfo{pages}{51--75}.
\newblock


\bibitem[\protect\citeauthoryear{Guidotti, Monreale, Ruggieri, Turini,
  Giannotti, and Pedreschi}{Guidotti et~al\mbox{.}}{2019}]%
        {guidotti2019survey}
\bibfield{author}{\bibinfo{person}{Riccardo Guidotti}, \bibinfo{person}{Anna
  Monreale}, \bibinfo{person}{Salvatore Ruggieri}, \bibinfo{person}{Franco
  Turini}, \bibinfo{person}{Fosca Giannotti}, {and} \bibinfo{person}{Dino
  Pedreschi}.} \bibinfo{year}{2019}\natexlab{}.
\newblock \showarticletitle{A survey of methods for explaining black box
  models}.
\newblock \bibinfo{journal}{\emph{ACM Computing Surveys (CSUR)}}
  \bibinfo{volume}{51}, \bibinfo{number}{5} (\bibinfo{year}{2019}),
  \bibinfo{pages}{1--42}.
\newblock


\bibitem[\protect\citeauthoryear{Guo, Zhang, Mu, Heng, Liu, Wei, and Sun}{Guo
  et~al\mbox{.}}{2020}]%
        {guo2020single}
\bibfield{author}{\bibinfo{person}{Zichao Guo}, \bibinfo{person}{Xiangyu
  Zhang}, \bibinfo{person}{Haoyuan Mu}, \bibinfo{person}{Wen Heng},
  \bibinfo{person}{Zechun Liu}, \bibinfo{person}{Yichen Wei}, {and}
  \bibinfo{person}{Jian Sun}.} \bibinfo{year}{2020}\natexlab{}.
\newblock \showarticletitle{Single path one-shot neural architecture search
  with uniform sampling}. In \bibinfo{booktitle}{\emph{Proceedings of the
  European Conference on Computer Vision (ECCV)}}. Springer,
  \bibinfo{pages}{544--560}.
\newblock


\bibitem[\protect\citeauthoryear{Ha, Dai, and Le}{Ha et~al\mbox{.}}{2017}]%
        {ha2017hypernetworks}
\bibfield{author}{\bibinfo{person}{David Ha}, \bibinfo{person}{Andrew Dai},
  {and} \bibinfo{person}{Quoc~V Le}.} \bibinfo{year}{2017}\natexlab{}.
\newblock \showarticletitle{Hypernetworks}. In
  \bibinfo{booktitle}{\emph{International Conference on Learning
  Representations (ICLR)}}.
\newblock


\bibitem[\protect\citeauthoryear{Han, Liu, Mao, Pu, Pedram, Horowitz, and
  Dally}{Han et~al\mbox{.}}{2016a}]%
        {han2016eie}
\bibfield{author}{\bibinfo{person}{Song Han}, \bibinfo{person}{Xingyu Liu},
  \bibinfo{person}{Huizi Mao}, \bibinfo{person}{Jing Pu},
  \bibinfo{person}{Ardavan Pedram}, \bibinfo{person}{Mark~A Horowitz}, {and}
  \bibinfo{person}{William~J Dally}.} \bibinfo{year}{2016}\natexlab{a}.
\newblock \showarticletitle{EIE: Efficient inference engine on compressed deep
  neural network}.
\newblock \bibinfo{journal}{\emph{ACM SIGARCH Computer Architecture News}}
  \bibinfo{volume}{44}, \bibinfo{number}{3} (\bibinfo{year}{2016}),
  \bibinfo{pages}{243--254}.
\newblock


\bibitem[\protect\citeauthoryear{Han, Mao, and Dally}{Han
  et~al\mbox{.}}{2016b}]%
        {han2016deep}
\bibfield{author}{\bibinfo{person}{Song Han}, \bibinfo{person}{Huizi Mao},
  {and} \bibinfo{person}{William~J Dally}.} \bibinfo{year}{2016}\natexlab{b}.
\newblock \showarticletitle{Deep compression: Compressing deep neural networks
  with pruning, trained quantization and huffman coding}. In
  \bibinfo{booktitle}{\emph{International Conference on Learning
  Representations (ICLR)}}.
\newblock


\bibitem[\protect\citeauthoryear{Han, Huang, Song, Yang, Wang, and Wang}{Han
  et~al\mbox{.}}{2021}]%
        {han2021dynamic}
\bibfield{author}{\bibinfo{person}{Yizeng Han}, \bibinfo{person}{Gao Huang},
  \bibinfo{person}{Shiji Song}, \bibinfo{person}{Le Yang},
  \bibinfo{person}{Honghui Wang}, {and} \bibinfo{person}{Yulin Wang}.}
  \bibinfo{year}{2021}\natexlab{}.
\newblock \showarticletitle{Dynamic neural networks: A survey}.
\newblock \bibinfo{journal}{\emph{arXiv preprint arXiv:2102.04906}}
  (\bibinfo{year}{2021}).
\newblock


\bibitem[\protect\citeauthoryear{Hansen and Ostermeier}{Hansen and
  Ostermeier}{2001}]%
        {hansen2001completely}
\bibfield{author}{\bibinfo{person}{Nikolaus Hansen} {and}
  \bibinfo{person}{Andreas Ostermeier}.} \bibinfo{year}{2001}\natexlab{}.
\newblock \showarticletitle{Completely derandomized self-adaptation in
  evolution strategies}.
\newblock \bibinfo{journal}{\emph{Evolutionary Computation}}
  \bibinfo{volume}{9}, \bibinfo{number}{2} (\bibinfo{year}{2001}),
  \bibinfo{pages}{159--195}.
\newblock


\bibitem[\protect\citeauthoryear{He, Gkioxari, Doll{\'a}r, and Girshick}{He
  et~al\mbox{.}}{2017a}]%
        {he2017mask}
\bibfield{author}{\bibinfo{person}{Kaiming He}, \bibinfo{person}{Georgia
  Gkioxari}, \bibinfo{person}{Piotr Doll{\'a}r}, {and} \bibinfo{person}{Ross
  Girshick}.} \bibinfo{year}{2017}\natexlab{a}.
\newblock \showarticletitle{Mask r-cnn}. In
  \bibinfo{booktitle}{\emph{Proceedings of the IEEE International Conference
  Computer Vision (ICCV)}}. \bibinfo{pages}{2961--2969}.
\newblock


\bibitem[\protect\citeauthoryear{He, Zhang, Ren, and Sun}{He
  et~al\mbox{.}}{2016}]%
        {he2016deep}
\bibfield{author}{\bibinfo{person}{Kaiming He}, \bibinfo{person}{Xiangyu
  Zhang}, \bibinfo{person}{Shaoqing Ren}, {and} \bibinfo{person}{Jian Sun}.}
  \bibinfo{year}{2016}\natexlab{}.
\newblock \showarticletitle{Deep residual learning for image recognition}. In
  \bibinfo{booktitle}{\emph{Proceedings of the IEEE Conference Computer Vision
  Pattern Recognition (CVPR)}}. \bibinfo{pages}{770--778}.
\newblock


\bibitem[\protect\citeauthoryear{He, Liao, Zhang, Nie, Hu, and Chua}{He
  et~al\mbox{.}}{2017b}]%
        {he2017neural}
\bibfield{author}{\bibinfo{person}{Xiangnan He}, \bibinfo{person}{Lizi Liao},
  \bibinfo{person}{Hanwang Zhang}, \bibinfo{person}{Liqiang Nie},
  \bibinfo{person}{Xia Hu}, {and} \bibinfo{person}{Tat-Seng Chua}.}
  \bibinfo{year}{2017}\natexlab{b}.
\newblock \showarticletitle{Neural collaborative filtering}. In
  \bibinfo{booktitle}{\emph{Proceedings of the International Conference on
  World Wide Web (WWW)}}. \bibinfo{pages}{173--182}.
\newblock


\bibitem[\protect\citeauthoryear{Hendrycks and Gimpel}{Hendrycks and
  Gimpel}{2017}]%
        {hendrycks2017baseline}
\bibfield{author}{\bibinfo{person}{Dan Hendrycks} {and} \bibinfo{person}{Kevin
  Gimpel}.} \bibinfo{year}{2017}\natexlab{}.
\newblock \showarticletitle{A baseline for detecting misclassified and
  out-of-distribution examples in neural networks}. In
  \bibinfo{booktitle}{\emph{International Conference on Learning
  Representations (ICLR)}}.
\newblock


\bibitem[\protect\citeauthoryear{Hiesinger}{Hiesinger}{2021}]%
        {hi21}
\bibfield{author}{\bibinfo{person}{Peter~Robin Hiesinger}.}
  \bibinfo{year}{2021}\natexlab{}.
\newblock \bibinfo{booktitle}{\emph{The Self-Assembling Brain: How Neural
  Networks Grow Smarter}}.
\newblock \bibinfo{publisher}{Princeton University Press}.
\newblock
\urldef\tempurl%
\url{https://doi.org/doi:10.1515/9780691215518}
\showDOI{\tempurl}


\bibitem[\protect\citeauthoryear{Hinton, Vinyals, and Dean}{Hinton
  et~al\mbox{.}}{2014}]%
        {hinton2014distilling}
\bibfield{author}{\bibinfo{person}{Geoffrey Hinton}, \bibinfo{person}{Oriol
  Vinyals}, {and} \bibinfo{person}{Jeff Dean}.}
  \bibinfo{year}{2014}\natexlab{}.
\newblock \showarticletitle{Distilling the knowledge in a neural network}. In
  \bibinfo{booktitle}{\emph{Proceedings of the International Conference on
  Neural Information Processing Systems (NeurIPS) Workshop}}.
\newblock


\bibitem[\protect\citeauthoryear{Hinton and Plaut}{Hinton and Plaut}{1987}]%
        {hinton1987using}
\bibfield{author}{\bibinfo{person}{Geoffrey~E Hinton} {and}
  \bibinfo{person}{David~C Plaut}.} \bibinfo{year}{1987}\natexlab{}.
\newblock \showarticletitle{Using fast weights to deblur old memories}. In
  \bibinfo{booktitle}{\emph{Proceedings of the 9th Annual Conference of the
  Cognitive Science Society}}. \bibinfo{pages}{177--186}.
\newblock


\bibitem[\protect\citeauthoryear{Hochreiter and Schmidhuber}{Hochreiter and
  Schmidhuber}{1997}]%
        {hochreiter1997long}
\bibfield{author}{\bibinfo{person}{Sepp Hochreiter} {and}
  \bibinfo{person}{J{\"u}rgen Schmidhuber}.} \bibinfo{year}{1997}\natexlab{}.
\newblock \showarticletitle{Long short-term memory}.
\newblock \bibinfo{journal}{\emph{Neural Computation}} \bibinfo{volume}{9},
  \bibinfo{number}{8} (\bibinfo{year}{1997}), \bibinfo{pages}{1735--1780}.
\newblock


\bibitem[\protect\citeauthoryear{Hochreiter, Younger, and Conwell}{Hochreiter
  et~al\mbox{.}}{2001}]%
        {hochreiter2001learning}
\bibfield{author}{\bibinfo{person}{Sepp Hochreiter}, \bibinfo{person}{A~Steven
  Younger}, {and} \bibinfo{person}{Peter~R Conwell}.}
  \bibinfo{year}{2001}\natexlab{}.
\newblock \showarticletitle{Learning to learn using gradient descent}. In
  \bibinfo{booktitle}{\emph{Proceedings of the International Conference on
  Artificial Neural Networks (ICANN)}}. \bibinfo{pages}{87--94}.
\newblock


\bibitem[\protect\citeauthoryear{Hofmann, Sch{\"o}lkopf, and Smola}{Hofmann
  et~al\mbox{.}}{2008}]%
        {hofmann2008kernel}
\bibfield{author}{\bibinfo{person}{Thomas Hofmann}, \bibinfo{person}{Bernhard
  Sch{\"o}lkopf}, {and} \bibinfo{person}{Alexander~J Smola}.}
  \bibinfo{year}{2008}\natexlab{}.
\newblock \showarticletitle{Kernel methods in machine learning}.
\newblock \bibinfo{journal}{\emph{The Annals of Statistics}}
  (\bibinfo{year}{2008}), \bibinfo{pages}{1171--1220}.
\newblock


\bibitem[\protect\citeauthoryear{Hopfield}{Hopfield}{1982}]%
        {hopfield1982neural}
\bibfield{author}{\bibinfo{person}{John~J Hopfield}.}
  \bibinfo{year}{1982}\natexlab{}.
\newblock \showarticletitle{Neural networks and physical systems with emergent
  collective computational abilities}.
\newblock \bibinfo{journal}{\emph{Proceedings of the National Academy of
  Sciences}} \bibinfo{volume}{79}, \bibinfo{number}{8} (\bibinfo{year}{1982}),
  \bibinfo{pages}{2554--2558}.
\newblock


\bibitem[\protect\citeauthoryear{Hornik, Stinchcombe, and White}{Hornik
  et~al\mbox{.}}{1989}]%
        {hornik1989multilayer}
\bibfield{author}{\bibinfo{person}{Kurt Hornik}, \bibinfo{person}{Maxwell
  Stinchcombe}, {and} \bibinfo{person}{Halbert White}.}
  \bibinfo{year}{1989}\natexlab{}.
\newblock \showarticletitle{Multilayer feedforward networks are universal
  approximators}.
\newblock \bibinfo{journal}{\emph{Neural Networks}} \bibinfo{volume}{2},
  \bibinfo{number}{5} (\bibinfo{year}{1989}), \bibinfo{pages}{359--366}.
\newblock


\bibitem[\protect\citeauthoryear{Hotelling}{Hotelling}{1933}]%
        {hotelling1933analysis}
\bibfield{author}{\bibinfo{person}{Harold Hotelling}.}
  \bibinfo{year}{1933}\natexlab{}.
\newblock \showarticletitle{Analysis of a complex of statistical variables into
  principal components.}
\newblock \bibinfo{journal}{\emph{Journal of Educational Psychology}}
  \bibinfo{volume}{24}, \bibinfo{number}{6} (\bibinfo{year}{1933}),
  \bibinfo{pages}{417}.
\newblock


\bibitem[\protect\citeauthoryear{Houthooft, Chen, Isola, Stadie, Wolski, Ho,
  and Abbeel}{Houthooft et~al\mbox{.}}{2018}]%
        {houthooft2018evolved}
\bibfield{author}{\bibinfo{person}{Rein Houthooft}, \bibinfo{person}{Richard~Y
  Chen}, \bibinfo{person}{Phillip Isola}, \bibinfo{person}{Bradly~C Stadie},
  \bibinfo{person}{Filip Wolski}, \bibinfo{person}{Jonathan Ho}, {and}
  \bibinfo{person}{Pieter Abbeel}.} \bibinfo{year}{2018}\natexlab{}.
\newblock \showarticletitle{Evolved policy gradients}. In
  \bibinfo{booktitle}{\emph{Proceedings of the International Conference on
  Neural Information Processing Systems (NeurIPS)}}.
\newblock


\bibitem[\protect\citeauthoryear{Hu, Langford, Caruana, Horvitz, and Dey}{Hu
  et~al\mbox{.}}{2018}]%
        {hula18}
\bibfield{author}{\bibinfo{person}{Hanzhang Hu}, \bibinfo{person}{John
  Langford}, \bibinfo{person}{Rich Caruana}, \bibinfo{person}{Eric Horvitz},
  {and} \bibinfo{person}{Debadeepta Dey}.} \bibinfo{year}{2018}\natexlab{}.
\newblock \showarticletitle{Macro Neural Architecture Search Revisited}. In
  \bibinfo{booktitle}{\emph{Proceedings of the International Conference on
  Neural Information Processing Systems (NeurIPS) Workshop}}.
\newblock


\bibitem[\protect\citeauthoryear{Huang, Liu, Van Der~Maaten, and
  Weinberger}{Huang et~al\mbox{.}}{2017}]%
        {huang2017densely}
\bibfield{author}{\bibinfo{person}{Gao Huang}, \bibinfo{person}{Zhuang Liu},
  \bibinfo{person}{Laurens Van Der~Maaten}, {and} \bibinfo{person}{Kilian~Q
  Weinberger}.} \bibinfo{year}{2017}\natexlab{}.
\newblock \showarticletitle{Densely connected convolutional networks}. In
  \bibinfo{booktitle}{\emph{Proceedings of the IEEE Conference Computer Vision
  Pattern Recognition (CVPR)}}. \bibinfo{pages}{4700--4708}.
\newblock


\bibitem[\protect\citeauthoryear{Hutter, Hoos, and Leyton-Brown}{Hutter
  et~al\mbox{.}}{2011}]%
        {hutter2011sequential}
\bibfield{author}{\bibinfo{person}{Frank Hutter}, \bibinfo{person}{Holger~H
  Hoos}, {and} \bibinfo{person}{Kevin Leyton-Brown}.}
  \bibinfo{year}{2011}\natexlab{}.
\newblock \showarticletitle{Sequential model-based optimization for general
  algorithm configuration}. In \bibinfo{booktitle}{\emph{Proceedings of the
  International Conference on Learning and Intelligent Optimization (LION)}}.
  \bibinfo{pages}{507--523}.
\newblock


\bibitem[\protect\citeauthoryear{Hutter, Hoos, Leyton-Brown, and
  St{\"u}tzle}{Hutter et~al\mbox{.}}{2009}]%
        {hutter2009paramils}
\bibfield{author}{\bibinfo{person}{Frank Hutter}, \bibinfo{person}{Holger~H
  Hoos}, \bibinfo{person}{Kevin Leyton-Brown}, {and} \bibinfo{person}{Thomas
  St{\"u}tzle}.} \bibinfo{year}{2009}\natexlab{}.
\newblock \showarticletitle{{ParamILS}: an automatic algorithm configuration
  framework}.
\newblock \bibinfo{journal}{\emph{Journal of Artificial Intelligence Research
  (JAIR)}}  \bibinfo{volume}{36} (\bibinfo{year}{2009}),
  \bibinfo{pages}{267--306}.
\newblock


\bibitem[\protect\citeauthoryear{Hwang and Masud}{Hwang and Masud}{2012}]%
        {hwang2012multiple}
\bibfield{author}{\bibinfo{person}{Ching-Lai Hwang} {and} \bibinfo{person}{Abu
  Syed~Md Masud}.} \bibinfo{year}{2012}\natexlab{}.
\newblock \bibinfo{booktitle}{\emph{Multiple objective decision
  making—methods and applications: a state-of-the-art survey}}.
  Vol.~\bibinfo{volume}{164}.
\newblock \bibinfo{publisher}{Springer Science \& Business Media}.
\newblock


\bibitem[\protect\citeauthoryear{Ioffe and Szegedy}{Ioffe and Szegedy}{2015}]%
        {ioffe2015batch}
\bibfield{author}{\bibinfo{person}{Sergey Ioffe} {and}
  \bibinfo{person}{Christian Szegedy}.} \bibinfo{year}{2015}\natexlab{}.
\newblock \showarticletitle{Batch normalization: Accelerating deep network
  training by reducing internal covariate shift}. In
  \bibinfo{booktitle}{\emph{The International Conference on Machine Learning
  (ICML)}}. \bibinfo{pages}{448--456}.
\newblock


\bibitem[\protect\citeauthoryear{Ivakhnenko}{Ivakhnenko}{1971}]%
        {ivakhnenko1971polynomial}
\bibfield{author}{\bibinfo{person}{Alexey~Grigorevich Ivakhnenko}.}
  \bibinfo{year}{1971}\natexlab{}.
\newblock \showarticletitle{Polynomial theory of complex systems}.
\newblock \bibinfo{journal}{\emph{IEEE Transactions on Systems, Man, and
  Cybernetics (SMC)}}  \bibinfo{volume}{4} (\bibinfo{year}{1971}),
  \bibinfo{pages}{364--378}.
\newblock


\bibitem[\protect\citeauthoryear{Jacot, Gabriel, and Hongler}{Jacot
  et~al\mbox{.}}{2018}]%
        {jacot2018neural}
\bibfield{author}{\bibinfo{person}{Arthur Jacot}, \bibinfo{person}{Franck
  Gabriel}, {and} \bibinfo{person}{Cl{\'e}ment Hongler}.}
  \bibinfo{year}{2018}\natexlab{}.
\newblock \showarticletitle{Neural tangent kernel: convergence and
  generalization in neural networks}. In \bibinfo{booktitle}{\emph{Proceedings
  of the International Conference on Neural Information Processing Systems
  (NeurIPS)}}. \bibinfo{pages}{8580--8589}.
\newblock


\bibitem[\protect\citeauthoryear{Jaderberg, Dalibard, Osindero, Czarnecki,
  Donahue, Razavi, Vinyals, Green, Dunning, Simonyan, et~al\mbox{.}}{Jaderberg
  et~al\mbox{.}}{2017}]%
        {jaderberg2017population}
\bibfield{author}{\bibinfo{person}{Max Jaderberg}, \bibinfo{person}{Valentin
  Dalibard}, \bibinfo{person}{Simon Osindero}, \bibinfo{person}{Wojciech~M
  Czarnecki}, \bibinfo{person}{Jeff Donahue}, \bibinfo{person}{Ali Razavi},
  \bibinfo{person}{Oriol Vinyals}, \bibinfo{person}{Tim Green},
  \bibinfo{person}{Iain Dunning}, \bibinfo{person}{Karen Simonyan},
  {et~al\mbox{.}}} \bibinfo{year}{2017}\natexlab{}.
\newblock \showarticletitle{Population based training of neural networks}.
\newblock \bibinfo{journal}{\emph{arXiv preprint arXiv:1711.09846}}
  (\bibinfo{year}{2017}).
\newblock


\bibitem[\protect\citeauthoryear{Jamieson and Talwalkar}{Jamieson and
  Talwalkar}{2016}]%
        {jamieson2016non}
\bibfield{author}{\bibinfo{person}{Kevin Jamieson} {and} \bibinfo{person}{Ameet
  Talwalkar}.} \bibinfo{year}{2016}\natexlab{}.
\newblock \showarticletitle{Non-stochastic best arm identification and
  hyperparameter optimization}. In \bibinfo{booktitle}{\emph{The International
  Conference on Artificial Intelligence and Statistics (AISTATS)}}.
  \bibinfo{pages}{240--248}.
\newblock


\bibitem[\protect\citeauthoryear{Jang, Gu, and Poole}{Jang
  et~al\mbox{.}}{2017}]%
        {jang2017categorical}
\bibfield{author}{\bibinfo{person}{Eric Jang}, \bibinfo{person}{Shixiang Gu},
  {and} \bibinfo{person}{Ben Poole}.} \bibinfo{year}{2017}\natexlab{}.
\newblock \showarticletitle{Categorical reparameterization with
  gumbel-softmax}. In \bibinfo{booktitle}{\emph{International Conference on
  Learning Representations (ICLR)}}.
\newblock


\bibitem[\protect\citeauthoryear{Jenatton, Archambeau, Gonz{\'a}lez, and
  Seeger}{Jenatton et~al\mbox{.}}{2017}]%
        {jenatton2017bayesian}
\bibfield{author}{\bibinfo{person}{Rodolphe Jenatton}, \bibinfo{person}{Cedric
  Archambeau}, \bibinfo{person}{Javier Gonz{\'a}lez}, {and}
  \bibinfo{person}{Matthias Seeger}.} \bibinfo{year}{2017}\natexlab{}.
\newblock \showarticletitle{Bayesian optimization with tree-structured
  dependencies}. In \bibinfo{booktitle}{\emph{The International Conference on
  Machine Learning (ICML)}}. \bibinfo{pages}{1655--1664}.
\newblock


\bibitem[\protect\citeauthoryear{Ji, Xu, Yang, and Yu}{Ji
  et~al\mbox{.}}{2012}]%
        {ji20123d}
\bibfield{author}{\bibinfo{person}{Shuiwang Ji}, \bibinfo{person}{Wei Xu},
  \bibinfo{person}{Ming Yang}, {and} \bibinfo{person}{Kai Yu}.}
  \bibinfo{year}{2012}\natexlab{}.
\newblock \showarticletitle{{3D} convolutional neural networks for human action
  recognition}.
\newblock \bibinfo{journal}{\emph{IEEE Transactions on Pattern Analysis and
  Machine Intelligence (TPAMI)}} \bibinfo{volume}{35}, \bibinfo{number}{1}
  (\bibinfo{year}{2012}), \bibinfo{pages}{221--231}.
\newblock


\bibitem[\protect\citeauthoryear{Jiang, Yang, Sha, Zhuge, Gu, Dasgupta, Shi,
  and Hu}{Jiang et~al\mbox{.}}{2020}]%
        {jiang2020hardware}
\bibfield{author}{\bibinfo{person}{Weiwen Jiang}, \bibinfo{person}{Lei Yang},
  \bibinfo{person}{Edwin Hsing-Mean Sha}, \bibinfo{person}{Qingfeng Zhuge},
  \bibinfo{person}{Shouzhen Gu}, \bibinfo{person}{Sakyasingha Dasgupta},
  \bibinfo{person}{Yiyu Shi}, {and} \bibinfo{person}{Jingtong Hu}.}
  \bibinfo{year}{2020}\natexlab{}.
\newblock \showarticletitle{Hardware/software co-exploration of neural
  architectures}.
\newblock \bibinfo{journal}{\emph{IEEE Transactions on Computer-Aided Design of
  Integrated Circuits and Systems (TCAD)}} \bibinfo{volume}{39},
  \bibinfo{number}{12} (\bibinfo{year}{2020}), \bibinfo{pages}{4805--4815}.
\newblock


\bibitem[\protect\citeauthoryear{Jin, Song, and Hu}{Jin et~al\mbox{.}}{2019}]%
        {jin2019auto}
\bibfield{author}{\bibinfo{person}{Haifeng Jin}, \bibinfo{person}{Qingquan
  Song}, {and} \bibinfo{person}{Xia Hu}.} \bibinfo{year}{2019}\natexlab{}.
\newblock \showarticletitle{Auto-keras: An efficient neural architecture search
  system}. In \bibinfo{booktitle}{\emph{Proceedings of the 25th ACM SIGKDD
  International Conference on Knowledge Discovery \& Data Mining}}.
  \bibinfo{pages}{1946--1956}.
\newblock


\bibitem[\protect\citeauthoryear{Jomaa, Grabocka, and Schmidt-Thieme}{Jomaa
  et~al\mbox{.}}{2019}]%
        {jomaa2019hyp}
\bibfield{author}{\bibinfo{person}{Hadi~S Jomaa}, \bibinfo{person}{Josif
  Grabocka}, {and} \bibinfo{person}{Lars Schmidt-Thieme}.}
  \bibinfo{year}{2019}\natexlab{}.
\newblock \showarticletitle{{Hyp-RL}: Hyperparameter optimization by
  reinforcement learning}.
\newblock \bibinfo{journal}{\emph{arXiv preprint arXiv:1906.11527}}
  (\bibinfo{year}{2019}).
\newblock


\bibitem[\protect\citeauthoryear{Jozefowicz, Zaremba, and Sutskever}{Jozefowicz
  et~al\mbox{.}}{2015}]%
        {jozefowicz2015empirical}
\bibfield{author}{\bibinfo{person}{Rafal Jozefowicz}, \bibinfo{person}{Wojciech
  Zaremba}, {and} \bibinfo{person}{Ilya Sutskever}.}
  \bibinfo{year}{2015}\natexlab{}.
\newblock \showarticletitle{An empirical exploration of recurrent network
  architectures}. In \bibinfo{booktitle}{\emph{The International Conference on
  Machine Learning (ICML)}}. \bibinfo{pages}{2342--2350}.
\newblock


\bibitem[\protect\citeauthoryear{Kadlec and Gabrys}{Kadlec and Gabrys}{2009}]%
        {kadlec2009architecture}
\bibfield{author}{\bibinfo{person}{Petr Kadlec} {and} \bibinfo{person}{Bogdan
  Gabrys}.} \bibinfo{year}{2009}\natexlab{}.
\newblock \showarticletitle{Architecture for development of adaptive on-line
  prediction models}.
\newblock \bibinfo{journal}{\emph{Memetic Computing}} \bibinfo{volume}{1},
  \bibinfo{number}{4} (\bibinfo{year}{2009}), \bibinfo{pages}{241--269}.
\newblock


\bibitem[\protect\citeauthoryear{Kandasamy, Dasarathy, Oliva, Schneider, and
  P{\'o}czos}{Kandasamy et~al\mbox{.}}{2016}]%
        {kandasamy2016gaussian}
\bibfield{author}{\bibinfo{person}{Kirthevasan Kandasamy},
  \bibinfo{person}{Gautam Dasarathy}, \bibinfo{person}{Junier Oliva},
  \bibinfo{person}{Jeff Schneider}, {and} \bibinfo{person}{Barnab{\'a}s
  P{\'o}czos}.} \bibinfo{year}{2016}\natexlab{}.
\newblock \showarticletitle{Gaussian Process Bandit Optimisation with
  Multi-fidelity Evaluations}. In \bibinfo{booktitle}{\emph{Proceedings of the
  International Conference on Neural Information Processing Systems
  (NeurIPS)}}. \bibinfo{pages}{1000--1008}.
\newblock


\bibitem[\protect\citeauthoryear{Kandasamy, Neiswanger, Schneider, Poczos, and
  Xing}{Kandasamy et~al\mbox{.}}{2018}]%
        {kandasamy2018neural}
\bibfield{author}{\bibinfo{person}{Kirthevasan Kandasamy},
  \bibinfo{person}{Willie Neiswanger}, \bibinfo{person}{Jeff Schneider},
  \bibinfo{person}{Barnabas Poczos}, {and} \bibinfo{person}{Eric~P Xing}.}
  \bibinfo{year}{2018}\natexlab{}.
\newblock \showarticletitle{Neural Architecture Search with Bayesian
  Optimisation and Optimal Transport}. In \bibinfo{booktitle}{\emph{Proceedings
  of the International Conference on Neural Information Processing Systems
  (NeurIPS)}}.
\newblock


\bibitem[\protect\citeauthoryear{Karnin, Koren, and Somekh}{Karnin
  et~al\mbox{.}}{2013}]%
        {karnin2013almost}
\bibfield{author}{\bibinfo{person}{Zohar Karnin}, \bibinfo{person}{Tomer
  Koren}, {and} \bibinfo{person}{Oren Somekh}.}
  \bibinfo{year}{2013}\natexlab{}.
\newblock \showarticletitle{Almost optimal exploration in multi-armed bandits}.
  In \bibinfo{booktitle}{\emph{The International Conference on Machine Learning
  (ICML)}}. \bibinfo{pages}{1238--1246}.
\newblock


\bibitem[\protect\citeauthoryear{Kedziora, Musial, and Gabrys}{Kedziora
  et~al\mbox{.}}{2020}]%
        {kedziora2020autonoml}
\bibfield{author}{\bibinfo{person}{David~Jacob Kedziora},
  \bibinfo{person}{Katarzyna Musial}, {and} \bibinfo{person}{Bogdan Gabrys}.}
  \bibinfo{year}{2020}\natexlab{}.
\newblock \showarticletitle{{AutonoML}: Towards an Integrated Framework for
  Autonomous Machine Learning}.
\newblock \bibinfo{journal}{\emph{arXiv preprint arXiv:2012.12600}}
  (\bibinfo{year}{2020}).
\newblock


\bibitem[\protect\citeauthoryear{Kingma and Ba}{Kingma and Ba}{2015}]%
        {kingma2015adam}
\bibfield{author}{\bibinfo{person}{Diederik~P Kingma} {and}
  \bibinfo{person}{Jimmy Ba}.} \bibinfo{year}{2015}\natexlab{}.
\newblock \showarticletitle{Adam: A method for stochastic optimization}. In
  \bibinfo{booktitle}{\emph{International Conference on Learning
  Representations (ICLR)}}.
\newblock


\bibitem[\protect\citeauthoryear{Kirkpatrick, Pascanu, Rabinowitz, Veness,
  Desjardins, Rusu, Milan, Quan, Ramalho, Grabska-Barwinska, Hassabis, Clopath,
  Kumaran, and Hadsell}{Kirkpatrick et~al\mbox{.}}{2017}]%
        {kirkpatrick2017overcoming}
\bibfield{author}{\bibinfo{person}{James Kirkpatrick}, \bibinfo{person}{Razvan
  Pascanu}, \bibinfo{person}{Neil Rabinowitz}, \bibinfo{person}{Joel Veness},
  \bibinfo{person}{Guillaume Desjardins}, \bibinfo{person}{Andrei~A. Rusu},
  \bibinfo{person}{Kieran Milan}, \bibinfo{person}{John Quan},
  \bibinfo{person}{Tiago Ramalho}, \bibinfo{person}{Agnieszka
  Grabska-Barwinska}, \bibinfo{person}{Demis Hassabis},
  \bibinfo{person}{Claudia Clopath}, \bibinfo{person}{Dharshan Kumaran}, {and}
  \bibinfo{person}{Raia Hadsell}.} \bibinfo{year}{2017}\natexlab{}.
\newblock \showarticletitle{Overcoming catastrophic forgetting in neural
  networks}.
\newblock \bibinfo{journal}{\emph{Proceedings of the National Academy of
  Sciences}} \bibinfo{volume}{114}, \bibinfo{number}{13}
  (\bibinfo{year}{2017}), \bibinfo{pages}{3521--3526}.
\newblock
\urldef\tempurl%
\url{https://doi.org/10.1073/pnas.1611835114}
\showDOI{\tempurl}


\bibitem[\protect\citeauthoryear{Kirsch, van Steenkiste, and
  Schmidhuber}{Kirsch et~al\mbox{.}}{2020}]%
        {kirsch2020improving}
\bibfield{author}{\bibinfo{person}{Louis Kirsch}, \bibinfo{person}{Sjoerd van
  Steenkiste}, {and} \bibinfo{person}{J{\"u}rgen Schmidhuber}.}
  \bibinfo{year}{2020}\natexlab{}.
\newblock \showarticletitle{Improving generalization in meta reinforcement
  learning using learned objectives}. In
  \bibinfo{booktitle}{\emph{International Conference on Learning
  Representations (ICLR)}}.
\newblock


\bibitem[\protect\citeauthoryear{Klein and Hutter}{Klein and Hutter}{2019}]%
        {klein2019tabular}
\bibfield{author}{\bibinfo{person}{Aaron Klein} {and} \bibinfo{person}{Frank
  Hutter}.} \bibinfo{year}{2019}\natexlab{}.
\newblock \showarticletitle{Tabular benchmarks for joint architecture and
  hyperparameter optimization}.
\newblock \bibinfo{journal}{\emph{arXiv preprint arXiv:1905.04970}}
  (\bibinfo{year}{2019}).
\newblock


\bibitem[\protect\citeauthoryear{Konyushkova, Raphael, and Fua}{Konyushkova
  et~al\mbox{.}}{2017}]%
        {konyushkova2017learning}
\bibfield{author}{\bibinfo{person}{Ksenia Konyushkova},
  \bibinfo{person}{Sznitman Raphael}, {and} \bibinfo{person}{Pascal Fua}.}
  \bibinfo{year}{2017}\natexlab{}.
\newblock \showarticletitle{Learning active learning from data}. In
  \bibinfo{booktitle}{\emph{Proceedings of the International Conference on
  Neural Information Processing Systems (NeurIPS)}}.
  \bibinfo{pages}{4228--4238}.
\newblock


\bibitem[\protect\citeauthoryear{Krawczyk, Minku, Gama, Stefanowski, and
  Wo{\'z}niak}{Krawczyk et~al\mbox{.}}{2017}]%
        {krawczyk2017ensemble}
\bibfield{author}{\bibinfo{person}{Bartosz Krawczyk},
  \bibinfo{person}{Leandro~L Minku}, \bibinfo{person}{Joao Gama},
  \bibinfo{person}{Jerzy Stefanowski}, {and} \bibinfo{person}{Micha{\l}
  Wo{\'z}niak}.} \bibinfo{year}{2017}\natexlab{}.
\newblock \showarticletitle{Ensemble learning for data stream analysis: A
  survey}.
\newblock \bibinfo{journal}{\emph{Information Fusion}}  \bibinfo{volume}{37}
  (\bibinfo{year}{2017}), \bibinfo{pages}{132--156}.
\newblock


\bibitem[\protect\citeauthoryear{Krizhevsky, Hinton, et~al\mbox{.}}{Krizhevsky
  et~al\mbox{.}}{2009}]%
        {krizhevsky2009learning}
\bibfield{author}{\bibinfo{person}{Alex Krizhevsky}, \bibinfo{person}{Geoffrey
  Hinton}, {et~al\mbox{.}}} \bibinfo{year}{2009}\natexlab{}.
\newblock \bibinfo{title}{Learning multiple layers of features from tiny
  images}.
\newblock
\newblock


\bibitem[\protect\citeauthoryear{Krizhevsky, Sutskever, and Hinton}{Krizhevsky
  et~al\mbox{.}}{2012}]%
        {krizhevsky2012imagenet}
\bibfield{author}{\bibinfo{person}{Alex Krizhevsky}, \bibinfo{person}{Ilya
  Sutskever}, {and} \bibinfo{person}{Geoffrey~E Hinton}.}
  \bibinfo{year}{2012}\natexlab{}.
\newblock \showarticletitle{Imagenet classification with deep convolutional
  neural networks}. In \bibinfo{booktitle}{\emph{Proceedings of the
  International Conference on Neural Information Processing Systems
  (NeurIPS)}}. \bibinfo{pages}{1097--1105}.
\newblock


\bibitem[\protect\citeauthoryear{Larsen, Hansen, Svarer, and Ohlsson}{Larsen
  et~al\mbox{.}}{1996}]%
        {larsen1996design}
\bibfield{author}{\bibinfo{person}{Jan Larsen}, \bibinfo{person}{Lars~Kai
  Hansen}, \bibinfo{person}{Claus Svarer}, {and} \bibinfo{person}{M Ohlsson}.}
  \bibinfo{year}{1996}\natexlab{}.
\newblock \showarticletitle{Design and regularization of neural networks: the
  optimal use of a validation set}. In \bibinfo{booktitle}{\emph{IEEE Workshop
  on Neural Networks for Signal Processing (NNSP)}}. \bibinfo{pages}{62--71}.
\newblock


\bibitem[\protect\citeauthoryear{Larsen, Svarer, Andersen, and Hansen}{Larsen
  et~al\mbox{.}}{2002}]%
        {larsen2002adaptive}
\bibfield{author}{\bibinfo{person}{Jan Larsen}, \bibinfo{person}{Claus Svarer},
  \bibinfo{person}{Lars~Nonboe Andersen}, {and} \bibinfo{person}{Lars~Kai
  Hansen}.} \bibinfo{year}{2002}\natexlab{}.
\newblock \showarticletitle{Adaptive Regularization in Neural Network
  Modeling}. In \bibinfo{booktitle}{\emph{Neural Networks: Tricks of the
  Trade}}.
\newblock
\urldef\tempurl%
\url{https://doi.org/10.1007/3-540-49430-8_6}
\showDOI{\tempurl}


\bibitem[\protect\citeauthoryear{Lazzaro, Ryckebusch, Mahowald, and
  Mead}{Lazzaro et~al\mbox{.}}{1988}]%
        {lazzaro1988winner}
\bibfield{author}{\bibinfo{person}{John Lazzaro}, \bibinfo{person}{Sylvie
  Ryckebusch}, \bibinfo{person}{Misha~Anne Mahowald}, {and}
  \bibinfo{person}{Caver~A Mead}.} \bibinfo{year}{1988}\natexlab{}.
\newblock \showarticletitle{Winner-take-all networks of O (n) complexity}. In
  \bibinfo{booktitle}{\emph{Proceedings of the International Conference on
  Neural Information Processing Systems (NeurIPS)}}, Vol.~\bibinfo{volume}{1}.
\newblock


\bibitem[\protect\citeauthoryear{LeCun}{LeCun}{1985}]%
        {lecun1985learning}
\bibfield{author}{\bibinfo{person}{Yann LeCun}.}
  \bibinfo{year}{1985}\natexlab{}.
\newblock \showarticletitle{A learning scheme for asymmetric threshold
  networks}.
\newblock \bibinfo{journal}{\emph{Proceedings of Cognitiva}}
  \bibinfo{volume}{85}, \bibinfo{number}{537} (\bibinfo{year}{1985}),
  \bibinfo{pages}{599--604}.
\newblock


\bibitem[\protect\citeauthoryear{LeCun, Bengio, and Hinton}{LeCun
  et~al\mbox{.}}{2015}]%
        {lecun2015deep}
\bibfield{author}{\bibinfo{person}{Yann LeCun}, \bibinfo{person}{Yoshua
  Bengio}, {and} \bibinfo{person}{Geoffrey Hinton}.}
  \bibinfo{year}{2015}\natexlab{}.
\newblock \showarticletitle{Deep learning}.
\newblock \bibinfo{journal}{\emph{Nature}} \bibinfo{volume}{521},
  \bibinfo{number}{7553} (\bibinfo{year}{2015}), \bibinfo{pages}{436--444}.
\newblock


\bibitem[\protect\citeauthoryear{LeCun, Boser, Denker, Henderson, Howard,
  Hubbard, and Jackel}{LeCun et~al\mbox{.}}{1989}]%
        {lecun1989backpropagation}
\bibfield{author}{\bibinfo{person}{Yann LeCun}, \bibinfo{person}{Bernhard
  Boser}, \bibinfo{person}{John~S Denker}, \bibinfo{person}{Donnie Henderson},
  \bibinfo{person}{Richard~E Howard}, \bibinfo{person}{Wayne Hubbard}, {and}
  \bibinfo{person}{Lawrence~D Jackel}.} \bibinfo{year}{1989}\natexlab{}.
\newblock \showarticletitle{Backpropagation applied to handwritten zip code
  recognition}.
\newblock \bibinfo{journal}{\emph{Neural Computation}} \bibinfo{volume}{1},
  \bibinfo{number}{4} (\bibinfo{year}{1989}), \bibinfo{pages}{541--551}.
\newblock


\bibitem[\protect\citeauthoryear{Li, Spyra, Perel, Dalibard, Jaderberg, Gu,
  Budden, Harley, and Gupta}{Li et~al\mbox{.}}{2019}]%
        {li2019generalized}
\bibfield{author}{\bibinfo{person}{Ang Li}, \bibinfo{person}{Ola Spyra},
  \bibinfo{person}{Sagi Perel}, \bibinfo{person}{Valentin Dalibard},
  \bibinfo{person}{Max Jaderberg}, \bibinfo{person}{Chenjie Gu},
  \bibinfo{person}{David Budden}, \bibinfo{person}{Tim Harley}, {and}
  \bibinfo{person}{Pramod Gupta}.} \bibinfo{year}{2019}\natexlab{}.
\newblock \showarticletitle{A generalized framework for population based
  training}. In \bibinfo{booktitle}{\emph{Proceedings of the ACM SIGKDD
  International Conference on Knowledge Discovery and Data Mining (KDD)}}.
  \bibinfo{pages}{1791--1799}.
\newblock


\bibitem[\protect\citeauthoryear{Li, Jiang, Bai, Zhang, Zheng, Dong, Liu, Yang,
  and Li}{Li et~al\mbox{.}}{2021a}]%
        {li2021full}
\bibfield{author}{\bibinfo{person}{Bo Li}, \bibinfo{person}{Xinyang Jiang},
  \bibinfo{person}{Donglin Bai}, \bibinfo{person}{Yuge Zhang},
  \bibinfo{person}{Ningxin Zheng}, \bibinfo{person}{Xuanyi Dong},
  \bibinfo{person}{Lu Liu}, \bibinfo{person}{Yuqing Yang}, {and}
  \bibinfo{person}{Dongsheng Li}.} \bibinfo{year}{2021}\natexlab{a}.
\newblock \showarticletitle{Full-Cycle Energy Consumption Benchmark for
  Low-Carbon Computer Vision}.
\newblock \bibinfo{journal}{\emph{arXiv preprint arXiv:2108.13465}}
  (\bibinfo{year}{2021}).
\newblock


\bibitem[\protect\citeauthoryear{Li, Yu, Fu, Zhang, Zhao, You, Yu, Wang, and
  Lin}{Li et~al\mbox{.}}{2021b}]%
        {li2021hw}
\bibfield{author}{\bibinfo{person}{Chaojian Li}, \bibinfo{person}{Zhongzhi Yu},
  \bibinfo{person}{Yonggan Fu}, \bibinfo{person}{Yongan Zhang},
  \bibinfo{person}{Yang Zhao}, \bibinfo{person}{Haoran You},
  \bibinfo{person}{Qixuan Yu}, \bibinfo{person}{Yue Wang}, {and}
  \bibinfo{person}{Yingyan Lin}.} \bibinfo{year}{2021}\natexlab{b}.
\newblock \showarticletitle{{HW-NAS-Bench}: Hardware-Aware Neural Architecture
  Search Benchmark}. In \bibinfo{booktitle}{\emph{International Conference on
  Learning Representations (ICLR)}}.
\newblock


\bibitem[\protect\citeauthoryear{Li and Malik}{Li and Malik}{2017}]%
        {li2017learning}
\bibfield{author}{\bibinfo{person}{Ke Li} {and} \bibinfo{person}{Jitendra
  Malik}.} \bibinfo{year}{2017}\natexlab{}.
\newblock \showarticletitle{Learning to optimize}. In
  \bibinfo{booktitle}{\emph{International Conference on Learning
  Representations (ICLR)}}.
\newblock


\bibitem[\protect\citeauthoryear{Li, Jamieson, DeSalvo, Rostamizadeh, and
  Talwalkar}{Li et~al\mbox{.}}{2018}]%
        {li2018hyperband}
\bibfield{author}{\bibinfo{person}{Lisha Li}, \bibinfo{person}{Kevin Jamieson},
  \bibinfo{person}{Giulia DeSalvo}, \bibinfo{person}{Afshin Rostamizadeh},
  {and} \bibinfo{person}{Ameet Talwalkar}.} \bibinfo{year}{2018}\natexlab{}.
\newblock \showarticletitle{Hyperband: A novel bandit-based approach to
  hyperparameter optimization}.
\newblock \bibinfo{journal}{\emph{Journal of Machine Learning Research (JMLR)}}
  \bibinfo{volume}{18}, \bibinfo{number}{1} (\bibinfo{year}{2018}),
  \bibinfo{pages}{6765--6816}.
\newblock


\bibitem[\protect\citeauthoryear{Li and Talwalkar}{Li and Talwalkar}{2020}]%
        {li2020random}
\bibfield{author}{\bibinfo{person}{Liam Li} {and} \bibinfo{person}{Ameet
  Talwalkar}.} \bibinfo{year}{2020}\natexlab{}.
\newblock \showarticletitle{Random search and reproducibility for neural
  architecture search}. In \bibinfo{booktitle}{\emph{Uncertainty in Artificial
  Intelligence}}. \bibinfo{pages}{367--377}.
\newblock


\bibitem[\protect\citeauthoryear{Li, Hu, Wang, Hospedales, Robertson, and
  Yang}{Li et~al\mbox{.}}{2020}]%
        {li2020differentiable}
\bibfield{author}{\bibinfo{person}{Yonggang Li}, \bibinfo{person}{Guosheng Hu},
  \bibinfo{person}{Yongtao Wang}, \bibinfo{person}{Timothy Hospedales},
  \bibinfo{person}{Neil~M Robertson}, {and} \bibinfo{person}{Yongxin Yang}.}
  \bibinfo{year}{2020}\natexlab{}.
\newblock \showarticletitle{{DADA}: Differentiable Automatic Data
  Augmentation}. In \bibinfo{booktitle}{\emph{Proceedings of the European
  Conference on Computer Vision (ECCV)}}. Springer, \bibinfo{pages}{580--595}.
\newblock


\bibitem[\protect\citeauthoryear{Li and Hoiem}{Li and Hoiem}{2017}]%
        {li2017learning_without_forget}
\bibfield{author}{\bibinfo{person}{Zhizhong Li} {and} \bibinfo{person}{Derek
  Hoiem}.} \bibinfo{year}{2017}\natexlab{}.
\newblock \showarticletitle{Learning without forgetting}.
\newblock \bibinfo{journal}{\emph{IEEE Transactions on Pattern Analysis and
  Machine Intelligence (TPAMI)}} \bibinfo{volume}{40}, \bibinfo{number}{12}
  (\bibinfo{year}{2017}), \bibinfo{pages}{2935--2947}.
\newblock


\bibitem[\protect\citeauthoryear{Lim, Kim, Kim, Kim, and Kim}{Lim
  et~al\mbox{.}}{2019}]%
        {lim2019fast}
\bibfield{author}{\bibinfo{person}{Sungbin Lim}, \bibinfo{person}{Ildoo Kim},
  \bibinfo{person}{Taesup Kim}, \bibinfo{person}{Chiheon Kim}, {and}
  \bibinfo{person}{Sungwoong Kim}.} \bibinfo{year}{2019}\natexlab{}.
\newblock \showarticletitle{Fast autoaugment}. In
  \bibinfo{booktitle}{\emph{Proceedings of the International Conference on
  Neural Information Processing Systems (NeurIPS)}}.
\newblock


\bibitem[\protect\citeauthoryear{Lindauer and Hutter}{Lindauer and
  Hutter}{2020}]%
        {lindauer2020best}
\bibfield{author}{\bibinfo{person}{Marius Lindauer} {and}
  \bibinfo{person}{Frank Hutter}.} \bibinfo{year}{2020}\natexlab{}.
\newblock \showarticletitle{Best practices for scientific research on neural
  architecture search}.
\newblock \bibinfo{journal}{\emph{Journal of Machine Learning Research (JMLR)}}
  \bibinfo{volume}{21}, \bibinfo{number}{243} (\bibinfo{year}{2020}),
  \bibinfo{pages}{1--18}.
\newblock


\bibitem[\protect\citeauthoryear{Linnainmaa}{Linnainmaa}{1970}]%
        {linnainmaa1970representation}
\bibfield{author}{\bibinfo{person}{Seppo Linnainmaa}.}
  \bibinfo{year}{1970}\natexlab{}.
\newblock \showarticletitle{The representation of the cumulative rounding error
  of an algorithm as a Taylor expansion of the local rounding errors}.
\newblock \bibinfo{journal}{\emph{Master's Thesis (in Finnish), Univ.
  Helsinki}} (\bibinfo{year}{1970}), \bibinfo{pages}{6--7}.
\newblock


\bibitem[\protect\citeauthoryear{Little}{Little}{1974}]%
        {little1974existence}
\bibfield{author}{\bibinfo{person}{William~A Little}.}
  \bibinfo{year}{1974}\natexlab{}.
\newblock \showarticletitle{The existence of persistent states in the brain}.
\newblock \bibinfo{journal}{\emph{Mathematical Biosciences}}
  \bibinfo{volume}{19}, \bibinfo{number}{1-2} (\bibinfo{year}{1974}),
  \bibinfo{pages}{101--120}.
\newblock


\bibitem[\protect\citeauthoryear{Liu, Wang, Foroosh, Tappen, and Pensky}{Liu
  et~al\mbox{.}}{2015}]%
        {liu2015sparse}
\bibfield{author}{\bibinfo{person}{Baoyuan Liu}, \bibinfo{person}{Min Wang},
  \bibinfo{person}{Hassan Foroosh}, \bibinfo{person}{Marshall Tappen}, {and}
  \bibinfo{person}{Marianna Pensky}.} \bibinfo{year}{2015}\natexlab{}.
\newblock \showarticletitle{Sparse convolutional neural networks}. In
  \bibinfo{booktitle}{\emph{Proceedings of the IEEE Conference Computer Vision
  Pattern Recognition (CVPR)}}. \bibinfo{pages}{806--814}.
\newblock


\bibitem[\protect\citeauthoryear{Liu, Chen, Schroff, Adam, Hua, Yuille, and
  Fei-Fei}{Liu et~al\mbox{.}}{2019a}]%
        {liu2019auto}
\bibfield{author}{\bibinfo{person}{Chenxi Liu}, \bibinfo{person}{Liang-Chieh
  Chen}, \bibinfo{person}{Florian Schroff}, \bibinfo{person}{Hartwig Adam},
  \bibinfo{person}{Wei Hua}, \bibinfo{person}{Alan~L Yuille}, {and}
  \bibinfo{person}{Li Fei-Fei}.} \bibinfo{year}{2019}\natexlab{a}.
\newblock \showarticletitle{{Auto-DeepLab}: Hierarchical neural architecture
  search for semantic image segmentation}. In
  \bibinfo{booktitle}{\emph{Proceedings of the IEEE Conference Computer Vision
  Pattern Recognition (CVPR)}}. \bibinfo{pages}{82--92}.
\newblock


\bibitem[\protect\citeauthoryear{Liu, Zoph, Neumann, Shlens, Hua, Li, Fei-Fei,
  Yuille, Huang, and Murphy}{Liu et~al\mbox{.}}{2018}]%
        {Liu_2018_ECCV}
\bibfield{author}{\bibinfo{person}{Chenxi Liu}, \bibinfo{person}{Barret Zoph},
  \bibinfo{person}{Maxim Neumann}, \bibinfo{person}{Jonathon Shlens},
  \bibinfo{person}{Wei Hua}, \bibinfo{person}{Li-Jia Li}, \bibinfo{person}{Li
  Fei-Fei}, \bibinfo{person}{Alan Yuille}, \bibinfo{person}{Jonathan Huang},
  {and} \bibinfo{person}{Kevin Murphy}.} \bibinfo{year}{2018}\natexlab{}.
\newblock \showarticletitle{Progressive Neural Architecture Search}. In
  \bibinfo{booktitle}{\emph{Proceedings of the European Conference on Computer
  Vision (ECCV)}}. \bibinfo{pages}{19--34}.
\newblock


\bibitem[\protect\citeauthoryear{Liu, Brock, Simonyan, and Le}{Liu
  et~al\mbox{.}}{2020}]%
        {liu2020evolving}
\bibfield{author}{\bibinfo{person}{Hanxiao Liu}, \bibinfo{person}{Andrew
  Brock}, \bibinfo{person}{Karen Simonyan}, {and} \bibinfo{person}{Quoc~V Le}.}
  \bibinfo{year}{2020}\natexlab{}.
\newblock \showarticletitle{Evolving normalization-activation layers}. In
  \bibinfo{booktitle}{\emph{Proceedings of the International Conference on
  Neural Information Processing Systems (NeurIPS)}}.
\newblock


\bibitem[\protect\citeauthoryear{Liu, Simonyan, and Yang}{Liu
  et~al\mbox{.}}{2019b}]%
        {liu2019darts}
\bibfield{author}{\bibinfo{person}{Hanxiao Liu}, \bibinfo{person}{Karen
  Simonyan}, {and} \bibinfo{person}{Yiming Yang}.}
  \bibinfo{year}{2019}\natexlab{b}.
\newblock \showarticletitle{{DARTS}: Differentiable Architecture Search}. In
  \bibinfo{booktitle}{\emph{International Conference on Learning
  Representations (ICLR)}}.
\newblock


\bibitem[\protect\citeauthoryear{Liu, Zhou, Long, Jiang, Dong, and Zhang}{Liu
  et~al\mbox{.}}{2021}]%
        {liu2021isometric}
\bibfield{author}{\bibinfo{person}{Lu Liu}, \bibinfo{person}{Tianyi Zhou},
  \bibinfo{person}{Guodong Long}, \bibinfo{person}{Jing Jiang},
  \bibinfo{person}{Xuanyi Dong}, {and} \bibinfo{person}{Chengqi Zhang}.}
  \bibinfo{year}{2021}\natexlab{}.
\newblock \showarticletitle{Isometric propagation network for generalized
  zero-shot learning}. In \bibinfo{booktitle}{\emph{International Conference on
  Learning Representations (ICLR)}}.
\newblock


\bibitem[\protect\citeauthoryear{Liu, Zhou, Long, Jiang, and Zhang}{Liu
  et~al\mbox{.}}{2019c}]%
        {liu2019learning}
\bibfield{author}{\bibinfo{person}{Lu Liu}, \bibinfo{person}{Tianyi Zhou},
  \bibinfo{person}{Guodong Long}, \bibinfo{person}{Jing Jiang}, {and}
  \bibinfo{person}{Chengqi Zhang}.} \bibinfo{year}{2019}\natexlab{c}.
\newblock \showarticletitle{Learning to propagate for graph meta-learning}. In
  \bibinfo{booktitle}{\emph{Proceedings of the International Conference on
  Neural Information Processing Systems (NeurIPS)}}.
\newblock


\bibitem[\protect\citeauthoryear{Lorraine, Vicol, and Duvenaud}{Lorraine
  et~al\mbox{.}}{2020}]%
        {lorraine2020optimizing}
\bibfield{author}{\bibinfo{person}{Jonathan Lorraine}, \bibinfo{person}{Paul
  Vicol}, {and} \bibinfo{person}{David Duvenaud}.}
  \bibinfo{year}{2020}\natexlab{}.
\newblock \showarticletitle{Optimizing millions of hyperparameters by implicit
  differentiation}. In \bibinfo{booktitle}{\emph{The International Conference
  on Artificial Intelligence and Statistics (AISTATS)}}.
  \bibinfo{pages}{1540--1552}.
\newblock


\bibitem[\protect\citeauthoryear{Loshchilov and Hutter}{Loshchilov and
  Hutter}{2016}]%
        {loshchilov2016cma}
\bibfield{author}{\bibinfo{person}{Ilya Loshchilov} {and}
  \bibinfo{person}{Frank Hutter}.} \bibinfo{year}{2016}\natexlab{}.
\newblock \showarticletitle{{CMA-ES} for hyperparameter optimization of deep
  neural networks}.
\newblock In \bibinfo{booktitle}{\emph{International Conference on Learning
  Representations (ICLR) Workshop}}.
\newblock


\bibitem[\protect\citeauthoryear{Luketina, Berglund, Greff, and Raiko}{Luketina
  et~al\mbox{.}}{2016}]%
        {luketina2016scalable}
\bibfield{author}{\bibinfo{person}{Jelena Luketina}, \bibinfo{person}{Mathias
  Berglund}, \bibinfo{person}{Klaus Greff}, {and} \bibinfo{person}{Tapani
  Raiko}.} \bibinfo{year}{2016}\natexlab{}.
\newblock \showarticletitle{Scalable gradient-based tuning of continuous
  regularization hyperparameters}. In \bibinfo{booktitle}{\emph{The
  International Conference on Machine Learning (ICML)}}.
  \bibinfo{pages}{2952--2960}.
\newblock


\bibitem[\protect\citeauthoryear{Luo, Tian, Qin, Chen, and Liu}{Luo
  et~al\mbox{.}}{2018}]%
        {luo2018neural}
\bibfield{author}{\bibinfo{person}{Renqian Luo}, \bibinfo{person}{Fei Tian},
  \bibinfo{person}{Tao Qin}, \bibinfo{person}{Enhong Chen}, {and}
  \bibinfo{person}{Tie-Yan Liu}.} \bibinfo{year}{2018}\natexlab{}.
\newblock \showarticletitle{Neural architecture optimization}. In
  \bibinfo{booktitle}{\emph{Proceedings of the International Conference on
  Neural Information Processing Systems (NeurIPS)}}.
  \bibinfo{pages}{7827--7838}.
\newblock


\bibitem[\protect\citeauthoryear{Ma}{Ma}{2019}]%
        {ma2019nlpaug}
\bibfield{author}{\bibinfo{person}{Edward Ma}.}
  \bibinfo{year}{2019}\natexlab{}.
\newblock \bibinfo{title}{{NLP} Augmentation}.
\newblock \bibinfo{howpublished}{https://github.com/makcedward/nlpaug}.
\newblock


\bibitem[\protect\citeauthoryear{MacKay, Vicol, Lorraine, Duvenaud, and
  Grosse}{MacKay et~al\mbox{.}}{2019}]%
        {mackay2019self}
\bibfield{author}{\bibinfo{person}{Matthew MacKay}, \bibinfo{person}{Paul
  Vicol}, \bibinfo{person}{Jon Lorraine}, \bibinfo{person}{David Duvenaud},
  {and} \bibinfo{person}{Roger Grosse}.} \bibinfo{year}{2019}\natexlab{}.
\newblock \showarticletitle{Self-tuning networks: Bilevel optimization of
  hyperparameters using structured best-response functions}. In
  \bibinfo{booktitle}{\emph{International Conference on Learning
  Representations (ICLR)}}.
\newblock


\bibitem[\protect\citeauthoryear{Maclaurin, Duvenaud, and Adams}{Maclaurin
  et~al\mbox{.}}{2015}]%
        {maclaurin2015gradient}
\bibfield{author}{\bibinfo{person}{Dougal Maclaurin}, \bibinfo{person}{David
  Duvenaud}, {and} \bibinfo{person}{Ryan Adams}.}
  \bibinfo{year}{2015}\natexlab{}.
\newblock \showarticletitle{Gradient-based hyperparameter optimization through
  reversible learning}. In \bibinfo{booktitle}{\emph{The International
  Conference on Machine Learning (ICML)}}. \bibinfo{pages}{2113--2122}.
\newblock


\bibitem[\protect\citeauthoryear{Madrid, Escalante, Morales, Tu, Yu,
  Sun-Hosoya, Guyon, and Sebag}{Madrid et~al\mbox{.}}{2018}]%
        {madrid2019towards}
\bibfield{author}{\bibinfo{person}{Jorge~G Madrid}, \bibinfo{person}{Hugo~Jair
  Escalante}, \bibinfo{person}{Eduardo~F Morales}, \bibinfo{person}{Wei-Wei
  Tu}, \bibinfo{person}{Yang Yu}, \bibinfo{person}{Lisheng Sun-Hosoya},
  \bibinfo{person}{Isabelle Guyon}, {and} \bibinfo{person}{Mich{\`e}le Sebag}.}
  \bibinfo{year}{2018}\natexlab{}.
\newblock \showarticletitle{Towards AutoML in the presence of Drift: first
  results}. In \bibinfo{booktitle}{\emph{The International Conference on
  Machine Learning (ICML) Workshop}}.
\newblock


\bibitem[\protect\citeauthoryear{Marler and Arora}{Marler and Arora}{2004}]%
        {marler2004survey}
\bibfield{author}{\bibinfo{person}{R~Timothy Marler} {and}
  \bibinfo{person}{Jasbir~S Arora}.} \bibinfo{year}{2004}\natexlab{}.
\newblock \showarticletitle{Survey of multi-objective optimization methods for
  engineering}.
\newblock \bibinfo{journal}{\emph{Structural and multidisciplinary
  optimization}} \bibinfo{volume}{26}, \bibinfo{number}{6}
  (\bibinfo{year}{2004}), \bibinfo{pages}{369--395}.
\newblock


\bibitem[\protect\citeauthoryear{Martel, Ro{\ss}mann, Sultanow, Wei{\ss},
  Wissel, Pelzel, and Se{\ss}ler}{Martel et~al\mbox{.}}{2021}]%
        {martel2021software}
\bibfield{author}{\bibinfo{person}{Yannick Martel}, \bibinfo{person}{Arne
  Ro{\ss}mann}, \bibinfo{person}{Eldar Sultanow}, \bibinfo{person}{Oliver
  Wei{\ss}}, \bibinfo{person}{Matthias Wissel}, \bibinfo{person}{Frank Pelzel},
  {and} \bibinfo{person}{Matthias Se{\ss}ler}.}
  \bibinfo{year}{2021}\natexlab{}.
\newblock \showarticletitle{Software Architecture Best Practices for Enterprise
  Artificial Intelligence}.
\newblock \bibinfo{journal}{\emph{INFORMATIK}} (\bibinfo{year}{2021}).
\newblock


\bibitem[\protect\citeauthoryear{McCloskey and Cohen}{McCloskey and
  Cohen}{1989}]%
        {mccloskey1989catastrophic}
\bibfield{author}{\bibinfo{person}{Michael McCloskey} {and}
  \bibinfo{person}{Neal~J Cohen}.} \bibinfo{year}{1989}\natexlab{}.
\newblock \showarticletitle{Catastrophic interference in connectionist
  networks: The sequential learning problem}.
\newblock In \bibinfo{booktitle}{\emph{Psychology of learning and motivation}}.
  Vol.~\bibinfo{volume}{24}. \bibinfo{publisher}{Elsevier},
  \bibinfo{pages}{109--165}.
\newblock


\bibitem[\protect\citeauthoryear{McCulloch and Pitts}{McCulloch and
  Pitts}{1943}]%
        {mcculloch1943logical}
\bibfield{author}{\bibinfo{person}{Warren~S McCulloch} {and}
  \bibinfo{person}{Walter Pitts}.} \bibinfo{year}{1943}\natexlab{}.
\newblock \showarticletitle{A logical calculus of the ideas immanent in nervous
  activity}.
\newblock \bibinfo{journal}{\emph{The Bulletin of Mathematical Biophysics}}
  \bibinfo{volume}{5}, \bibinfo{number}{4} (\bibinfo{year}{1943}),
  \bibinfo{pages}{115--133}.
\newblock


\bibitem[\protect\citeauthoryear{Mellor, Turner, Storkey, and Crowley}{Mellor
  et~al\mbox{.}}{2021}]%
        {mellor2021neural}
\bibfield{author}{\bibinfo{person}{Joseph Mellor}, \bibinfo{person}{Jack
  Turner}, \bibinfo{person}{Amos Storkey}, {and} \bibinfo{person}{Elliot~J.
  Crowley}.} \bibinfo{year}{2021}\natexlab{}.
\newblock \showarticletitle{Neural Architecture Search without Training}. In
  \bibinfo{booktitle}{\emph{The International Conference on Machine Learning
  (ICML)}}.
\newblock


\bibitem[\protect\citeauthoryear{Michie, Spiegelhalter, Taylor,
  et~al\mbox{.}}{Michie et~al\mbox{.}}{1994}]%
        {michie1994machine}
\bibfield{author}{\bibinfo{person}{Donald Michie}, \bibinfo{person}{David~J
  Spiegelhalter}, \bibinfo{person}{CC Taylor}, {et~al\mbox{.}}}
  \bibinfo{year}{1994}\natexlab{}.
\newblock \showarticletitle{Machine Learning}.
\newblock \bibinfo{journal}{\emph{Neural and Statistical Classification}}
  \bibinfo{volume}{13}, \bibinfo{number}{1994} (\bibinfo{year}{1994}),
  \bibinfo{pages}{1--298}.
\newblock


\bibitem[\protect\citeauthoryear{Microsoft Corporation}{Microsoft
  Corporation}{2018}]%
        {NNI}
Microsoft Corporation \bibinfo{year}{2018}\natexlab{}.
\newblock \bibinfo{title}{Neural Network Intelligence}.
\newblock \bibinfo{howpublished}{https://github.com/microsoft/nni}.
\newblock


\bibitem[\protect\citeauthoryear{Milutinovic, Baydin, Zinkov, Harvey, Song,
  Wood, and Shen}{Milutinovic et~al\mbox{.}}{2017}]%
        {milutinovic2017end}
\bibfield{author}{\bibinfo{person}{Mitar Milutinovic},
  \bibinfo{person}{At{\i}l{\i}m~G{\"u}ne{\c{s}} Baydin},
  \bibinfo{person}{Robert Zinkov}, \bibinfo{person}{William Harvey},
  \bibinfo{person}{Dawn Song}, \bibinfo{person}{Frank Wood}, {and}
  \bibinfo{person}{Wade Shen}.} \bibinfo{year}{2017}\natexlab{}.
\newblock \showarticletitle{End-to-end training of differentiable pipelines
  across machine learning frameworks}. In \bibinfo{booktitle}{\emph{Proceedings
  of the International Conference on Neural Information Processing Systems
  (NeurIPS) Workshop}}.
\newblock


\bibitem[\protect\citeauthoryear{Mirhoseini, Goldie, Pham, Steiner, Le, and
  Dean}{Mirhoseini et~al\mbox{.}}{2018}]%
        {mirhoseini2018hierarchical}
\bibfield{author}{\bibinfo{person}{Azalia Mirhoseini}, \bibinfo{person}{Anna
  Goldie}, \bibinfo{person}{Hieu Pham}, \bibinfo{person}{Benoit Steiner},
  \bibinfo{person}{Quoc~V Le}, {and} \bibinfo{person}{Jeff Dean}.}
  \bibinfo{year}{2018}\natexlab{}.
\newblock \showarticletitle{A hierarchical model for device placement}. In
  \bibinfo{booktitle}{\emph{International Conference on Learning
  Representations (ICLR)}}.
\newblock


\bibitem[\protect\citeauthoryear{Mirhoseini, Goldie, Yazgan, Jiang, Songhori,
  Wang, Lee, Johnson, Pathak, Nazi, et~al\mbox{.}}{Mirhoseini
  et~al\mbox{.}}{2021}]%
        {mirhoseini2021graph}
\bibfield{author}{\bibinfo{person}{Azalia Mirhoseini}, \bibinfo{person}{Anna
  Goldie}, \bibinfo{person}{Mustafa Yazgan}, \bibinfo{person}{Joe~Wenjie
  Jiang}, \bibinfo{person}{Ebrahim Songhori}, \bibinfo{person}{Shen Wang},
  \bibinfo{person}{Young-Joon Lee}, \bibinfo{person}{Eric Johnson},
  \bibinfo{person}{Omkar Pathak}, \bibinfo{person}{Azade Nazi},
  {et~al\mbox{.}}} \bibinfo{year}{2021}\natexlab{}.
\newblock \showarticletitle{A graph placement methodology for fast chip
  design}.
\newblock \bibinfo{journal}{\emph{Nature}} \bibinfo{volume}{594},
  \bibinfo{number}{7862} (\bibinfo{year}{2021}), \bibinfo{pages}{207--212}.
\newblock


\bibitem[\protect\citeauthoryear{Mirhoseini, Pham, Le, Steiner, Larsen, Zhou,
  Kumar, Norouzi, Bengio, and Dean}{Mirhoseini et~al\mbox{.}}{2017}]%
        {mirhoseini2017device}
\bibfield{author}{\bibinfo{person}{Azalia Mirhoseini}, \bibinfo{person}{Hieu
  Pham}, \bibinfo{person}{Quoc~V Le}, \bibinfo{person}{Benoit Steiner},
  \bibinfo{person}{Rasmus Larsen}, \bibinfo{person}{Yuefeng Zhou},
  \bibinfo{person}{Naveen Kumar}, \bibinfo{person}{Mohammad Norouzi},
  \bibinfo{person}{Samy Bengio}, {and} \bibinfo{person}{Jeff Dean}.}
  \bibinfo{year}{2017}\natexlab{}.
\newblock \showarticletitle{Device placement optimization with reinforcement
  learning}. In \bibinfo{booktitle}{\emph{The International Conference on
  Machine Learning (ICML)}}. \bibinfo{pages}{2430--2439}.
\newblock


\bibitem[\protect\citeauthoryear{Mo{\v{c}}kus}{Mo{\v{c}}kus}{1975}]%
        {movckus1975bayesian}
\bibfield{author}{\bibinfo{person}{Jonas Mo{\v{c}}kus}.}
  \bibinfo{year}{1975}\natexlab{}.
\newblock \showarticletitle{On Bayesian methods for seeking the extremum}. In
  \bibinfo{booktitle}{\emph{Optimization Techniques IFIP Technical
  Conference}}. Springer, \bibinfo{pages}{400--404}.
\newblock


\bibitem[\protect\citeauthoryear{Nardi, Koeplinger, and Olukotun}{Nardi
  et~al\mbox{.}}{2019}]%
        {nardi2019practical}
\bibfield{author}{\bibinfo{person}{Luigi Nardi}, \bibinfo{person}{David
  Koeplinger}, {and} \bibinfo{person}{Kunle Olukotun}.}
  \bibinfo{year}{2019}\natexlab{}.
\newblock \showarticletitle{Practical design space exploration}. In
  \bibinfo{booktitle}{\emph{IEEE International Symposium on Modeling, Analysis,
  and Simulation of Computer and Telecommunication Systems (MASCOTS)}}.
  \bibinfo{pages}{347--358}.
\newblock


\bibitem[\protect\citeauthoryear{Nguyen, Kedziora, Musial, and Gabrys}{Nguyen
  et~al\mbox{.}}{2021}]%
        {nguyen2021exploring}
\bibfield{author}{\bibinfo{person}{Tien-Dung Nguyen},
  \bibinfo{person}{David~Jacob Kedziora}, \bibinfo{person}{Katarzyna Musial},
  {and} \bibinfo{person}{Bogdan Gabrys}.} \bibinfo{year}{2021}\natexlab{}.
\newblock \showarticletitle{Exploring Opportunistic Meta-knowledge to Reduce
  Search Spaces for Automated Machine Learning}. In
  \bibinfo{booktitle}{\emph{International Joint Conference on Neural Network
  (IJCNN)}}.
\newblock


\bibitem[\protect\citeauthoryear{Niu and Bansal}{Niu and Bansal}{2019}]%
        {niu2019automatically}
\bibfield{author}{\bibinfo{person}{Tong Niu} {and} \bibinfo{person}{Mohit
  Bansal}.} \bibinfo{year}{2019}\natexlab{}.
\newblock \showarticletitle{Automatically Learning Data Augmentation Policies
  for Dialogue Tasks}. In \bibinfo{booktitle}{\emph{Proceedings of the
  Conference on Empirical Methods in Natural Language Processing (EMNLP)}}.
  \bibinfo{pages}{1317--1323}.
\newblock


\bibitem[\protect\citeauthoryear{Panayotov, Chen, Povey, and
  Khudanpur}{Panayotov et~al\mbox{.}}{2015}]%
        {panayotov2015librispeech}
\bibfield{author}{\bibinfo{person}{Vassil Panayotov}, \bibinfo{person}{Guoguo
  Chen}, \bibinfo{person}{Daniel Povey}, {and} \bibinfo{person}{Sanjeev
  Khudanpur}.} \bibinfo{year}{2015}\natexlab{}.
\newblock \showarticletitle{{Librispeech}: an asr corpus based on public domain
  audio books}. In \bibinfo{booktitle}{\emph{IEEE International conference on
  acoustics, speech and signal processing (ICASSP)}}.
  \bibinfo{pages}{5206--5210}.
\newblock


\bibitem[\protect\citeauthoryear{Parisi, Kemker, Part, Kanan, and
  Wermter}{Parisi et~al\mbox{.}}{2019}]%
        {pake19}
\bibfield{author}{\bibinfo{person}{German~I. Parisi}, \bibinfo{person}{Ronald
  Kemker}, \bibinfo{person}{Jose~L. Part}, \bibinfo{person}{Christopher Kanan},
  {and} \bibinfo{person}{Stefan Wermter}.} \bibinfo{year}{2019}\natexlab{}.
\newblock \showarticletitle{Continual lifelong learning with neural networks: A
  review}.
\newblock \bibinfo{journal}{\emph{Neural Networks}}  \bibinfo{volume}{113}
  (\bibinfo{year}{2019}), \bibinfo{pages}{54--71}.
\newblock
\showISSN{0893-6080}
\urldef\tempurl%
\url{https://doi.org/10.1016/j.neunet.2019.01.012}
\showDOI{\tempurl}


\bibitem[\protect\citeauthoryear{Parkhi, Vedaldi, and Zisserman}{Parkhi
  et~al\mbox{.}}{2015}]%
        {parkhi2015deep}
\bibfield{author}{\bibinfo{person}{Omkar~M Parkhi}, \bibinfo{person}{Andrea
  Vedaldi}, {and} \bibinfo{person}{Andrew Zisserman}.}
  \bibinfo{year}{2015}\natexlab{}.
\newblock \showarticletitle{Deep face recognition}. In
  \bibinfo{booktitle}{\emph{Proceedings of the British Machine Vision
  Conference (BMVC)}}. \bibinfo{pages}{41.1--41.12}.
\newblock


\bibitem[\protect\citeauthoryear{Parsa, Ankit, Ziabari, and Roy}{Parsa
  et~al\mbox{.}}{2019}]%
        {parsa2019pabo}
\bibfield{author}{\bibinfo{person}{Maryam Parsa}, \bibinfo{person}{Aayush
  Ankit}, \bibinfo{person}{Amirkoushyar Ziabari}, {and}
  \bibinfo{person}{Kaushik Roy}.} \bibinfo{year}{2019}\natexlab{}.
\newblock \showarticletitle{{PABO}: Pseudo agent-based multi-objective bayesian
  hyperparameter optimization for efficient neural accelerator design}. In
  \bibinfo{booktitle}{\emph{The IEEE/ACM International Conference on
  Computer-Aided Design (ICCAD)}}. \bibinfo{pages}{1--8}.
\newblock


\bibitem[\protect\citeauthoryear{Patterson, Gonzalez, Le, Liang, Munguia,
  Rothchild, So, Texier, and Dean}{Patterson et~al\mbox{.}}{2021}]%
        {patterson2021carbon}
\bibfield{author}{\bibinfo{person}{David Patterson}, \bibinfo{person}{Joseph
  Gonzalez}, \bibinfo{person}{Quoc Le}, \bibinfo{person}{Chen Liang},
  \bibinfo{person}{Lluis-Miquel Munguia}, \bibinfo{person}{Daniel Rothchild},
  \bibinfo{person}{David So}, \bibinfo{person}{Maud Texier}, {and}
  \bibinfo{person}{Jeff Dean}.} \bibinfo{year}{2021}\natexlab{}.
\newblock \showarticletitle{Carbon emissions and large neural network
  training}.
\newblock \bibinfo{journal}{\emph{arXiv preprint arXiv:2104.10350}}
  (\bibinfo{year}{2021}).
\newblock


\bibitem[\protect\citeauthoryear{Pearson}{Pearson}{1901}]%
        {pearson1901on}
\bibfield{author}{\bibinfo{person}{Karl Pearson}.}
  \bibinfo{year}{1901}\natexlab{}.
\newblock \showarticletitle{On lines and planes of closest fit to systems of
  points in space}.
\newblock \bibinfo{journal}{\emph{The London, Edinburgh, and Dublin
  Philosophical Magazine and Journal of Science}} \bibinfo{volume}{2},
  \bibinfo{number}{11} (\bibinfo{year}{1901}), \bibinfo{pages}{559--572}.
\newblock


\bibitem[\protect\citeauthoryear{Pedregosa}{Pedregosa}{2016}]%
        {pedregosa2016hyperparameter}
\bibfield{author}{\bibinfo{person}{Fabian Pedregosa}.}
  \bibinfo{year}{2016}\natexlab{}.
\newblock \showarticletitle{Hyperparameter optimization with approximate
  gradient}. In \bibinfo{booktitle}{\emph{The International Conference on
  Machine Learning (ICML)}}. \bibinfo{pages}{737--746}.
\newblock


\bibitem[\protect\citeauthoryear{Peng, Dong, Real, Tan, Lu, Bender, Liu, Kraft,
  Liang, and Le}{Peng et~al\mbox{.}}{2020}]%
        {peng2020pyglove}
\bibfield{author}{\bibinfo{person}{Daiyi Peng}, \bibinfo{person}{Xuanyi Dong},
  \bibinfo{person}{Esteban Real}, \bibinfo{person}{Mingxing Tan},
  \bibinfo{person}{Yifeng Lu}, \bibinfo{person}{Gabriel Bender},
  \bibinfo{person}{Hanxiao Liu}, \bibinfo{person}{Adam Kraft},
  \bibinfo{person}{Chen Liang}, {and} \bibinfo{person}{Quoc Le}.}
  \bibinfo{year}{2020}\natexlab{}.
\newblock \showarticletitle{{PyGlove}: Symbolic Programming for Automated
  Machine Learning}. In \bibinfo{booktitle}{\emph{Proceedings of the
  International Conference on Neural Information Processing Systems
  (NeurIPS)}}.
\newblock


\bibitem[\protect\citeauthoryear{Peng, Bao, Chen, Wu, and Guo}{Peng
  et~al\mbox{.}}{2018}]%
        {peng2018optimus}
\bibfield{author}{\bibinfo{person}{Yanghua Peng}, \bibinfo{person}{Yixin Bao},
  \bibinfo{person}{Yangrui Chen}, \bibinfo{person}{Chuan Wu}, {and}
  \bibinfo{person}{Chuanxiong Guo}.} \bibinfo{year}{2018}\natexlab{}.
\newblock \showarticletitle{Optimus: an efficient dynamic resource scheduler
  for deep learning clusters}. In \bibinfo{booktitle}{\emph{Proceedings of the
  EuroSys Conference (EuroSys)}}. \bibinfo{pages}{1--14}.
\newblock


\bibitem[\protect\citeauthoryear{Pham, Guan, Zoph, Le, and Dean}{Pham
  et~al\mbox{.}}{2018}]%
        {pham2018efficient}
\bibfield{author}{\bibinfo{person}{Hieu Pham}, \bibinfo{person}{Melody Guan},
  \bibinfo{person}{Barret Zoph}, \bibinfo{person}{Quoc Le}, {and}
  \bibinfo{person}{Jeff Dean}.} \bibinfo{year}{2018}\natexlab{}.
\newblock \showarticletitle{Efficient Neural Architecture Search via Parameters
  Sharing}. In \bibinfo{booktitle}{\emph{The International Conference on
  Machine Learning (ICML)}}. \bibinfo{pages}{4095--4104}.
\newblock


\bibitem[\protect\citeauthoryear{Pineau, Vincent-Lamarre, Sinha, Larivi{\`e}re,
  Beygelzimer, d’Alch{\'e} Buc, Fox, and Larochelle}{Pineau
  et~al\mbox{.}}{2021}]%
        {pineau2020improving}
\bibfield{author}{\bibinfo{person}{Joelle Pineau}, \bibinfo{person}{Philippe
  Vincent-Lamarre}, \bibinfo{person}{Koustuv Sinha}, \bibinfo{person}{Vincent
  Larivi{\`e}re}, \bibinfo{person}{Alina Beygelzimer},
  \bibinfo{person}{Florence d’Alch{\'e} Buc}, \bibinfo{person}{Emily Fox},
  {and} \bibinfo{person}{Hugo Larochelle}.} \bibinfo{year}{2021}\natexlab{}.
\newblock \showarticletitle{Improving reproducibility in machine learning
  research: a report from the NeurIPS 2019 reproducibility program}.
\newblock \bibinfo{journal}{\emph{Journal of Machine Learning Research (JMLR)}}
   \bibinfo{volume}{22} (\bibinfo{year}{2021}).
\newblock


\bibitem[\protect\citeauthoryear{Pollack}{Pollack}{1989}]%
        {pollack1989implications}
\bibfield{author}{\bibinfo{person}{Jordan~B Pollack}.}
  \bibinfo{year}{1989}\natexlab{}.
\newblock \showarticletitle{Implications of recursive distributed
  representations}. In \bibinfo{booktitle}{\emph{Proceedings of the
  International Conference on Neural Information Processing Systems
  (NeurIPS)}}. \bibinfo{pages}{527--536}.
\newblock


\bibitem[\protect\citeauthoryear{Quinlan}{Quinlan}{1986}]%
        {quinlan1986induction}
\bibfield{author}{\bibinfo{person}{J.~Ross Quinlan}.}
  \bibinfo{year}{1986}\natexlab{}.
\newblock \showarticletitle{Induction of decision trees}.
\newblock \bibinfo{journal}{\emph{Machine learning}} \bibinfo{volume}{1},
  \bibinfo{number}{1} (\bibinfo{year}{1986}), \bibinfo{pages}{81--106}.
\newblock


\bibitem[\protect\citeauthoryear{Radosavovic, Kosaraju, Girshick, He, and
  Doll{\'a}r}{Radosavovic et~al\mbox{.}}{2020}]%
        {radosavovic2020designing}
\bibfield{author}{\bibinfo{person}{Ilija Radosavovic},
  \bibinfo{person}{Raj~Prateek Kosaraju}, \bibinfo{person}{Ross Girshick},
  \bibinfo{person}{Kaiming He}, {and} \bibinfo{person}{Piotr Doll{\'a}r}.}
  \bibinfo{year}{2020}\natexlab{}.
\newblock \showarticletitle{Designing network design spaces}. In
  \bibinfo{booktitle}{\emph{Proceedings of the IEEE Conference Computer Vision
  Pattern Recognition (CVPR)}}. \bibinfo{pages}{10428--10436}.
\newblock


\bibitem[\protect\citeauthoryear{Rajpurkar, Zhang, Lopyrev, and
  Liang}{Rajpurkar et~al\mbox{.}}{2016}]%
        {rajpurkar2016squad}
\bibfield{author}{\bibinfo{person}{Pranav Rajpurkar}, \bibinfo{person}{Jian
  Zhang}, \bibinfo{person}{Konstantin Lopyrev}, {and} \bibinfo{person}{Percy
  Liang}.} \bibinfo{year}{2016}\natexlab{}.
\newblock \showarticletitle{Squad: 100,000+ questions for machine comprehension
  of text}.
\newblock \bibinfo{journal}{\emph{arXiv preprint arXiv:1606.05250}}
  (\bibinfo{year}{2016}).
\newblock


\bibitem[\protect\citeauthoryear{Ramachandran, Zoph, and Le}{Ramachandran
  et~al\mbox{.}}{2017}]%
        {ramachandran2017searching}
\bibfield{author}{\bibinfo{person}{Prajit Ramachandran},
  \bibinfo{person}{Barret Zoph}, {and} \bibinfo{person}{Quoc~V Le}.}
  \bibinfo{year}{2017}\natexlab{}.
\newblock \showarticletitle{Searching for activation functions}.
\newblock \bibinfo{journal}{\emph{arXiv preprint arXiv:1710.05941}}
  (\bibinfo{year}{2017}).
\newblock


\bibitem[\protect\citeauthoryear{Ravi and Larochelle}{Ravi and
  Larochelle}{2017}]%
        {ravi2017optimization}
\bibfield{author}{\bibinfo{person}{Sachin Ravi} {and} \bibinfo{person}{Hugo
  Larochelle}.} \bibinfo{year}{2017}\natexlab{}.
\newblock \showarticletitle{Optimization as a model for few-shot learning}. In
  \bibinfo{booktitle}{\emph{International Conference on Learning
  Representations (ICLR)}}.
\newblock


\bibitem[\protect\citeauthoryear{Reagen, Hern{\'a}ndez-Lobato, Adolf, Gelbart,
  Whatmough, Wei, and Brooks}{Reagen et~al\mbox{.}}{2017}]%
        {reagen2017case}
\bibfield{author}{\bibinfo{person}{Brandon Reagen},
  \bibinfo{person}{Jos{\'e}~Miguel Hern{\'a}ndez-Lobato},
  \bibinfo{person}{Robert Adolf}, \bibinfo{person}{Michael Gelbart},
  \bibinfo{person}{Paul Whatmough}, \bibinfo{person}{Gu-Yeon Wei}, {and}
  \bibinfo{person}{David Brooks}.} \bibinfo{year}{2017}\natexlab{}.
\newblock \showarticletitle{A case for efficient accelerator design space
  exploration via bayesian optimization}. In \bibinfo{booktitle}{\emph{IEEE/ACM
  International Symposium on Low Power Electronics and Design (ISLPED)}}.
  \bibinfo{pages}{1--6}.
\newblock


\bibitem[\protect\citeauthoryear{Real, Aggarwal, Huang, and Le}{Real
  et~al\mbox{.}}{2019}]%
        {real2019regularized}
\bibfield{author}{\bibinfo{person}{Esteban Real}, \bibinfo{person}{Alok
  Aggarwal}, \bibinfo{person}{Yanping Huang}, {and} \bibinfo{person}{Quoc~V
  Le}.} \bibinfo{year}{2019}\natexlab{}.
\newblock \showarticletitle{Regularized evolution for image classifier
  architecture search}. In \bibinfo{booktitle}{\emph{AAAI Conference on
  Artificial Intelligence (AAAI)}}. \bibinfo{pages}{4780--4789}.
\newblock


\bibitem[\protect\citeauthoryear{Real, Liang, So, and Le}{Real
  et~al\mbox{.}}{2020}]%
        {real2020automl}
\bibfield{author}{\bibinfo{person}{Esteban Real}, \bibinfo{person}{Chen Liang},
  \bibinfo{person}{David So}, {and} \bibinfo{person}{Quoc Le}.}
  \bibinfo{year}{2020}\natexlab{}.
\newblock \showarticletitle{{AutoML-Zero}: evolving machine learning algorithms
  from scratch}. In \bibinfo{booktitle}{\emph{The International Conference on
  Machine Learning (ICML)}}. \bibinfo{pages}{8007--8019}.
\newblock


\bibitem[\protect\citeauthoryear{Real, Moore, Selle, Saxena, Suematsu, Tan, Le,
  and Kurakin}{Real et~al\mbox{.}}{2017}]%
        {real2017large}
\bibfield{author}{\bibinfo{person}{Esteban Real}, \bibinfo{person}{Sherry
  Moore}, \bibinfo{person}{Andrew Selle}, \bibinfo{person}{Saurabh Saxena},
  \bibinfo{person}{Yutaka~Leon Suematsu}, \bibinfo{person}{Jie Tan},
  \bibinfo{person}{Quoc~V Le}, {and} \bibinfo{person}{Alexey Kurakin}.}
  \bibinfo{year}{2017}\natexlab{}.
\newblock \showarticletitle{Large-Scale Evolution of Image Classifiers}. In
  \bibinfo{booktitle}{\emph{The International Conference on Machine Learning
  (ICML)}}. \bibinfo{pages}{2902--2911}.
\newblock


\bibitem[\protect\citeauthoryear{Ren, Zeng, Yang, and Urtasun}{Ren
  et~al\mbox{.}}{2018}]%
        {ren2018learning}
\bibfield{author}{\bibinfo{person}{Mengye Ren}, \bibinfo{person}{Wenyuan Zeng},
  \bibinfo{person}{Bin Yang}, {and} \bibinfo{person}{Raquel Urtasun}.}
  \bibinfo{year}{2018}\natexlab{}.
\newblock \showarticletitle{Learning to reweight examples for robust deep
  learning}. In \bibinfo{booktitle}{\emph{The International Conference on
  Machine Learning (ICML)}}. \bibinfo{pages}{4334--4343}.
\newblock


\bibitem[\protect\citeauthoryear{Ren, Xiao, Chang, Huang, Li, Chen, and
  Wang}{Ren et~al\mbox{.}}{2021}]%
        {ren2021comprehensive}
\bibfield{author}{\bibinfo{person}{Pengzhen Ren}, \bibinfo{person}{Yun Xiao},
  \bibinfo{person}{Xiaojun Chang}, \bibinfo{person}{Po-Yao Huang},
  \bibinfo{person}{Zhihui Li}, \bibinfo{person}{Xiaojiang Chen}, {and}
  \bibinfo{person}{Xin Wang}.} \bibinfo{year}{2021}\natexlab{}.
\newblock \showarticletitle{A comprehensive survey of neural architecture
  search: Challenges and solutions}.
\newblock \bibinfo{journal}{\emph{ACM Computing Surveys (CSUR)}}
  \bibinfo{volume}{54}, \bibinfo{number}{4} (\bibinfo{year}{2021}),
  \bibinfo{pages}{1--34}.
\newblock


\bibitem[\protect\citeauthoryear{Ricci, Rokach, and Shapira}{Ricci
  et~al\mbox{.}}{2011}]%
        {ricci2011introduction}
\bibfield{author}{\bibinfo{person}{Francesco Ricci}, \bibinfo{person}{Lior
  Rokach}, {and} \bibinfo{person}{Bracha Shapira}.}
  \bibinfo{year}{2011}\natexlab{}.
\newblock \showarticletitle{Introduction to recommender systems handbook}.
\newblock In \bibinfo{booktitle}{\emph{Recommender systems handbook}}.
  \bibinfo{publisher}{Springer}, \bibinfo{pages}{1--35}.
\newblock


\bibitem[\protect\citeauthoryear{Riedmiller and Braun}{Riedmiller and
  Braun}{1993}]%
        {riedmiller1993direct}
\bibfield{author}{\bibinfo{person}{Martin Riedmiller} {and}
  \bibinfo{person}{Heinrich Braun}.} \bibinfo{year}{1993}\natexlab{}.
\newblock \showarticletitle{A direct adaptive method for faster backpropagation
  learning: The RPROP algorithm}. In \bibinfo{booktitle}{\emph{IEEE
  International Conference on Neural Networks}}. \bibinfo{pages}{586--591}.
\newblock


\bibitem[\protect\citeauthoryear{Romera-Paredes and Torr}{Romera-Paredes and
  Torr}{2015}]%
        {romera2015embarrassingly}
\bibfield{author}{\bibinfo{person}{Bernardino Romera-Paredes} {and}
  \bibinfo{person}{Philip Torr}.} \bibinfo{year}{2015}\natexlab{}.
\newblock \showarticletitle{An embarrassingly simple approach to zero-shot
  learning}. In \bibinfo{booktitle}{\emph{The International Conference on
  Machine Learning (ICML)}}. \bibinfo{pages}{2152--2161}.
\newblock


\bibitem[\protect\citeauthoryear{Ru, Alvi, Nguyen, Osborne, and Roberts}{Ru
  et~al\mbox{.}}{2020a}]%
        {ru2020bayesian}
\bibfield{author}{\bibinfo{person}{Binxin Ru}, \bibinfo{person}{Ahsan Alvi},
  \bibinfo{person}{Vu Nguyen}, \bibinfo{person}{Michael~A Osborne}, {and}
  \bibinfo{person}{Stephen Roberts}.} \bibinfo{year}{2020}\natexlab{a}.
\newblock \showarticletitle{Bayesian optimisation over multiple continuous and
  categorical inputs}. In \bibinfo{booktitle}{\emph{The International
  Conference on Machine Learning (ICML)}}. \bibinfo{pages}{8276--8285}.
\newblock


\bibitem[\protect\citeauthoryear{Ru, Esperanca, and Carlucci}{Ru
  et~al\mbox{.}}{2020b}]%
        {ru2020neural}
\bibfield{author}{\bibinfo{person}{Binxin Ru}, \bibinfo{person}{Pedro
  Esperanca}, {and} \bibinfo{person}{Fabio Carlucci}.}
  \bibinfo{year}{2020}\natexlab{b}.
\newblock \showarticletitle{Neural Architecture Generator Optimization}. In
  \bibinfo{booktitle}{\emph{Proceedings of the International Conference on
  Neural Information Processing Systems (NeurIPS)}}.
\newblock


\bibitem[\protect\citeauthoryear{Rumelhart, Hinton, and Williams}{Rumelhart
  et~al\mbox{.}}{1985}]%
        {rumelhart1985learning}
\bibfield{author}{\bibinfo{person}{David~E Rumelhart},
  \bibinfo{person}{Geoffrey~E Hinton}, {and} \bibinfo{person}{Ronald~J
  Williams}.} \bibinfo{year}{1985}\natexlab{}.
\newblock \bibinfo{booktitle}{\emph{Learning internal representations by error
  propagation}}.
\newblock \bibinfo{type}{{T}echnical {R}eport}.
  \bibinfo{institution}{California Univ San Diego La Jolla Inst for Cognitive
  Science}.
\newblock


\bibitem[\protect\citeauthoryear{Russakovsky, Deng, Su, Krause, Satheesh, Ma,
  Huang, Karpathy, Khosla, Bernstein, Berg, and Fei{-}Fei}{Russakovsky
  et~al\mbox{.}}{2015}]%
        {russakovsky2015imagenet}
\bibfield{author}{\bibinfo{person}{Olga Russakovsky}, \bibinfo{person}{Jia
  Deng}, \bibinfo{person}{Hao Su}, \bibinfo{person}{Jonathan Krause},
  \bibinfo{person}{Sanjeev Satheesh}, \bibinfo{person}{Sean Ma},
  \bibinfo{person}{Zhiheng Huang}, \bibinfo{person}{Andrej Karpathy},
  \bibinfo{person}{Aditya Khosla}, \bibinfo{person}{Michael Bernstein},
  \bibinfo{person}{Alexander Berg}, {and} \bibinfo{person}{Li Fei{-}Fei}.}
  \bibinfo{year}{2015}\natexlab{}.
\newblock \showarticletitle{Imagenet large scale visual recognition challenge}.
\newblock \bibinfo{journal}{\emph{International Journal of Computer Vision
  (IJCV)}} \bibinfo{volume}{115}, \bibinfo{number}{3} (\bibinfo{year}{2015}),
  \bibinfo{pages}{211--252}.
\newblock


\bibitem[\protect\citeauthoryear{Rusu, Rabinowitz, Desjardins, Soyer,
  Kirkpatrick, Kavukcuoglu, Pascanu, and Hadsell}{Rusu et~al\mbox{.}}{2016}]%
        {rusu2016progressive}
\bibfield{author}{\bibinfo{person}{Andrei~A Rusu}, \bibinfo{person}{Neil~C
  Rabinowitz}, \bibinfo{person}{Guillaume Desjardins}, \bibinfo{person}{Hubert
  Soyer}, \bibinfo{person}{James Kirkpatrick}, \bibinfo{person}{Koray
  Kavukcuoglu}, \bibinfo{person}{Razvan Pascanu}, {and} \bibinfo{person}{Raia
  Hadsell}.} \bibinfo{year}{2016}\natexlab{}.
\newblock \showarticletitle{Progressive neural networks}.
\newblock \bibinfo{journal}{\emph{arXiv preprint arXiv:1606.04671}}
  (\bibinfo{year}{2016}).
\newblock


\bibitem[\protect\citeauthoryear{Ruta, Gabrys, and Lemke}{Ruta
  et~al\mbox{.}}{2011}]%
        {ruta2011generic}
\bibfield{author}{\bibinfo{person}{Dymitr Ruta}, \bibinfo{person}{Bogdan
  Gabrys}, {and} \bibinfo{person}{Christiane Lemke}.}
  \bibinfo{year}{2011}\natexlab{}.
\newblock \showarticletitle{A generic multilevel architecture for time series
  prediction}.
\newblock \bibinfo{journal}{\emph{IEEE Transactions on Knowledge and Data
  Engineering (TKDE)}} \bibinfo{volume}{23}, \bibinfo{number}{3}
  (\bibinfo{year}{2011}), \bibinfo{pages}{350--359}.
\newblock


\bibitem[\protect\citeauthoryear{Salimans, Ho, Chen, Sidor, and
  Sutskever}{Salimans et~al\mbox{.}}{2017}]%
        {salimans2017evolution}
\bibfield{author}{\bibinfo{person}{Tim Salimans}, \bibinfo{person}{Jonathan
  Ho}, \bibinfo{person}{Xi Chen}, \bibinfo{person}{Szymon Sidor}, {and}
  \bibinfo{person}{Ilya Sutskever}.} \bibinfo{year}{2017}\natexlab{}.
\newblock \showarticletitle{Evolution strategies as a scalable alternative to
  reinforcement learning}.
\newblock \bibinfo{journal}{\emph{arXiv preprint arXiv:1703.03864}}
  (\bibinfo{year}{2017}).
\newblock


\bibitem[\protect\citeauthoryear{Salvador, Budka, and Gabrys}{Salvador
  et~al\mbox{.}}{2016}]%
        {sabu16}
\bibfield{author}{\bibinfo{person}{Manuel~Martin Salvador},
  \bibinfo{person}{Marcin Budka}, {and} \bibinfo{person}{Bogdan Gabrys}.}
  \bibinfo{year}{2016}\natexlab{}.
\newblock \showarticletitle{Towards Automatic Composition of Multicomponent
  Predictive Systems}.
\newblock In \bibinfo{booktitle}{\emph{International Conference on Hybrid
  Artificial Intelligence Systems}}. \bibinfo{pages}{27--39}.
\newblock


\bibitem[\protect\citeauthoryear{Salvador, Budka, and Gabrys}{Salvador
  et~al\mbox{.}}{2019}]%
        {sabu19}
\bibfield{author}{\bibinfo{person}{Manuel~Martin Salvador},
  \bibinfo{person}{Marcin Budka}, {and} \bibinfo{person}{Bogdan Gabrys}.}
  \bibinfo{year}{2019}\natexlab{}.
\newblock \showarticletitle{Automatic Composition and Optimization of
  Multicomponent Predictive Systems With an Extended Auto-{WEKA}}.
\newblock \bibinfo{journal}{\emph{IEEE Transactions on Automation Science and
  Engineering (TASE)}} \bibinfo{volume}{16}, \bibinfo{number}{2}
  (\bibinfo{date}{apr} \bibinfo{year}{2019}), \bibinfo{pages}{946--959}.
\newblock
\urldef\tempurl%
\url{https://doi.org/10.1109/tase.2018.2876430}
\showDOI{\tempurl}


\bibitem[\protect\citeauthoryear{Sandler, Howard, Zhu, Zhmoginov, and
  Chen}{Sandler et~al\mbox{.}}{2018}]%
        {sandler2018mobilenetv2}
\bibfield{author}{\bibinfo{person}{Mark Sandler}, \bibinfo{person}{Andrew
  Howard}, \bibinfo{person}{Menglong Zhu}, \bibinfo{person}{Andrey Zhmoginov},
  {and} \bibinfo{person}{Liang-Chieh Chen}.} \bibinfo{year}{2018}\natexlab{}.
\newblock \showarticletitle{Mobilenetv2: Inverted residuals and linear
  bottlenecks}. In \bibinfo{booktitle}{\emph{Proceedings of the IEEE Conference
  Computer Vision Pattern Recognition (CVPR)}}. \bibinfo{pages}{4510--4520}.
\newblock


\bibitem[\protect\citeauthoryear{Schaul and Schmidhuber}{Schaul and
  Schmidhuber}{2010}]%
        {schaul2010metalearning}
\bibfield{author}{\bibinfo{person}{Tom Schaul} {and}
  \bibinfo{person}{J{\"u}rgen Schmidhuber}.} \bibinfo{year}{2010}\natexlab{}.
\newblock \showarticletitle{Metalearning}.
\newblock \bibinfo{journal}{\emph{Scholarpedia}} \bibinfo{volume}{5},
  \bibinfo{number}{6} (\bibinfo{year}{2010}), \bibinfo{pages}{4650}.
\newblock


\bibitem[\protect\citeauthoryear{Schmidhuber}{Schmidhuber}{1987}]%
        {schmidhuber1987evolutionary}
\bibfield{author}{\bibinfo{person}{J{\"u}rgen Schmidhuber}.}
  \bibinfo{year}{1987}\natexlab{}.
\newblock \emph{\bibinfo{title}{Evolutionary Principles in Self-referential
  Learning}}.
\newblock \bibinfo{thesistype}{Ph.\,D. Dissertation}.
\newblock


\bibitem[\protect\citeauthoryear{Schmidhuber}{Schmidhuber}{1992}]%
        {schmidhuber1992learning}
\bibfield{author}{\bibinfo{person}{J{\"u}rgen Schmidhuber}.}
  \bibinfo{year}{1992}\natexlab{}.
\newblock \showarticletitle{Learning to control fast-weight memories: An
  alternative to dynamic recurrent networks}.
\newblock \bibinfo{journal}{\emph{Neural Computation}} \bibinfo{volume}{4},
  \bibinfo{number}{1} (\bibinfo{year}{1992}), \bibinfo{pages}{131--139}.
\newblock


\bibitem[\protect\citeauthoryear{Schmidhuber}{Schmidhuber}{2002}]%
        {schmidhuber2002bias}
\bibfield{author}{\bibinfo{person}{Juergen Schmidhuber}.}
  \bibinfo{year}{2002}\natexlab{}.
\newblock \showarticletitle{Bias-optimal incremental problem solving}. In
  \bibinfo{booktitle}{\emph{Proceedings of the International Conference on
  Neural Information Processing Systems (NeurIPS)}}.
\newblock


\bibitem[\protect\citeauthoryear{Schmidhuber}{Schmidhuber}{2015}]%
        {schmidhuber2015deep}
\bibfield{author}{\bibinfo{person}{J{\"u}rgen Schmidhuber}.}
  \bibinfo{year}{2015}\natexlab{}.
\newblock \showarticletitle{Deep learning in neural networks: An overview}.
\newblock \bibinfo{journal}{\emph{Neural Networks}}  \bibinfo{volume}{61}
  (\bibinfo{year}{2015}), \bibinfo{pages}{85--117}.
\newblock


\bibitem[\protect\citeauthoryear{Schmidhuber, Zhao, and
  Schraudolph}{Schmidhuber et~al\mbox{.}}{1998}]%
        {schmidhuber1998reinforcement}
\bibfield{author}{\bibinfo{person}{J{\"u}rgen Schmidhuber},
  \bibinfo{person}{Jieyu Zhao}, {and} \bibinfo{person}{Nicol~N Schraudolph}.}
  \bibinfo{year}{1998}\natexlab{}.
\newblock \showarticletitle{Reinforcement learning with self-modifying
  policies}.
\newblock In \bibinfo{booktitle}{\emph{Learning to learn}}.
  \bibinfo{publisher}{Springer}, \bibinfo{pages}{293--309}.
\newblock


\bibitem[\protect\citeauthoryear{Schulman, Wolski, Dhariwal, Radford, and
  Klimov}{Schulman et~al\mbox{.}}{2017}]%
        {schulman2017proximal}
\bibfield{author}{\bibinfo{person}{John Schulman}, \bibinfo{person}{Filip
  Wolski}, \bibinfo{person}{Prafulla Dhariwal}, \bibinfo{person}{Alec Radford},
  {and} \bibinfo{person}{Oleg Klimov}.} \bibinfo{year}{2017}\natexlab{}.
\newblock \showarticletitle{Proximal policy optimization algorithms}.
\newblock \bibinfo{journal}{\emph{arXiv preprint arXiv:1707.06347}}
  (\bibinfo{year}{2017}).
\newblock


\bibitem[\protect\citeauthoryear{Sculley, Holt, Golovin, Davydov, Phillips,
  Ebner, Chaudhary, Young, Crespo, and Dennison}{Sculley et~al\mbox{.}}{2015}]%
        {sculley2015hidden}
\bibfield{author}{\bibinfo{person}{David Sculley}, \bibinfo{person}{Gary Holt},
  \bibinfo{person}{Daniel Golovin}, \bibinfo{person}{Eugene Davydov},
  \bibinfo{person}{Todd Phillips}, \bibinfo{person}{Dietmar Ebner},
  \bibinfo{person}{Vinay Chaudhary}, \bibinfo{person}{Michael Young},
  \bibinfo{person}{Jean-Francois Crespo}, {and} \bibinfo{person}{Dan
  Dennison}.} \bibinfo{year}{2015}\natexlab{}.
\newblock \showarticletitle{Hidden technical debt in machine learning systems}.
  In \bibinfo{booktitle}{\emph{Proceedings of the International Conference on
  Neural Information Processing Systems (NeurIPS)}}.
  \bibinfo{pages}{2503--2511}.
\newblock


\bibitem[\protect\citeauthoryear{Shaban, Cheng, Hatch, and Boots}{Shaban
  et~al\mbox{.}}{2019}]%
        {shaban2019truncated}
\bibfield{author}{\bibinfo{person}{Amirreza Shaban}, \bibinfo{person}{Ching-An
  Cheng}, \bibinfo{person}{Nathan Hatch}, {and} \bibinfo{person}{Byron Boots}.}
  \bibinfo{year}{2019}\natexlab{}.
\newblock \showarticletitle{Truncated back-propagation for bilevel
  optimization}. In \bibinfo{booktitle}{\emph{The International Conference on
  Artificial Intelligence and Statistics (AISTATS)}}.
  \bibinfo{pages}{1723--1732}.
\newblock


\bibitem[\protect\citeauthoryear{Shorten and Khoshgoftaar}{Shorten and
  Khoshgoftaar}{2019}]%
        {shorten2019survey}
\bibfield{author}{\bibinfo{person}{Connor Shorten} {and}
  \bibinfo{person}{Taghi~M Khoshgoftaar}.} \bibinfo{year}{2019}\natexlab{}.
\newblock \showarticletitle{A survey on image data augmentation for deep
  learning}.
\newblock \bibinfo{journal}{\emph{Journal of Big Data}} \bibinfo{volume}{6},
  \bibinfo{number}{1} (\bibinfo{year}{2019}), \bibinfo{pages}{1--48}.
\newblock


\bibitem[\protect\citeauthoryear{Shu, Xie, Yi, Zhao, Zhou, Xu, and Meng}{Shu
  et~al\mbox{.}}{2019}]%
        {shu2019meta}
\bibfield{author}{\bibinfo{person}{Jun Shu}, \bibinfo{person}{Qi Xie},
  \bibinfo{person}{Lixuan Yi}, \bibinfo{person}{Qian Zhao},
  \bibinfo{person}{Sanping Zhou}, \bibinfo{person}{Zongben Xu}, {and}
  \bibinfo{person}{Deyu Meng}.} \bibinfo{year}{2019}\natexlab{}.
\newblock \showarticletitle{Meta-weight-net: Learning an explicit mapping for
  sample weighting}. In \bibinfo{booktitle}{\emph{Proceedings of the
  International Conference on Neural Information Processing Systems
  (NeurIPS)}}.
\newblock


\bibitem[\protect\citeauthoryear{Siems, Zimmer, Zela, Lukasik, Keuper, and
  Hutter}{Siems et~al\mbox{.}}{2020}]%
        {siems2020bench}
\bibfield{author}{\bibinfo{person}{Julien Siems}, \bibinfo{person}{Lucas
  Zimmer}, \bibinfo{person}{Arber Zela}, \bibinfo{person}{Jovita Lukasik},
  \bibinfo{person}{Margret Keuper}, {and} \bibinfo{person}{Frank Hutter}.}
  \bibinfo{year}{2020}\natexlab{}.
\newblock \showarticletitle{{NAS-Bench-301} and the case for surrogate
  benchmarks for neural architecture search}.
\newblock \bibinfo{journal}{\emph{arXiv preprint arXiv:2008.09777}}
  (\bibinfo{year}{2020}).
\newblock


\bibitem[\protect\citeauthoryear{Silver, Huang, Maddison, Guez, Sifre, Van
  Den~Driessche, Schrittwieser, Antonoglou, Panneershelvam, Lanctot,
  et~al\mbox{.}}{Silver et~al\mbox{.}}{2016}]%
        {silver2016mastering}
\bibfield{author}{\bibinfo{person}{David Silver}, \bibinfo{person}{Aja Huang},
  \bibinfo{person}{Chris~J Maddison}, \bibinfo{person}{Arthur Guez},
  \bibinfo{person}{Laurent Sifre}, \bibinfo{person}{George Van Den~Driessche},
  \bibinfo{person}{Julian Schrittwieser}, \bibinfo{person}{Ioannis Antonoglou},
  \bibinfo{person}{Veda Panneershelvam}, \bibinfo{person}{Marc Lanctot},
  {et~al\mbox{.}}} \bibinfo{year}{2016}\natexlab{}.
\newblock \showarticletitle{Mastering the game of Go with deep neural networks
  and tree search}.
\newblock \bibinfo{journal}{\emph{Nature}} \bibinfo{volume}{529},
  \bibinfo{number}{7587} (\bibinfo{year}{2016}), \bibinfo{pages}{484--489}.
\newblock


\bibitem[\protect\citeauthoryear{Simonyan and Zisserman}{Simonyan and
  Zisserman}{2015}]%
        {simonyan2015very}
\bibfield{author}{\bibinfo{person}{Karen Simonyan} {and}
  \bibinfo{person}{Andrew Zisserman}.} \bibinfo{year}{2015}\natexlab{}.
\newblock \showarticletitle{Very deep convolutional networks for large-scale
  image recognition}. In \bibinfo{booktitle}{\emph{International Conference on
  Learning Representations (ICLR)}}.
\newblock


\bibitem[\protect\citeauthoryear{Smith, Kindermans, Ying, and Le}{Smith
  et~al\mbox{.}}{2018}]%
        {smith2018don}
\bibfield{author}{\bibinfo{person}{Samuel~L Smith}, \bibinfo{person}{Pieter-Jan
  Kindermans}, \bibinfo{person}{Chris Ying}, {and} \bibinfo{person}{Quoc~V
  Le}.} \bibinfo{year}{2018}\natexlab{}.
\newblock \showarticletitle{Don't decay the learning rate, increase the batch
  size}. In \bibinfo{booktitle}{\emph{International Conference on Learning
  Representations (ICLR)}}.
\newblock


\bibitem[\protect\citeauthoryear{Smith-Miles}{Smith-Miles}{2009}]%
        {smith2009cross}
\bibfield{author}{\bibinfo{person}{Kate~A Smith-Miles}.}
  \bibinfo{year}{2009}\natexlab{}.
\newblock \showarticletitle{Cross-disciplinary perspectives on meta-learning
  for algorithm selection}.
\newblock \bibinfo{journal}{\emph{ACM Computing Surveys (CSUR)}}
  \bibinfo{volume}{41}, \bibinfo{number}{1} (\bibinfo{year}{2009}),
  \bibinfo{pages}{1--25}.
\newblock


\bibitem[\protect\citeauthoryear{Snell, Swersky, and Zemel}{Snell
  et~al\mbox{.}}{2017}]%
        {snell2017prototypical}
\bibfield{author}{\bibinfo{person}{Jake Snell}, \bibinfo{person}{Kevin
  Swersky}, {and} \bibinfo{person}{Richard Zemel}.}
  \bibinfo{year}{2017}\natexlab{}.
\newblock \showarticletitle{Prototypical Networks for Few-shot Learning}. In
  \bibinfo{booktitle}{\emph{Proceedings of the International Conference on
  Neural Information Processing Systems (NeurIPS)}}.
  \bibinfo{pages}{4077--4087}.
\newblock


\bibitem[\protect\citeauthoryear{Snoek, Larochelle, and Adams}{Snoek
  et~al\mbox{.}}{2012}]%
        {snoek2012practical}
\bibfield{author}{\bibinfo{person}{Jasper Snoek}, \bibinfo{person}{Hugo
  Larochelle}, {and} \bibinfo{person}{Ryan~P Adams}.}
  \bibinfo{year}{2012}\natexlab{}.
\newblock \showarticletitle{Practical Bayesian optimization of machine learning
  algorithms}. In \bibinfo{booktitle}{\emph{Proceedings of the International
  Conference on Neural Information Processing Systems (NeurIPS)}}.
  \bibinfo{pages}{2951--2959}.
\newblock


\bibitem[\protect\citeauthoryear{So, Le, and Liang}{So et~al\mbox{.}}{2019}]%
        {so2019evolved}
\bibfield{author}{\bibinfo{person}{David So}, \bibinfo{person}{Quoc Le}, {and}
  \bibinfo{person}{Chen Liang}.} \bibinfo{year}{2019}\natexlab{}.
\newblock \showarticletitle{The evolved transformer}. In
  \bibinfo{booktitle}{\emph{The International Conference on Machine Learning
  (ICML)}}. \bibinfo{pages}{5877--5886}.
\newblock


\bibitem[\protect\citeauthoryear{So, Ma{\'n}ke, Liu, Dai, Shazeer, and Le}{So
  et~al\mbox{.}}{2021}]%
        {so2021primer}
\bibfield{author}{\bibinfo{person}{David~R So}, \bibinfo{person}{Wojciech
  Ma{\'n}ke}, \bibinfo{person}{Hanxiao Liu}, \bibinfo{person}{Zihang Dai},
  \bibinfo{person}{Noam Shazeer}, {and} \bibinfo{person}{Quoc~V Le}.}
  \bibinfo{year}{2021}\natexlab{}.
\newblock \showarticletitle{Primer: Searching for efficient transformers for
  language modeling}. In \bibinfo{booktitle}{\emph{Proceedings of the
  International Conference on Neural Information Processing Systems
  (NeurIPS)}}.
\newblock


\bibitem[\protect\citeauthoryear{Song, Jin, and Hu}{Song et~al\mbox{.}}{2022}]%
        {song2022automated}
\bibfield{author}{\bibinfo{person}{Qingquan Song}, \bibinfo{person}{Haifeng
  Jin}, {and} \bibinfo{person}{Xia Hu}.} \bibinfo{year}{2022}\natexlab{}.
\newblock \bibinfo{booktitle}{\emph{Automated Machine Learning in Action}}.
\newblock \bibinfo{publisher}{Manning Publications Co.}
\newblock


\bibitem[\protect\citeauthoryear{Sprechmann, Jayakumar, Rae, Pritzel, Badia,
  Uria, Vinyals, Hassabis, Pascanu, and Blundell}{Sprechmann
  et~al\mbox{.}}{2018}]%
        {sprechmann2018memory}
\bibfield{author}{\bibinfo{person}{Pablo Sprechmann},
  \bibinfo{person}{Siddhant~M Jayakumar}, \bibinfo{person}{Jack~W Rae},
  \bibinfo{person}{Alexander Pritzel}, \bibinfo{person}{Adria~Puigdomenech
  Badia}, \bibinfo{person}{Benigno Uria}, \bibinfo{person}{Oriol Vinyals},
  \bibinfo{person}{Demis Hassabis}, \bibinfo{person}{Razvan Pascanu}, {and}
  \bibinfo{person}{Charles Blundell}.} \bibinfo{year}{2018}\natexlab{}.
\newblock \showarticletitle{Memory-based parameter adaptation}. In
  \bibinfo{booktitle}{\emph{International Conference on Learning
  Representations (ICLR)}}.
\newblock


\bibitem[\protect\citeauthoryear{Strubell, Ganesh, and McCallum}{Strubell
  et~al\mbox{.}}{2019}]%
        {strubell2019energy}
\bibfield{author}{\bibinfo{person}{Emma Strubell}, \bibinfo{person}{Ananya
  Ganesh}, {and} \bibinfo{person}{Andrew McCallum}.}
  \bibinfo{year}{2019}\natexlab{}.
\newblock \showarticletitle{Energy and policy considerations for deep learning
  in NLP}. In \bibinfo{booktitle}{\emph{The Annual Meeting of the Association
  for Computational Linguistics}}.
\newblock


\bibitem[\protect\citeauthoryear{Such, Rawal, Lehman, Stanley, and Clune}{Such
  et~al\mbox{.}}{2020}]%
        {such2020generative}
\bibfield{author}{\bibinfo{person}{Felipe~Petroski Such},
  \bibinfo{person}{Aditya Rawal}, \bibinfo{person}{Joel Lehman},
  \bibinfo{person}{Kenneth Stanley}, {and} \bibinfo{person}{Jeffrey Clune}.}
  \bibinfo{year}{2020}\natexlab{}.
\newblock \showarticletitle{Generative teaching networks: Accelerating neural
  architecture search by learning to generate synthetic training data}. In
  \bibinfo{booktitle}{\emph{The International Conference on Machine Learning
  (ICML)}}. \bibinfo{pages}{9206--9216}.
\newblock


\bibitem[\protect\citeauthoryear{Sutton}{Sutton}{1992}]%
        {sutton1992adapting}
\bibfield{author}{\bibinfo{person}{Richard~S Sutton}.}
  \bibinfo{year}{1992}\natexlab{}.
\newblock \showarticletitle{Adapting bias by gradient descent: An incremental
  version of delta-bar-delta}. In \bibinfo{booktitle}{\emph{AAAI Conference on
  Artificial Intelligence (AAAI)}}. \bibinfo{pages}{171--176}.
\newblock


\bibitem[\protect\citeauthoryear{Sutton and Barto}{Sutton and Barto}{2018}]%
        {sutton2018reinforcement}
\bibfield{author}{\bibinfo{person}{Richard~S Sutton} {and}
  \bibinfo{person}{Andrew~G Barto}.} \bibinfo{year}{2018}\natexlab{}.
\newblock \bibinfo{booktitle}{\emph{Reinforcement learning: An introduction}}.
\newblock \bibinfo{publisher}{MIT press}.
\newblock


\bibitem[\protect\citeauthoryear{S{\"u}zen}{S{\"u}zen}{2020}]%
        {suzen2020equivalence}
\bibfield{author}{\bibinfo{person}{Mehmet S{\"u}zen}.}
  \bibinfo{year}{2020}\natexlab{}.
\newblock \showarticletitle{Equivalence in Deep Neural Networks via Conjugate
  Matrix Ensembles}.
\newblock \bibinfo{journal}{\emph{arXiv preprint arXiv:2006.13687}}
  (\bibinfo{year}{2020}).
\newblock


\bibitem[\protect\citeauthoryear{S{\"u}zen, Cerd{\`a}, and Weber}{S{\"u}zen
  et~al\mbox{.}}{2019}]%
        {suzen2019periodic}
\bibfield{author}{\bibinfo{person}{Mehmet S{\"u}zen}, \bibinfo{person}{Joan~J
  Cerd{\`a}}, {and} \bibinfo{person}{Cornelius Weber}.}
  \bibinfo{year}{2019}\natexlab{}.
\newblock \showarticletitle{Periodic Spectral Ergodicity: A Complexity Measure
  for Deep Neural Networks and Neural Architecture Search}.
\newblock \bibinfo{journal}{\emph{arXiv preprint arXiv:1911.07831}}
  (\bibinfo{year}{2019}).
\newblock


\bibitem[\protect\citeauthoryear{Swersky, Snoek, and Adams}{Swersky
  et~al\mbox{.}}{2014}]%
        {swersky2014freeze}
\bibfield{author}{\bibinfo{person}{Kevin Swersky}, \bibinfo{person}{Jasper
  Snoek}, {and} \bibinfo{person}{Ryan~Prescott Adams}.}
  \bibinfo{year}{2014}\natexlab{}.
\newblock \showarticletitle{Freeze-thaw bayesian optimization}.
\newblock \bibinfo{journal}{\emph{arXiv preprint arXiv:1406.3896}}
  (\bibinfo{year}{2014}).
\newblock


\bibitem[\protect\citeauthoryear{Szegedy, Liu, Jia, Sermanet, Reed, Anguelov,
  Erhan, Vanhoucke, and Rabinovich}{Szegedy et~al\mbox{.}}{2015}]%
        {szegedy2015going}
\bibfield{author}{\bibinfo{person}{Christian Szegedy}, \bibinfo{person}{Wei
  Liu}, \bibinfo{person}{Yangqing Jia}, \bibinfo{person}{Pierre Sermanet},
  \bibinfo{person}{Scott Reed}, \bibinfo{person}{Dragomir Anguelov},
  \bibinfo{person}{Dumitru Erhan}, \bibinfo{person}{Vincent Vanhoucke}, {and}
  \bibinfo{person}{Andrew Rabinovich}.} \bibinfo{year}{2015}\natexlab{}.
\newblock \showarticletitle{Going deeper with convolutions}. In
  \bibinfo{booktitle}{\emph{Proceedings of the IEEE Conference Computer Vision
  Pattern Recognition (CVPR)}}. \bibinfo{pages}{1--9}.
\newblock


\bibitem[\protect\citeauthoryear{Tan, Chen, Pang, Vasudevan, Sandler, Howard,
  and Le}{Tan et~al\mbox{.}}{2019}]%
        {tan2019mnasnet}
\bibfield{author}{\bibinfo{person}{Mingxing Tan}, \bibinfo{person}{Bo Chen},
  \bibinfo{person}{Ruoming Pang}, \bibinfo{person}{Vijay Vasudevan},
  \bibinfo{person}{Mark Sandler}, \bibinfo{person}{Andrew Howard}, {and}
  \bibinfo{person}{Quoc~V Le}.} \bibinfo{year}{2019}\natexlab{}.
\newblock \showarticletitle{{MnasNet}: Platform-aware neural architecture
  search for mobile}. In \bibinfo{booktitle}{\emph{Proceedings of the IEEE
  Conference Computer Vision Pattern Recognition (CVPR)}}.
  \bibinfo{pages}{2820--2828}.
\newblock


\bibitem[\protect\citeauthoryear{Tan and Le}{Tan and Le}{2021}]%
        {tan2021efficientnetv2}
\bibfield{author}{\bibinfo{person}{Mingxing Tan} {and} \bibinfo{person}{Quoc~V
  Le}.} \bibinfo{year}{2021}\natexlab{}.
\newblock \showarticletitle{{EfficientNetV2}: Smaller models and faster
  training}. In \bibinfo{booktitle}{\emph{The International Conference on
  Machine Learning (ICML)}}, Vol.~\bibinfo{volume}{139}.
  \bibinfo{pages}{10096--10106}.
\newblock


\bibitem[\protect\citeauthoryear{Thornton, Hutter, Hoos, and
  Leyton-Brown}{Thornton et~al\mbox{.}}{2013}]%
        {thornton2013auto}
\bibfield{author}{\bibinfo{person}{Chris Thornton}, \bibinfo{person}{Frank
  Hutter}, \bibinfo{person}{Holger~H Hoos}, {and} \bibinfo{person}{Kevin
  Leyton-Brown}.} \bibinfo{year}{2013}\natexlab{}.
\newblock \showarticletitle{{Auto-WEKA}: Combined selection and hyperparameter
  optimization of classification algorithms}. In
  \bibinfo{booktitle}{\emph{Proceedings of the ACM SIGKDD International
  Conference on Knowledge Discovery and Data Mining (KDD)}}.
  \bibinfo{pages}{847--855}.
\newblock


\bibitem[\protect\citeauthoryear{Thrun and Mitchell}{Thrun and
  Mitchell}{1995}]%
        {thrun1995lifelong}
\bibfield{author}{\bibinfo{person}{Sebastian Thrun} {and}
  \bibinfo{person}{Tom~M Mitchell}.} \bibinfo{year}{1995}\natexlab{}.
\newblock \showarticletitle{Lifelong robot learning}.
\newblock \bibinfo{journal}{\emph{Robotics and autonomous systems}}
  \bibinfo{volume}{15}, \bibinfo{number}{1-2} (\bibinfo{year}{1995}),
  \bibinfo{pages}{25--46}.
\newblock


\bibitem[\protect\citeauthoryear{Vaswani, Shazeer, Parmar, Uszkoreit, Jones,
  Gomez, Kaiser, and Polosukhin}{Vaswani et~al\mbox{.}}{2017}]%
        {vaswani2017attention}
\bibfield{author}{\bibinfo{person}{Ashish Vaswani}, \bibinfo{person}{Noam
  Shazeer}, \bibinfo{person}{Niki Parmar}, \bibinfo{person}{Jakob Uszkoreit},
  \bibinfo{person}{Llion Jones}, \bibinfo{person}{Aidan~N Gomez},
  \bibinfo{person}{{\L}ukasz Kaiser}, {and} \bibinfo{person}{Illia
  Polosukhin}.} \bibinfo{year}{2017}\natexlab{}.
\newblock \showarticletitle{Attention is all you need}. In
  \bibinfo{booktitle}{\emph{Proceedings of the International Conference on
  Neural Information Processing Systems (NeurIPS)}}.
  \bibinfo{pages}{6000--6010}.
\newblock


\bibitem[\protect\citeauthoryear{Veeriah, Hessel, Xu, Lewis, Rajendran, Oh, van
  Hasselt, Silver, and Singh}{Veeriah et~al\mbox{.}}{2019}]%
        {veeriah2019discovery}
\bibfield{author}{\bibinfo{person}{Vivek Veeriah}, \bibinfo{person}{Matteo
  Hessel}, \bibinfo{person}{Zhongwen Xu}, \bibinfo{person}{Richard Lewis},
  \bibinfo{person}{Janarthanan Rajendran}, \bibinfo{person}{Junhyuk Oh},
  \bibinfo{person}{Hado van Hasselt}, \bibinfo{person}{David Silver}, {and}
  \bibinfo{person}{Satinder Singh}.} \bibinfo{year}{2019}\natexlab{}.
\newblock \showarticletitle{Discovery of useful questions as auxiliary tasks}.
  In \bibinfo{booktitle}{\emph{Proceedings of the International Conference on
  Neural Information Processing Systems (NeurIPS)}}.
\newblock


\bibitem[\protect\citeauthoryear{Vincent, Larochelle, Bengio, and
  Manzagol}{Vincent et~al\mbox{.}}{2008}]%
        {vincent2008extracting}
\bibfield{author}{\bibinfo{person}{Pascal Vincent}, \bibinfo{person}{Hugo
  Larochelle}, \bibinfo{person}{Yoshua Bengio}, {and}
  \bibinfo{person}{Pierre-Antoine Manzagol}.} \bibinfo{year}{2008}\natexlab{}.
\newblock \showarticletitle{Extracting and composing robust features with
  denoising autoencoders}. In \bibinfo{booktitle}{\emph{The International
  Conference on Machine Learning (ICML)}}. \bibinfo{pages}{1096--1103}.
\newblock


\bibitem[\protect\citeauthoryear{Wan, Dai, Zhang, He, Tian, Xie, Wu, Yu, Xu,
  Chen, et~al\mbox{.}}{Wan et~al\mbox{.}}{2020}]%
        {wan2020fbnetv2}
\bibfield{author}{\bibinfo{person}{Alvin Wan}, \bibinfo{person}{Xiaoliang Dai},
  \bibinfo{person}{Peizhao Zhang}, \bibinfo{person}{Zijian He},
  \bibinfo{person}{Yuandong Tian}, \bibinfo{person}{Saining Xie},
  \bibinfo{person}{Bichen Wu}, \bibinfo{person}{Matthew Yu},
  \bibinfo{person}{Tao Xu}, \bibinfo{person}{Kan Chen}, {et~al\mbox{.}}}
  \bibinfo{year}{2020}\natexlab{}.
\newblock \showarticletitle{{FBNetV2}: Differentiable neural architecture
  search for spatial and channel dimensions}. In
  \bibinfo{booktitle}{\emph{Proceedings of the IEEE Conference Computer Vision
  Pattern Recognition (CVPR)}}. \bibinfo{pages}{12965--12974}.
\newblock


\bibitem[\protect\citeauthoryear{Wang, Lan, Liu, Ouyang, Zeng, and Qin}{Wang
  et~al\mbox{.}}{2021}]%
        {wang2021generalizing}
\bibfield{author}{\bibinfo{person}{Jindong Wang}, \bibinfo{person}{Cuiling
  Lan}, \bibinfo{person}{Chang Liu}, \bibinfo{person}{Yidong Ouyang},
  \bibinfo{person}{Wenjun Zeng}, {and} \bibinfo{person}{Tao Qin}.}
  \bibinfo{year}{2021}\natexlab{}.
\newblock \showarticletitle{Generalizing to Unseen Domains: A Survey on Domain
  Generalization}.
\newblock \bibinfo{journal}{\emph{arXiv preprint arXiv:2103.03097}}
  (\bibinfo{year}{2021}).
\newblock


\bibitem[\protect\citeauthoryear{Wang, Kurth-Nelson, Soyer, Leibo, Tirumala,
  Munos, Blundell, Kumaran, and Botvinick}{Wang et~al\mbox{.}}{2017}]%
        {wang2017learning}
\bibfield{author}{\bibinfo{person}{Jane~X Wang}, \bibinfo{person}{Zeb
  Kurth-Nelson}, \bibinfo{person}{Hubert Soyer}, \bibinfo{person}{Joel~Z
  Leibo}, \bibinfo{person}{Dhruva Tirumala}, \bibinfo{person}{R{\'e}mi Munos},
  \bibinfo{person}{Charles Blundell}, \bibinfo{person}{Dharshan Kumaran}, {and}
  \bibinfo{person}{Matt~M Botvinick}.} \bibinfo{year}{2017}\natexlab{}.
\newblock \showarticletitle{Learning to reinforcement learn}. In
  \bibinfo{booktitle}{\emph{Cognitive Science Society (CogSci)}}.
\newblock


\bibitem[\protect\citeauthoryear{Wang, Liu, Lin, Lin, and Han}{Wang
  et~al\mbox{.}}{2019}]%
        {wang2019haq}
\bibfield{author}{\bibinfo{person}{Kuan Wang}, \bibinfo{person}{Zhijian Liu},
  \bibinfo{person}{Yujun Lin}, \bibinfo{person}{Ji Lin}, {and}
  \bibinfo{person}{Song Han}.} \bibinfo{year}{2019}\natexlab{}.
\newblock \showarticletitle{{HAQ}: Hardware-aware automated quantization with
  mixed precision}. In \bibinfo{booktitle}{\emph{Proceedings of the IEEE
  Conference Computer Vision Pattern Recognition (CVPR)}}.
  \bibinfo{pages}{8612--8620}.
\newblock


\bibitem[\protect\citeauthoryear{Wang, Wang, Cai, Lin, Liu, Wang, Lin, and
  Han}{Wang et~al\mbox{.}}{2020}]%
        {wang2020apq}
\bibfield{author}{\bibinfo{person}{Tianzhe Wang}, \bibinfo{person}{Kuan Wang},
  \bibinfo{person}{Han Cai}, \bibinfo{person}{Ji Lin}, \bibinfo{person}{Zhijian
  Liu}, \bibinfo{person}{Hanrui Wang}, \bibinfo{person}{Yujun Lin}, {and}
  \bibinfo{person}{Song Han}.} \bibinfo{year}{2020}\natexlab{}.
\newblock \showarticletitle{{APQ}: Joint search for network architecture,
  pruning and quantization policy}. In \bibinfo{booktitle}{\emph{Proceedings of
  the IEEE Conference Computer Vision Pattern Recognition (CVPR)}}.
  \bibinfo{pages}{2078--2087}.
\newblock


\bibitem[\protect\citeauthoryear{Wang, Zhu, Torralba, and Efros}{Wang
  et~al\mbox{.}}{2018}]%
        {wang2018dataset}
\bibfield{author}{\bibinfo{person}{Tongzhou Wang}, \bibinfo{person}{Jun-Yan
  Zhu}, \bibinfo{person}{Antonio Torralba}, {and} \bibinfo{person}{Alexei~A
  Efros}.} \bibinfo{year}{2018}\natexlab{}.
\newblock \showarticletitle{Dataset distillation}.
\newblock \bibinfo{journal}{\emph{arXiv preprint arXiv:1811.10959}}
  (\bibinfo{year}{2018}).
\newblock


\bibitem[\protect\citeauthoryear{Watkins}{Watkins}{1989}]%
        {watkins1989learning}
\bibfield{author}{\bibinfo{person}{Christopher John Cornish~Hellaby Watkins}.}
  \bibinfo{year}{1989}\natexlab{}.
\newblock \showarticletitle{Learning from delayed rewards}.
\newblock \bibinfo{journal}{\emph{PhD thesis, University of Cambridge}}
  (\bibinfo{year}{1989}).
\newblock


\bibitem[\protect\citeauthoryear{Weigend and Gershenfeld}{Weigend and
  Gershenfeld}{1993}]%
        {weigend1993results}
\bibfield{author}{\bibinfo{person}{Andreas~S Weigend} {and}
  \bibinfo{person}{Neil~A Gershenfeld}.} \bibinfo{year}{1993}\natexlab{}.
\newblock \showarticletitle{Results of the time series prediction competition
  at the Santa Fe Institute}. In \bibinfo{booktitle}{\emph{IEEE International
  Conference on Neural Networks}}. \bibinfo{pages}{1786--1793}.
\newblock


\bibitem[\protect\citeauthoryear{Wen, Liu, Chen, Li, Bender, and
  Kindermans}{Wen et~al\mbox{.}}{2020}]%
        {wen2020neural}
\bibfield{author}{\bibinfo{person}{Wei Wen}, \bibinfo{person}{Hanxiao Liu},
  \bibinfo{person}{Yiran Chen}, \bibinfo{person}{Hai Li},
  \bibinfo{person}{Gabriel Bender}, {and} \bibinfo{person}{Pieter-Jan
  Kindermans}.} \bibinfo{year}{2020}\natexlab{}.
\newblock \showarticletitle{Neural predictor for neural architecture search}.
  In \bibinfo{booktitle}{\emph{European Conference on Computer Vision}}.
  Springer, \bibinfo{pages}{660--676}.
\newblock


\bibitem[\protect\citeauthoryear{White, Neiswanger, and Savani}{White
  et~al\mbox{.}}{2021}]%
        {white2021bananas}
\bibfield{author}{\bibinfo{person}{Colin White}, \bibinfo{person}{Willie
  Neiswanger}, {and} \bibinfo{person}{Yash Savani}.}
  \bibinfo{year}{2021}\natexlab{}.
\newblock \showarticletitle{{BANANAS}: Bayesian optimization with neural
  architectures for neural architecture search}. In
  \bibinfo{booktitle}{\emph{AAAI Conference on Artificial Intelligence
  (AAAI)}}.
\newblock


\bibitem[\protect\citeauthoryear{White and White}{White and White}{2016}]%
        {white2016greedy}
\bibfield{author}{\bibinfo{person}{Martha White} {and} \bibinfo{person}{Adam
  White}.} \bibinfo{year}{2016}\natexlab{}.
\newblock \showarticletitle{A Greedy Approach to Adapting the Trace Parameter
  for Temporal Difference Learning}. In \bibinfo{booktitle}{\emph{Proceedings
  of the International Conference on Autonomous Agents and Multiagent Systems
  (AAMAS)}}. \bibinfo{pages}{557--565}.
\newblock


\bibitem[\protect\citeauthoryear{Williams}{Williams}{1992}]%
        {williams1992simple}
\bibfield{author}{\bibinfo{person}{Ronald~J Williams}.}
  \bibinfo{year}{1992}\natexlab{}.
\newblock \showarticletitle{Simple statistical gradient-following algorithms
  for connectionist reinforcement learning}.
\newblock \bibinfo{journal}{\emph{Machine learning}} \bibinfo{volume}{8},
  \bibinfo{number}{3-4} (\bibinfo{year}{1992}), \bibinfo{pages}{229--256}.
\newblock


\bibitem[\protect\citeauthoryear{Wistuba, Rawat, and Pedapati}{Wistuba
  et~al\mbox{.}}{2019}]%
        {wistuba2019survey}
\bibfield{author}{\bibinfo{person}{Martin Wistuba}, \bibinfo{person}{Ambrish
  Rawat}, {and} \bibinfo{person}{Tejaswini Pedapati}.}
  \bibinfo{year}{2019}\natexlab{}.
\newblock \showarticletitle{A survey on neural architecture search}.
\newblock \bibinfo{journal}{\emph{arXiv preprint arXiv:1905.01392}}
  (\bibinfo{year}{2019}).
\newblock


\bibitem[\protect\citeauthoryear{Wolpert and Macready}{Wolpert and
  Macready}{1997}]%
        {wolpert1997no}
\bibfield{author}{\bibinfo{person}{David~H Wolpert} {and}
  \bibinfo{person}{William~G Macready}.} \bibinfo{year}{1997}\natexlab{}.
\newblock \showarticletitle{No free lunch theorems for optimization}.
\newblock \bibinfo{journal}{\emph{IEEE Transactions on Evolutionary
  Computation}} \bibinfo{volume}{1}, \bibinfo{number}{1}
  (\bibinfo{year}{1997}), \bibinfo{pages}{67--82}.
\newblock


\bibitem[\protect\citeauthoryear{Wu, Dai, Zhang, Wang, Sun, Wu, Tian, Vajda,
  Jia, and Keutzer}{Wu et~al\mbox{.}}{2019}]%
        {wu2019fbnet}
\bibfield{author}{\bibinfo{person}{Bichen Wu}, \bibinfo{person}{Xiaoliang Dai},
  \bibinfo{person}{Peizhao Zhang}, \bibinfo{person}{Yanghan Wang},
  \bibinfo{person}{Fei Sun}, \bibinfo{person}{Yiming Wu},
  \bibinfo{person}{Yuandong Tian}, \bibinfo{person}{Peter Vajda},
  \bibinfo{person}{Yangqing Jia}, {and} \bibinfo{person}{Kurt Keutzer}.}
  \bibinfo{year}{2019}\natexlab{}.
\newblock \showarticletitle{Fbnet: Hardware-aware efficient convnet design via
  differentiable neural architecture search}. In
  \bibinfo{booktitle}{\emph{Proceedings of the IEEE Conference Computer Vision
  Pattern Recognition (CVPR)}}. \bibinfo{pages}{10734--10742}.
\newblock


\bibitem[\protect\citeauthoryear{Wu, Wang, Zhang, Tian, Vajda, and Keutzer}{Wu
  et~al\mbox{.}}{2018}]%
        {wu2018mixed}
\bibfield{author}{\bibinfo{person}{Bichen Wu}, \bibinfo{person}{Yanghan Wang},
  \bibinfo{person}{Peizhao Zhang}, \bibinfo{person}{Yuandong Tian},
  \bibinfo{person}{Peter Vajda}, {and} \bibinfo{person}{Kurt Keutzer}.}
  \bibinfo{year}{2018}\natexlab{}.
\newblock \showarticletitle{Mixed precision quantization of convnets via
  differentiable neural architecture search}.
\newblock \bibinfo{journal}{\emph{arXiv preprint arXiv:1812.00090}}
  (\bibinfo{year}{2018}).
\newblock


\bibitem[\protect\citeauthoryear{Xian, Lampert, Schiele, and Akata}{Xian
  et~al\mbox{.}}{2018}]%
        {xian2018zero}
\bibfield{author}{\bibinfo{person}{Yongqin Xian}, \bibinfo{person}{Christoph~H
  Lampert}, \bibinfo{person}{Bernt Schiele}, {and} \bibinfo{person}{Zeynep
  Akata}.} \bibinfo{year}{2018}\natexlab{}.
\newblock \showarticletitle{Zero-shot learning—a comprehensive evaluation of
  the good, the bad and the ugly}.
\newblock \bibinfo{journal}{\emph{IEEE Transactions on Pattern Analysis and
  Machine Intelligence (TPAMI)}} \bibinfo{volume}{41}, \bibinfo{number}{9}
  (\bibinfo{year}{2018}), \bibinfo{pages}{2251--2265}.
\newblock


\bibitem[\protect\citeauthoryear{Xie, Kirillov, Girshick, and He}{Xie
  et~al\mbox{.}}{2019a}]%
        {xie2019exploring}
\bibfield{author}{\bibinfo{person}{Saining Xie}, \bibinfo{person}{Alexander
  Kirillov}, \bibinfo{person}{Ross Girshick}, {and} \bibinfo{person}{Kaiming
  He}.} \bibinfo{year}{2019}\natexlab{a}.
\newblock \showarticletitle{Exploring randomly wired neural networks for image
  recognition}. In \bibinfo{booktitle}{\emph{Proceedings of the IEEE
  International Conference Computer Vision (ICCV)}}.
  \bibinfo{pages}{1284--1293}.
\newblock


\bibitem[\protect\citeauthoryear{Xie, Zheng, Liu, and Lin}{Xie
  et~al\mbox{.}}{2019b}]%
        {xie2019snas}
\bibfield{author}{\bibinfo{person}{Sirui Xie}, \bibinfo{person}{Hehui Zheng},
  \bibinfo{person}{Chunxiao Liu}, {and} \bibinfo{person}{Liang Lin}.}
  \bibinfo{year}{2019}\natexlab{b}.
\newblock \showarticletitle{{SNAS}: stochastic neural architecture search}. In
  \bibinfo{booktitle}{\emph{International Conference on Learning
  Representations (ICLR)}}.
\newblock


\bibitem[\protect\citeauthoryear{Xiong, Huang, Yu, Liu, Zhu, and Shao}{Xiong
  et~al\mbox{.}}{2020}]%
        {xiong2020number}
\bibfield{author}{\bibinfo{person}{Huan Xiong}, \bibinfo{person}{Lei Huang},
  \bibinfo{person}{Mengyang Yu}, \bibinfo{person}{Li Liu}, \bibinfo{person}{Fan
  Zhu}, {and} \bibinfo{person}{Ling Shao}.} \bibinfo{year}{2020}\natexlab{}.
\newblock \showarticletitle{On the number of linear regions of convolutional
  neural networks}. In \bibinfo{booktitle}{\emph{The International Conference
  on Machine Learning (ICML)}}. \bibinfo{pages}{10514--10523}.
\newblock


\bibitem[\protect\citeauthoryear{Xiong, Mehta, and Singh}{Xiong
  et~al\mbox{.}}{2019}]%
        {xiong2019resource}
\bibfield{author}{\bibinfo{person}{Yunyang Xiong}, \bibinfo{person}{Ronak
  Mehta}, {and} \bibinfo{person}{Vikas Singh}.}
  \bibinfo{year}{2019}\natexlab{}.
\newblock \showarticletitle{Resource constrained neural network architecture
  search: Will a submodularity assumption help?}. In
  \bibinfo{booktitle}{\emph{Proceedings of the IEEE International Conference
  Computer Vision (ICCV)}}. \bibinfo{pages}{1901--1910}.
\newblock


\bibitem[\protect\citeauthoryear{Xu, Xie, Zhang, Chen, Qi, Tian, and Xiong}{Xu
  et~al\mbox{.}}{2020b}]%
        {xu2020pc}
\bibfield{author}{\bibinfo{person}{Yuhui Xu}, \bibinfo{person}{Lingxi Xie},
  \bibinfo{person}{Xiaopeng Zhang}, \bibinfo{person}{Xin Chen},
  \bibinfo{person}{Guo-Jun Qi}, \bibinfo{person}{Qi Tian}, {and}
  \bibinfo{person}{Hongkai Xiong}.} \bibinfo{year}{2020}\natexlab{b}.
\newblock \showarticletitle{{PC-DARTS}: Partial channel connections for
  memory-efficient architecture search}. In
  \bibinfo{booktitle}{\emph{International Conference on Learning
  Representations (ICLR)}}.
\newblock


\bibitem[\protect\citeauthoryear{Xu, van Hasselt, Hessel, Oh, Singh, and
  Silver}{Xu et~al\mbox{.}}{2020a}]%
        {xu2020meta}
\bibfield{author}{\bibinfo{person}{Zhongwen Xu}, \bibinfo{person}{Hado van
  Hasselt}, \bibinfo{person}{Matteo Hessel}, \bibinfo{person}{Junhyuk Oh},
  \bibinfo{person}{Satinder Singh}, {and} \bibinfo{person}{David Silver}.}
  \bibinfo{year}{2020}\natexlab{a}.
\newblock \showarticletitle{Meta-gradient reinforcement learning with an
  objective discovered online}. In \bibinfo{booktitle}{\emph{Proceedings of the
  International Conference on Neural Information Processing Systems
  (NeurIPS)}}.
\newblock


\bibitem[\protect\citeauthoryear{Xu, van Hasselt, and Silver}{Xu
  et~al\mbox{.}}{2018}]%
        {xu2018meta}
\bibfield{author}{\bibinfo{person}{Zhongwen Xu}, \bibinfo{person}{Hado van
  Hasselt}, {and} \bibinfo{person}{David Silver}.}
  \bibinfo{year}{2018}\natexlab{}.
\newblock \showarticletitle{Meta-gradient reinforcement learning}. In
  \bibinfo{booktitle}{\emph{Proceedings of the International Conference on
  Neural Information Processing Systems (NeurIPS)}}.
\newblock


\bibitem[\protect\citeauthoryear{Yan, White, Savani, and Hutter}{Yan
  et~al\mbox{.}}{2021}]%
        {NAS-Bench-x11}
\bibfield{author}{\bibinfo{person}{Shen Yan}, \bibinfo{person}{Colin White},
  \bibinfo{person}{Yash Savani}, {and} \bibinfo{person}{Frank Hutter}.}
  \bibinfo{year}{2021}\natexlab{}.
\newblock \showarticletitle{NAS-Bench-x11 and the Power of Learning Curves}. In
  \bibinfo{booktitle}{\emph{Proceedings of the International Conference on
  Neural Information Processing Systems (NeurIPS)}}.
\newblock


\bibitem[\protect\citeauthoryear{Yang and Shami}{Yang and Shami}{2020}]%
        {yang2020hyperparameter}
\bibfield{author}{\bibinfo{person}{Li Yang} {and} \bibinfo{person}{Abdallah
  Shami}.} \bibinfo{year}{2020}\natexlab{}.
\newblock \showarticletitle{On hyperparameter optimization of machine learning
  algorithms: Theory and practice}.
\newblock \bibinfo{journal}{\emph{Neurocomputing}}  \bibinfo{volume}{415}
  (\bibinfo{year}{2020}), \bibinfo{pages}{295--316}.
\newblock


\bibitem[\protect\citeauthoryear{Yang, Yan, Li, Kwon, Lai, Krishna, Chandra,
  Jiang, and Shi}{Yang et~al\mbox{.}}{2020c}]%
        {yang2020co}
\bibfield{author}{\bibinfo{person}{Lei Yang}, \bibinfo{person}{Zheyu Yan},
  \bibinfo{person}{Meng Li}, \bibinfo{person}{Hyoukjun Kwon},
  \bibinfo{person}{Liangzhen Lai}, \bibinfo{person}{Tushar Krishna},
  \bibinfo{person}{Vikas Chandra}, \bibinfo{person}{Weiwen Jiang}, {and}
  \bibinfo{person}{Yiyu Shi}.} \bibinfo{year}{2020}\natexlab{c}.
\newblock \showarticletitle{Co-exploration of neural architectures and
  heterogeneous asic accelerator designs targeting multiple tasks}. In
  \bibinfo{booktitle}{\emph{The ACM/ESDA/IEEE Design Automation Conference
  (DAC)}}. \bibinfo{pages}{1--6}.
\newblock


\bibitem[\protect\citeauthoryear{Yang, Ji, Xu, Wang, Lv, Yu, Gong, Dikmen, Lin,
  and Huang}{Yang et~al\mbox{.}}{2009}]%
        {yang2009detecting}
\bibfield{author}{\bibinfo{person}{Ming Yang}, \bibinfo{person}{Shuiwang Ji},
  \bibinfo{person}{Wei Xu}, \bibinfo{person}{Jinjun Wang},
  \bibinfo{person}{Fengjun Lv}, \bibinfo{person}{Kai Yu},
  \bibinfo{person}{Yihong Gong}, \bibinfo{person}{Mert Dikmen},
  \bibinfo{person}{Dennis~J Lin}, {and} \bibinfo{person}{Thomas~S Huang}.}
  \bibinfo{year}{2009}\natexlab{}.
\newblock \showarticletitle{Detecting human actions in surveillance videos}. In
  \bibinfo{booktitle}{\emph{TREC Video Retrieval Evaluation Workshop}}.
  Citeseer.
\newblock


\bibitem[\protect\citeauthoryear{Yang, Chen, and Sze}{Yang
  et~al\mbox{.}}{2017}]%
        {yang2017designing}
\bibfield{author}{\bibinfo{person}{Tien-Ju Yang}, \bibinfo{person}{Yu-Hsin
  Chen}, {and} \bibinfo{person}{Vivienne Sze}.}
  \bibinfo{year}{2017}\natexlab{}.
\newblock \showarticletitle{Designing energy-efficient convolutional neural
  networks using energy-aware pruning}. In
  \bibinfo{booktitle}{\emph{Proceedings of the IEEE Conference Computer Vision
  Pattern Recognition (CVPR)}}. \bibinfo{pages}{5687--5695}.
\newblock


\bibitem[\protect\citeauthoryear{Yang, Howard, Chen, Zhang, Go, Sandler, Sze,
  and Adam}{Yang et~al\mbox{.}}{2018}]%
        {yang2018netadapt}
\bibfield{author}{\bibinfo{person}{Tien-Ju Yang}, \bibinfo{person}{Andrew
  Howard}, \bibinfo{person}{Bo Chen}, \bibinfo{person}{Xiao Zhang},
  \bibinfo{person}{Alec Go}, \bibinfo{person}{Mark Sandler},
  \bibinfo{person}{Vivienne Sze}, {and} \bibinfo{person}{Hartwig Adam}.}
  \bibinfo{year}{2018}\natexlab{}.
\newblock \showarticletitle{{NetAdapt}: Platform-aware neural network
  adaptation for mobile applications}. In \bibinfo{booktitle}{\emph{Proceedings
  of the European Conference on Computer Vision (ECCV)}}.
  \bibinfo{pages}{285--300}.
\newblock


\bibitem[\protect\citeauthoryear{Yang, Liu, Zhou, Bian, and Liu}{Yang
  et~al\mbox{.}}{2020a}]%
        {yang2020qlib}
\bibfield{author}{\bibinfo{person}{Xiao Yang}, \bibinfo{person}{Weiqing Liu},
  \bibinfo{person}{Dong Zhou}, \bibinfo{person}{Jiang Bian}, {and}
  \bibinfo{person}{Tie-Yan Liu}.} \bibinfo{year}{2020}\natexlab{a}.
\newblock \showarticletitle{Qlib: An AI-oriented Quantitative Investment
  Platform}.
\newblock \bibinfo{journal}{\emph{arXiv preprint arXiv:2009.11189}}
  (\bibinfo{year}{2020}).
\newblock


\bibitem[\protect\citeauthoryear{Yang, Wang, Han, Xu, Xu, Tao, and Xu}{Yang
  et~al\mbox{.}}{2020b}]%
        {yang2020searching}
\bibfield{author}{\bibinfo{person}{Zhaohui Yang}, \bibinfo{person}{Yunhe Wang},
  \bibinfo{person}{Kai Han}, \bibinfo{person}{Chunjing Xu},
  \bibinfo{person}{Chao Xu}, \bibinfo{person}{Dacheng Tao}, {and}
  \bibinfo{person}{Chang Xu}.} \bibinfo{year}{2020}\natexlab{b}.
\newblock \showarticletitle{Searching for Low-Bit Weights in Quantized Neural
  Networks}. In \bibinfo{booktitle}{\emph{Proceedings of the International
  Conference on Neural Information Processing Systems (NeurIPS)}}.
\newblock


\bibitem[\protect\citeauthoryear{Ying}{Ying}{2019}]%
        {ying2019enumerating}
\bibfield{author}{\bibinfo{person}{Chris Ying}.}
  \bibinfo{year}{2019}\natexlab{}.
\newblock \showarticletitle{Enumerating unique computational graphs via an
  iterative graph invariant}.
\newblock \bibinfo{journal}{\emph{arXiv preprint arXiv:1902.06192}}
  (\bibinfo{year}{2019}).
\newblock


\bibitem[\protect\citeauthoryear{Ying, Klein, Christiansen, Real, Murphy, and
  Hutter}{Ying et~al\mbox{.}}{2019}]%
        {ying2019bench}
\bibfield{author}{\bibinfo{person}{Chris Ying}, \bibinfo{person}{Aaron Klein},
  \bibinfo{person}{Eric Christiansen}, \bibinfo{person}{Esteban Real},
  \bibinfo{person}{Kevin Murphy}, {and} \bibinfo{person}{Frank Hutter}.}
  \bibinfo{year}{2019}\natexlab{}.
\newblock \showarticletitle{{NAS-Bench-101}: Towards reproducible neural
  architecture search}. In \bibinfo{booktitle}{\emph{The International
  Conference on Machine Learning (ICML)}}. \bibinfo{pages}{7105--7114}.
\newblock


\bibitem[\protect\citeauthoryear{Yu, Jin, Liu, Bender, Kindermans, Tan, Huang,
  Song, Pang, and Le}{Yu et~al\mbox{.}}{2020a}]%
        {yu2020bignas}
\bibfield{author}{\bibinfo{person}{Jiahui Yu}, \bibinfo{person}{Pengchong Jin},
  \bibinfo{person}{Hanxiao Liu}, \bibinfo{person}{Gabriel Bender},
  \bibinfo{person}{Pieter-Jan Kindermans}, \bibinfo{person}{Mingxing Tan},
  \bibinfo{person}{Thomas Huang}, \bibinfo{person}{Xiaodan Song},
  \bibinfo{person}{Ruoming Pang}, {and} \bibinfo{person}{Quoc Le}.}
  \bibinfo{year}{2020}\natexlab{a}.
\newblock \showarticletitle{{BigNAS}: Scaling up neural architecture search
  with big single-stage models}. In \bibinfo{booktitle}{\emph{Proceedings of
  the European Conference on Computer Vision (ECCV)}}.
  \bibinfo{pages}{702--717}.
\newblock


\bibitem[\protect\citeauthoryear{Yu, Yang, Xu, Yang, and Huang}{Yu
  et~al\mbox{.}}{2019}]%
        {yu2019slimmable}
\bibfield{author}{\bibinfo{person}{Jiahui Yu}, \bibinfo{person}{Linjie Yang},
  \bibinfo{person}{Ning Xu}, \bibinfo{person}{Jianchao Yang}, {and}
  \bibinfo{person}{Thomas Huang}.} \bibinfo{year}{2019}\natexlab{}.
\newblock \showarticletitle{Slimmable neural networks}. In
  \bibinfo{booktitle}{\emph{International Conference on Learning
  Representations (ICLR)}}.
\newblock


\bibitem[\protect\citeauthoryear{Yu, Sciuto, Jaggi, Musat, and Salzmann}{Yu
  et~al\mbox{.}}{2020b}]%
        {yu2020evaluating}
\bibfield{author}{\bibinfo{person}{Kaicheng Yu}, \bibinfo{person}{Christian
  Sciuto}, \bibinfo{person}{Martin Jaggi}, \bibinfo{person}{Claudiu Musat},
  {and} \bibinfo{person}{Mathieu Salzmann}.} \bibinfo{year}{2020}\natexlab{b}.
\newblock \showarticletitle{Evaluating the search phase of neural architecture
  search}. In \bibinfo{booktitle}{\emph{International Conference on Learning
  Representations (ICLR)}}.
\newblock


\bibitem[\protect\citeauthoryear{Yu and Zhu}{Yu and Zhu}{2020}]%
        {yu2020hyper}
\bibfield{author}{\bibinfo{person}{Tong Yu} {and} \bibinfo{person}{Hong Zhu}.}
  \bibinfo{year}{2020}\natexlab{}.
\newblock \showarticletitle{Hyper-parameter optimization: A review of
  algorithms and applications}.
\newblock \bibinfo{journal}{\emph{arXiv preprint arXiv:2003.05689}}
  (\bibinfo{year}{2020}).
\newblock


\bibitem[\protect\citeauthoryear{Zaidi, Zela, Elsken, Holmes, Hutter, and
  Teh}{Zaidi et~al\mbox{.}}{2020}]%
        {zaidi2020neural}
\bibfield{author}{\bibinfo{person}{Sheheryar Zaidi}, \bibinfo{person}{Arber
  Zela}, \bibinfo{person}{Thomas Elsken}, \bibinfo{person}{Chris Holmes},
  \bibinfo{person}{Frank Hutter}, {and} \bibinfo{person}{Yee~Whye Teh}.}
  \bibinfo{year}{2020}\natexlab{}.
\newblock \showarticletitle{Neural ensemble search for performant and
  calibrated predictions}. In \bibinfo{booktitle}{\emph{The International
  Conference on Machine Learning (ICML) Workshop}}.
\newblock


\bibitem[\protect\citeauthoryear{Zela, Elsken, Saikia, Marrakchi, Brox, and
  Hutter}{Zela et~al\mbox{.}}{2020a}]%
        {zela2020understanding}
\bibfield{author}{\bibinfo{person}{Arber Zela}, \bibinfo{person}{Thomas
  Elsken}, \bibinfo{person}{Tonmoy Saikia}, \bibinfo{person}{Yassine
  Marrakchi}, \bibinfo{person}{Thomas Brox}, {and} \bibinfo{person}{Frank
  Hutter}.} \bibinfo{year}{2020}\natexlab{a}.
\newblock \showarticletitle{Understanding and robustifying differentiable
  architecture search}. In \bibinfo{booktitle}{\emph{International Conference
  on Learning Representations (ICLR)}}.
\newblock


\bibitem[\protect\citeauthoryear{Zela, Klein, Falkner, and Hutter}{Zela
  et~al\mbox{.}}{2018}]%
        {zela2018towards}
\bibfield{author}{\bibinfo{person}{Arber Zela}, \bibinfo{person}{Aaron Klein},
  \bibinfo{person}{Stefan Falkner}, {and} \bibinfo{person}{Frank Hutter}.}
  \bibinfo{year}{2018}\natexlab{}.
\newblock \showarticletitle{Towards automated deep learning: Efficient joint
  neural architecture and hyperparameter search}. In
  \bibinfo{booktitle}{\emph{The International Conference on Machine Learning
  (ICML) Workshop}}.
\newblock


\bibitem[\protect\citeauthoryear{Zela, Siems, and Hutter}{Zela
  et~al\mbox{.}}{2020b}]%
        {zela2020bench}
\bibfield{author}{\bibinfo{person}{Arber Zela}, \bibinfo{person}{Julien Siems},
  {and} \bibinfo{person}{Frank Hutter}.} \bibinfo{year}{2020}\natexlab{b}.
\newblock \showarticletitle{{NAS-Bench-1Shot1}: Benchmarking and dissecting
  one-shot neural architecture search}. In
  \bibinfo{booktitle}{\emph{International Conference on Learning
  Representations (ICLR)}}.
\newblock


\bibitem[\protect\citeauthoryear{Zhang, Han, Yang, Zhang, Liu, Yang, and
  Zhou}{Zhang et~al\mbox{.}}{2020}]%
        {zhang2020retiarii}
\bibfield{author}{\bibinfo{person}{Quanlu Zhang}, \bibinfo{person}{Zhenhua
  Han}, \bibinfo{person}{Fan Yang}, \bibinfo{person}{Yuge Zhang},
  \bibinfo{person}{Zhe Liu}, \bibinfo{person}{Mao Yang}, {and}
  \bibinfo{person}{Lidong Zhou}.} \bibinfo{year}{2020}\natexlab{}.
\newblock \showarticletitle{Retiarii: A Deep Learning Exploratory-Training
  Framework}. In \bibinfo{booktitle}{\emph{The USENIX Symposium on Operating
  Systems Design and Implementation (OSDI)}}. \bibinfo{pages}{919--936}.
\newblock


\bibitem[\protect\citeauthoryear{Zhang and Dietterich}{Zhang and
  Dietterich}{1995}]%
        {zhang1995reinforcement}
\bibfield{author}{\bibinfo{person}{Wei Zhang} {and} \bibinfo{person}{Thomas~G
  Dietterich}.} \bibinfo{year}{1995}\natexlab{}.
\newblock \showarticletitle{A reinforcement learning approach to job-shop
  scheduling}. In \bibinfo{booktitle}{\emph{International Joint Conferences on
  Artificial Intelligence (IJCAI)}}, Vol.~\bibinfo{volume}{95}.
  \bibinfo{pages}{1114--1120}.
\newblock


\bibitem[\protect\citeauthoryear{Zhong, Yan, Wu, Shao, and Liu}{Zhong
  et~al\mbox{.}}{2018}]%
        {zhong2018practical}
\bibfield{author}{\bibinfo{person}{Zhao Zhong}, \bibinfo{person}{Junjie Yan},
  \bibinfo{person}{Wei Wu}, \bibinfo{person}{Jing Shao}, {and}
  \bibinfo{person}{Cheng-Lin Liu}.} \bibinfo{year}{2018}\natexlab{}.
\newblock \showarticletitle{Practical block-wise neural network architecture
  generation}. In \bibinfo{booktitle}{\emph{Proceedings of the IEEE Conference
  Computer Vision Pattern Recognition (CVPR)}}. \bibinfo{pages}{2423--2432}.
\newblock


\bibitem[\protect\citeauthoryear{Zhou, Yang, Wang, and Pan}{Zhou
  et~al\mbox{.}}{2019}]%
        {zhou2019bayesnas}
\bibfield{author}{\bibinfo{person}{Hongpeng Zhou}, \bibinfo{person}{Minghao
  Yang}, \bibinfo{person}{Jun Wang}, {and} \bibinfo{person}{Wei Pan}.}
  \bibinfo{year}{2019}\natexlab{}.
\newblock \showarticletitle{{BayesNAS}: A bayesian approach for neural
  architecture search}. In \bibinfo{booktitle}{\emph{The International
  Conference on Machine Learning (ICML)}}. \bibinfo{pages}{7603--7613}.
\newblock


\bibitem[\protect\citeauthoryear{Zhou, Dong, Akin, Tan, Peng, Meng,
  Yazdanbakhsh, Huang, Narayanaswami, and Laudon}{Zhou et~al\mbox{.}}{2022}]%
        {zhou2022rethinking}
\bibfield{author}{\bibinfo{person}{Yanqi Zhou}, \bibinfo{person}{Xuanyi Dong},
  \bibinfo{person}{Berkin Akin}, \bibinfo{person}{Mingxing Tan},
  \bibinfo{person}{Daiyi Peng}, \bibinfo{person}{Tianjian Meng},
  \bibinfo{person}{Amir Yazdanbakhsh}, \bibinfo{person}{Da Huang},
  \bibinfo{person}{Ravi Narayanaswami}, {and} \bibinfo{person}{James Laudon}.}
  \bibinfo{year}{2022}\natexlab{}.
\newblock \showarticletitle{Rethinking Co-design of Neural Architectures and
  Hardware Accelerators}. In \bibinfo{booktitle}{\emph{The Conference on
  Machine Learning and Systems (MLSys)}}.
\newblock


\bibitem[\protect\citeauthoryear{Zhou, Sun, and Li}{Zhou et~al\mbox{.}}{2009}]%
        {zhou2009multi}
\bibfield{author}{\bibinfo{person}{Zhi-Hua Zhou}, \bibinfo{person}{Yu-Yin Sun},
  {and} \bibinfo{person}{Yu-Feng Li}.} \bibinfo{year}{2009}\natexlab{}.
\newblock \showarticletitle{Multi-instance learning by treating instances as
  non-iid samples}. In \bibinfo{booktitle}{\emph{The International Conference
  on Machine Learning (ICML)}}. \bibinfo{pages}{1249--1256}.
\newblock


\bibitem[\protect\citeauthoryear{Zhu, Park, Isola, and Efros}{Zhu
  et~al\mbox{.}}{2017}]%
        {zhu2017unpaired}
\bibfield{author}{\bibinfo{person}{Jun-Yan Zhu}, \bibinfo{person}{Taesung
  Park}, \bibinfo{person}{Phillip Isola}, {and} \bibinfo{person}{Alexei~A
  Efros}.} \bibinfo{year}{2017}\natexlab{}.
\newblock \showarticletitle{Unpaired image-to-image translation using
  cycle-consistent adversarial networks}. In
  \bibinfo{booktitle}{\emph{Proceedings of the IEEE International Conference
  Computer Vision (ICCV)}}. \bibinfo{pages}{2223--2232}.
\newblock


\bibitem[\protect\citeauthoryear{Zimmer, Lindauer, and Hutter}{Zimmer
  et~al\mbox{.}}{2021}]%
        {zimmer-tpami21a}
\bibfield{author}{\bibinfo{person}{Lucas Zimmer}, \bibinfo{person}{Marius
  Lindauer}, {and} \bibinfo{person}{Frank Hutter}.}
  \bibinfo{year}{2021}\natexlab{}.
\newblock \showarticletitle{{Auto-PyTorch Tabular}: Multi-Fidelity MetaLearning
  for Efficient and Robust AutoDL}.
\newblock \bibinfo{journal}{\emph{IEEE Transactions on Pattern Analysis and
  Machine Intelligence (TPAMI)}} (\bibinfo{year}{2021}).
\newblock


\bibitem[\protect\citeauthoryear{Zliobaite, Bifet, Gaber, Gabrys, Gama, Minku,
  and Musial}{Zliobaite et~al\mbox{.}}{2012}]%
        {zliobaite2012next}
\bibfield{author}{\bibinfo{person}{Indre Zliobaite}, \bibinfo{person}{Albert
  Bifet}, \bibinfo{person}{Mohamed Gaber}, \bibinfo{person}{Bogdan Gabrys},
  \bibinfo{person}{Joao Gama}, \bibinfo{person}{Leandro Minku}, {and}
  \bibinfo{person}{Katarzyna Musial}.} \bibinfo{year}{2012}\natexlab{}.
\newblock \showarticletitle{Next challenges for adaptive learning systems}.
\newblock \bibinfo{journal}{\emph{ACM SIGKDD Explorations Newsletter}}
  \bibinfo{volume}{14}, \bibinfo{number}{1} (\bibinfo{year}{2012}),
  \bibinfo{pages}{48--55}.
\newblock


\bibitem[\protect\citeauthoryear{{\v{Z}}liobait{\.{e}}, Budka, and
  Stahl}{{\v{Z}}liobait{\.{e}} et~al\mbox{.}}{2015}]%
        {zlbu15}
\bibfield{author}{\bibinfo{person}{Indr{\.{e}} {\v{Z}}liobait{\.{e}}},
  \bibinfo{person}{Marcin Budka}, {and} \bibinfo{person}{Frederic Stahl}.}
  \bibinfo{year}{2015}\natexlab{}.
\newblock \showarticletitle{Towards cost-sensitive adaptation: When is it worth
  updating your predictive model?}
\newblock \bibinfo{journal}{\emph{Neurocomputing}}  \bibinfo{volume}{150}
  (\bibinfo{date}{feb} \bibinfo{year}{2015}), \bibinfo{pages}{240--249}.
\newblock
\urldef\tempurl%
\url{https://doi.org/10.1016/j.neucom.2014.05.084}
\showDOI{\tempurl}


\bibitem[\protect\citeauthoryear{Zliobaite and Gabrys}{Zliobaite and
  Gabrys}{2014}]%
        {zliobaite2014adaptive}
\bibfield{author}{\bibinfo{person}{Indre Zliobaite} {and}
  \bibinfo{person}{Bogdan Gabrys}.} \bibinfo{year}{2014}\natexlab{}.
\newblock \showarticletitle{Adaptive preprocessing for streaming data}.
\newblock \bibinfo{journal}{\emph{IEEE Transactions on Knowledge and Data
  Engineering (TKDE)}} \bibinfo{volume}{26}, \bibinfo{number}{2}
  (\bibinfo{year}{2014}), \bibinfo{pages}{309--321}.
\newblock


\bibitem[\protect\citeauthoryear{Zoph, Cubuk, Ghiasi, Lin, Shlens, and Le}{Zoph
  et~al\mbox{.}}{2020}]%
        {zoph2020learning}
\bibfield{author}{\bibinfo{person}{Barret Zoph}, \bibinfo{person}{Ekin~D
  Cubuk}, \bibinfo{person}{Golnaz Ghiasi}, \bibinfo{person}{Tsung-Yi Lin},
  \bibinfo{person}{Jonathon Shlens}, {and} \bibinfo{person}{Quoc~V Le}.}
  \bibinfo{year}{2020}\natexlab{}.
\newblock \showarticletitle{Learning data augmentation strategies for object
  detection}. In \bibinfo{booktitle}{\emph{Proceedings of the European
  Conference on Computer Vision (ECCV)}}. Springer, \bibinfo{pages}{566--583}.
\newblock


\bibitem[\protect\citeauthoryear{Zoph and Le}{Zoph and Le}{2017}]%
        {zoph2017NAS}
\bibfield{author}{\bibinfo{person}{Barret Zoph} {and} \bibinfo{person}{Quoc~V
  Le}.} \bibinfo{year}{2017}\natexlab{}.
\newblock \showarticletitle{Neural architecture search with reinforcement
  learning}. In \bibinfo{booktitle}{\emph{International Conference on Learning
  Representations (ICLR)}}.
\newblock


\bibitem[\protect\citeauthoryear{Zoph, Vasudevan, Shlens, and Le}{Zoph
  et~al\mbox{.}}{2018}]%
        {zoph2018learning}
\bibfield{author}{\bibinfo{person}{Barret Zoph}, \bibinfo{person}{Vijay
  Vasudevan}, \bibinfo{person}{Jonathon Shlens}, {and} \bibinfo{person}{Quoc~V
  Le}.} \bibinfo{year}{2018}\natexlab{}.
\newblock \showarticletitle{Learning transferable architectures for scalable
  image recognition}. In \bibinfo{booktitle}{\emph{Proceedings of the IEEE
  Conference Computer Vision Pattern Recognition (CVPR)}}.
  \bibinfo{pages}{8697--8710}.
\newblock


\end{thebibliography}

\end{document}

% \title{Automated Deep Learning: A Survey}
% \title{A Comprehensive Survey of Automated Deep Learning}
% \title{A Comprehensive Survey of Automated Deep Learning: Think Outside NAS and HPO}
\title{Supplementary Material for ``Automated Deep Learning: Think Outside NAS and HPO''}

% @article{dong2021automated,
%   title  = {Automated Deep Learning: Think Outside NAS and HPO},
%   author = {Dong, Xuanyi and David, Kedziora and Musial, Katarzyna and Gabrys, Bogdan},
%   journal= {Technical Report},
%   year   = {2021}
% }

\author{Xuanyi Dong, David Jacob Kedziora, Katarzyna Musial, Bogdan Gabrys
\IEEEcompsocitemizethanks{
% \IEEEcompsocthanksitem
\IEEEcompsocthanksitem Xuanyi Dong, David Jacob Kedziora, Katarzyna Musial and Bogdan Gabrys are with Faculty of Engineering and IT, University of Technology Sydney, NSW, Australia. (e-mail: Xuanyi.Dxy@gmail.com, David.Kedziora@uts.edu.au, Katarzyna.Musial-Gabrys@uts.edu.au, Bogdan.Gabrys@uts.edu.au)\protect
}% % <-this % stops an unwanted space
}

\markboth{Technical Report}{}

% Recap
% Analysis  / Comments / 

% Global Problem:
% 1, The concrete novelty / the main contribution of this survey.
%      - more broader scope yet less biased than existing surveys (just focused on academy)
%      - two complementary perspectives to dissect AutoDL
%        -- top-down,
%        -- bottom-up,
% new perspective, how close the gap between the current state vs. final goal.
% // some reviews are technical but narrowed, somehow misleading,
% // technical depth,
% // Most recent AutoML contributions, solely driven by xx. not just model selection / training
% // End-to-end automated. Obstacles-vs-opportunities, discussion 
% // CASH, NAS, HPO, environment changed, 
% // Outside-of-the-box
% [Why] not-needed / alternative / too difficulty
%
% generate codes
%
%
% Could we unify some concepts and terminology ? 
%
% 2, Motivation for AutoDL:
%      - currently narrows down to accuracy gain.
%      - 
%
% 3, term,
%
% Minor [1], UK/Australian vs. US spelling
% Minor [2], Why DL is a thing,
%
% I hope to make ManualDL vs. AutoDL very clear in this manuscript.
%
% How to define the auto ????

\IEEEtitleabstractindextext{
\begin{abstract}
In this document, we provide supplementary material for ``Automated Deep Learning: Think Outside NAS and HPO''
\end{abstract}
}

% \begin{IEEEkeywords}
% Automated Deep Learning, Survey, Neural Architecture Search, Hyperparameter Optimization, Hardware Search
% \end{IEEEkeywords}
% }

% make the title area
\maketitle

\IEEEdisplaynontitleabstractindextext
\IEEEpeerreviewmaketitle

\IEEEraisesectionheading{\section{Introduction}\label{sec:introduction}}

\begin{table}[!htp]
\centering
\setlength{\tabcolsep}{3pt}
\begin{tabular}{ l | l }
\toprule
Term                & Explanation \\
\midrule
Problem Context     & \\
ML Task             & \\
Convolution Kernel  & the weight tensor for convolutional layers \\
NAS                 & Neural Architecture Search \\
HPO                 & Hyperparamter Optimization \\
AutoML              & Automated Machine Learning \\
AutoDL              & Automated Deep Learning    \\
\bottomrule
\end{tabular}
\caption{
We.
}
\label{table:terminology}
\end{table}

\subsection{AutoDL Software}\label{sec:software}

\begin{table}[!ht]
\centering
\setlength{\tabcolsep}{1pt}
\begin{tabular}{l l l}
\toprule
            & Dependency & Link \\
\midrule
Keras Tuner~\cite{keras_tuner}          & Tensorflow        & keras-team.github.io/keras-tuner \\
NNI~\cite{NNI}                          &  No DL Platform   & github.com/microsoft/nni         \\
AutoGluon                               & Apache MXNet      & auto.gluon.ai                    \\
PyGlove~\cite{peng2020pyglove}          & No DL Platform    & Google Internal                   \\
TPOT~\cite{le2020scaling}               & XGBoost & epistasislab.github.io/tpot                \\
Optuna~\cite{optuna_2019}               & & https://optuna.org                                 \\
H2O                                     & & www.h2o.ai                                         \\
TPOT                                    & & epistasislab.github.io/tpot                        \\
Auto-Sklearn~\cite{Feurer2015_11d0e628} & & automl.github.io/auto-sklearn                      \\
HPOLib~\cite{eggensperger2013towards}   & & github.com/automl/HPOlib                           \\
NASLib                                  & & github.com/automl/NASLib                           \\
Auto-WEKA                               & & github.com/automl/autoweka                 \\
\bottomrule
\end{tabular}
\caption{
A summary for AutoDL softwares.
}
\label{table:software}
\end{table}

\newpage

\section{Checklist}

% For the AutoDL checklist
\newcommand{\answerYes}[1][]{\textcolor{blue}{[Yes] #1}}
\newcommand{\answerNo}[1][]{\textcolor{orange}{[No] #1}}
\newcommand{\answerNA}[1][]{\textcolor{gray}{[N/A] #1}}
\newcommand{\answerTODO}[1][]{\textcolor{red}{\bf [TODO]}}

\begin{enumerate}

\item {\uppercase\expandafter{\romannumeral1}. Reproducibility}
\begin{enumerate}
  \item Did you release code for the training and evaluation pipeline for the base DL model?
    \answerTODO{}
  \item Did you release code for the search space?
    \answerTODO{}
  \item Did you release the hyperparameters used for the final evaluation pipeline, as well as random seeds?
    \answerTODO{}
  \item Did you release code for your NAS method?
    \answerTODO{}
  \item Did you release hyperparameters for your NAS method, as well as random seeds?
    \answerTODO{}
\end{enumerate}

\item {\uppercase\expandafter{\romannumeral2}. Engineering}
\begin{enumerate}
  \item Did you paste the report from [xxx]?
    \answerTODO{}
\end{enumerate}

\item {\uppercase\expandafter{\romannumeral3}. Accuracy \& Efficiency}
\begin{enumerate}
  \item Did you report the final performance of your AutoDL algorithm?
    \answerTODO{}
  \item Did you report the computational costs?
    \answerTODO{}
\end{enumerate}

\item {\uppercase\expandafter{\romannumeral4}. Stability \& Sensitivity}
\begin{enumerate}
  \item Did you run multiple times of the algorithm and report mean and variance?
    \answerTODO{}
  \item Did you analyze the sensitivity of the algorithm regarding meta hyperparameters?
    \answerTODO{}
\end{enumerate}

\item {\uppercase\expandafter{\romannumeral5}. Generalizability}
\begin{enumerate}
  \item How many datasets are used to perform search?
    \answerTODO{}
  \item How many datasets are used to perform search?
    \answerTODO{}
  \item Did you analyze the sensitivity of the algorithm regarding meta hyperparameters?
    \answerTODO{}
\end{enumerate}

\end{enumerate}

\bibliographystyle{IEEEtran}
\bibliography{IEEEabrv,egbib}